\documentclass{article}

     \usepackage[preprint]{neurips_2025}

\usepackage[utf8]{inputenc} 
\usepackage[T1]{fontenc}    
\usepackage{hyperref}       
\usepackage{url}            
\usepackage{booktabs}       
\usepackage{amsfonts}       
\usepackage{nicefrac}       
\usepackage{microtype}      
\usepackage{xcolor}         
\usepackage{wrapfig}   
\usepackage{amssymb}

\usepackage{tablefootnote}

\usepackage{multirow}
\usepackage{caption}
\usepackage{amsmath}
\usepackage{amsthm}
\usepackage{dsfont}
\usepackage{algorithm}
\usepackage{algorithmic}
\usepackage[pdftex]{graphicx}
\usepackage{threeparttable}
\usepackage{subcaption}
\usepackage{mathtools}
\usepackage{placeins}
\usepackage{adjustbox}

\theoremstyle{plain}
\newtheorem{theorem}{Theorem}[section]

\theoremstyle{definition}

\newtheorem{example}[theorem]{Example}
\theoremstyle{remark}

\newcommand\cM{\mathcal M}
\newcommand\cL{\mathcal L}

\newcommand\cD{\mathcal D}

\newcommand\bbR{\mathbb R}
\newcommand\bbE{\mathbb E}

\newcommand\sY{\mathsf Y}
\newcommand\sTheta{\mathsf \Theta}

\newcommand{\blue}[1]{\textcolor{blue}{#1}}

\title{Scalable Bayesian Monte Carlo:\\
fast uncertainty estimation beyond deep ensembles}

\author{%
  Xinzhu Liang\\
  Mathematics Department, University of Manchester\\
  Manchester, M13 9PL, UK \\
  \texttt{xinzhu.liang@postgrad.manchester.ac.uk} \\
  \And
  Joseph M. Lukens\thanks{Quantum Information Science Section, Oak Ridge National Laboratory, Oak Ridge, Tennessee 37831, USA} \\
  School of Electrical and Computer Engineering\\
  Purdue University, West Lafayette, Indiana 47907, USA\\
  \AND
  Sanjaya Lohani \\
  Department of Electrical and Computer Engineering, Southern Methodist University \\
  Dallas, Texas 75205, USA\\
  \And
  Brian T. Kirby\thanks{Tulane University, New Orleans, Louisiana 70118, USA} \\
  DEVCOM US Army Research Laboratory\\ Adelphi, Maryland 20783, USA\\
  \And
  Thomas~A. Searles \\
  Department of Electrical and Computer Engineering, University of Illinois Chicago\\
  Chicago, Illinois 60607, USA\\
  \And
  Xin Qiu \\
  Cognizant AI Labs\\
  San Francisco, California 94105, USA
  \And
  Kody J.~H. Law \\
  Mathematics Department, University of Manchester\\
  Manchester, M13 9PL, United Kingdom\\
}

\begin{document}

\maketitle

\begin{abstract}
This work introduces a {\em new method} designed for Bayesian deep learning
called scalable Bayesian Monte Carlo (SBMC).
The method is comprised of a {\em model} and an {\em algorithm}.
The model interpolates between a point estimator and the 
posterior.
The algorithm is a
{\em parallel} implementation of 
sequential Monte Carlo sampler (SMC$_\parallel$) or Markov chain Monte Carlo (MCMC$_\parallel$).
We collectively refer to these {\em consistent} (asymptotically unbiased) algorithms as
{\em Bayesian Monte Carlo} (BMC), and any such algorithm can be used in our SBMC method.
The utility of the method is demonstrated on practical examples: MNIST, CIFAR, IMDb.
A systematic numerical study reveals that 
{\em for the same wall-clock time as state-of-the-art (SOTA) 
methods like deep ensembles (DE)},
SBMC achieves comparable or better accuracy and substantially improved 
uncertainty quantification (UQ)--{\em in particular, epistemic UQ}.
This is demonstrated on the downstream task of estimating the confidence in predictions, 
which can be used for reliability assessment or abstention decisions. 
\end{abstract}

\section{Introduction}
\label{sec:intro}

Uncertainty quantification (UQ) in deep learning is critical for 
safe and reliable deployment, yet remains a core challenge.
The Bayesian formulation provides UQ in addition to Bayes optimal accuracy, by averaging 
realizations from the posterior distribution, rather than relying on a single point estimator.
%
Fully Bayesian approaches like consistent 
Markov chain Monte Carlo (MCMC) and sequential Monte Carlo (SMC)
offer asymptotically unbiased posterior estimates, 
but at the cost of prohibitive compute time compared to simple point estimators like the maximum a posteriori (MAP).
Bayesian deep learning (BDL) often rely on scalable approximations
such as Monte Carlo Dropout \citep{gal2016dropout}, 
deep ensemble (DE) \citep{lakshminarayanan2017simple},
Laplace approximation \citep{daxberger2021laplace,eschenhagen2021mixtures}, 
Stochastic Weight Averaging (SWA) \citep{izmailov2018averaging}, 
SWA-Gaussian (SWAG) \citep{maddox2019simple,wilson2020bayesian}, 
which are fast and provide strong empirical performance,
but lack formal consistency guarantees.

\begin{figure}        
        \centering
        \includegraphics[width=\textwidth]
        {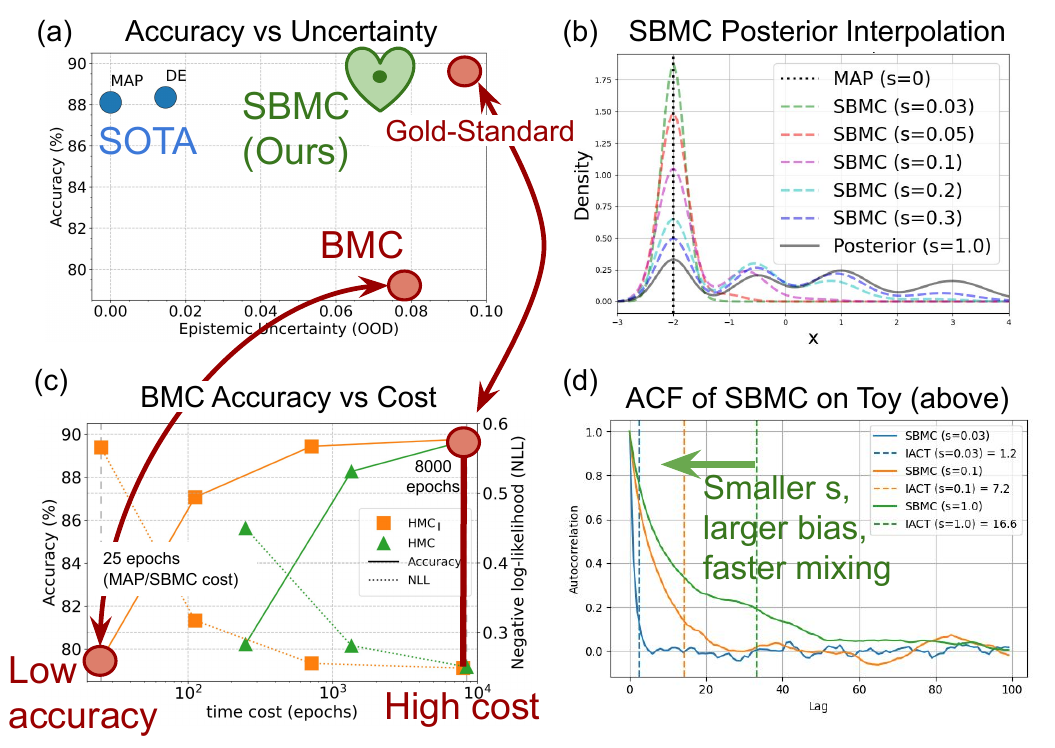}
        \caption{\footnotesize
        {\bf Left panels}: IMDb sentiment classification.
        {\bf (a)} SBMC provides a good {\em balance} of 
        accuracy and UQ (quantified by epistemic entropy on OOD data),
        {\em for the same cost as deep ensembles} 
        (every method runs for 25 epochs except the Gold-Standard (GS) 
        BMC solution, which runs for 8000 epochs).
        {\bf (c)} Standard implementation of HMC and HMC$_\parallel$.
        BMC methods typically deliver high accuracy for high cost (GS)
        and low accuracy for low cost.
        {\bf Right panels}: SBMC {\em approximate models}, 
        on a simple toy example. 
        {\bf (b)} The original 
        posterior ($s=1$) and the approximations for a range of $s$.
        {\bf (d)} The autocorrelation function \protect\footnotemark (ACF: correlation between samples separated by `Lag' steps) 
        of SBMC for very long NUTS \cite{hoffman2014no} chains for a few choicse of $s$. 
        As $s$ decreases the target becomes simpler and hence easier to explore, but the bias (with respect to the posterior) increases.}
        \label{fig:sbmc_cartoon}
\end{figure}

Here we present a new approximate inference method called Scalable Bayesian Monte Carlo (SBMC), 
which bridges the gap between fast but heuristic methods and principled yet expensive samplers.
It is a general method comprised of an approximate {\em model} and {\em algorithm} 
to simulate from the model. 
Our key insight is the model approximation $\pi_s$ which features a 
scalar interpolation parameter $s\in[0,1]$ that allows tuning between
the MAP estimator ($s=0$) and the full Bayesian posterior ($s=1$). 
For smaller $s$ the target is {\em easier to simulate from}, albeit with a larger bias
with respect to the posterior.
See the right panels of Figure \ref{fig:sbmc_cartoon}.  
By simulating from this approximate target with {\em parallel} implementations of BMC algorithms, 
which we will denote by S-SMC$_\parallel$ and S-MCMC$_\parallel$, 
SBMC delivers strong performance in accuracy and UQ {\em at a comparable cost to SOTA methods like DE}. 
The prefix ``S-" is for ``scalable'', and the scalability comes from the model approximation
{\em in tandem} with the parallelism, denoted by the subscript $_\parallel$.
Without the model approximation, the required simulation time is prohibitive.

Given data \(\cD\), the Bayesian posterior distribution over \(\theta \in \sTheta \subseteq \bbR^{d}\) is given by 
\begin{equation}\label{eq:posterior}
    \pi(\theta) \propto \cL(\theta) \pi_0(\theta) \, , 
\end{equation}
where \(\cL(\theta) := \cL(\theta; \cD)\) is the likelihood of
the data \(\cD\) and \(\pi_0(\theta)\) is the prior. 
The Bayes estimator of a quantity of interest \(\varphi:\sTheta \rightarrow \bbR\) is \(\bbE[\varphi|\cD]=\int_\sTheta \varphi(\theta) \pi(\theta) d\theta\). It minimizes the appropriate 
Bayes risk at the population level and as such is
Bayes optimal \citep{mackay1992practical, 
neal2012bayesian, 
andrieu2003introduction, bishop2006pattern}.

 %
\footnotetext{Autocorrelation function (ACF) and integrated autocorrelation time (IACT) are defined in Appendix \ref{app:IACT}.}
In general the posterior (target) distribution can only be evaluated up-to a constant of proportionality, 
and the available consistent methods for inference (learning) are of Monte Carlo type: 
notably Markov chain Monte Carlo (MCMC) 
\citep{metropolis1953equation, hastings1970monte,duane1987hybrid,gelfand1990sampling,geyer1992practical,robert1999monte,roberts1996exponential}
and sequential Monte Carlo (SMC) samplers \citep{jarzynski1997equilibrium,berzuini2001resample,del2006sequential,dai2022invitation,chopin2020introduction}.
The past several decades have seen enormous progress in methodology as well as practical applications
\citep{galison2022first,mohan2024evaluating}, 
however standard implementations of these algorithms are still too expensive for practical 
BDL, and so BMC algorithms are typically
 used only as a benchmark for cheaper approximations \citep{izmailov2021bayesian}.
 See e.g. \citep{angelino2016patterns,papamarkou2024position} 
 for recent reviews and further references. 
The present work aims to address this computational intractability by 
(i) targeting an {\em approximation} of \eqref{eq:posterior},
and (ii) distributing the BMC workload across many workers in parallel.
We will show that these two things together provide a {\em practical and scalable method}.
The focus of the present work is on demonstrating the value of the SBMC method itself,
independently of the particular BMC algorithm used,
and so we mostly focus on standard implementations of HMC and SMC.
But one of the virtues of SBMC is its extensibility: stochastic gradient MCMC methods 
\cite{welling2011bayesian,chen2014stochastic} 
and/or other data-parallel techniques 
\cite{angelino2016patterns,maclaurin2014firefly,rendell2020global}
and more sophisticated adaptive methods \cite{hoffman2021adaptive}
can be swapped in later for additional gains.




The contributions of the present work are concisely summarized as follows:
\begin{itemize}
\item New SBMC method (e.g. S-SMC$_\parallel$ and S-MCMC$_\parallel$) which allows the practitioner to 
{\em interpolate} between the MAP (or another point) estimator for $s=0$ (0 additional simulation time) 
and the full posterior for $s=1$ (long simulation time), thus balancing their UQ demands against their budget.
\item A thorough systematic empirical evaluation of 
SBMC on several benchmarks demonstrates that it 
achieves excellent performance on both accuracy and UQ 
{\em at a cost comparable to DE, where traditional BMC methods fail severely}, 
demonstrating its strong scalability and robustness. 
See the top left panel of Figure \ref{fig:sbmc_cartoon}. 
\item This benefit is illustrated on the downstream task of 
estimating prediction confidence, which can be used 
to improve safety and reliability. 
To that end, a meta-classifier is built using seven features of the SBMC posterior.
\end{itemize}

The paper is organized as follows. 
In Section~\ref{sec:map-sbmc}, 
we introduce the SBMC method.
In Section~\ref{sec:uq} we discuss 
its UQ abilities and the downstream 
task of output confidence prediction, as motivation,
and present the main results.
Section~\ref{sec:related} discusses related literature.
Section~\ref{sec:conclusion} presents the conclusion and additional discussion.

\section{Scalable Bayesian Monte Carlo (SBMC) method}
 \label{sec:map-sbmc}

We define {\em time cost} as the required {\em simulation time per chain/particle},
and we will measure this by {\em epochs}, i.e. likelihood plus gradient evaluations,
as a {\em hardware-agnostic proxy} for wall-clock time.
Parallel implementations of consistent BMC algorithms like SMC$_\parallel$ and HMC$_\parallel$
improve time cost
\begin{wrapfigure}[14]{r}{0.5\textwidth}
\vspace{-10pt}
\begin{minipage}{0.5\textwidth}
\begin{algorithm}[H]
    \caption{
    SBMC method}
    \label{alg:sbmc}
\begin{algorithmic}
    \STATE {\bf Inputs}: \(\cL, \pi_0, s,N,P\). 
\vspace{4pt}
    \STATE {\bf Compute} $\theta_{\sf MAP}$, and create $\overline\pi_0,\overline\pi$ as in \eqref{eq:sbmc0},\eqref{eq:sbmc}.
\vspace{4pt}
\FOR{\(p=1\) {\bfseries to} \(P\) (in parallel)} 
\vspace{4pt}
    \STATE {\bf Run} Algorithm~\ref{alg:smc} (S-SMC) or \ref{alg:mcmc} (S-MCMC).
\vspace{4pt}
\STATE {\bf Output}: \(\{\theta^{i,p}\}_{i=1}^N\) and \(Z^{N,p}\).
    \vspace{4pt}
    \ENDFOR
    \vspace{4pt}
\STATE Build $\hat{\varphi}_{\sf SBMC} =\frac{\sum_{p=1}^P Z^{N,p} \frac1N \sum_{n=1}^N \varphi(\theta^{i,p})}{\sum_{p=1}^P Z^{N,p}} \, .$
\end{algorithmic}
\end{algorithm}
\end{minipage}
\end{wrapfigure}
with near linear speed-up \citep{ours}, but each process still needs to run for a long time, 
as seen in Figure \ref{fig:sbmc_cartoon} (c).
This section introduces the model and algorithm choices that define the SBMC method in Algorithm \ref{alg:sbmc}, which delivers improved performance 
on metrics of interest {\em for a comparable time cost
to deep ensembles}.

\textbf{The model. } 
Assume the prior is $\pi_0=\mathcal{N}(0,V)$ for simplicity and
define the MAP estimator as 
$$\theta_{\sf MAP} = {\sf argmax}_\theta \cL(\theta;\cD)\pi_0(\theta) \, ,$$
where $\cL$ is the likelihood defined in equation \eqref{eq:posterior}.
For a fixed tuning parameter $s \in (0,1)$, we define $0\prec \Sigma(s)=\Sigma(s)^{\sf T}\in \bbR^{d\times d}$
and $\alpha(s) \in [0,1]$, such that $\Sigma(0)=0$ and $\Sigma(1)=V$, and $\alpha(0)=1$ and $\alpha(1)=0$.
Define the new prior as
\begin{equation}\label{eq:sbmc0}
\overline{\pi}_0(\theta) =
\mathcal{N} (\theta ; \alpha(s) \theta_{\sf MAP}, \Sigma(s)) \, .
\end{equation}
The SBMC method then targets the following distribution
\begin{equation}\label{eq:sbmc}
    \overline{\pi}(\theta) \propto \cL(\theta) \overline{\pi}_0(\theta) \, ,
\end{equation}
which we will refer to as the {\em anchored} posterior.
We will refer to $\theta_{\sf MAP}$ as the {\em anchor}.

For $s \rightarrow 0$, we recover a Dirac measure concentrated on 
the MAP estimator, 
which means no sampling is required.
Conversely, as $s \rightarrow 1$, we recover the {\em original posterior}. 
Hence $s$ is a scalar interpolation parameter which allows us to tune between these limits.
For simplicity we will typically consider only the standard isotropic case $V=v {\sf Id}$ and
let $\alpha(s)={\bf 1}_{\{s<\frac12\}}$ and $\Sigma = sv {\sf Id}$.

We show that this approximate model
balances the complementary strengths of the two approaches for small $s$,  
and enables BMC methods to deliver 
{\em scalable gains over alternatives like deep ensembles at a comparable cost.} 
The method is relatively insensitive to the exact value of $s$ and we recommend a default value of $s=0.1$.
We will use the notation 
S-SMC$_\parallel$ and S-MCMC$_\parallel$ to distinguish the SBMC method from standard implementations of the algorithms 
targeting \eqref{eq:posterior}.
For example, S-SMC$_\parallel$ means the SMC$_\parallel$ algorithm 
is used to sample from \eqref{eq:sbmc}.

\begin{wrapfigure}[18]{l}{0.5\textwidth}
  \vspace{-15pt}
  \begin{minipage}{\linewidth}
  \begin{algorithm}[H]
    \caption{S-SMC sampler}
    \label{alg:smc}
    \begin{algorithmic}
       \STATE {\bf Inputs}: \(\cL, \overline \pi_0,N\).
       \vspace{4pt}
       \STATE 
       Init. \(\theta^i_0 \sim \overline \pi_0\) for \(i=1,\dots,N\). \(Z^N=1\). 
       \FOR{\(j=1\) {\bfseries to} \(J\) (in serial)}
         \STATE (Optional) Select 
         $\lambda_j$ s.t. ESS$=\rho N, \rho < 1$.
         \STATE Store \(Z^N *= 
         \frac{1}{N} \sum_{k=1}^N \cL(\theta_{j-1}^k)^{\lambda_j-\lambda_{j-1}}\).
         \vspace{3pt}
         \FOR{\(i=1\) {\bfseries to} \(N\) (in parallel)}
            \vspace{2pt}
            \STATE Define \(w^i_j \propto \cL(\theta_{j-1}^i)^{\lambda_j-\lambda_{j-1}}\).
           \STATE {\bf Selection}: 
           \(I_j^i \sim \{w_j^1,\dots,w_j^N\}\).
           \STATE {\bf Mutation}: 
           \(\theta_j^i \sim \cM_j(\theta_{j-1}^{I_j^i},
           \cdot )\).
         \ENDFOR
       \ENDFOR
       \vspace{4pt}
    \STATE {\bf Outputs}: \(\{\theta^{i}=\theta^{i}_J\}_{i=1}^N\) and \(Z^{N}\).
    \end{algorithmic}
  \end{algorithm}
\end{minipage}
\end{wrapfigure}
\textbf{The algorithm} can be {\em any BMC method}.
In the present work we will focus on SMC sampler and MCMC,
but any alternative is admissible. 
For example, SG-MCMC or other methods which allow mini-batch gradients
may be quite convenient for managing the memory requirements of very large problems.

\noindent{\bf SMC sampler. } Define a sequence of intermediate targets 
    $\overline \pi_j(\theta) \propto \cL(\theta)^{\lambda_j}\overline \pi_0(\theta)\, ,$
according to a tempering schedule $0=\lambda_0,\dots,\lambda_J=1$,
which will be chosen adaptively according to the effective sample size (ESS), 
as described in 
\ref{ex:adaptive_temp} in the Appendix.
The SMC sampler \citep{del2004feynman} 
alternates between {\em selection} by importance re-sampling,
and {\em mutation} according to an appropriate intermediate MCMC transition kernel \(\cM_{j}\),
such that \((\overline \pi_{j} \cM_{j})(d\theta) = \overline \pi_{j}(d\theta)\) \citep{geyer1992practical}.
This operation must sufficiently de-correlate the samples, 
and as such we typically define the MCMC kernels \({\cM_{j}}\) by several 
steps of some basic MCMC kernel, leading to 
$L_j$ epochs (likelihood/gradient evaluations).
We will employ two standard MCMC kernels: preconditioned Crank-Nicolson (pCN) 
\citep{bernardo1998regression,cotter2013mcmc}
and Hamiltonian Monte Carlo (HMC) \citep{duane1987hybrid, neal2011mcmc}.
In the latter case, there are also 
several leapfrog steps for each HMC step contributing to $L_j$.

For a quantity of interest \(\varphi: \sTheta \rightarrow \bbR\), 
the S-SMC estimator from Algorithm 
\ref{alg:smc} is given by 
\begin{equation}\label{eq:smc}
\overline \pi^N(\varphi)  :=  \frac{1}{N} \sum_{i=1}^{N} \varphi(\theta^{i})
\stackrel{N\to\infty}{\longrightarrow}
\bbE_{\overline \pi}[\varphi] \approx \bbE_{\pi}[\varphi] = \bbE[\varphi \mid \cD ] \, .
\end{equation}

\begin{wrapfigure}[13]{r}{0.4\textwidth}
  \vspace{-15pt}
  \begin{minipage}{\linewidth}
  \begin{algorithm}[H]
   \caption{
   S-MCMC}
   \label{alg:mcmc}
\begin{algorithmic}
       \STATE {\bf Inputs}: \(\cL, \overline \pi_0,N\).
       \vspace{4pt}
       \STATE 
   \(\theta^i_0 \sim \overline \pi_0\) for \(i=1,...,N\). 
    \FOR{\(i=1\) {\bfseries to} \(N\) (in parallel)} 
    \FOR{\(j=1\) {\bfseries to} \(J\) (in serial)}
    \STATE Draw \(\theta_{j}^i \sim \cM_J(\theta_{j-1}^i, \cdot )\).
    \ENDFOR
    \ENDFOR
       \vspace{4pt}
        \STATE {\bf Outputs}: \(\{\theta^{i}_J\}_{i=1}^N\) and \(Z^{N}\equiv 1\).
       \end{algorithmic}
\end{algorithm}
\end{minipage}
\end{wrapfigure}
\noindent {\bf S-SMC$_\parallel$ } refers to $P$ parallel executions of
Algorithm \ref{alg:smc}, each with \(N\) particles, 
leading to a \(P\) times lower 
communication and memory overhead than a single S-SMC sampler with \(NP\) samples.
This simplification 
is crucial for massive problems such as BDL, which require distributed architectures.
Synchronous Single Instruction, Multiple Data (SIMD) resources can be used for the $N$ communicating particles
(and model- and data-parallel likelihood calculations),
while all communication between the $P$ processes is eliminated.
The S-SMC$_\parallel$ ratio estimator is given as follows, 
for $p$ i.i.d. realizations $\overline \pi^{N,p}(\varphi)$ of \eqref{eq:smc}
    \begin{equation}\label{eq:ssmc}
        \hat{\varphi}_{\text{S-SMC$_\parallel$}}  
    = \sum_{p=1}^{P} \omega_p \overline \pi^{N,p}(\varphi) \, , \quad 
    \omega_p = \frac{Z^{N,p}}{\sum_{p'=1}^P Z^{N,p'}} \, .
    \end{equation}

{\bf S-MCMC$_\parallel$ }
refers to $P$ parallel executions of
Algorithm \ref{alg:mcmc}, 
which already features $N$ parallel short chains free from any communication.
The purpose of formulating MCMC in this way is to match SMC, which  
itself features $N$ parallel chains that need to communicate intermittently
at the selection stage. The estimator is built exactly as \eqref{eq:ssmc}, 
with \(\{\theta^{i,p}=\theta^{i,p}_J\}_{i=1}^N\) in \eqref{eq:smc} and $Z^{N,p}=1$.

\section{Motivation and Results}
\label{sec:uq}

\textbf{UQ} is a crucial pain-point for neural networks, 
and BDL is one of the leading contenders to deliver it.
Our primary UQ metric will be {\em epistemic entropy, } which is the difference 
between {\em total} and aleatoric entropy, defined as follows 
\citep{hullermeier2021aleatoric,depeweg2018decomposition,shaker2020aleatoric,krause2025probabilistic}
\begin{equation}\label{eq:entropy}
\setlength\abovedisplayskip{5pt}%
 \setlength\belowdisplayskip{3pt}%
H_{\sf ep}(x) =
\underbrace{- \sum_{y\in {\sf Y}} \bbE[p(y|x,\theta) | \cD]
\log \bbE[p(y|x,\theta) | \cD]}_{H_{\sf tot}(x)} 
- \underbrace{\bbE
\Big [ -\sum_{y\in {\sf Y}} p(y|x,\theta )\log( p(y|x,\theta) ) 
| \cD \Big ] }_{H_{\sf al}(x)} \, .
\end{equation}
Aleatoric uncertainty is irreducible and can be thought of as label error 
(people may sometimes disagree on the label of a given hand-written digit), 
whereas epistemic entropy quantifies uncertainty which can be reduced with more data.  
Our focus is the latter, as it is 
{\em only} captured by Bayesian methods.
It can be viewed as the mutual information between parameter and predictive posterior random variables for input $x$,
and as such is $0$ by definition for point estimators that yield deterministic predictive estimators.


\begin{wrapfigure}[14]{l}{0.6\textwidth}
    \vspace{-5pt}
    \centering
    \includegraphics[width=0.6\columnwidth]{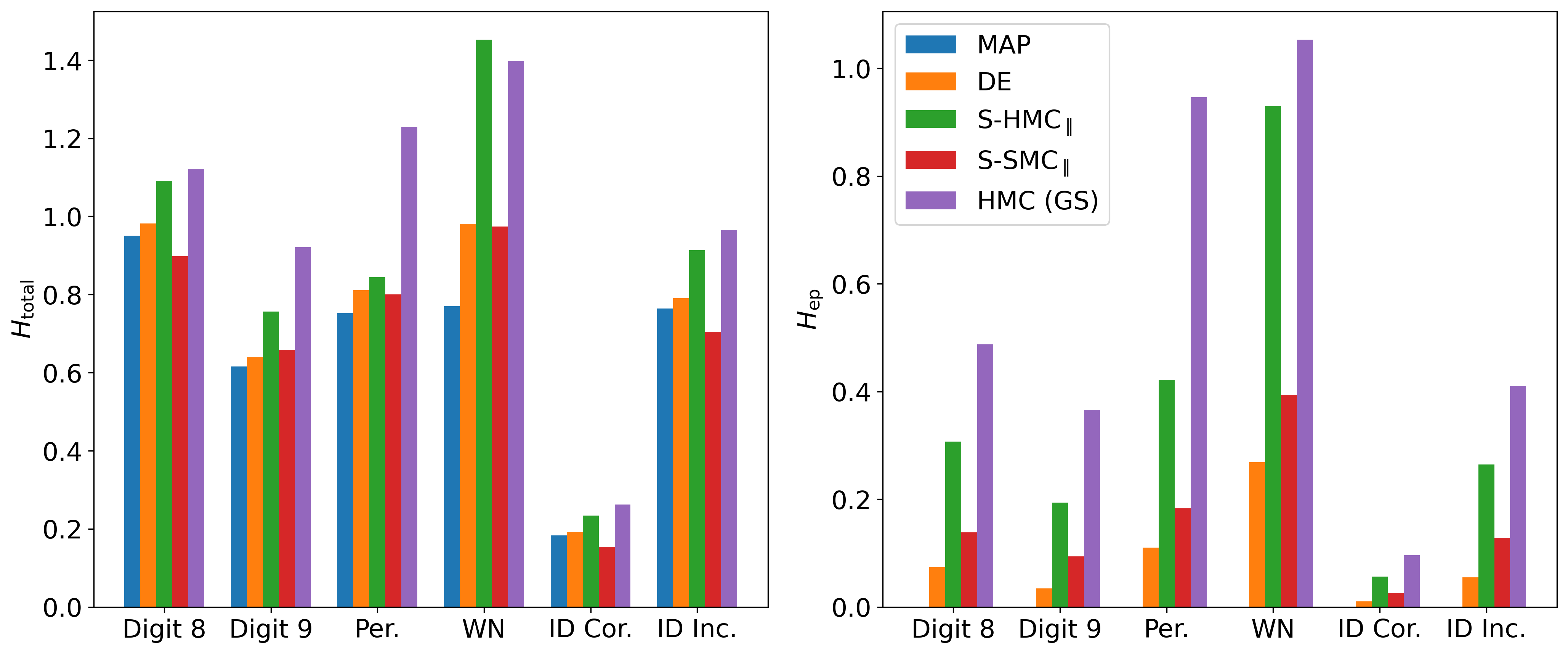}
    \caption{Average total and epistemic entropy over four OOD
    classes and correct and incorrect predictions ID
    for MNIST7 ($P=1$).}
\label{fig:entropyAll}
\end{wrapfigure} 
To illustrate the properties of SBMC, we conduct an experiment. 
The dataset is a subset of 1200 MNIST \citep{lecun2010mnist} data trained on digits $0,\dots,7$ (MNIST7)
with $8,9$ held out as (similar but) out-of-domain (OOD).
White noise inputs and randomly selected in-domain (ID) digits corrupted
with white noise are considered as 
far OOD classes. 
The architecture is 
described in Appendix \ref{app:mnist}. The prior variance
is $v=0.1$.
The average total and epistemic entropy for various categories of OOD
test data are presented in Figure \ref{fig:entropyAll} 
{(per-digit result is given in Appendix \ref{app:mnist_uq})}.
The average epistemic entropy over the test data, split by correct and incorrect predictions, 
are presented in Figure \ref{fig:entropyAll}. 
This quantity is clearly predictive of misclassifications, 
and this downstream task will be revisited below.

\begin{table}[H]
\scriptsize
\caption{Comparison of methods 
on MNIST7 test data. SBMC methods are bold. For metrics, the best gold-standard (GS) value is bold, 
along with others within 
1\% for accuracy and NLL, and 50\% for entropies (entropy is harder to estimate, and also high precision is less critical).
SMC$_\parallel$, HMC$_\parallel$, and SGHMC$_\parallel$ 
are highlighted in \textcolor{red}{red} as they are particularly bad 
for these very short chains, and this is precisely the problem the approximate methods like SBMC address.}
\label{tab:map-sbmc}
\centering
\begin{tabular}{l|c|c|c|c|c|c|c|c}
\toprule
&  & {\bf Time Cost} & {Total Cost} &  & & H$_{\mathsf {ep}}$ & H$_{\mathsf {ep}}$ &  \\
 Method & P & (epochs) $\downarrow$ & (epochs) $\downarrow$ & Accuracy $\uparrow$ & NLL $\downarrow$ & correct & incorrect & H$_{\mathsf {ep}}$ OOD \\
\midrule
\blue{MAP} & 1 & \blue{160} & 160 & 92.3$\pm$0.366 & 0.253$\pm$0.012 & 0 & 0 & 0 \\

SWA & 1
& 160 & 160 & 92.3$\pm$0.387 & 0.27$\pm$0.017 & 0 & 0 & 0 \\

MC Drop & 1
& 160 & 160 & {\bf 93.9$\pm$0.626}  & {\bf 0.214$\pm$0.021} & 0.049$\pm$0.007 & {\bf 0.269$\pm$0.008}  & 0.267$\pm$0.01  \\

Laplace & 1
& \blue{160} & {160} & 88.2$\pm$0.235 & 0.539$\pm$0.022 & 0.504$\pm$0.036 & 0.901$\pm$0.038 & { 1.22$\pm$0.033} \\

Deep Ens & 1 & 176 & 1760 & 92.4$\pm$0.150& 0.245$\pm$0.004 & 0.011$\pm$0.000 & 0.057$\pm$0.001 & 0.123$\pm$0.011 \\

Deep Ens & 8 & 178 & 14,240 & 92.5$\pm$0.059 & 0.239$\pm$0.001 & 0.011$\pm$0.000 & 0.059$\pm$0.001 & 0.134$\pm$0.004 \\

\textcolor{red}{SGHMC$_\parallel$} & 1 & 160 & 1600 & \textcolor{red}{87.7$\pm$0.742} & \textcolor{red}{0.974$\pm$0.05} &0.652$\pm$0.031 & { 0.725$\pm$0.031} & {\bf 0.883$\pm$0.036} \\

{\bf \blue{S-}SGHMC$_\parallel$} & 1 & 160 \blue{+ 160} & 1760 & 90.3$\pm$0.758 & 0.409$\pm$0.014 & 0.342$\pm$0.015 & { 0.687$\pm$0.027} & {\bf 0.836$\pm$0.039} \\

{\bf \blue{S-}SGHMC$_\parallel$} & 8 & 160 \blue{+ 160} & 12,960 & 92.3$\pm$0.160 & 0.388$\pm$0.001 & 0.434$\pm$0.005 & { 0.782$\pm$0.007} & {\bf 0.965$\pm$0.003} \\

\textcolor{red}{SMC$_\parallel$} & 1 & 173 & 1730 & \textcolor{red}{79.7$\pm$2.71} & \textcolor{red}{0.623$\pm$0.091} & 0.013$\pm$0.002 & 0.033$\pm$0.009 & 0.045$\pm$0.011 \\

{\bf \blue{S-}SMC$_\parallel$} & 1 & 170 \blue{+ 160} & 1860 & 92.2$\pm$0.371 & 0.267$\pm$0.014 & 0.026$\pm$0.003 & 0.129$\pm$0.014 & 0.202$\pm$0.028 \\

{\bf \blue{S-}SMC$_\parallel$} & 8 & 178 \blue{+ 160} & 14,400 & {\bf 93.3$\pm$0.160} & {\bf 0.226$\pm$0.004} & {\bf 0.059$\pm$0.001} & {\bf 0.272$\pm$0.003} & {\bf 0.378$\pm$0.03} \\

\textcolor{red}{HMC$_\parallel$} & 1 & 160 & 1600 & \textcolor{red}{78.4$\pm$2.38} & \textcolor{red}{1.27$\pm$0.085} & 0.303$\pm$0.025 & {\bf 0.325$\pm$0.026} & {\bf 0.594$\pm$0.021} \\

{\bf \blue{S-}HMC$_\parallel$} & 1 & 160 \blue{+ 160} & 1760 & {\bf 93.0$\pm$0.166} & {\bf 0.232$\pm$0.002} & {\bf 0.056$\pm$0.001} & {\bf 0.264$\pm$0.002} & {\bf 0.463$\pm$0.009} \\

{\bf \blue{S-}HMC$_\parallel$} & 8 & 160 \blue{+ 160} & 12,960 & {\bf 93.1$\pm$0.085} & {\bf 0.231$\pm$0.002} & {\bf 0.070$\pm$0.000} & {\bf 0.299$\pm$0.002} & {\bf 0.531$\pm$0.011}  \\
\midrule

HMC (GS) & 1 & \textcolor{red}{20,000} & \textcolor{red}{20,000} & {\bf 93.6$\pm$0.415} & {\bf 0.222$\pm$0.009} & {\bf 0.096$\pm$0.004} & {\bf 0.410$\pm$0.013} & {\bf 0.768$\pm$0.084} \\
HMC (GS) & 1 & \textcolor{red}{200,000} & \textcolor{red}{200,000} &  {\bf 94.8\(\pm\)0.211} & {\bf 1.94$\pm$0.004} & {\bf 0.120$\pm$0.004} & {\bf 0.493$\pm$0.008} & {\bf 1.04$\pm$0.122} \\
\bottomrule
\end{tabular}
\end{table}

In Table \ref{tab:map-sbmc} we compare several SOTA competitors, 
including the \blue{MAP}
(computed with SGD and early stopping on validation data), DE, MC Dropout,
Laplace approximation, and SWA,
with \blue{(S-)}HMC$_\parallel$
\blue{(S-)}SMC$_\parallel$, and \blue{(S-)}SGHMC$_\parallel$
all with {\em approximately equal time cost of $\approx 170$ epochs}, 
according to SGD early stopping.
These can be further parallelized with model- and data-parallel techniques, but we do not consider that here.
Note that SBMC methods {\em require the \blue{MAP} estimator}, 
so their total time cost is roughly \blue{double}.
A single HMC run using $2e4-2e5$ epochs is also included as a GS baseline.
Convergence for all methods is verified by running $5$ chains with dispersed initial conditions and measuring the 
standard error.
Ensemble methods use $P$ independent ensembles of $N=10$ particles, 
and all particles are used for estimating posterior expectations.

\begin{figure}
    \centering
    \includegraphics[width=0.49\textwidth]{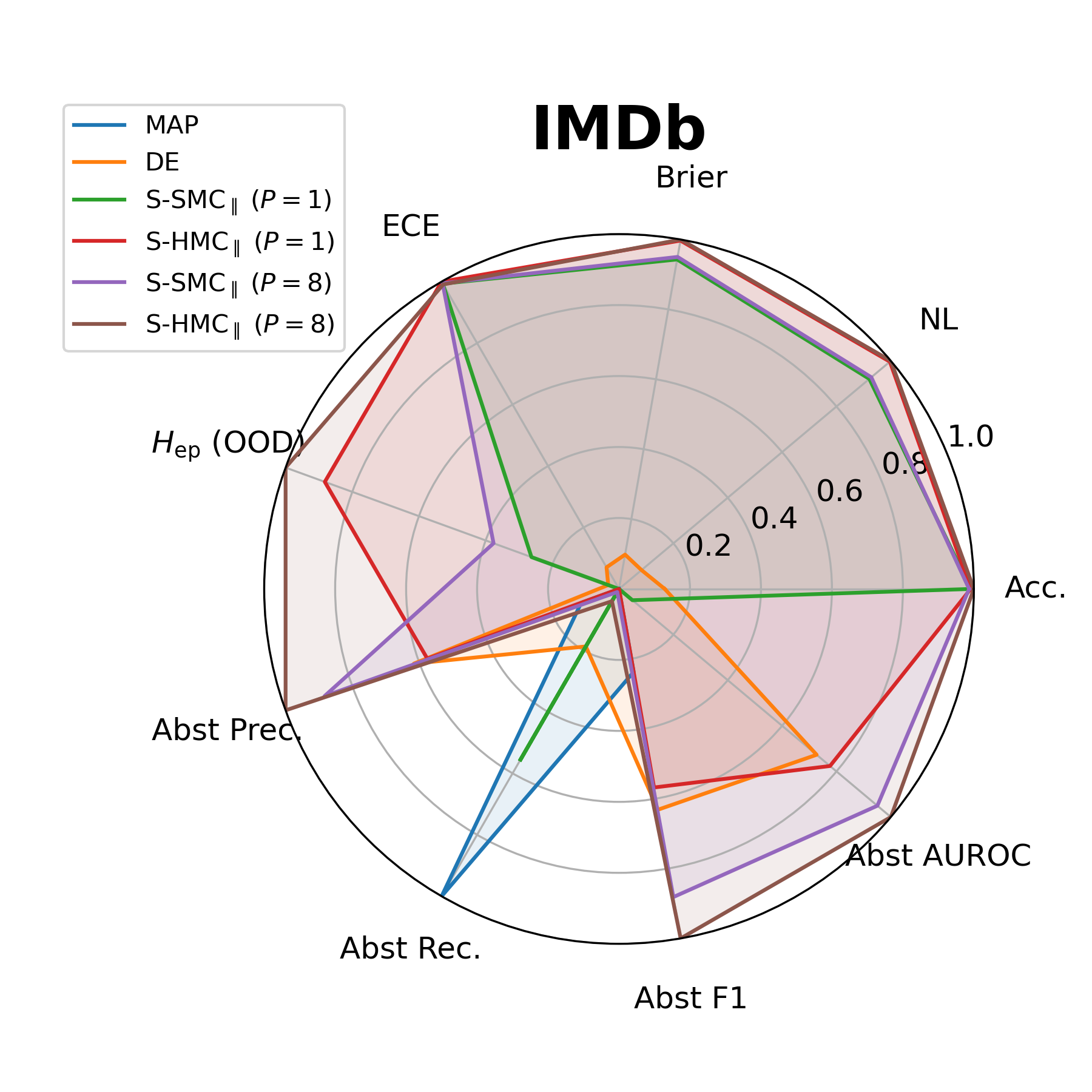}
    \includegraphics[width=0.445\textwidth]{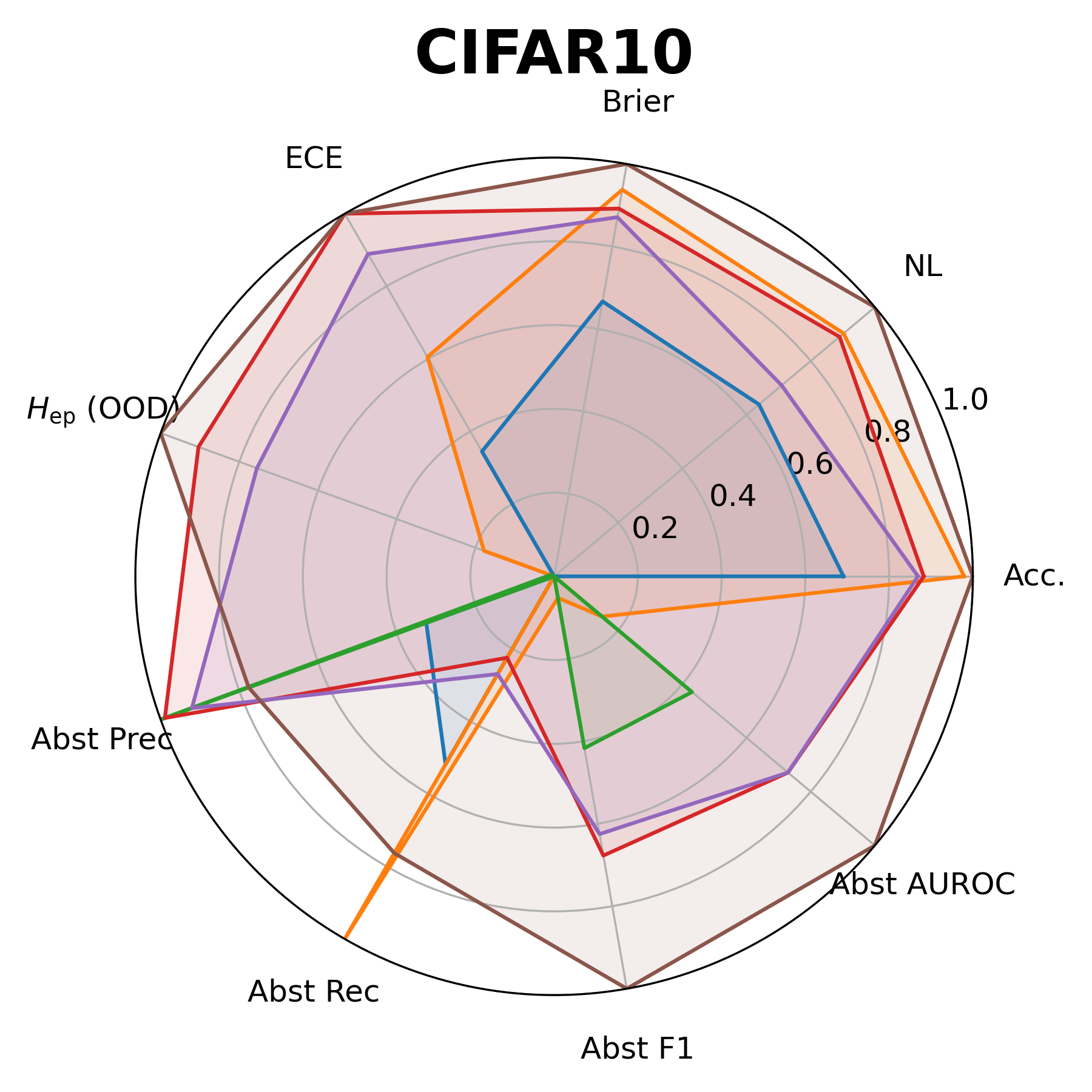}
  \caption{
    Accuracy, UQ, and confidence meta-classifier abstention (Abst) metrics (re-normalized so $1$ is best) 
    for IMDb (left) and CIFAR10 (right).}
    \label{fig:spiders}
\vspace{-20pt}
\end{figure}
The results show that when directly targeting \eqref{eq:posterior}, 
SMC$_\parallel$, HMC$_\parallel$, and SGHMC$_\parallel$, degrade rapidly away from convergence, 
to the point of \textcolor{red}{catastrophic failure} in first order metrics at this cost level.
However, their UQ performance is still adequate--for example, 
some of the HMC$_\parallel$ and SGHMC$_\parallel$ $H_{\sf ep}$ estimators 
are within our tolerance of $50\%$ of the GS solution (in bold).
The MAP and DE quickly deliver good accuracy, but 
do not accurately estimate $H_{\sf ep}$.
These failure modes are perfectly {\em complementary}. 
To achieve the ``best of both worlds'', 
{SBMC anchors to the MAP} estimator to preserve accuracy, 
and then uses an ensemble of short parallel runs of BMC to augment that with uncertainty.
MC Dropout performs particularly well and is notably the only non-BMC method which achieves
an $H_{\sf ep}$ estimator within our tolerance of $50\%$ of the GS solution (in bold), 
and for a low total cost.

For the next experiments, we look at the 
IMDb sentiment classification dataset
\citep{maas2011}
and the CIFAR dataset \citep{krizhevsky2009learning}.
Results for these cases 
are comparable and summarized in Figure \ref{fig:spiders} and  
Appendix \ref{app:uq}, along with further figures and tables.
Further details on all model architectures are given in Appendix \ref{app:BNN}.
The next step is to apply the approach to LLMs--
preliminary simulation results for 
next-token prediction with GPT-2 are given in Appendix \ref{app:gpt2}.


  \begin{wraptable}[13]{l}{0.565\textwidth}
   \vspace{-5pt}
   \centering
  \caption{Confidence meta-classifier results (optimal \(F_1\) decision threshold, MNIST7).}
  \label{tab:ood_optimal}
  \begin{tabular}{ll|c|c|c|c}
    \toprule
    \(P\) & {\footnotesize Method} & {\footnotesize Precision} & {\footnotesize Recall} & F1 & \tiny{AUC-ROC} \\
\midrule
– & MAP & 0.707 & 0.898 & 0.791 & 0.828 \\
– & DE  & 0.734 & 0.897 & 0.807 & 0.855 \\
1 & S-SMC$_\parallel$ & 0.701 & 0.890 & 0.783 & 0.824 \\
8 & S-SMC$_\parallel$ & \textbf{0.753} & \textbf{0.915} & \textbf{0.826} & \textbf{0.884} \\
1 & S-HMC$_\parallel$ & \textbf{0.750} & \textbf{0.906} & \textbf{0.820} & \textbf{0.885} \\
8 & S-HMC$_\parallel$ & \textbf{0.752} & \textbf{0.913} & \textbf{0.825} & \textbf{0.892} \\
\bottomrule
  \end{tabular}
\end{wraptable}
\textbf{Out-of-domain inference.}
One clear application of UQ is inferring confidence in model predictions, 
i.e. whether the output is reliable or ``hallucinated", to borrow the vernacular from  modern LLMs 
\citep{ji2023survey,deepseek_r1_2025}
This information can be used to decide whether the model should abstain from responding 
or provided to the user so they can make their own decision about whether to trust the response.
To that end, we propose to build a {\em confidence meta-classifier} of incorrect/OOD data,
as follows. 
First, we fit our model to 1000 training data and 
{200 validation data for early stop procedure in MAP and DE methods/prior} from MNIST7,
and label incorrect predictions as $z=1$ and correct predictions as $z=0$.
Then, we generate {$2000$} additional OOD meta-training data as described above,
all of which get label $z=1$.
Let
$p_{\sf max}(x,\theta) := \max_y p(y|x,\theta) = p(y^*|x,\theta)$ 
denote the maximum probability, and denote the difference between the top two as
$
\Delta_{\sf max}(x,\theta) := p_{\sf max}(x,\theta) - \max_{y'\in{\sf Y}\backslash y^* } {p}(y|x,\theta) \, .
$
Consider as features: 
$$
p_{\sf max}(x,\theta), H_{\sf total}(x), 
\bbE[p_{\sf max}(x,\theta)|\cD], \bbE[\Delta_{\sf max}(x,\theta)|\cD],
H_{\sf ep}(x), {\sf Var}[p_{\sf max}(x,\theta)|\cD], 
{\sf Var}[\Delta_{\sf max}(x,\theta)|\cD] \, .
$$
\begin{wrapfigure}[18]{r}{0.55\textwidth}
    \vspace{-10pt}
    \centering
    \includegraphics[width=0.55\textwidth]{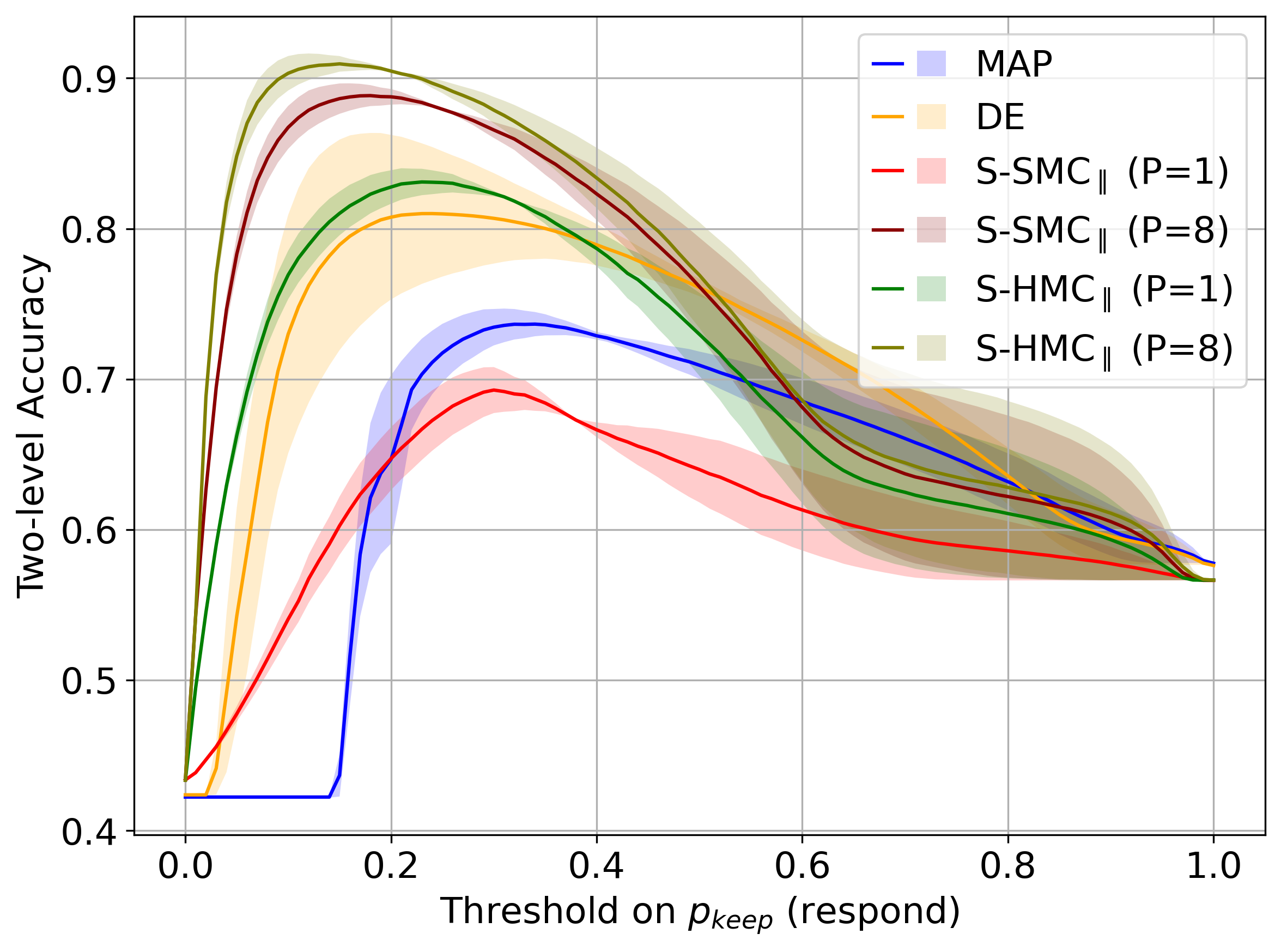}
  \caption{
    2-level estimator (using confidence meta-classifier for abstention) accuracy on IMDb.}
    \label{fig:roc-ood}
\end{wrapfigure} 
Note that the last 3 are identically 0 for the MAP estimator, 
as they capture the epistemic uncertainty in the data.
We build and standardize these features for each of our 4 models--MAP, DE, 
S-SMC$_\parallel$, S-HMC$_\parallel$--
and train the binary meta-classifier $x \mapsto z$, using 
a single hidden layer MLP with $50$ neurons.

The results on {2000} ID test data plus {2000} newly-generated OOD test data
are presented in Table \ref{tab:ood_optimal}. 
Accuracy is the ratio of true positives (not correct) and true negatives (correct) 
to the total testing dataset size. 
All estimators do surprisingly well, and our SBMC methods are the best.
AUC-ROC is perhaps the most useful metric, as it measures the ability of
the score $p_{\sf incorrect}(x)$ to rank in-correctness of the subsequent inference:
the probability that a randomly selected incorrect example from the test set,
$x_{\sf incorrect} \in \cD_{\sf incorrect}$
will have a higher score than a randomly selected correct example,
$x_{\sf correct} \in \cD_{\sf correct}$: 
$\mathbb{P}[ p_{\sf incorrect}(x_{\sf incorrect}) > p_{\sf incorrect}(x_{\sf correct}) ]$.

Figure \ref{fig:roc-ood} 
shows the accuracy of a 2-level
estimator over thresholds for IMDb. 
See Appendix {\ref{app:ood}} for further details.
First we infer with the meta-classifier whether the model inference will be correct,
under the assumption that 
OOD inputs implies incorrect inference.
If yes, we infer with the original model.
If no, we {\em abstain} from inference 
(abstentions are counted as correct decisions when the original model would be incorrect).
Figure \ref{fig:spiders} presents summary metrics for IMDb and CIFAR10.

\begin{wrapfigure}[24]{l}{0.55\textwidth}
\vspace{-5pt}
    \centering
    \includegraphics[width=0.27\columnwidth]{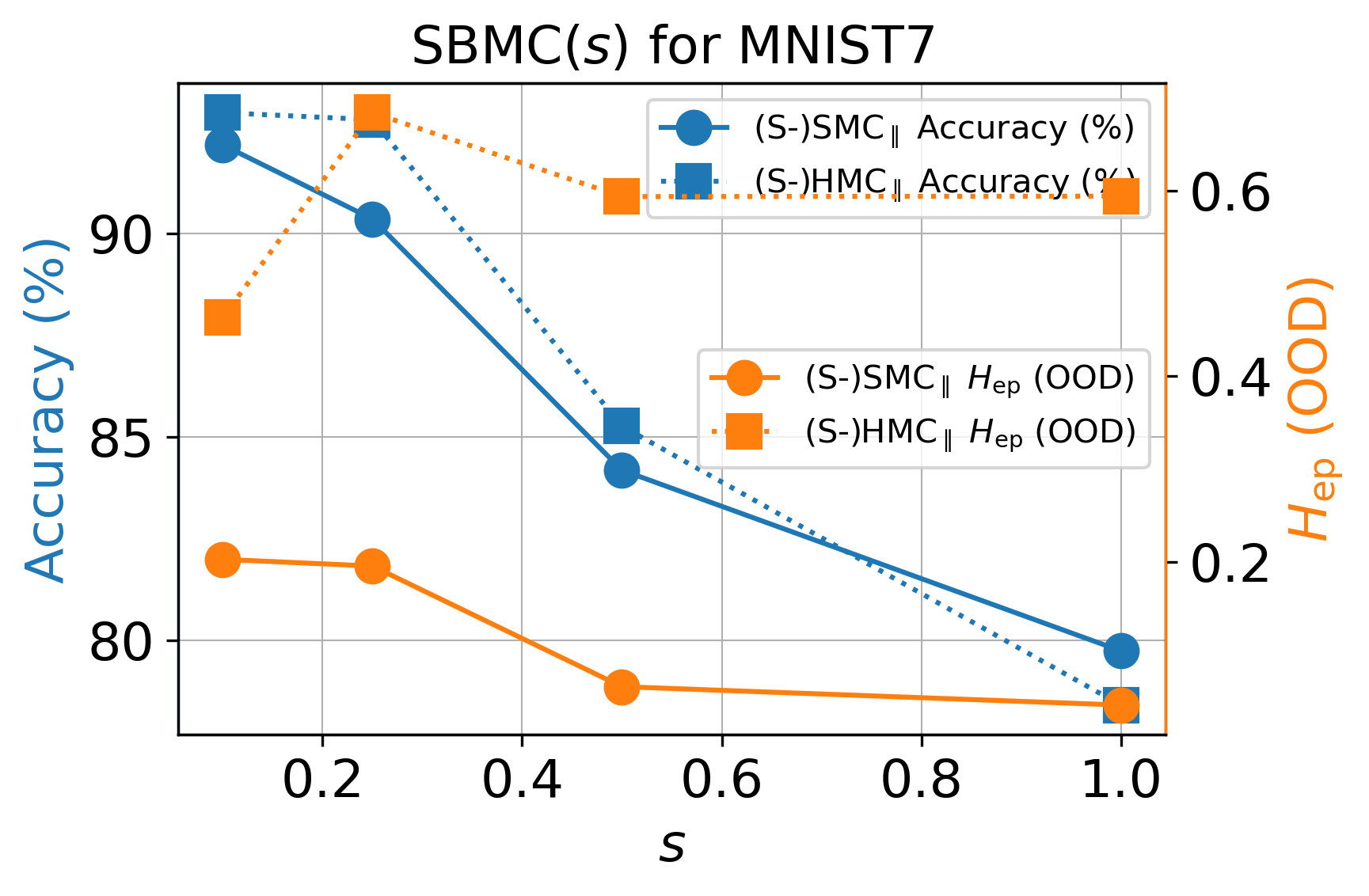}
    \includegraphics[width=0.27\columnwidth]{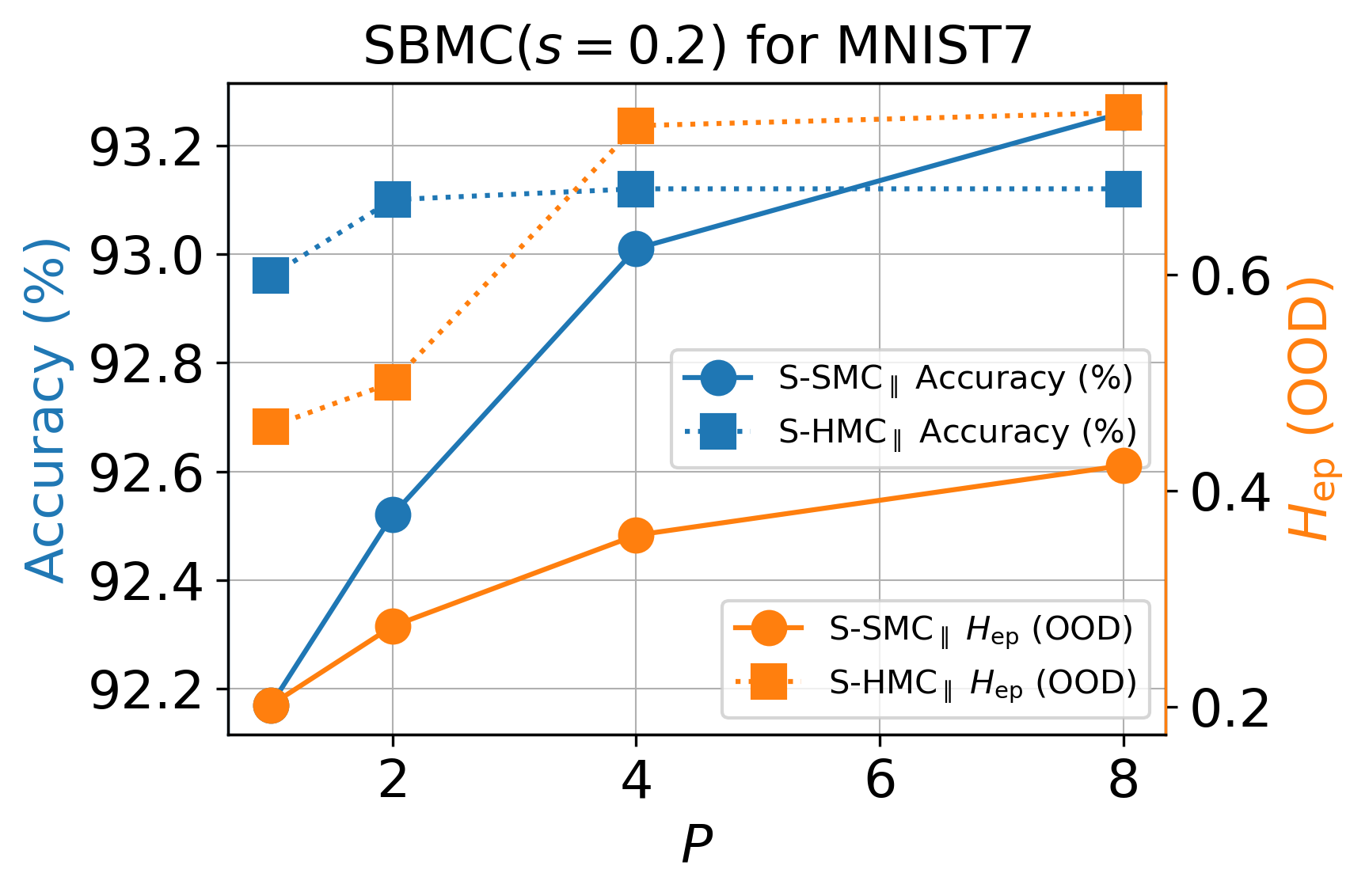} \\
    \includegraphics[width=0.27\columnwidth]{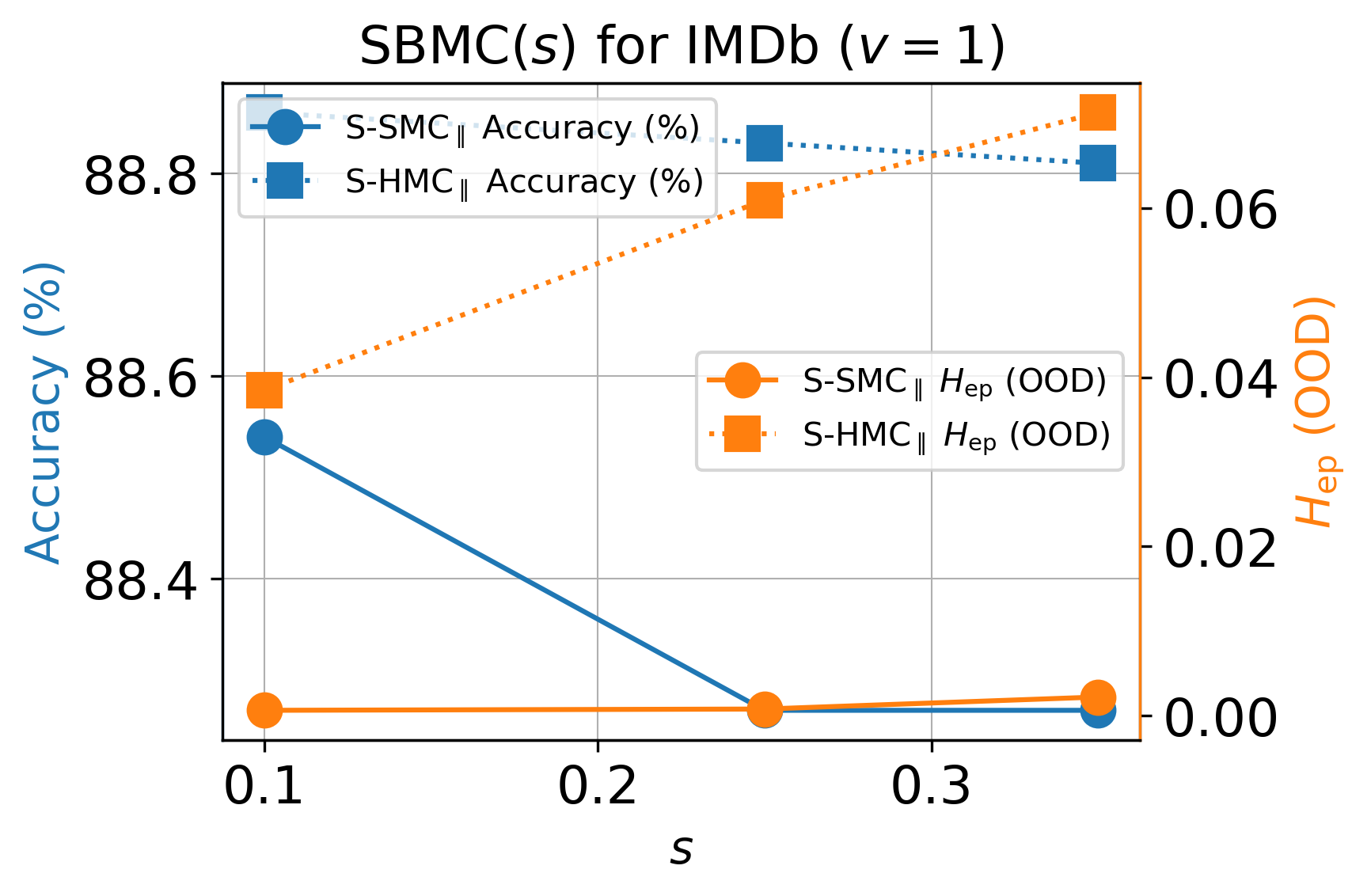}
    \includegraphics[width=0.27\columnwidth]{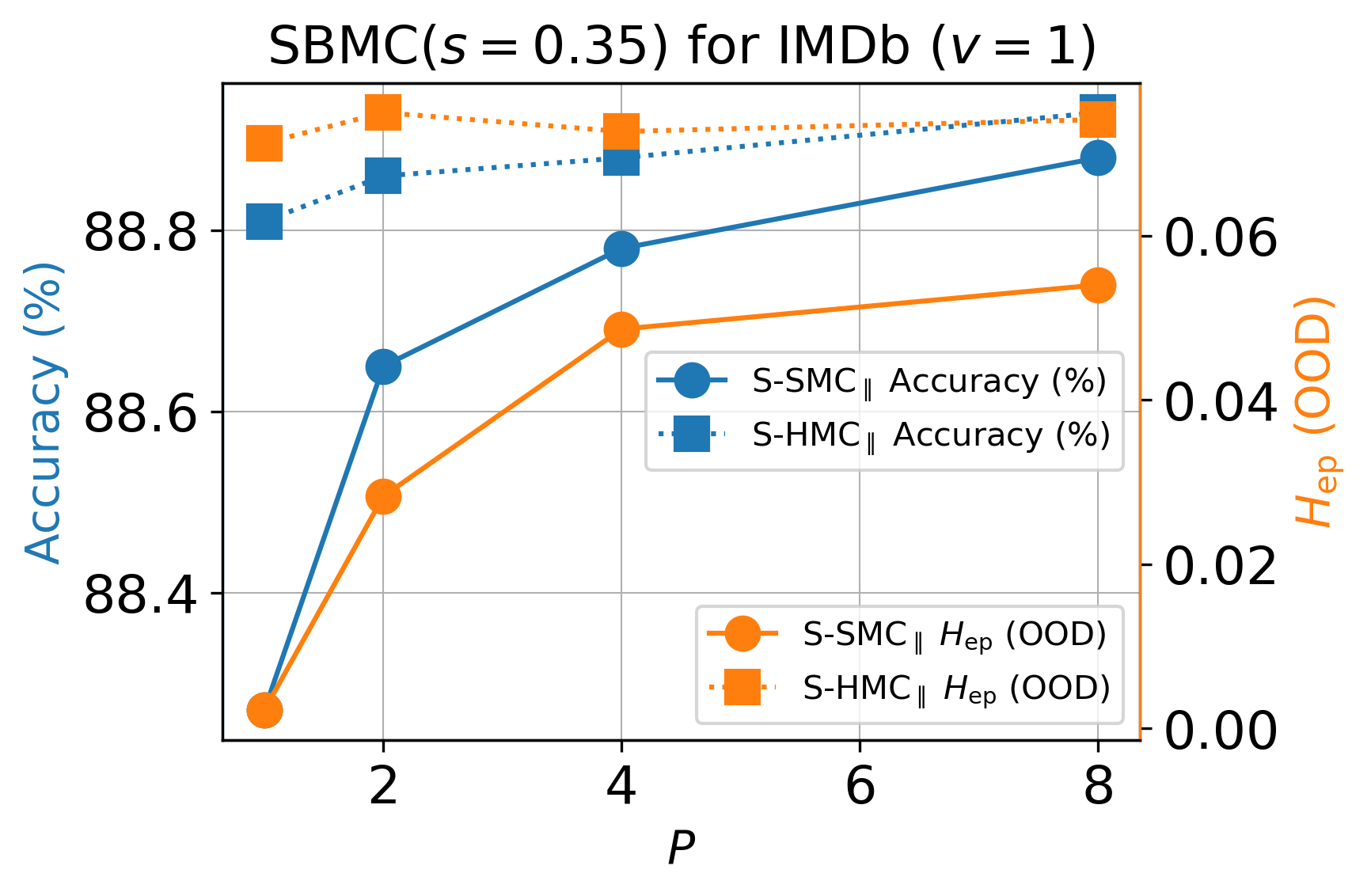} \\    
    \includegraphics[width=0.27\columnwidth]{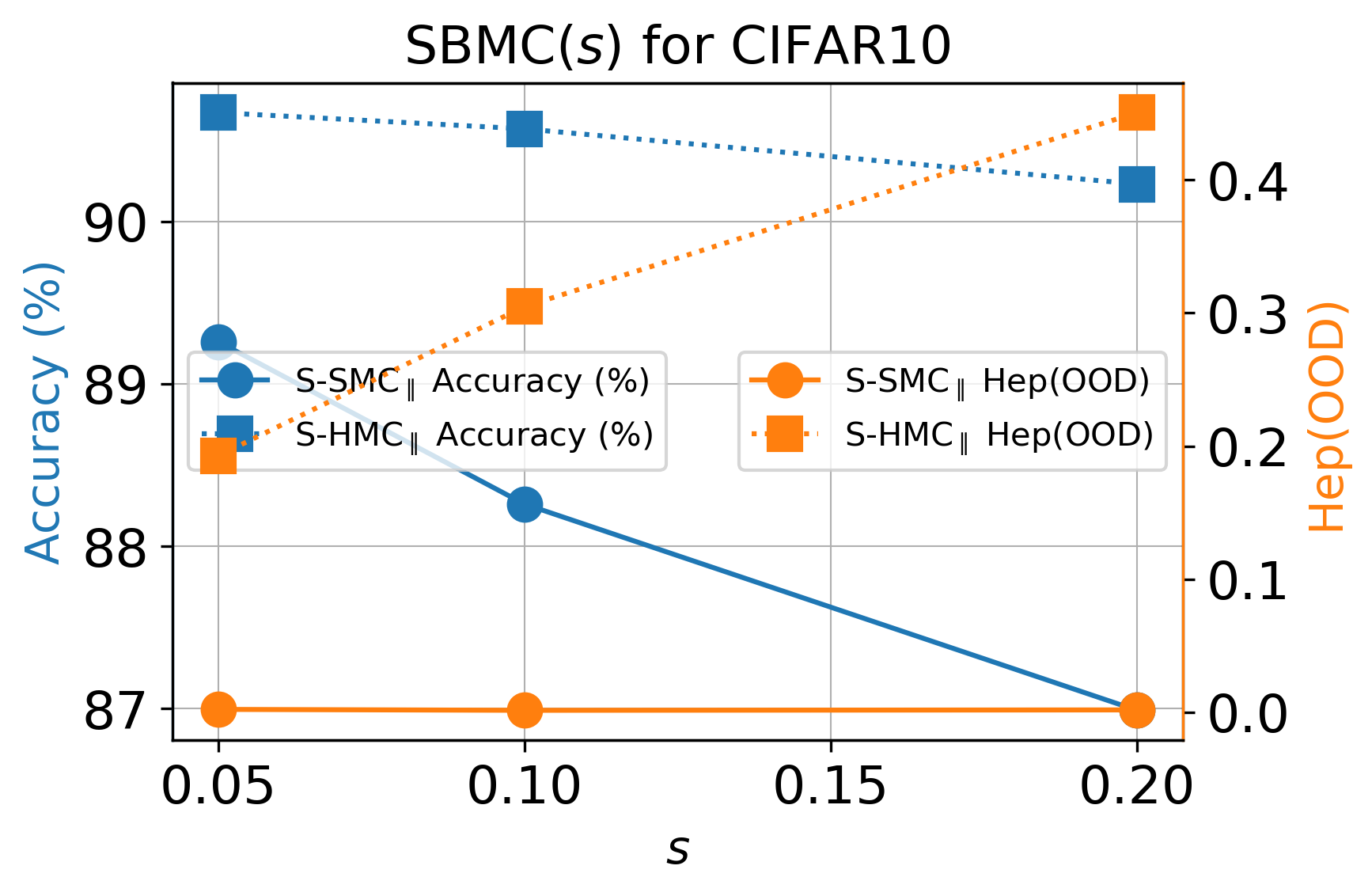}
    \includegraphics[width=0.27\columnwidth]{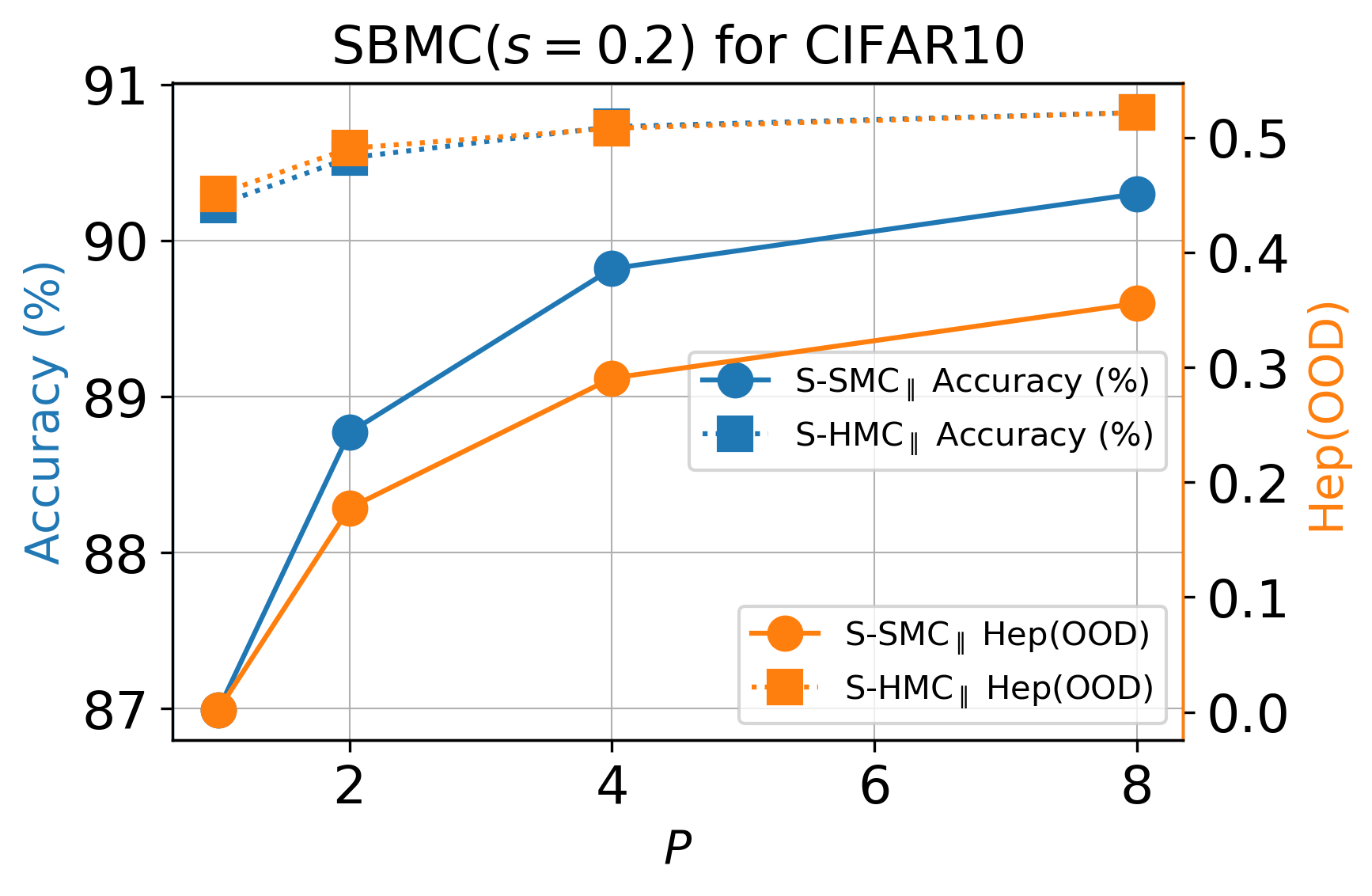} \\ 
    \caption{Dual axis Accuracy and H$_{\sf ep}$(OOD) ablations 
    over $s$ in column 1 and $P$ in column 2 for 
    MNIST7 (row 1), IMDb (row 2), and CIFAR10 (row 3).}
  \label{fig:ablations}
\end{wrapfigure}
\noindent 
\textbf{Ablations} are considered by varying $s$ and $P$.
Small $s$ improves mixing, as shown in Figure \ref{fig:sbmc_cartoon} (d),
but also introduces bias because
$\overline \pi \neq \pi$ (see \eqref{eq:sbmc}, \eqref{eq:posterior}).
In this short-chain setting, smaller $s$ typically increases accuracy and decreases $H_{\sf ep}$(OOD).
Both algorithms improve with $P$, but it is particularly notable that SMC$_\parallel$
improves much more.
{Comprehensive results 
are given in Appendix \ref{app:ablation}.}

\noindent 
\textbf{Tuning.} 
Firstly, we would like to emphasize that the results are fairly insensitive to $s\in [0.05,0.3]$, 
and we recommend selecting $s=0.1$ as a good default choice.
It is worth noting that $v$ is an important hyper-parameter {\em a priori},
at the level of the original Bayesian model. 
For example, changing from $v=1/40$ 
to $v=0.1$ for $s=0.1$ changes (accuracy, NLL) from $(0.867,0.365)$ to $(0.889,0.271)$.
\footnote{Prior tuning is relevant for {\em all methods}, and is not particular to SBMC.}
If desired, one should first select $v$ optimally for the MAP/DE,
and then select $s$.
Both can be done with CV.
See Section \ref{sec:related} for further discussion.


\section{Discussion of Related Work}
\label{sec:related}


There has been a growing amount of work recently in many-short-chain MCMC, e.g.  
~\citep{wilkinson2006parallel,chen2016accelerated,sommer2024connecting,margossian2024nested,
nguyen2025unorthodox,sommermicrocanonical,duffieldscalable}. 
The island-SMC method \cite{verge2015parallel} considers interacting SMCs, 
which is necessary for consistency {\em unless the estimator is 
carefully constructed with appropriate weights}
\cite{whiteley2016role,dai2022invitation}.
See e.g. \cite{ours} for further discussion.

The idea of \textit{MAP-anchored priors} is intuitive, 
and closely related to a number of successful methods.
In addition to augmenting data with adversarial perturbations of the inputs,
as proposed in the original DE paper,
another intuitive idea to promote spread and generalization is to randomize the model itself.
Randomized maximum likelihood (RML) approaches
do this by anchoring each ensemble member to random draws from the prior and/or data
~\citep{gu2007iterative,bardsley2014randomize,pearce2020uncertainty}. 
SBMC can easily bootstrap DE or RML ideas
by initializing each process from a different MAP estimator.
It is worth noting that MC Dropout \cite{gal2016dropout} could also do this.

The method most closely related to our work is \cite{paulin2025sampling}, 
who anchor to the SWA estimator by adding a Gaussian factor, and simulate an ensemble of ULAs.
They also observed an extreme speedup in mixing time. 
We experimented with a similar formulation with a factor of $\mathcal{N}(\theta_{\sf MAP},s{\sf Id})$, 
$s\in(0,\infty)$, which also interpolates between the posterior and the MAP estimator
and is arguably more elegant and theoretically appealing.
But, the effective prior centers on $\frac{v}{s+v}\theta_{\sf MAP}$ (or SWA),
and in practice we found that this version did not perform as well 
as centering the prior on $\theta_{\sf MAP}$ itself. 

\begin{figure}
\includegraphics[width=.32\textwidth]{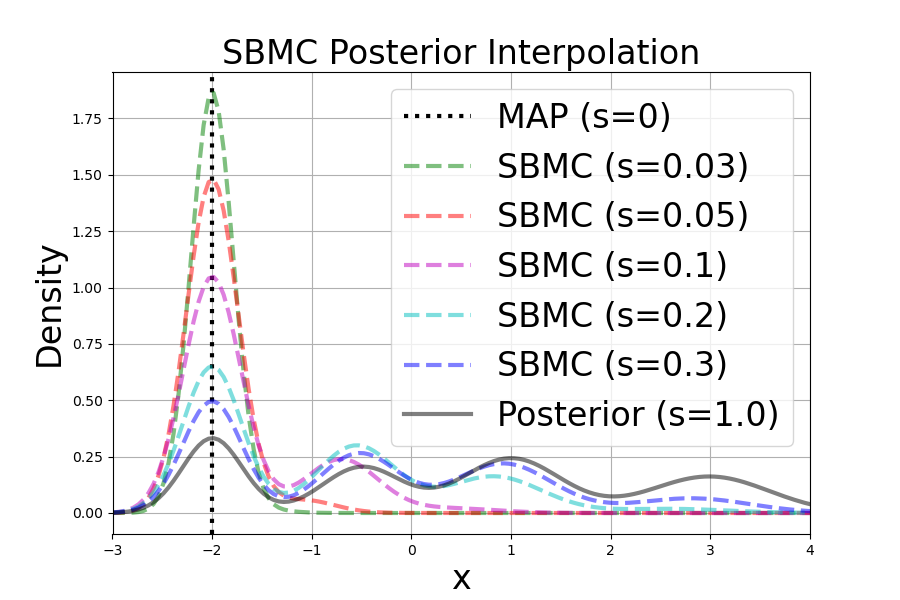}
\includegraphics[width=.32\textwidth]{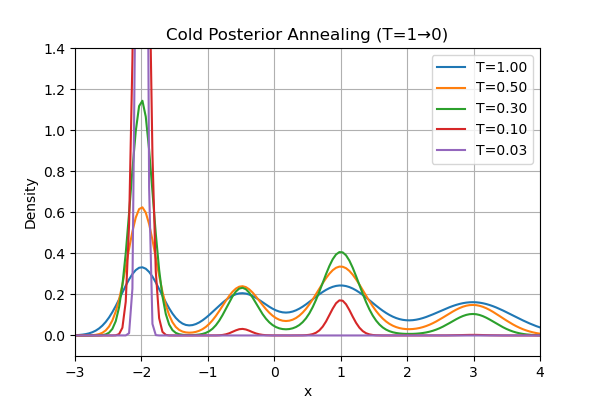}
\includegraphics[width=.32\textwidth]{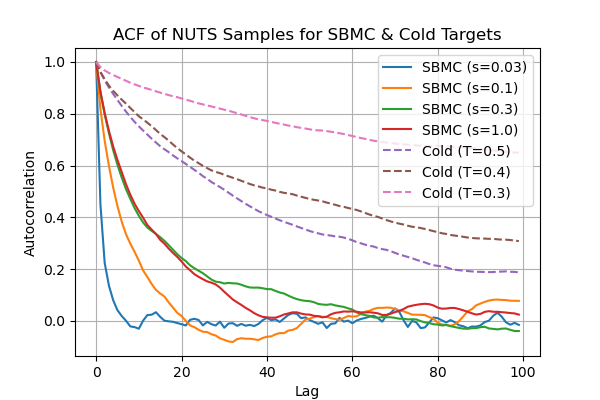}
\caption{Left: SBMC for various $s$. Middle: Cold posterior for various $T$.
Right: Autocorrelation functions using NUTS sampler, showing that SBMC {\em improves} mixing,
while CP {\em hinders} mixing.}
\label{fig:cpe}
\end{figure}

\noindent{\bf Cold posteriors} \citep{wenzel2020good}
also interpolate between $\delta_{\theta_{\sf MAP}}$
and the posterior
via {\em annealing} (or `tempering') the posterior \eqref{eq:posterior}
with an inverse temperature $T<1$, 
as 
 $
  \tilde\pi_{T}(\theta)\;\propto\; \cL(\theta; \cD )^{1/T} \pi_0^{1/T} \, .
$
This \emph{sharpens} the posterior, as shown in Figure \ref{fig:cpe} (middle) 
which typically makes the target distribution {\em more difficult to simulate} and {\em slows down} MCMC mixing,
as shown in Figure \ref{fig:cpe} (right).

The SBMC likelihood is \emph{effectively flattened relative to the Gaussian prior}
by the factor $s$, while the missing information is represented in the sharper prior, as shown in Figure \ref{fig:cpe} (left).
In practice, this means that the nonlinear and irregular component of the gradients has a smaller relative magnitude, 
the total Hessian of the posterior is better conditioned, and the 
chains mix faster.
See Figure \ref{fig:cpe} (right) for an illustration of the mixing behavior, 
and Appendix \ref{app:theory} for further discussion, including a sketch of the mathematics.
See Appendix \ref{app:related} for discussion of other related works.

\section{Conclusion}
\label{sec:conclusion}

The SBMC method has been introduced and
shown to be within reach of modern practical applications.
It comprises a judicious {\bf model} which uses a scalar parameter $s$ to 
interpolate between $\delta_{\theta_{\sf MAP}}$ ($s=0$)
and the posterior ($s=1$), and is hence able to balance the benefits of each and achieve 
strong performance in accuracy and UQ metrics {\em at a cost comparable to SOTA approaches like DE}.
Both MCMC$_\parallel$ and SMC$_\parallel$ are attractive {\bf algorithm} options,
which are consistent for the given target model. Therefore we have a mechanism for controlling 
the approximation between two reasonable choices if convergence is ensured. 
However, since the method no longer targets the posterior for any $s<1$, we would recommend
adopting a heuristic approach to convergence as with other SOTA methods, 
rather than chasing more rigorous convergence guarantees. 
Any BMC algorithm can be used, 
and SG-MCMC methods are particularly attractive since they are amenable to mini-batching 
and close to SGD, for which ample deep learning tooling is readily available.
The next step is to apply the method to modern generative AI models 
\citep{deepseek_r1_2025,openai_gpt_oss_model_card_2025,bai2025_qwen25vl_report,huang2025_otter} 
for controlling hallucinations 
and improving robustness and reliability. 
See Appendix \ref{app:gpt2} for preliminary results on next-token prediction with GPT-2
and further discussion.

\begin{ack}

KJHL and XL gratefully acknowledge the support of IBM and EPSRC in the form of an Industrial Case Doctoral Studentship Award. 
JML acknowledges funding from the U.S. Department of Energy, Office of Science, Advanced Scientific Computing Research (Early Career Research Program, ReACT-QISE).

\end{ack}

\bibliography{refs}

\begin{thebibliography}{91}
\providecommand{\natexlab}[1]{#1}
\providecommand{\url}[1]{\texttt{#1}}
\expandafter\ifx\csname urlstyle\endcsname\relax
  \providecommand{\doi}[1]{doi: #1}\else
  \providecommand{\doi}{doi: \begingroup \urlstyle{rm}\Url}\fi

\bibitem[Andrieu et~al.(2003)Andrieu, De~Freitas, Doucet, and Jordan]{andrieu2003introduction}
Christophe Andrieu, Nando De~Freitas, Arnaud Doucet, and Michael~I Jordan.
\newblock An introduction to {MCMC} for machine learning.
\newblock \emph{Machine learning}, 50:\penalty0 5--43, 2003.

\bibitem[Angelino et~al.(2016)Angelino, Johnson, and Adams]{angelino2016patterns}
Elaine Angelino, Matthew~James Johnson, and Ryan~P Adams.
\newblock Patterns of scalable {B}ayesian inference.
\newblock \emph{Foundations and Trends{\textregistered} in Machine Learning}, 9\penalty0 (2-3):\penalty0 119--247, 2016.

\bibitem[Arteaga et~al.(2025)Arteaga, Sch{\"o}n, and Pielawski]{arteaga2025hallucination}
Gabriel~Y Arteaga, Thomas~B Sch{\"o}n, and Nicolas Pielawski.
\newblock Hallucination detection in {LLM}s: Fast and memory-efficient finetuned models.
\newblock In \emph{Northern Lights Deep Learning Conference}, pages 1--15. PMLR, 2025.

\bibitem[Bai et~al.(2025)Bai, Chen, Liu, and {et al.}]{bai2025_qwen25vl_report}
Shuai Bai, Keqin Chen, Xuejing Liu, and {et al.}
\newblock Qwen2.5-vl technical report.
\newblock arXiv preprint arXiv:2502.13923, February 2025.
\newblock URL \url{https://arxiv.org/abs/2502.13923}.

\bibitem[Bardsley et~al.(2014)Bardsley, Solonen, Haario, and Laine]{bardsley2014randomize}
Johnathan~M. Bardsley, Antti Solonen, Heikki Haario, and Marko Laine.
\newblock Randomize–then–optimize: A method for sampling from posterior distributions in nonlinear inverse problems.
\newblock \emph{SIAM Journal on Scientific Computing}, 36\penalty0 (4):\penalty0 A1895--A1910, 2014.
\newblock \doi{10.1137/140964023}.

\bibitem[Berzuini and Gilks(2001)]{berzuini2001resample}
Carlo Berzuini and Walter Gilks.
\newblock Resample-move filtering with cross-model jumps.
\newblock \emph{Sequential Monte Carlo Methods in Practice}, pages 117--138, 2001.

\bibitem[Bishop(2006)]{bishop2006pattern}
Christopher~M. Bishop.
\newblock \emph{Pattern Recognition and Machine Learning}.
\newblock Springer, New York, NY, 2006.
\newblock ISBN 978-0-387-31073-2.

\bibitem[Black et~al.(2022)]{black2022gptneox}
Sid Black et~al.
\newblock {GPT-NeoX-20B}: An open-source autoregressive language model.
\newblock \emph{arXiv 2204.06745}, 2022.

\bibitem[Chen et~al.(2014)Chen, Fox, and Guestrin]{chen2014stochastic}
Tianqi Chen, Emily Fox, and Carlos Guestrin.
\newblock Stochastic gradient hamiltonian monte carlo.
\newblock In \emph{International conference on machine learning}, pages 1683--1691. PMLR, 2014.

\bibitem[Chen et~al.(2016)Chen, Keyes, Law, and Ltaief]{chen2016accelerated}
Yuxin Chen, David Keyes, Kody~JH Law, and Hatem Ltaief.
\newblock Accelerated dimension-independent adaptive {M}etropolis.
\newblock \emph{SIAM Journal on Scientific Computing}, 38\penalty0 (5):\penalty0 S539--S565, 2016.

\bibitem[Chopin et~al.(2020)Chopin, Papaspiliopoulos, et~al.]{chopin2020introduction}
Nicolas Chopin, Omiros Papaspiliopoulos, et~al.
\newblock \emph{An introduction to sequential {Monte C}arlo}, volume~4.
\newblock Springer, 2020.

\bibitem[Chowdhery et~al.(2022)]{chowdhery2022palm}
Aakanksha Chowdhery et~al.
\newblock Pa{LM}: Scaling language modeling with pathways.
\newblock In \emph{ICML}, 2022.

\bibitem[Cotter et~al.(2013)Cotter, Roberts, Stuart, and White]{cotter2013mcmc}
Simon~L Cotter, Gareth~O Roberts, Andrew~M Stuart, and David White.
\newblock {MCMC} methods for functions: Modifying old algorithms to make them faster.
\newblock \emph{Statistical Science}, pages 424--446, 2013.

\bibitem[Dai et~al.(2022)Dai, Heng, Jacob, and Whiteley]{dai2022invitation}
Chenguang Dai, Jeremy Heng, Pierre~E Jacob, and Nick Whiteley.
\newblock An invitation to sequential {Monte C}arlo samplers.
\newblock \emph{Journal of the American Statistical Association}, 117\penalty0 (539):\penalty0 1587--1600, 2022.

\bibitem[Daxberger et~al.(2021)Daxberger, Kristiadi, Immer, Eschenhagen, Bauer, and Hennig]{daxberger2021laplace}
Erik Daxberger, Agustinus Kristiadi, Alexander Immer, Runa Eschenhagen, Matthias Bauer, and Philipp Hennig.
\newblock Laplace redux: Effortless {B}ayesian deep learning.
\newblock In \emph{Advances in Neural Information Processing Systems}, volume~34, pages 16367--16380, 2021.

\bibitem[{DeepSeek-AI}(2025)]{deepseek_r1_2025}
{DeepSeek-AI}.
\newblock Deepseek-r1: Incentivizing reasoning capability in llms via reinforcement learning.
\newblock \emph{arXiv preprint arXiv:2501.12948}, 2025.
\newblock URL \url{https://arxiv.org/abs/2501.12948}.

\bibitem[Del~Moral(2004)]{del2004feynman}
Pierre Del~Moral.
\newblock \emph{Feynman-kac formulae}.
\newblock Springer, 2004.

\bibitem[Del~Moral et~al.(2006)Del~Moral, Doucet, and Jasra]{del2006sequential}
Pierre Del~Moral, Arnaud Doucet, and Ajay Jasra.
\newblock Sequential {Monte C}arlo samplers.
\newblock \emph{Journal of the Royal Statistical Society Series B: Statistical Methodology}, 68\penalty0 (3):\penalty0 411--436, 2006.

\bibitem[Deligiannidis et~al.(2020)Deligiannidis, Doucet, and Vihola]{vihola2020poissonSMC}
Giorgos Deligiannidis, Arnaud Doucet, and Matti Vihola.
\newblock Unbiased likelihood estimation for {SMC} via {P}oisson randomisation.
\newblock \emph{Statistics and Computing}, 30\penalty0 (4):\penalty0 783--810, 2020.
\newblock \doi{10.1007/s11222-019-09906-5}.
\newblock Uses Poisson–product (random replicate) estimators to obtain non‑negative, exactly unbiased weights in sequential Monte Carlo.

\bibitem[Depeweg et~al.(2018)Depeweg, Hernandez-Lobato, Doshi-Velez, and Udluft]{depeweg2018decomposition}
Stefan Depeweg, Jose-Miguel Hernandez-Lobato, Finale Doshi-Velez, and Steffen Udluft.
\newblock Decomposition of uncertainty in {B}ayesian deep learning for efficient and risk-sensitive learning.
\newblock In \emph{International conference on machine learning}, pages 1184--1193. PMLR, 2018.

\bibitem[Duane et~al.(1987)Duane, Kennedy, Pendleton, and Roweth]{duane1987hybrid}
Simon Duane, Anthony~D Kennedy, Brian~J Pendleton, and Duncan Roweth.
\newblock Hybrid {Monte C}arlo.
\newblock \emph{Physics letters B}, 195\penalty0 (2):\penalty0 216--222, 1987.

\bibitem[Duffield et~al.(2025)Duffield, Donatella, Chiu, Klett, and Simpson]{duffieldscalable}
Samuel Duffield, Kaelan Donatella, Johnathan Chiu, Phoebe Klett, and Daniel Simpson.
\newblock Scalable bayesian learning with posteriors.
\newblock In \emph{The Thirteenth International Conference on Learning Representations}, 2025.

\bibitem[Efron and Morris(1977)]{efron1977stein}
Bradley Efron and Carl Morris.
\newblock Stein's estimation rule and its competitors—an empirical {B}ayes approach.
\newblock \emph{Journal of the American Statistical Association}, 72\penalty0 (357):\penalty0 591--599, 1977.

\bibitem[Eschenhagen et~al.(2021)Eschenhagen, Daxberger, Hennig, and Kristiadi]{eschenhagen2021mixtures}
Runa Eschenhagen, Erik Daxberger, Philipp Hennig, and Agustinus Kristiadi.
\newblock Mixtures of laplace approximations for improved post-hoc uncertainty in deep learning.
\newblock \emph{arXiv preprint arXiv:2111.03577}, 2021.

\bibitem[Farquhar et~al.(2024)Farquhar, Kossen, Kuhn, and Gal]{farquhar2024detecting}
Sebastian Farquhar, Jannik Kossen, Lorenz Kuhn, and Yarin Gal.
\newblock Detecting hallucinations in large language models using semantic entropy.
\newblock \emph{Nature}, 630\penalty0 (8017):\penalty0 625--630, 2024.

\bibitem[Gal and Ghahramani(2016)]{gal2016dropout}
Yarin Gal and Zoubin Ghahramani.
\newblock Dropout as a {B}ayesian approximation: Representing model uncertainty in deep learning.
\newblock In \emph{international conference on machine learning}, pages 1050--1059. PMLR, 2016.

\bibitem[Galison et~al.(2022)Galison, Akiyama, Alberdi, Alef, Algaba, Anantua, Asada, Azulay, Bach, Baczko, et~al.]{galison2022first}
Peter Galison, Kazunori Akiyama, Antxon Alberdi, Walter Alef, Juan~Carlos Algaba, Richard Anantua, Keiichi Asada, Rebecca Azulay, Uwe Bach, Anne-Kathrin Baczko, et~al.
\newblock First {Sagittarius A} event horizon telescope results. iii. imaging of the galactic center supermassive black hole.
\newblock \emph{Astrophysical journal. Letters}, 930\penalty0 (2):\penalty0 L17, 2022.

\bibitem[Gelfand and Smith(1990)]{gelfand1990sampling}
Alan~E Gelfand and Adrian~FM Smith.
\newblock Sampling-based approaches to calculating marginal densities.
\newblock \emph{Journal of the American statistical association}, 85\penalty0 (410):\penalty0 398--409, 1990.

\bibitem[Geyer(1992)]{geyer1992practical}
Charles~J Geyer.
\newblock Practical {Markov chain Monte C}arlo.
\newblock \emph{Statistical science}, pages 473--483, 1992.

\bibitem[Gilks et~al.(1994)Gilks, Roberts, and George]{gilks1994adaptive}
Walter~R Gilks, Gareth~O Roberts, and Edward~I George.
\newblock Adaptive direction sampling.
\newblock \emph{Journal of the Royal Statistical Society: Series D (The Statistician)}, 43\penalty0 (1):\penalty0 179--189, 1994.

\bibitem[Goodman and Weare(2010)]{goodman2010ensemble}
Jonathan Goodman and Jonathan Weare.
\newblock Ensemble samplers with affine invariance.
\newblock \emph{Communications in applied mathematics and computational science}, 5\penalty0 (1):\penalty0 65--80, 2010.

\bibitem[Goyal et~al.(2017)]{goyal2017imagenet1h}
Priya Goyal et~al.
\newblock Accurate, large minibatch sgd: Training imagenet in 1 hour.
\newblock \emph{arXiv 1706.02677}, 2017.

\bibitem[Gu and Oliver(2007)]{gu2007iterative}
Yaqing Gu and Dean~S Oliver.
\newblock An iterative ensemble {K}alman filter for multiphase fluid flow data assimilation.
\newblock \emph{Spe Journal}, 12\penalty0 (04):\penalty0 438--446, 2007.

\bibitem[Gunawan et~al.(2021)Gunawan, Quiroz, Kohn, Tran, and Villani]{gunawan2021subsamplingSMC}
Donny Gunawan, Mauricio Quiroz, Robert Kohn, Minh‐Ngoc Tran, and Mattias Villani.
\newblock Subsampling sequential {Monte C}arlo for static {B}ayesian models.
\newblock \emph{Bayesian Analysis}, 16\penalty0 (3):\penalty0 721--749, 2021.
\newblock \doi{10.1214/20-BA1220}.
\newblock Introduces Bernoulli‐thinning (random inclusion) unbiased‐likelihood estimators inside SMC.

\bibitem[Gustafsson et~al.(2020)Gustafsson, Danelljan, and Schon]{gustafsson2020evaluating}
Fredrik~K Gustafsson, Martin Danelljan, and Thomas~B Schon.
\newblock Evaluating scalable {B}ayesian deep learning methods for robust computer vision.
\newblock In \emph{Proceedings of the IEEE/CVF conference on computer vision and pattern recognition workshops}, pages 318--319, 2020.

\bibitem[Hastings(1970)]{hastings1970monte}
W~Keith Hastings.
\newblock Monte {C}arlo sampling methods using {M}arkov chains and their applications.
\newblock \emph{Biometrika}, 57\penalty0 (1):\penalty0 97--109, 1970.

\bibitem[Hoffman et~al.(2021)Hoffman, Radul, and Sountsov]{hoffman2021adaptive}
Matthew Hoffman, Alexey Radul, and Pavel Sountsov.
\newblock An adaptive-{MCMC} scheme for setting trajectory lengths in {Hamiltonian Monte C}arlo.
\newblock In \emph{International Conference on Artificial Intelligence and Statistics}, pages 3907--3915. PMLR, 2021.

\bibitem[Hoffman and Sountsov(2022)]{hoffman2022tuning}
Matthew~D Hoffman and Pavel Sountsov.
\newblock Tuning-free generalized {Hamiltonian Monte C}arlo.
\newblock In \emph{International conference on artificial intelligence and statistics}, pages 7799--7813. PMLR, 2022.

\bibitem[Hoffman et~al.(2014)Hoffman, Gelman, et~al.]{hoffman2014no}
Matthew~D Hoffman, Andrew Gelman, et~al.
\newblock The {No-U-Turn} sampler: adaptively setting path lengths in {Hamiltonian Monte C}arlo.
\newblock \emph{J. Mach. Learn. Res.}, 15\penalty0 (1):\penalty0 1593--1623, 2014.

\bibitem[Hou et~al.(2024)Hou, Li, He, Yan, Chen, and McAuley]{hou2024bridging}
Yupeng Hou, Jiacheng Li, Zhankui He, An~Yan, Xiusi Chen, and Julian McAuley.
\newblock Bridging language and items for retrieval and recommendation.
\newblock \emph{arXiv preprint arXiv:2403.03952}, 2024.

\bibitem[Hu et~al.(2022)Hu, Shen, Wallis, Allen-Zhu, Li, Wang, Wang, Chen, et~al.]{hu2022lora}
Edward~J Hu, Yelong Shen, Phillip Wallis, Zeyuan Allen-Zhu, Yuanzhi Li, Shean Wang, Lu~Wang, Weizhu Chen, et~al.
\newblock Lora: Low-rank adaptation of large language models.
\newblock \emph{ICLR}, 1\penalty0 (2):\penalty0 3, 2022.

\bibitem[Huang et~al.(2025)Huang, Liu, Fu, Wu, Mukadam, Malik, Goldberg, and Abbeel]{huang2025_otter}
Huang Huang, Fangchen Liu, Letian Fu, Tingfan Wu, Mustafa Mukadam, Jitendra Malik, Ken Goldberg, and Pieter Abbeel.
\newblock Otter: A vision-language-action model with text-aware visual feature extraction.
\newblock \emph{arXiv preprint arXiv:2503.03734}, 2025.
\newblock URL \url{https://arxiv.org/abs/2503.03734}.

\bibitem[Huang et~al.(2019)]{huang2019gpipe}
Yanping Huang et~al.
\newblock {GP}ipe: Efficient training of giant neural networks using pipeline parallelism.
\newblock In \emph{NeurIPS}, 2019.

\bibitem[H{\"u}llermeier and Waegeman(2021)]{hullermeier2021aleatoric}
Eyke H{\"u}llermeier and Willem Waegeman.
\newblock Aleatoric and epistemic uncertainty in machine learning: An introduction to concepts and methods.
\newblock \emph{Machine learning}, 110\penalty0 (3):\penalty0 457--506, 2021.

\bibitem[Izmailov et~al.(2018)Izmailov, Podoprikhin, Garipov, Vetrov, and Wilson]{izmailov2018averaging}
Pavel Izmailov, Dmitrii Podoprikhin, Timur Garipov, Dmitry Vetrov, and Andrew~Gordon Wilson.
\newblock Averaging weights leads to wider optima and better generalization.
\newblock In \emph{34th Conference on Uncertainty in Artificial Intelligence 2018, UAI 2018}, pages 876--885. Association For Uncertainty in Artificial Intelligence (AUAI), 2018.

\bibitem[Izmailov et~al.(2021)Izmailov, Vikram, Hoffman, and Wilson]{izmailov2021bayesian}
Pavel Izmailov, Sharad Vikram, Matthew~D Hoffman, and Andrew Gordon~Gordon Wilson.
\newblock What are {B}ayesian neural network posteriors really like?
\newblock In \emph{International conference on machine learning}, pages 4629--4640. PMLR, 2021.

\bibitem[Jarzynski(1997)]{jarzynski1997equilibrium}
Christopher Jarzynski.
\newblock Equilibrium free-energy differences from nonequilibrium measurements: A master-equation approach.
\newblock \emph{Physical Review E}, 56\penalty0 (5):\penalty0 5018, 1997.

\bibitem[Ji et~al.(2023)Ji, Lee, Frieske, Yu, Su, Xu, Ishii, Bang, Madotto, and Fung]{ji2023survey}
Ziwei Ji, Nayeon Lee, Rita Frieske, Tiezheng Yu, Dan Su, Yanfei Xu, Eric Ishii, Yejin Bang, Andrea Madotto, and Pascale Fung.
\newblock Survey of hallucination in natural language generation.
\newblock \emph{ACM Computing Surveys}, 55\penalty0 (12):\penalty0 248:1--248:38, 2023.
\newblock \doi{10.1145/3571730}.

\bibitem[Krause and H{\"u}botter(2025)]{krause2025probabilistic}
Andreas Krause and Jonas H{\"u}botter.
\newblock Probabilistic {Artificial I}ntelligence.
\newblock \emph{arXiv preprint arXiv:2502.05244}, 2025.

\bibitem[Krizhevsky et~al.(2009)Krizhevsky, Hinton, et~al.]{krizhevsky2009learning}
Alex Krizhevsky, Geoffrey Hinton, et~al.
\newblock Learning multiple layers of features from tiny images.
\newblock 2009.

\bibitem[Lakshminarayanan et~al.(2017)Lakshminarayanan, Pritzel, and Blundell]{lakshminarayanan2017simple}
Balaji Lakshminarayanan, Alexander Pritzel, and Charles Blundell.
\newblock Simple and scalable predictive uncertainty estimation using deep ensembles.
\newblock \emph{Advances in neural information processing systems}, 30, 2017.

\bibitem[LeCun et~al.(2010)LeCun, Cortes, and Burges]{lecun2010mnist}
Yann LeCun, Corinna Cortes, and CJ~Burges.
\newblock Mnist handwritten digit database.
\newblock \emph{ATT Labs [Online]. Available: http://yann.lecun.com/exdb/mnist}, 2, 2010.

\bibitem[Liang et~al.(2025)Liang, Lukens, Lohani, Kirby, Searles, Qiu, and Law]{ours}
Xinzhu Liang, Joseph~M Lukens, Sanjaya Lohani, Brian~T Kirby, Thomas~A Searles, Xin Qiu, and Kody~JH Law.
\newblock Comparison of parallel {SMC} and {MCMC} for {B}ayesian deep learning.
\newblock \emph{arXiv preprint arXiv:2402.06173}, 2025.

\bibitem[Maas et~al.(2011)Maas, Daly, Pham, Huang, Ng, and Potts]{maas2011}
Andrew Maas, Raymond~E Daly, Peter~T Pham, Dan Huang, Andrew~Y Ng, and Christopher Potts.
\newblock Learning word vectors for sentiment analysis.
\newblock In \emph{Proceedings of the 49th annual meeting of the association for computational linguistics: Human language technologies}, pages 142--150, 2011.

\bibitem[MacKay(1992)]{mackay1992practical}
David~JC MacKay.
\newblock A practical {B}ayesian framework for backpropagation networks.
\newblock \emph{Neural computation}, 4\penalty0 (3):\penalty0 448--472, 1992.

\bibitem[Maclaurin and Adams(2014)]{maclaurin2014firefly}
Dougal Maclaurin and Ryan~P Adams.
\newblock Firefly {Monte C}arlo: Exact {MCMC} with subsets of data.
\newblock \emph{arXiv preprint arXiv:1403.5693}, 2014.

\bibitem[Maddox et~al.(2019)Maddox, Izmailov, Garipov, Vetrov, and Wilson]{maddox2019simple}
Wesley~J Maddox, Pavel Izmailov, Timur Garipov, Dmitry~P Vetrov, and Andrew~Gordon Wilson.
\newblock A simple baseline for bayesian uncertainty in deep learning.
\newblock \emph{Advances in neural information processing systems}, 32, 2019.

\bibitem[Margossian et~al.(2024)Margossian, Hoffman, Sountsov, Riou-Durand, Vehtari, and Gelman]{margossian2024nested}
Charles~C Margossian, Matthew~D Hoffman, Pavel Sountsov, Lionel Riou-Durand, Aki Vehtari, and Andrew Gelman.
\newblock Nested $\hat{R}$: assessing the convergence of {Markov chain Monte Carlo} when running many short chains.
\newblock \emph{Bayesian Analysis}, 1\penalty0 (1):\penalty0 1--28, 2024.

\bibitem[Metropolis et~al.(1953)Metropolis, Rosenbluth, Rosenbluth, Teller, and Teller]{metropolis1953equation}
Nicholas Metropolis, Arianna~W Rosenbluth, Marshall~N Rosenbluth, Augusta~H Teller, and Edward Teller.
\newblock Equation of state calculations by fast computing machines.
\newblock \emph{The journal of chemical physics}, 21\penalty0 (6):\penalty0 1087--1092, 1953.

\bibitem[Mohan and Scaife(2024)]{mohan2024evaluating}
Devina Mohan and Anna~MM Scaife.
\newblock Evaluating {B}ayesian deep learning for radio galaxy classification.
\newblock \emph{arXiv preprint arXiv:2405.18351}, 2024.

\bibitem[Neal(1998)]{bernardo1998regression}
Radford Neal.
\newblock Regression and classification using {G}aussian process priors.
\newblock \emph{Bayesian statistics}, 6:\penalty0 475, 1998.

\bibitem[Neal(1993)]{neal2012bayesian}
Radford~M Neal.
\newblock \emph{Bayesian learning for neural networks}, volume 118.
\newblock Springer Science \& Business Media, 1993.

\bibitem[Neal et~al.(2011)]{neal2011mcmc}
Radford~M Neal et~al.
\newblock {MCMC} using {H}amiltonian dynamics.
\newblock \emph{Handbook of markov chain monte carlo}, 2\penalty0 (11):\penalty0 2, 2011.

\bibitem[Nguyen et~al.(2025)Nguyen, Law, and Lukens]{nguyen2025unorthodox}
Hanson~H Nguyen, Kody~JH Law, and Joseph~M Lukens.
\newblock Unorthodox parallelization for bayesian quantum state estimation.
\newblock \emph{New Journal of Physics}, 27\penalty0 (5):\penalty0 054507, 2025.

\bibitem[{OpenAI}(2025)]{openai_gpt_oss_model_card_2025}
{OpenAI}.
\newblock gpt-oss-120b and gpt-oss-20b model card, August 2025.
\newblock URL \url{https://openai.com/index/gpt-oss-model-card/}.

\bibitem[Papamarkou et~al.(2024)Papamarkou, Skoularidou, Palla, Aitchison, Arbel, Dunson, Filippone, Fortuin, Hennig, Hubin, et~al.]{papamarkou2024position}
Theodore Papamarkou, Maria Skoularidou, Konstantina Palla, Laurence Aitchison, Julyan Arbel, David Dunson, Maurizio Filippone, Vincent Fortuin, Philipp Hennig, Aliaksandr Hubin, et~al.
\newblock Position paper: {B}ayesian deep learning in the age of large-scale {AI}.
\newblock \emph{arXiv preprint arXiv:2402.00809}, 2024.

\bibitem[Paulin et~al.(2025)Paulin, Whalley, Chada, and Leimkuhler]{paulin2025sampling}
Daniel Paulin, Peter~A Whalley, Neil~K Chada, and Benedict~J Leimkuhler.
\newblock Sampling from bayesian neural network posteriors with symmetric minibatch splitting langevin dynamics.
\newblock In \emph{International Conference on Artificial Intelligence and Statistics}, pages 5014--5022. PMLR, 2025.

\bibitem[Pearce et~al.(2020)Pearce, Leibfried, and Brintrup]{pearce2020uncertainty}
Tim Pearce, Felix Leibfried, and Alexandra Brintrup.
\newblock Uncertainty in neural networks: Approximately {B}ayesian ensembling.
\newblock In Silvia Chiappa and Roberto Calandra, editors, \emph{Proceedings of the Twenty Third International Conference on Artificial Intelligence and Statistics}, volume 108 of \emph{Proceedings of Machine Learning Research}, pages 234--244. PMLR, Aug 2020.
\newblock URL \url{https://proceedings.mlr.press/v108/pearce20a.html}.

\bibitem[Qiu and Miikkulainen()]{qiusemantic}
Xin Qiu and Risto Miikkulainen.
\newblock Semantic density: Uncertainty quantification for large language models through confidence measurement in semantic space.
\newblock In \emph{The Thirty-eighth Annual Conference on Neural Information Processing Systems}.

\bibitem[Qiu et~al.()Qiu, Meyerson, and Miikkulainen]{qiuquantifying}
Xin Qiu, Elliot Meyerson, and Risto Miikkulainen.
\newblock Quantifying point-prediction uncertainty in neural networks via residual estimation with an {I/O} kernel.
\newblock In \emph{International Conference on Learning Representations}.

\bibitem[Rajbhandari et~al.(2020)Rajbhandari, Rasley, Ruwase, and He]{rajbhandari2020zero}
Samyam Rajbhandari, Jeff Rasley, Olatunji Ruwase, and Yuxiong He.
\newblock {ZeRO}: Memory optimizations toward training trillion parameter models.
\newblock In \emph{SC (ACM/IEEE Intl. Conf. for High Performance Computing)}, 2020.
\newblock URL \url{https://arxiv.org/abs/1910.02054}.

\bibitem[Reimers and Gurevych(2019)]{reimers-2019-sentence-bert}
Nils Reimers and Iryna Gurevych.
\newblock Sentence-{BERT}: Sentence embeddings using siamese {BERT}-networks.
\newblock In \emph{Proceedings of the 2019 Conference on Empirical Methods in Natural Language Processing}. Association for Computational Linguistics, 11 2019.
\newblock URL \url{https://arxiv.org/abs/1908.10084}.

\bibitem[Rendell et~al.(2020)Rendell, Johansen, Lee, and Whiteley]{rendell2020global}
Lewis~J Rendell, Adam~M Johansen, Anthony Lee, and Nick Whiteley.
\newblock Global consensus {Monte C}arlo.
\newblock \emph{Journal of Computational and Graphical Statistics}, 30\penalty0 (2):\penalty0 249--259, 2020.

\bibitem[Robert et~al.(1999)Robert, Casella, and Casella]{robert1999monte}
Christian~P Robert, George Casella, and George Casella.
\newblock \emph{Monte {C}arlo statistical methods}, volume~2.
\newblock Springer, 1999.

\bibitem[Roberts and Tweedie(1996)]{roberts1996exponential}
Gareth~O Roberts and Richard~L Tweedie.
\newblock Exponential convergence of {L}angevin distributions and their discrete approximations.
\newblock \emph{Bernoulli}, pages 341--363, 1996.

\bibitem[Shaker and H{\"u}llermeier(2020)]{shaker2020aleatoric}
Mohammad~Hossein Shaker and Eyke H{\"u}llermeier.
\newblock Aleatoric and epistemic uncertainty with random forests.
\newblock In \emph{International Symposium on Intelligent Data Analysis}, pages 444--456. Springer, 2020.

\bibitem[Shoeybi et~al.(2019)]{shoeybi2019megatron}
Mohammad Shoeybi et~al.
\newblock Megatron-{LM}: Training multi-billion parameter language models using model parallelism.
\newblock In \emph{SC}, 2019.

\bibitem[Sommer et~al.(2024)Sommer, Wimmer, Papamarkou, Bothmann, Bischl, and R{\"u}gamer]{sommer2024connecting}
Emanuel Sommer, Lisa Wimmer, Theodore Papamarkou, Ludwig Bothmann, Bernd Bischl, and David R{\"u}gamer.
\newblock Connecting the dots: Is mode-connectedness the key to feasible sample-based inference in bayesian neural networks?
\newblock In \emph{International Conference on Machine Learning}, pages 45988--46018. PMLR, 2024.

\bibitem[Sommer et~al.(2025)Sommer, Robnik, Nozadze, Seljak, and R{\"u}gamer]{sommermicrocanonical}
Emanuel Sommer, Jakob Robnik, Giorgi Nozadze, Uros Seljak, and David R{\"u}gamer.
\newblock Microcanonical langevin ensembles: Advancing the sampling of bayesian neural networks.
\newblock In \emph{The Thirteenth International Conference on Learning Representations}, 2025.

\bibitem[Song et~al.(2020)Song, Tan, Qin, Lu, and Liu]{song2020mpnet}
Kaitao Song, Xu~Tan, Tao Qin, Jianfeng Lu, and Tie-Yan Liu.
\newblock Mpnet: Masked and permuted pre-training for language understanding.
\newblock \emph{Advances in neural information processing systems}, 33:\penalty0 16857--16867, 2020.

\bibitem[Syed et~al.(2024)Syed, Bouchard-C{\^o}t{\'e}, Chern, and Doucet]{syed2024optimised}
Saifuddin Syed, Alexandre Bouchard-C{\^o}t{\'e}, Kevin Chern, and Arnaud Doucet.
\newblock Optimised annealed {Sequential Monte C}arlo samplers.
\newblock \emph{arXiv preprint arXiv:2408.12057}, 2024.

\bibitem[Vashurin et~al.(2024)Vashurin, Fadeeva, Vazhentsev, Rvanova, Tsvigun, Vasilev, Xing, Sadallah, Grishchenkov, Petrakov, et~al.]{vashurin2024benchmarking}
Roman Vashurin, Ekaterina Fadeeva, Artem Vazhentsev, Lyudmila Rvanova, Akim Tsvigun, Daniil Vasilev, Rui Xing, Abdelrahman~Boda Sadallah, Kirill Grishchenkov, Sergey Petrakov, et~al.
\newblock Benchmarking uncertainty quantification methods for large language models with {LM}-polygraph.
\newblock \emph{arXiv preprint arXiv:2406.15627}, 2024.

\bibitem[Verg{\'e} et~al.(2015)Verg{\'e}, Dubarry, Del~Moral, and Moulines]{verge2015parallel}
Christelle Verg{\'e}, Cyrille Dubarry, Pierre Del~Moral, and Eric Moulines.
\newblock On parallel implementation of sequential {Monte C}arlo methods: the island particle model.
\newblock \emph{Statistics and Computing}, 25\penalty0 (2):\penalty0 243--260, 2015.

\bibitem[Vrugt et~al.(2009)Vrugt, ter Braak, Diks, Robinson, Hyman, and Higdon]{vrugt2009accelerating}
Jasper~A Vrugt, Cajo~JF ter Braak, Cees~GH Diks, Bruce~A Robinson, James~M Hyman, and Dave Higdon.
\newblock Accelerating {Markov chain Monte C}arlo simulation by differential evolution with self-adaptive randomized subspace sampling.
\newblock \emph{International journal of nonlinear sciences and numerical simulation}, 10\penalty0 (3):\penalty0 273--290, 2009.

\bibitem[Welling and Teh(2011)]{welling2011bayesian}
Max Welling and Yee~W Teh.
\newblock Bayesian learning via stochastic gradient {L}angevin dynamics.
\newblock In \emph{Proceedings of the 28th international conference on machine learning (ICML-11)}, pages 681--688. Citeseer, 2011.

\bibitem[Wenzel et~al.(2020)Wenzel, Roth, Veeling, {\'S}wi{\k{a}}tkowski, Tran, Mandt, Snoek, Salimans, Jenatton, and Nowozin]{wenzel2020good}
Florian Wenzel, Kevin Roth, Bastiaan~S Veeling, Jakub {\'S}wi{\k{a}}tkowski, Linh Tran, Stephan Mandt, Jasper Snoek, Tim Salimans, Rodolphe Jenatton, and Sebastian Nowozin.
\newblock How good is the {B}ayes posterior in deep neural networks really?
\newblock In \emph{Proceedings of the 37th International Conference on Machine Learning}, pages 10248--10259, 2020.

\bibitem[Whiteley et~al.(2015)Whiteley, Lee, and Heine]{whiteley2016role}
Nick Whiteley, Anthony Lee, and Kari Heine.
\newblock On the role of interaction in sequential {Monte C}arlo algorithms.
\newblock \emph{Bernoulli}, 22\penalty0 (1):\penalty0 494--529, 2015.

\bibitem[Wilkinson(2006)]{wilkinson2006parallel}
Darren~J Wilkinson.
\newblock Parallel {B}ayesian computation.
\newblock \emph{Statistics Textbooks and Monographs}, 184:\penalty0 477, 2006.

\bibitem[Wilson and Izmailov(2020)]{wilson2020bayesian}
Andrew~G Wilson and Pavel Izmailov.
\newblock Bayesian deep learning and a probabilistic perspective of generalization.
\newblock \emph{Advances in neural information processing systems}, 33:\penalty0 4697--4708, 2020.

\bibitem[You et~al.(2021)You, Liu, Wang, and Long]{you2021logme}
Kaichao You, Yong Liu, Jianmin Wang, and Mingsheng Long.
\newblock Log{ME}: Practical assessment of pre-trained models for transfer learning.
\newblock In Marina Meil{\`{a}} and Tong Zhang, editors, \emph{Proceedings of the 38th International Conference on Machine Learning}, volume 139 of \emph{Proceedings of Machine Learning Research}, pages 12133--12143. PMLR, Jul 2021.

\bibitem[You et~al.(2022)You, Liu, Zhang, Wang, Jordan, and Long]{you2022ranking}
Kaichao You, Yong Liu, Ziyang Zhang, Jianmin Wang, Michael~I Jordan, and Mingsheng Long.
\newblock Ranking and tuning pre-trained models: A new paradigm for exploiting model hubs.
\newblock \emph{Journal of Machine Learning Research}, 23\penalty0 (209):\penalty0 1--47, 2022.

\end{thebibliography}




\appendix

\section{Future Directions}

The most obvious next step is UQ for modern large language models (LLMs) \citep{deepseek_r1_2025},
where robustness and hallucination detection are crucial pain points \citep{vashurin2024benchmarking}.
It has been recently shown that high-quality entropy metrics are valuable for identifying 
untrustworthy outputs there \citep{gustafsson2020evaluating,arteaga2025hallucination,farquhar2024detecting}.
In the context of LLMs, where the training itself is extremely computationally expensive, 
it becomes particularly important to have add-on plug-in type methods that can be applied
post-training, such as \citep{qiuquantifying,farquhar2024detecting,qiusemantic}. 
But those methods are constrained to the uncertainty already encoded 
in the point estimator of model weights, which may already be under-estimated.
The work \citep{arteaga2025hallucination} has shown that
batch ensembles of fine-tuned LLMs can also work well for UQ hallucination detection, 
and also that epistemic uncertainty provides valuable information for that task. 
Based on existing benchmarks against deep ensembles,
we believe our SBMC($s$) approach will perform even better. 
Furthermore, an even simpler and cheaper version is 
to {\em learn the last layer} only, 
so all the data can be pre-processed once and for all by the frozen 
pre-trained LLM parameters, and then we simply run the last layer through SBMC. 
This can be applied at the pre-training or post-training stage,
although the value of the method on downstream tasks will be most clear at post-training,
while there would be some necessary design choices for 
how to leverage a pre-trained ensemble, instead of a point estimator, during post-training.

\subsection{GPT-2}
\label{app:gpt2}

Here we present some preliminary results on GPT-2, on consumer hardware 
(an old MacBook Pro with an M2 processor and 16GB RAM).
These are early results, just to further emphasize scalability
and potential utility in hallucination detection. 
The starting point is the pre-trained GPT-2 model fine-tuned on Shakespeare data
\footnote{\texttt{https://huggingface.co/sadia72/gpt2-shakespeare}}.
We then adopt a LoRA approach \cite{hu2022lora} to fine-tune an additive rank 50 adjustment 
with $\approx 2e5$ parameters at the last layer 
on the 3e5 token tiny Shakespeare dataset 
\footnote{\texttt{https://raw.githubusercontent.com/karpathy/char-rnn/master/data/tinyshakespeare/input.txt}}
(in 128-token blocks) for 100 epochs, 
and consider top-1 token-level predictions 
\footnote{We truncated to the 2500 most frequent tokens, which includes 
tokens that appeared 11 or more times.}. 
The results are presented in Table \ref{tab:gpt2}.

\begin{table}[h]
\centering
\caption{Comparison of methods on test accuracy, NLL, and various entropy metrics
for next-token prediction with GPT2 on tiny Shakespeare.}
\label{tab:gpt2}
\begin{tabular}{l|c|c|c|c|c|c}
\toprule
Methods & Accuracy (\%) & NLL & $H_{\sf tot}$ correct & $H_{\sf tot}$ incorrect 
& $H_{\sf ep}$ correct & $H_{\sf ep}$ incorrect \\
\midrule
MAP    & 38.66 & 3.166 & 1.554 & 3.605 & 0     & 0     \\
\midrule
S-HMC  & 39.36 & 3.083 & 1.571 & 3.612 & 0.047 & 0.077 \\
\bottomrule
\end{tabular}
\end{table}

\subsection{Overcoming other computational bottlenecks}

Our sampler relies only on forward/back‑prop evaluations, so every mainstream hardware scheme can be stacked on top of it: data‑parallel all‑reduce for moderate models\citep{goyal2017imagenet1h}; 
optimizer‑state sharding (ZeRO/FSDP) when 
parameters no longer fit\citep{rajbhandari2020zero}; 
tensor model‑parallelism for in‑layer splits \citep{shoeybi2019megatron} and pipeline model‑parallelism for depthwise splits \citep{huang2019gpipe}; 
and, finally, the full hybrid of DP/sharding/tensor/pipeline that is now routine in trillion‑parameter language models \citep{chowdhery2022palm,black2022gptneox}.

\subsection{Further directions}

Further directions at the methodological level include
\begin{itemize}
\item DE-SBMC: an obvious extension, 
would be to condition the HMC ensemble, or SMC ensemble ($P>1$) with the DE, in case there may be any gain to be had.
\item One could condition SGLD/SGHMC with the MAP(s) from SGD (or DE). 
In this way, there is an initial phase which aims to recover a good point estimator, 
and then a second phase of essentially the same method, which aims to quantify the spread.
\item $P-$parallelizing $N-$ensemble MCMC methods such as \cite{gilks1994adaptive,goodman2010ensemble,vrugt2009accelerating,hoffman2022tuning}.
\item Leveraging $N-$ensemble MCMC methods within SMC for better mutations (with the cost of more communication).
\item Parallel stochastic-gradient-MCMC methods like SGLD \citep{welling2011bayesian} and SG-HMC \citep{chen2014stochastic}, 
and ensemblized versions thereof.
\item Related to above, mini-batch gradients can be used in lieu of full 
gradients, which may have some advantages in terms of scalability and convergence.
For SMC samplers, we have unbiased estimators $\widehat{\ell w}$ of log weights using mini-batches, and could use $\exp(\widehat{\ell w})$ for a non-negative and biased estimator or
Bernoulli/Poisson augmentation to achieve (non-negative) unbiased weights \citep{gunawan2021subsamplingSMC,vihola2020poissonSMC}.
\end{itemize}

\section{More related work}
\label{app:related}

\textit{Empirical Bayes } methods \citep{efron1977stein}
fit higher level parameters in hierarchical models through 
optimization of the marginal likelihood. 
An SBMC model could be built in principle with a general prior 
$\pi_\phi(\theta)$, for example $\mathcal{N}(\theta; \mu, \Sigma)$ for $\phi=(\mu,\Sigma)$,
and solved by EB.
There is a significant cost overhead for optimizing 
the marginal-likelihood, but that could be offset in principle 
with Laplace approximation \cite{bishop2006pattern} or other approaches. 
The particular SBMC($s$) model considered here could 
also utilize EB for selecting $s$ and/or $\alpha$, as an alternative to cross-validation.
And even before this, EB could be used to define the prior variance $v$, or a more general prior.

The work LogME~\cite{you2021logme} use EB for fitting prior and likelihood variance
in the context of transfer learning for regression,
and then they extended this idea for building estimators 
from an ensemble of pre-trained models
\cite{you2022ranking}. The latter could naturally be combined with other ensemble approaches 
described above, and plugged into SBMC.
The \textit{Laplace Redux} work of 
~\cite{daxberger2021laplace} provides an off-the-shelf Laplace module with block diagonal
Hessian approximations to plug pre-trained models 
into for transfer learning. This could naturally augment SBMC in a number of ways,
from leveraging it in marginal likelihood calculations for EB, to
using it as a drop in alternative for the prior, or 
leveraging their Hessian approximation in various other ways.
It is worth momentarily digressing on an approach inspired by this observation.
If the data is split into $N_\alpha$ pre-training and $N-N_\alpha$ fine-tuning
sets, or if the likelihood is split by $\alpha$ and $1-\alpha$ scalar fractions, 
then one could build a Laplace approximation (or another variational approximation) 
of the original $\alpha$ posterior, 
and use that as a prior for the remaining likelihood fraction.
This posterior approximation may be closer to the original GS,
and since each problem features an explicitly tempered likelihood, they should both be easier solve.
This may help with potentially overfitting, although we did not observe much of an issue in that respect.

\section{A sketch of the theory}
\label{app:theory}

We can think about the SBMC model as an incremental incorporation of the data. So first let $\sigma^2=v/a$ and consider the original problem with prior $\mathcal{N}(0,\sigma^2)$. Now {\em split} the log-likelihood into $(1-a)\ell + a\ell$, and consider incorporating {\em only the $a\ell$ part}. It is easy to see that the MAP estimator $\hat{\theta}_{\sf MAP}$ for this problem is equivalent to the MAP estimator associated with the prior $\mathcal{N}(0,v)$. The {\em Laplace approximation} however, will differ, depending on which one we consider. Let us consider the $\mathcal{N}(0,\sigma^2)$ prior, and now it is time to incorporate the rest of the data $(1-a)\ell$. The Hessian of our Laplace approximation is 
$$
a \nabla^2 \ell(\theta) + \frac1{2(v/a)} {\sf Id}. 
$$
This could be carried through rigorously, but for the sake of the argument, 
let's suppose we wave our hands and swap out $N_{\rm train} {\sf Id}$ 
for $\nabla^2 \ell(\theta)=\sum_{i=1}^{N_{\rm train}} \nabla^2\ell_i(\theta)$ (at least it is of the right order). 
To get back to the SBMC prior we have to equate 
$$
(2vN_{\rm train} + 1)/(2v/a) = 1/(2sv) \quad \Rightarrow \quad 
a = s^{-1}/(2vN_{\rm train} + 1) \, .
$$
Suppose $v=0.1$ (common) and $s=0.1$ (our recommendation). Then $a = 10/(N_{\rm train}/5+1) \ll 1$, for large datasets. 
Therefore, $\ell \approx (1-a)\ell$, and we end up with a reasonable approximation. Note this is also typically a positive thing for the prior, which is effectively $\mathcal{N}(0,\sigma^2=v/a)$, since broader priors and less inductive bias typically deliver better performance, and small variance priors are often chosen more as a matter of convenience. 

There are also immediate opportunities for extension, for example using better approximations of the Laplace approximation, or working out a more careful analysis along these lines. 


Note that we can do things like this also to mitigate catastrophic forgetting in general when fine-tuning in a continual learning setting. Something similar was done for continual learning in \cite{duffieldscalable}. Whereas, we do this simply to improve the approximation quality for short chains. 

\paragraph{Hessian. }Note the Hessian of the posterior $\pi$ in \eqref{eq:posterior} with $\mathcal{N}(0,v)$ prior is 
$$
\nabla^2 \ell + \frac1v {\sf Id} \, . 
$$
If we assume that the minimum eigenvalue of $\nabla^2 \ell$ is $0$, and the maximum is $\lambda_{\sf max}$, 
then the {\em condition number} of the Hessian of the posterior is ${\lambda_{\sf max}v +1}$.
Meanwhile, the Hessian of the SBMC target $\overline{\pi}$ in \eqref{eq:sbmc} with $\mathcal{N}(\theta_{\sf MAP},vs)$ prior 
will have condition number ${\lambda_{\sf max}vs +1}$. 
Therefore, the parameter $s<1$ allows us to ``tune away" the ill-conditioning of the posterior.
See also the discussion in \cite{paulin2025sampling}.


\section{Techniques for SMC$_\parallel$ in practice}
\label{app:const_trick}

{\bf Adaptive tempering. } As mentioned, adaptive tempering is used to ensure a dense tempering regime and 
provide stability\citep{syed2024optimised}. 

\begin{example}[Adaptive tempering]
\label{ex:adaptive_temp}
    In order to keep the sufficient diversity of sample population, we let the effective sample size to be at least \(\text{ESS}_{\min} = N/2\) at each tempering \(\lambda_{j-1}\) and use it compute the next tempering \(\lambda_j\). For \(j\)th tempering, we have weight samples \(\{ w^{k}_{j-1}, \theta^{k}_{j-1} \}_{k=1}^{N}\), then the ESS is computed by
    \begin{equation*}
        \text{ESS} = \frac{1}{\sum_{k=1}^{N} (w^{k}_{j-1})^2},
    \end{equation*}
    where \(w^{k}_{j-1} = \cL(\theta^{k}_{j-1})^{\lambda_{j} - \lambda_{j-1}}/\sum_{k=1}^N \cL(\theta_{j-1}^k)^{\lambda_j-\lambda_{j-1}}\). Let \(h = \lambda_j - \lambda_{j-1}\), the effective sample size can be presented as a function of \(h\), ESS\((h)\). Using suitable root finding method, one can find \(h^{*}\) such that \(\text{ESS}(h^{*})=\text{ESS}_{\min}\), then set the next tempering \(\lambda_j = \lambda_{j-1}+h^{*}\).
\end{example}

{Note that the partition function estimator \(Z^N\) is no longer unbiased
once we introduce adaptation, which means that in principle we should do short pilot runs and 
then keep everything fixed to preserve the integrity of the theory, 
but we have found this does not make a difference in practice.

{\bf Adaptive number of mutation steps. } 
The number of mutation steps $M$ is chosen adaptively. After resampling at a given tempering step, let $\theta^{i,0}$ denote the $i$-th sample and $\theta^{i,m}$ its state after $m$ mutation steps. We monitor the mean displacement from the post-resampling state,
\begin{equation*}
\text{dist}_m \;=\; \frac{1}{N}\sum_{i=1}^N \bigl\lVert \theta^{i,m}-\theta^{i,0} \bigr\rVert_2,
\end{equation*}
and terminate the mutation update at the smallest \(M \ge 2\) for which the displacement has stabilized:
\begin{equation*}
\frac{\lvert \text{dist}_M - \text{dist}_{M-1} \rvert}{\text{dist}_{M-1}} \;\le\; \eta,
\end{equation*}
with tolerance $\eta>0$. This criterion automatically increases $M$ when the tempering increment is large or the target becomes tighter (requiring more mixing to decorrelate the resampled particles), and conversely saves computation when the resampled state is already close to stationary at the new tempering level.

{\bf Numerical stability: nested Log-sum-exp. } When computing likelihoods in Sequential Monte Carlo (SMC) algorithms, numerical underflow frequently arises because likelihood values can become extremely small, often beyond computational precision. To address this, one standard practice is to work with log-likelihoods rather than likelihoods directly. By operating in the log domain, the computer can safely store and manipulate extremely small values without loss of precision.

Specifically, the standard \emph{log-sum-exp} trick can be applied to stabilize computations. For instance, consider a scenario with nested sums and products in parallel SMC. For each processor \( p = 1, \dots, P \), we initially have:
\[
Z^{N,p} = \prod_{j=1}^{J}\sum_{i=1}^{N}\omega_{j}^{i,p}.
\]

To avoid numerical instability, each sum within the product is computed using the log-sum-exp trick:
\[
\sum_{i=1}^{N}\omega_{j}^{i,p} = \exp\left(\max_{i}\log(w_j^{i,p})\right)\sum_{i=1}^{N}\exp\left(\log(w_j^{i,p}) - \max_{i}\log(w_j^{i,p})\right).
\]

This procedure yields the decomposition:
\[
Z^{N,p} = K^p \hat{Z}^p,
\]
where
\[
K^p = \prod_{j=1}^{J}\exp\left(\max_{i}\log(w_j^{i,p})\right), \quad \text{and} \quad \hat{Z}^p = \prod_{j=1}^{J}\sum_{i=1}^{N}\exp\left(\log(w_j^{i,p}) - \max_{i}\log(w_j^{i,p})\right).
\]

In parallel SMC, an additional stabilization step is applied across processors. The global normalization constant across processors can also suffer from numerical instability. To address this, the log-sum-exp trick is applied again at the processor level:
\[
Z^{N,p} = \exp\left(\log(\hat{Z}^p) + \log(K^p) - \log(K)\right)K,
\]
with
\[
\log(K) = \max_{p}\left(\log(\hat{Z}^p) + \log(K^p)\right).
\]

Since the factor \( K \) cancels out when calculating the parallel SMC estimator, it suffices to compute only:
\[
\exp\left(\log(\hat{Z}^p) + \log(K^p) - \log(K)\right),
\]
which ensures numerical stability even when \(K\) itself is computationally very small.

Thus, by recursively applying the log-sum-exp trick at both the particle and processor levels, parallel SMC estimators can robustly handle computations involving extremely small numbers without numerical underflow.

\section{Complementary description of simulations}
\label{app:simos}

\subsection{Computation of Error bars}
\label{app:errorbar}

Assume running $R$ times of experiments to get $R$ square errors/loss between simulated estimator $\hat{\varphi}$ and the ground truth, SE$(\hat{\varphi})^{r}$ for $r=1,...,R$. Take the MSE as an example, the MSE is the mean of SE$(\hat{\varphi})^{r}$ over $R$ realizations, and the standard error of MSE (s.e.) is computed by
\begin{equation}
    \frac{\sqrt{\frac{1}{R}\sum_{r=1}^{R}(\text{SE}(\hat{\varphi})^{r}-\text{MSE})^2}}{\sqrt{R}}.
\end{equation}




\subsection{Integrated Autocorrelation Time}
\label{app:IACT}

Integrated Autocorrelation Time (IACT) means 
the time until the chain is uncorrelated with its initial condition.
The precise mathematical definition is as follows.

Let $\theta_0, \dots, \theta_t, \dots$ denote the Markov chain, and let $\varphi(\theta)$ be a scalar function of the state. 
We first define the \emph{autocovariance function} (ACF) at lag $s$:
\[
\gamma_s(\varphi) 
= \mathbb{E}\!\left[ \big(\varphi(\theta_{t+s}) - \mathbb{E}[\varphi(\theta)]\big)
                     \big(\varphi(\theta_t) - \mathbb{E}[\varphi(\theta)]\big) \right],
\]
and the ACF at lag $s$ as the normalized quantity
\[
\rho_s(\varphi) = \frac{\gamma_s(\varphi)}{\gamma_0(\varphi)},
\]
where $\gamma_0(\varphi)$ is the variance of $\varphi(\theta)$.

Then the \emph{integrated autocorrelation time} (IACT) of $\varphi$ is then defined in terms of the ACF by
\[
\text{IACT}(\varphi) = 1 + 2 \sum_{s=1}^{\infty} \rho_s(\varphi).
\]


\subsection{Details of the Bayesian Neural Networks}
\label{app:BNN}

Let weights be \(A_{i} \in \bbR^{n_i \times n_{i-1}}\) and biases be \(b_i \in \bbR^{n_i}\) for \(i \in \{1,...,D\}\), we denote 
\(\theta := ((A_1,b_1),...,(A_D,b_D))\). The layer is defined by
\begin{equation*}
\begin{aligned}
    g_1(x,\theta) & := A_1 x + b_1, \\
    g_d(x,\theta) & := A_i \sigma_{n_{i-1}}(g_{i-1}(x)) + b_{i}, \ \ i \in \{2,...,D-1\}, \\
    g(x,\theta) & := A_{D}\sigma_{n_{D-1}}(g_{D-1}(x)) + b_{D},
\end{aligned}
\end{equation*}
where \(\sigma_{i}(u) := (\nu(u_1),...,\nu(u_i))^{T}\) with ReLU activation \(\nu(u) = \max \{0,u\}\).

Consider the discrete data set in a classification problem, we have \(\sY = \{1,...,K\}\) and \(n_D=K\), then we instead define the so-called \textit{softmax} function as
\begin{equation}\label{eqn:bayes_class}
    h_k(x,\theta) = \frac{\exp(g_{k}(x,\theta))}{\sum_{j=1}^{K}\exp(g_{j}(x,\theta))}, \ k \in \sY,
\end{equation}
and define \(h(x,\theta) = (h_1(x,\theta),...,h_K(x,\theta))\) as a categorical distribution on \(K\) outcomes based on data \(x\). Then we assume that \(y_i \sim h(x_i)\) for \(i=\{1,...,m\}\).

Now we describe the various neural network architectures we use for the various datasets.

\subsubsection{MNIST7 Classification Example}
\label{app:mnist}

The architecture is a simple CNN with 
(i) one hidden layer with 
$4$ channels of $3\times3$ kernels with unit stride and padding,
followed by (ii) ReLU activation and 
(iii) $2\times2$ max pooling, (iv) a linear layer,
and (v) a softmax. 
The parameter prior and dataset is built as follows
    \begin{itemize}
        \item Training is conducted on a sub-dataset consisting of the first \(1200\) training samples with labels \(0\) through \(7\). Evaluation is performed on first \(N_{\sf id}\) in-domain test images with labels \(0\) through \(7\) and the on the four generated out-of-domain dataset (\(N_{\sf ood}\) total number of data). 
        \item The OOD dataset is generated as follows: two of the datasets are the first \(N_{\sf ood}/4\) out-of-domain test images with labels \(8\) and \(9\), respectively. The third dataset, the white noise image (wn), is a set of \(N_{\sf ood}/4\) synthetic $28\times28$ “images” with pixels drawn uniformly at random from $[0,1]$. The fourth dataset, the perturbed image (per.), is a set of the first \(N_{\sf ood}/4\) MNIST test images of digits0–7, each pixel perturbed by Gaussian noise (standard error as \(0.5\)) while retaining its original label. 
        \item MAP and DE are estimated using an initialization and regularization based on the prior \(N(0, v {\sf Id}) \), 
        where \(d=6320\) and \(v = 0.1\). The tuning parameter in SBMC methods is \(s\). The batchsize is $64$.
        \item The gold-standard is computed by the single HMC over $5$ realizations, called HMC (GS), with \(N=B\), \(T=1\) and \(L=1\).
        \item {SWA. We train with SGD (\texttt{momentum = 0.9}) for a 25-epoch warm-up, 
        then perform SWA weight averaging with 1 sample per epoch,
        at a fixed \texttt{swa\_lr = 0.0005}.}
        \item {MC Dropout. A fixed 30\% dropout in the fully-connected layer; $10$ samples are used.}
        \item {Laplace. We fit a Laplace approximation with a Kronecker-factored approximation of the Hessian\footnote{\texttt{https://github.com/aleximmer/Laplace}}\cite{daxberger2021laplace}; 
        $10$ samples are used.} 
    \end{itemize}

\subsubsection{IMDb Classification Example}
\label{app:imdb}

Here we use SBERT embeddings \cite{reimers-2019-sentence-bert} based on the model 
{\texttt all-mpnet-base-v2} \cite{song2020mpnet}
\footnote{https://huggingface.co/sentence-transformers/all-mpnet-base-v2}.
In other words, frozen weights from {\texttt all-mpnet-base-v2}
until the $768$ dimensional [CLS] output.
The NN model and parameter prior for IMDb\footnote{https://huggingface.co/datasets/stanfordnlp/imdb} are built as follows
    \begin{itemize}
        \item NN is followed by (i) no hidden layer, (ii) ReLU activation, (iii) a final linear layer, and (iv) softmax output.
        \item Training is conducted on the whole train set ($25000$ data). Evaluation is performed on the whole test images as the in-domain dataset ($25000$ data) and on the four generated out-of-domain datasets ($N_{\sf ood}$ total number of data).
        \item The OOD dataset is generated as follows: four of these datasets (each dataset has $N_{\sf ood}/5$ data) use textual data from the Appliances domain, which is distinct from the in-domain IMDb movie review data. Specifically, four OOD datasets were constructed from Amazon Reviews 2023 Appliances data \cite{hou2024bridging} \footnote{https://amazon-reviews-2023.github.io/}, containing customer reviews and product metadata. Two datasets directly used the two JSON files, and two text-based OOD datasets were generated as follows. From \texttt{Appliances.jsonl}, we extracted the review text, representing natural language expressions of user opinions but unrelated to movies; from \texttt{meta\_Appliances.jsonl}, we constructed meta descriptions by concatenating each product's title and listed features. The last dataset, Lipsum, is a collection of 100 very short, meaningless text strings, each consisting of between one and ten randomly selected words drawn from the classic “Lorem ipsum” filler vocabulary.
        \item MAP and DE are estimated using an initialization and regularization based on the prior \(N(0, v {\sf Id}) \), where \(d=1538\). The tuning parameter in SBMC methods is \(s\). The batchsize is $64$.
        \item The gold-standard is computed by the single HMC over $5$ realizations, called HMC (GS), with \(N=B\), \(T=1\) and \(L=1\).
    \end{itemize}

\subsubsection{CIFAR-10 Classification Example}
\label{app:cifar}

Here, the architecture is ResNet-50 pre-trained from ImageNet with 
all parameters frozen until the final pooled $2048$ dimensional features.
The NN model and parameter prior 
for CIFAR10 are as follows.
    \begin{itemize}
        \item NN is followed by (i) no hidden layer, (ii) ReLU activations, (iii) a final linear layer, and (iv) softmax output. 
        \item Training is conducted on the whole train set ($50000$ data). Evaluation is performed on the whole test images as the in-domain dataset ($10000$ data) and on the three generated out-of-domain datasets ($N_{\sf ood}$ total number of data). 
        \item The OOD dataset is generated as follows: 
        \begin{itemize}
            \item Close OOD (CIFAR-100 “not in CIFAR-10”). Drawn $N_{\sf ood}/3$ data from the $90$ fine‐grained CIFAR-100 classes that don’t overlap with the 10 classes inCIFAR-10. All images are \(32\times32\) RGB natural photographs with nearly identical color distribution and textures to CIFAR-10.
            \item Corrupt OOD (CIFAR-10-C). Select $N_{\sf ood}/3$ CIFAR-10 test images and subject them to 15 types of realistic distortions—Gaussian/impulse noise (motion/defocus blur, frost, fog, brightness/contrast shifts, JPEG compression, pixelation, etc.) at five different severity levels. The pixel-level statistics are methodically disturbed, yet the original labels stay the same. 
            \item Far OOD (SVHN). Select $N_{\sf ood}/3$ data from $26032$ 32x32 RGB test photos of house-number digits ($0$–$9$) that have been cut from Google Street View. The SVHN displays centred white numbers on colourful, frequently cluttered urban backgrounds, in contrast to CIFAR's multi-object array of natural-scene photos. 
        \end{itemize}
        \item MAP and DE are estimated using an initialization and regularization based on the prior \(N(0, v {\sf Id}) \), where \(d=20490\) and \(v=0.2\). The tuning parameter in SBMC methods is \(s\). The batchsize is $128$.
        \item The gold-standard is computed by the single HMC over $5$ realizations, called HMC (GS), with \(N=B\), \(T=1\) and \(L=1\).
    \end{itemize}

\subsection{Hardware description}

The main CPU cluster we access 
has nodes with 2 $\times$ 16-core Intel Skylake Gold 6130 CPU @ 2.10GHz, 192GB RAM
{\em without communication} in between,
so it can only run $N/P=32$ particles in parallel
with one particle per core. 
There are also unconnected AMD “Genoa” compute nodes, 
with 2 $\times$ 84-core AMD EPYC 9634 CPUs and  1.5TB RAM.

\section{Further results for UQ}
\label{app:uq}

Results in this section further support the statement mentioned in the main text, that is, (i) SBMC significantly outperforms the MAP estimator, as well as a DE of MAP estimators, (ii) DE systematically underestimates $H_{\sf ep}$ for the same ensemble size as SBMC.

\subsection{MNIST7}
\label{app:mnist_uq}

In the MNIST7 case, the full setting is described in Appendix \ref{app:mnist}, where we let \(N_{\sf id}=7000\) and $N_{\sf ood}=2000$, where each dataset has $500$ data. Selected results appear in the main text in Figure \ref{fig:entropyAll}, where the full data table is given in Table \ref{tab:metric_diffP_mnist_s01}.
Additional detailed results of the per‐digit analysis are provided below, see Figure \ref{fig:full_digits_mnist}, and the full data table in Table \ref{tab:full_digits_mnist}.


\begin{figure}[H]
    \centering
    \begin{subfigure}[b]{0.32\textwidth}
    \includegraphics[width=\linewidth]{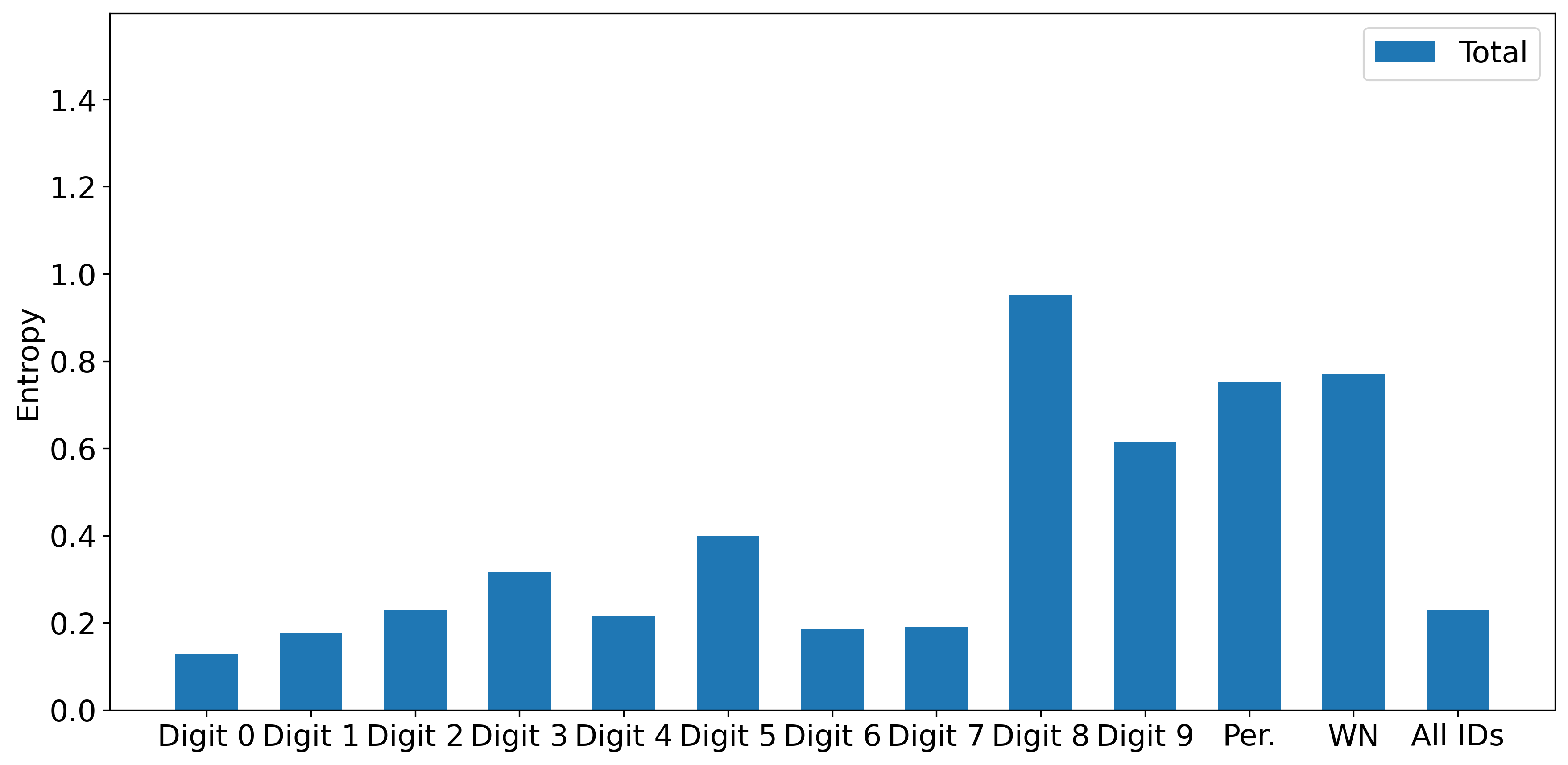}
    \caption{MAP}
  \end{subfigure}
  \begin{subfigure}[b]{0.32\textwidth}
    \includegraphics[width=\linewidth]{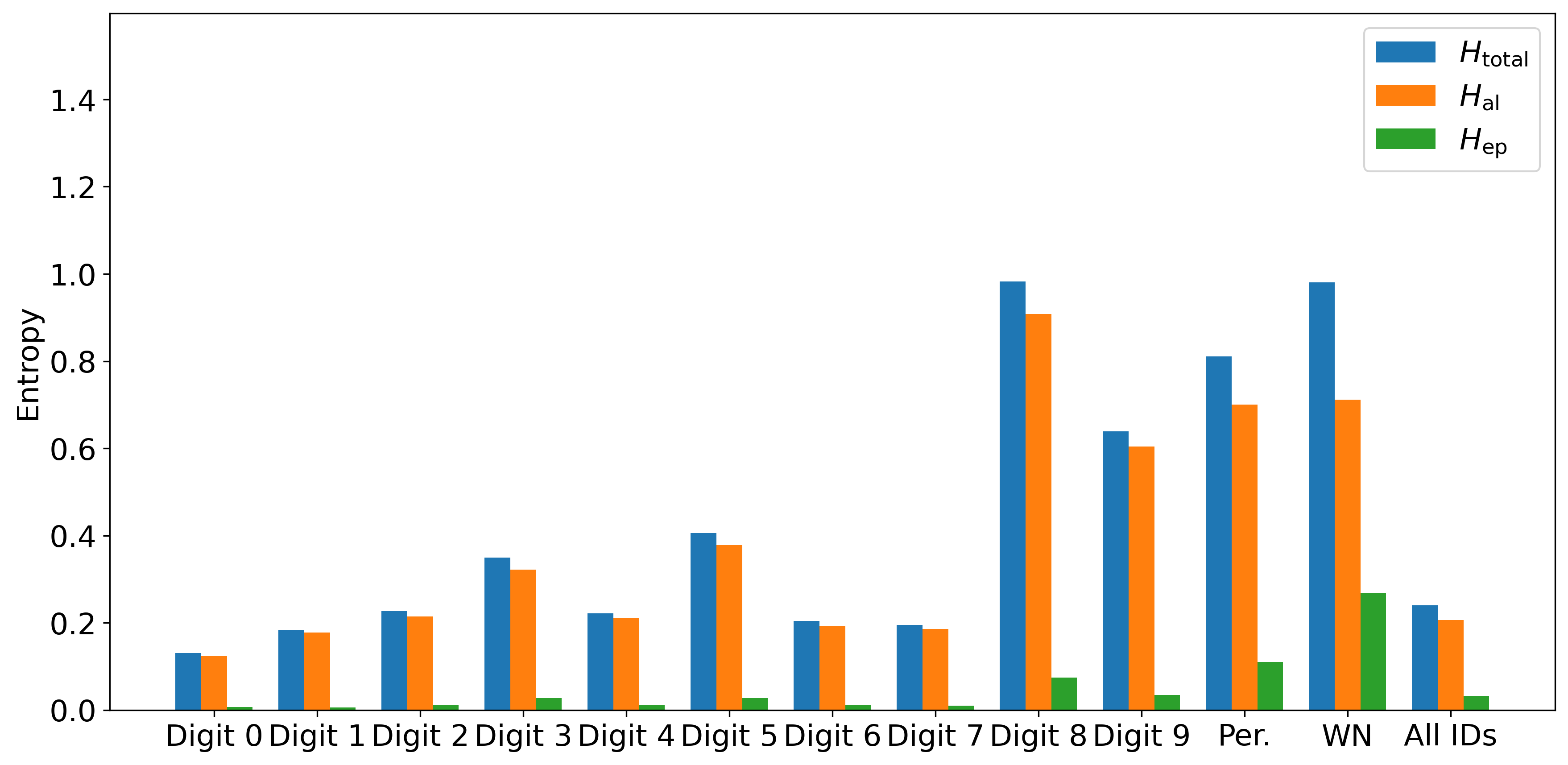}
    \caption{DE}
  \end{subfigure}
    \hfill
    \begin{subfigure}[b]{0.32\textwidth}
    \includegraphics[width=\linewidth]{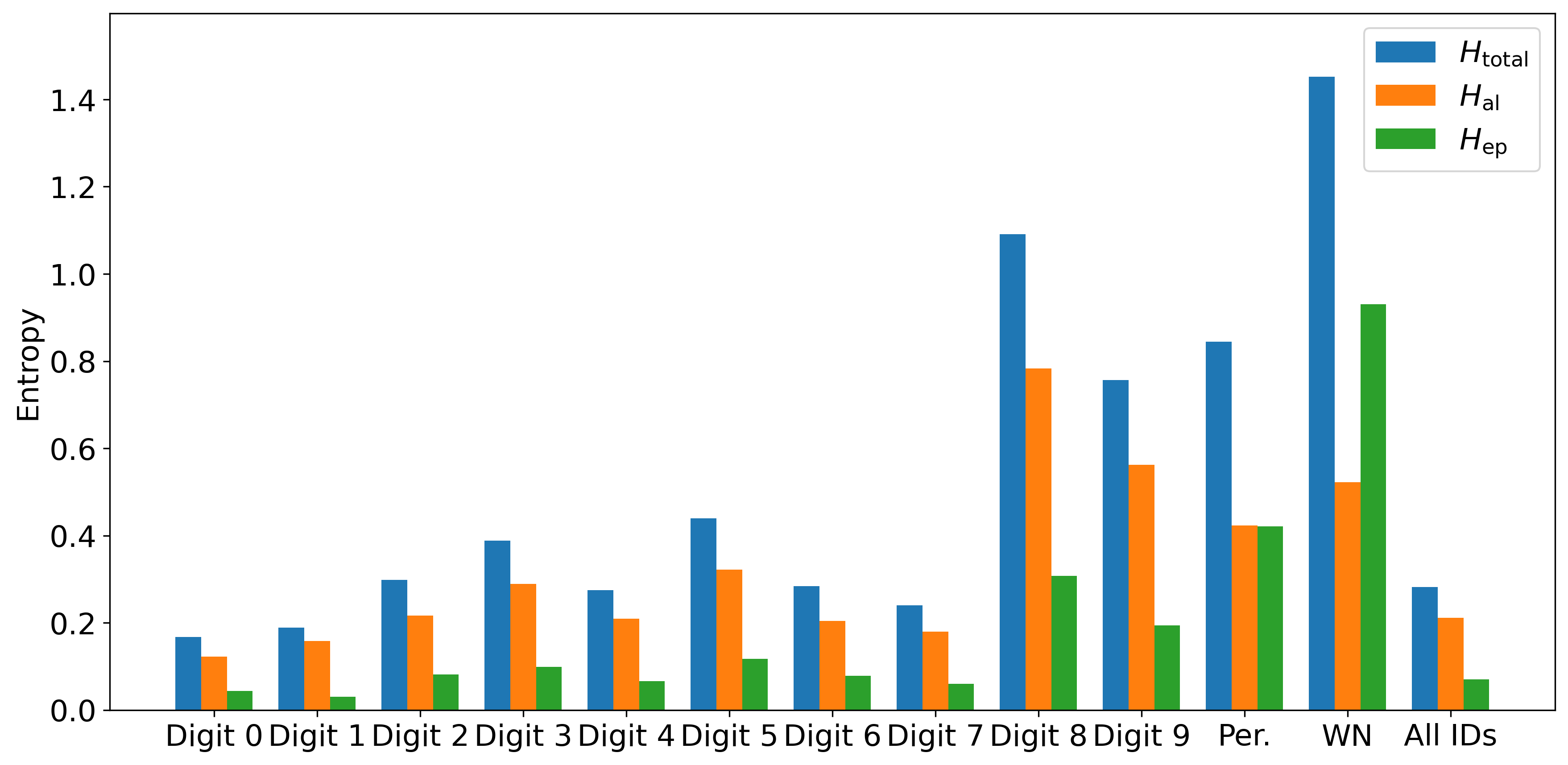}
    \caption{S-HMC\(_\parallel\)}
  \end{subfigure}
  \begin{subfigure}[b]{0.32\textwidth}
    \includegraphics[width=\linewidth]{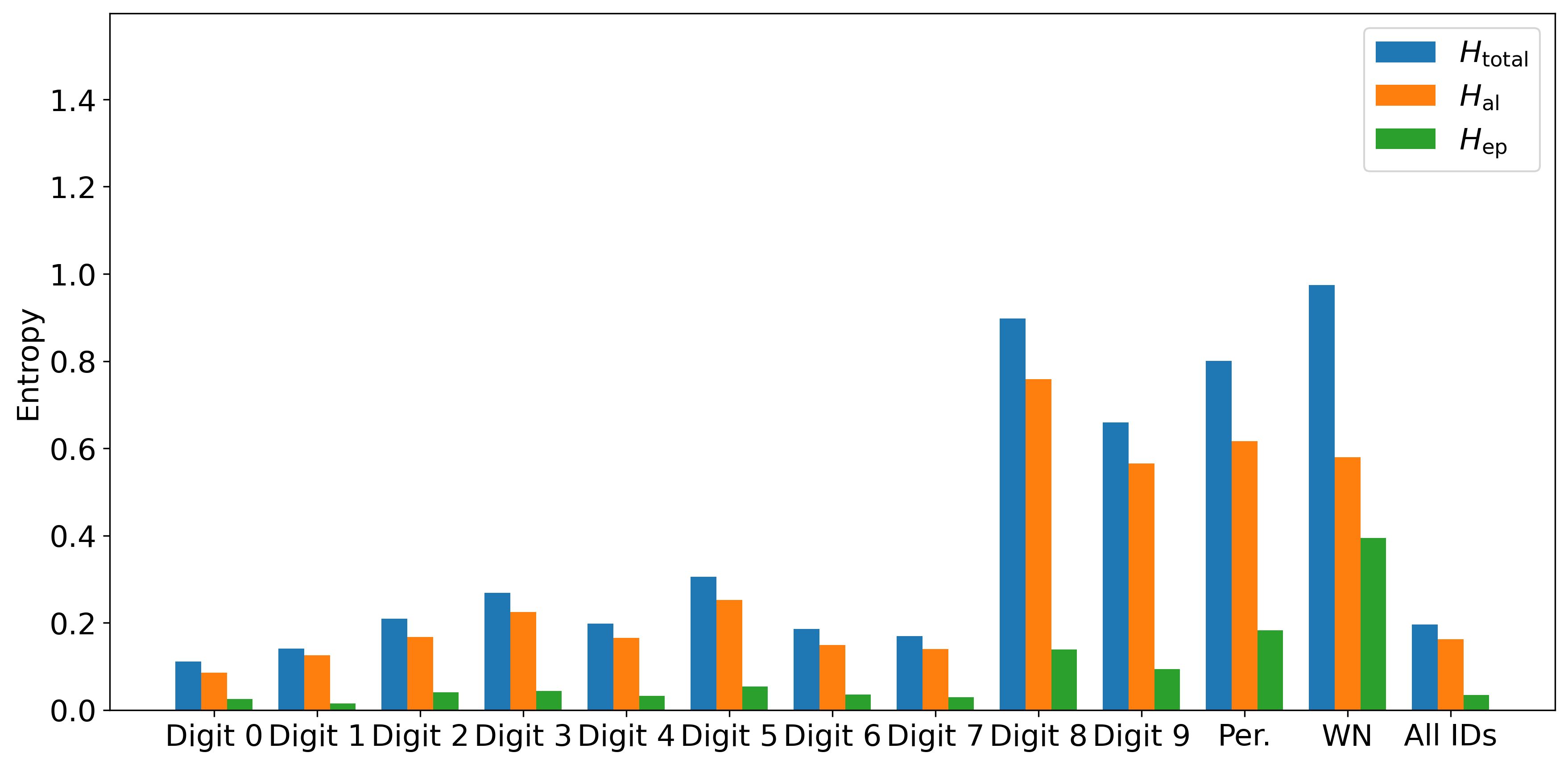}
    \caption{S-SMC\(_\parallel\)}
  \end{subfigure}
  \begin{subfigure}[b]{0.32\textwidth}
    \includegraphics[width=\linewidth]{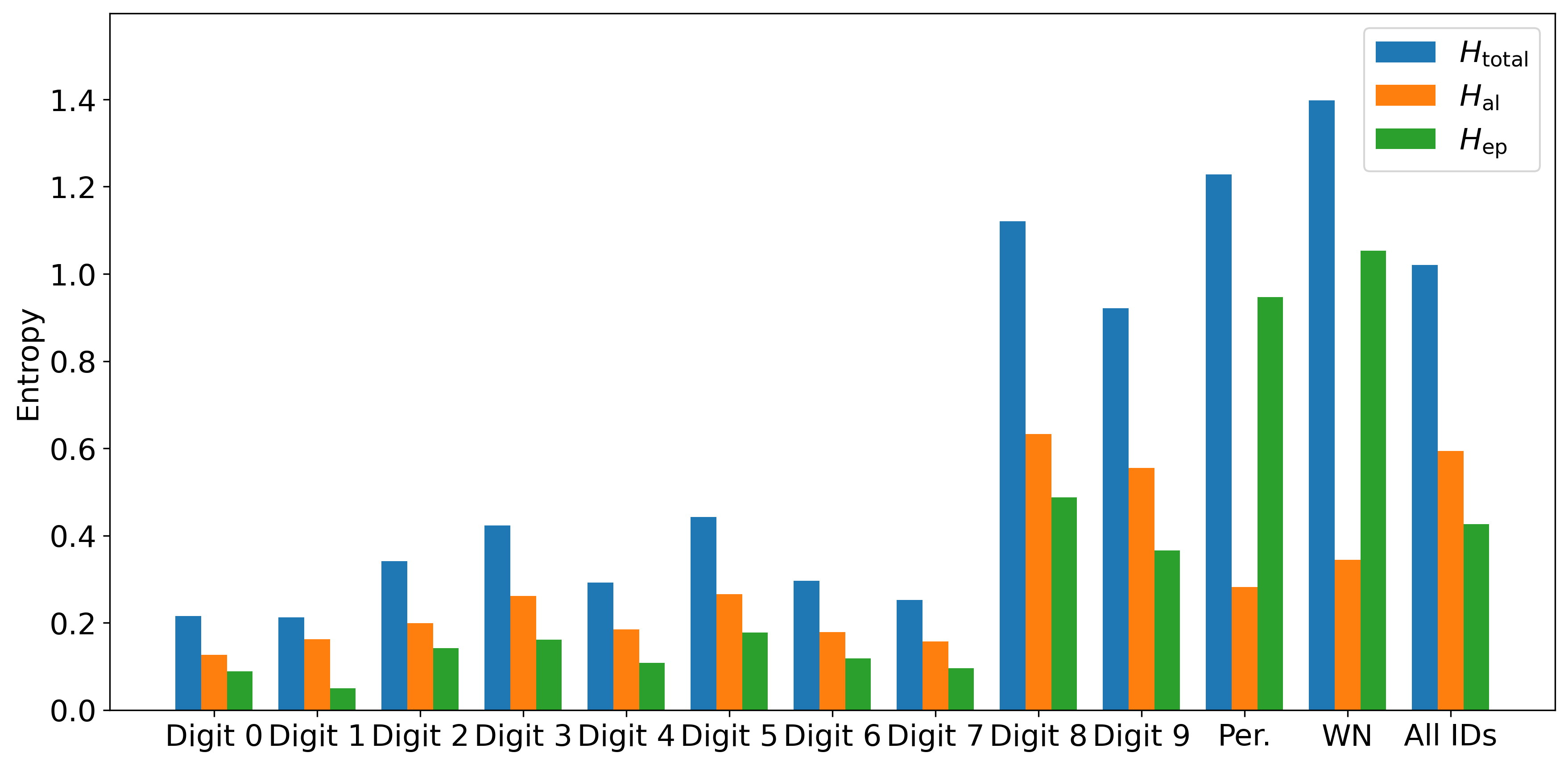}
    \caption{HMC (GS)}
  \end{subfigure}
    \caption{Comparison of entropy across groups for MNIST7. S-SMC$_\parallel$ ($P=1$ chain with $N=10$), S-HMC$_{\parallel}$ ($NP$ chains), HMC (GS) ($2e4$ samples), DE ($N$ models) and MAP, with fixed number of leapfrog $L=1$, $v=0.1$ and $s=0.1$ (5 realizations).}
\label{fig:full_digits_mnist}
\end{figure}

\subsection{IMDb}
\label{app:imdb_uq}

In the IMDb case, the full setting is described in Appendix \ref{app:imdb}, where we let $N_{\sf ood}=500$, and each dataset has $100$ data. 

\paragraph{Experiments with $v=\frac{1}{40}$.} Results of entropy comparison among MAP, DE and SBMCs are given in Figure \ref{fig:imdb_entropyAll}, showing comparison in the OOD datasets and the correct/incorrect predictions in the ID domain. Additional detailed results of the per‐digit analysis are provided below, see Figure \ref{fig:imdb_fullUQ}, and the full data table in Table \ref{tab:imdb_fullUQ}.

\begin{figure}[H]
    \centering
    \includegraphics[width=.8\linewidth]{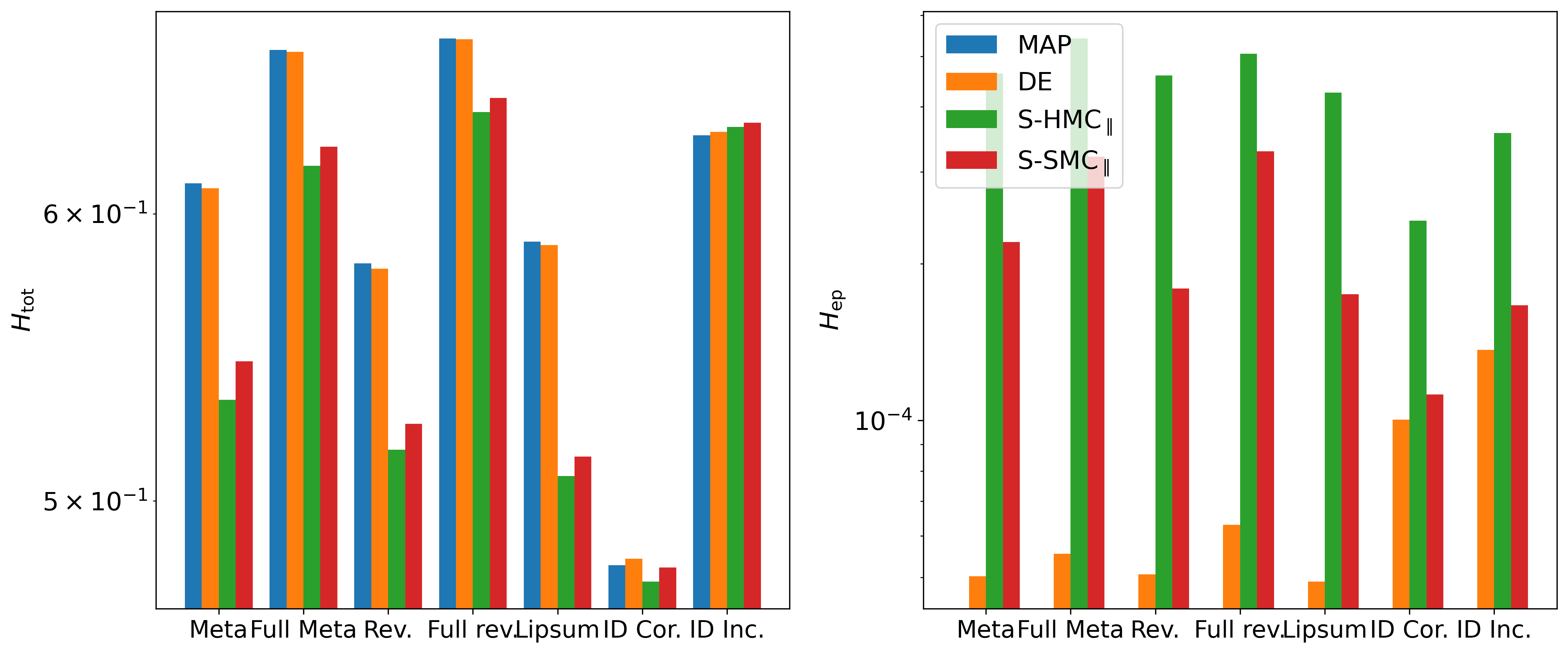}
    \caption{Comparison of average total and epistemic entropy over four out-of-domain classes and correct/incorrect predictions in-domain for IMDb. S-SMC\(_\parallel\) (\(P=1\) chain with \(N=10\)), S-HMC\(_{\parallel}\) (\(NP\) chains), DE (\(N\) models) and MAP, with fixed number of leapfrog \(L=1\), \(B=25\), \(M=1\), \(v=0.025\) and \(s=0.1\) (\(5\) realizations).}
\label{fig:imdb_entropyAll}
\end{figure}

\begin{figure}[H]
    \centering
    \begin{subfigure}[b]{0.24\textwidth}
    \includegraphics[width=\linewidth]{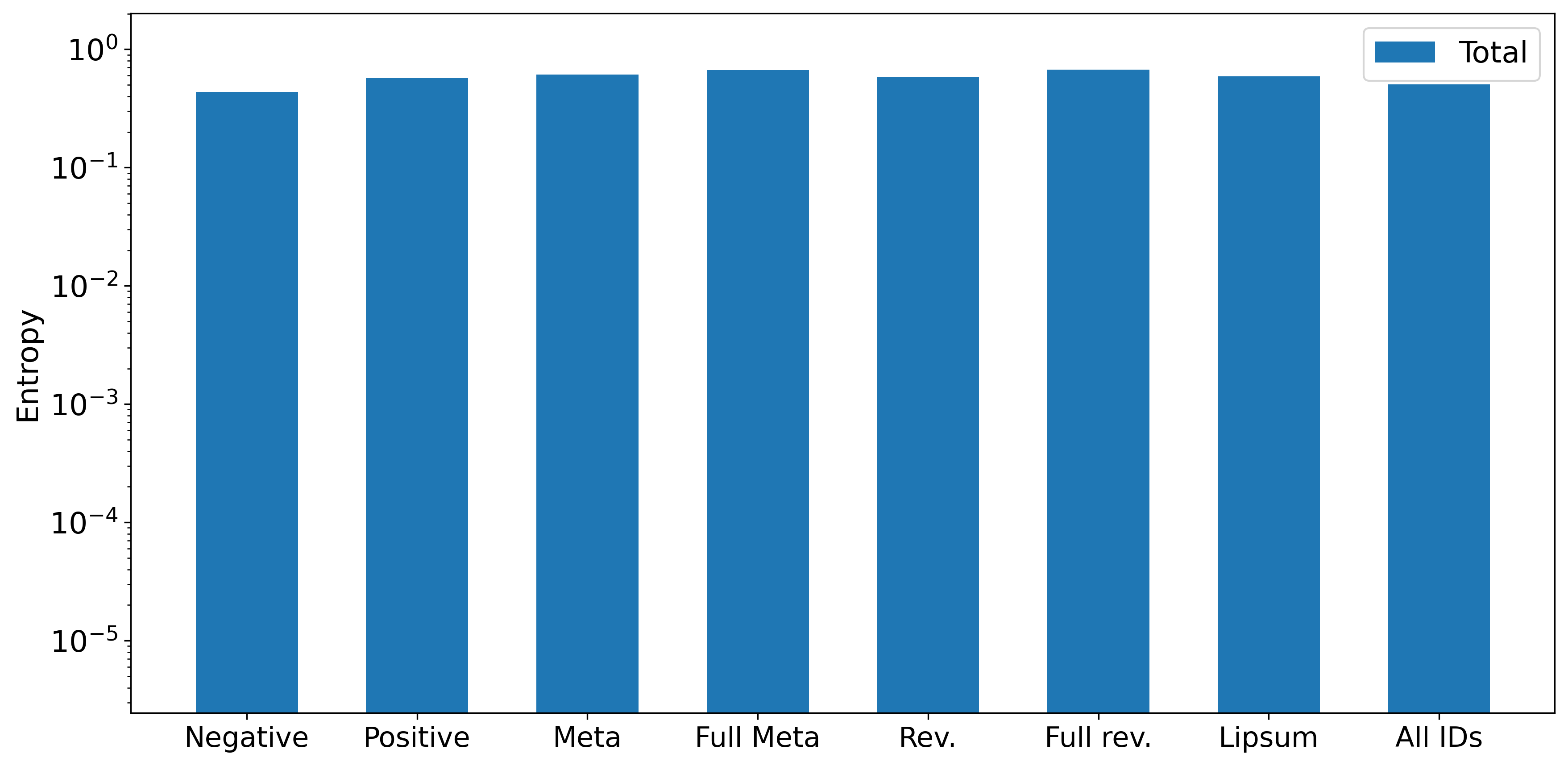}
    \caption{MAP}
  \end{subfigure}
  \begin{subfigure}[b]{0.24\textwidth}
    \includegraphics[width=\linewidth]{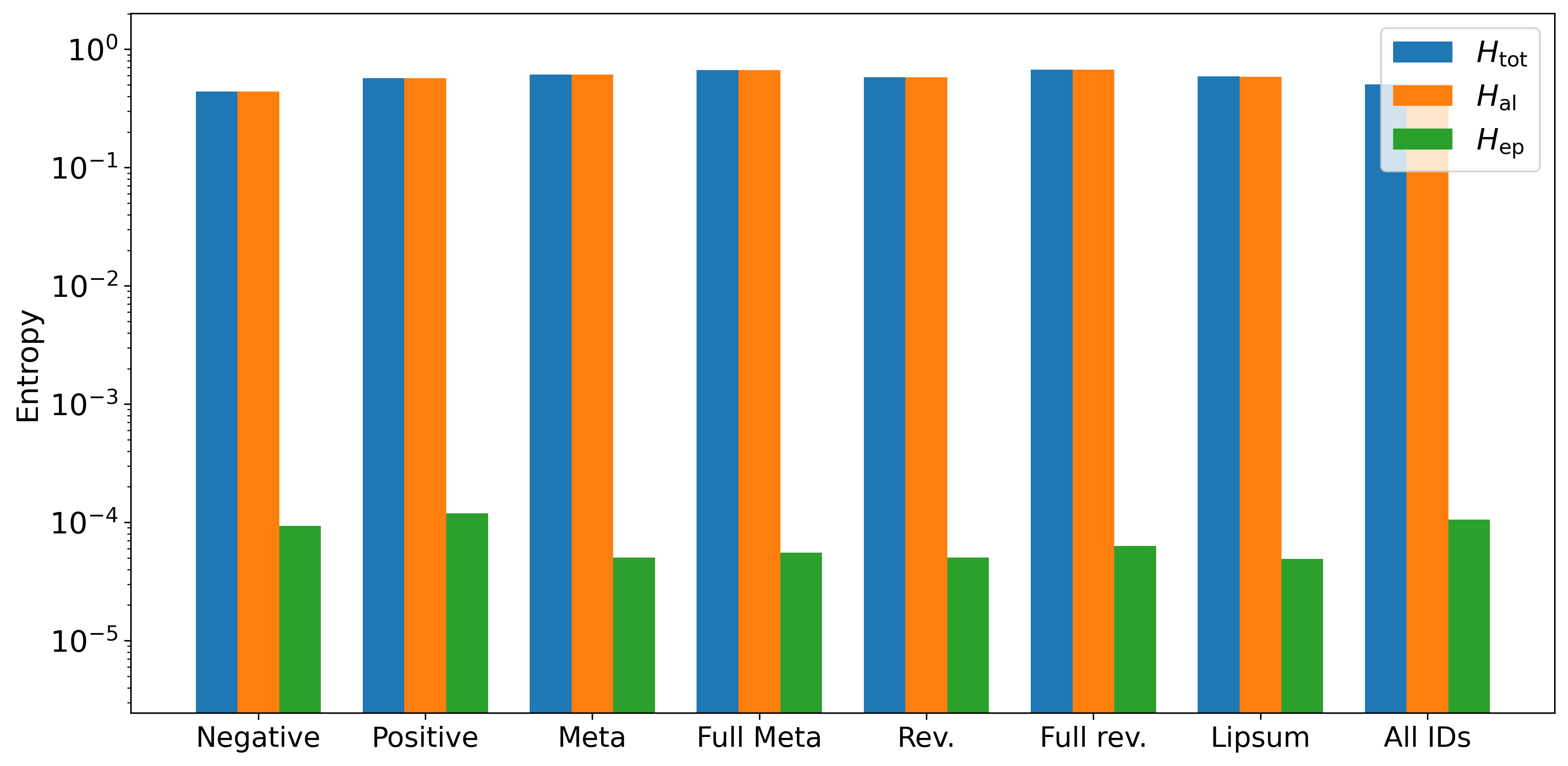}
    \caption{DE}
  \end{subfigure}
  \begin{subfigure}[b]{0.24\textwidth}
    \includegraphics[width=\linewidth]{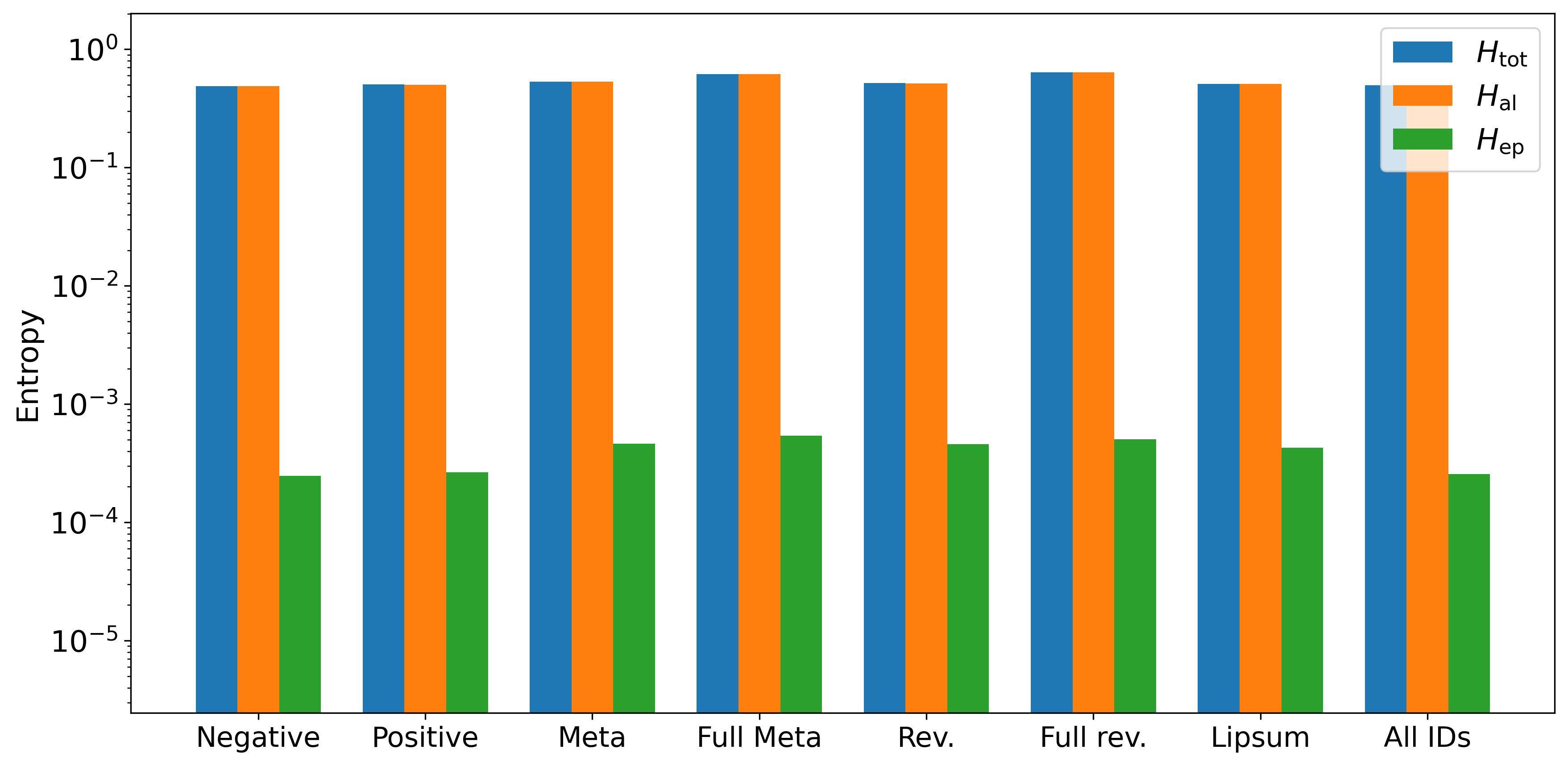}
    \caption{S-HMC\(_\parallel\)}
  \end{subfigure}
  \begin{subfigure}[b]{0.24\textwidth}
    \includegraphics[width=\linewidth]{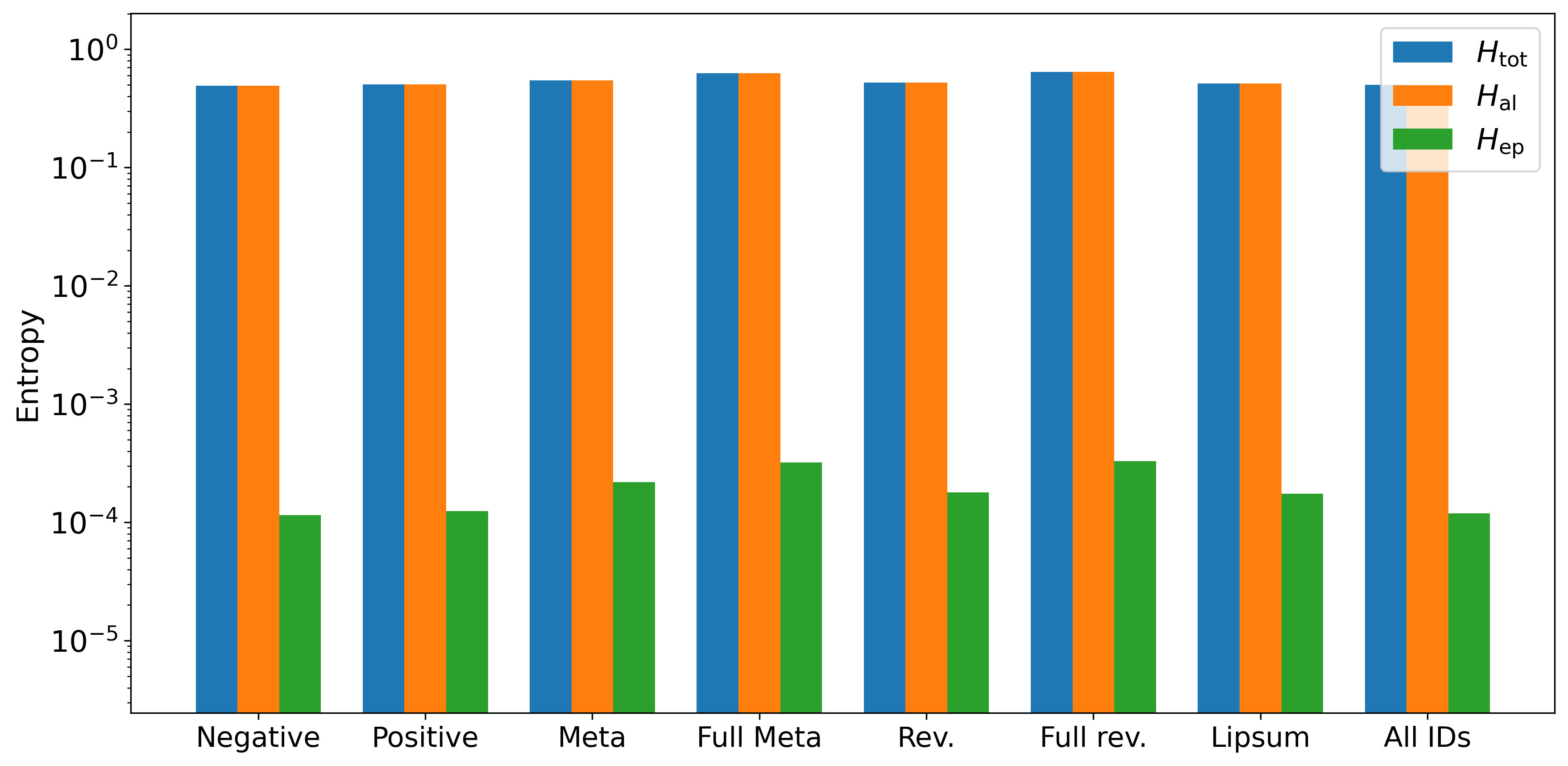}
    \caption{S-SMC\(_\parallel\)}
  \end{subfigure}
    \caption{Comparison of entropy across groups for IMDb. S-SMC\(_\parallel\) (\(P=1\) chain with \(N=10\)), S-HMC\(_{\parallel}\) (\(NP\) chains), DE (\(N\) models) and MAP, with fixed number of leapfrog \(L=1\), \(B=25\), \(M=1\), \(v=0.025\) and \(s=0.1\) (\(5\) realizations).}
\label{fig:imdb_fullUQ}
\end{figure}

\paragraph{Experiments with $v=1$.} Results of entropy comparison among MAP, DE and SBMCs are given in Figure \ref{fig:imdb_entropyAll_v1}, showing comparison in the OOD datasets and the correct/incorrect predictions in the ID domain. Additional detailed results of the per‐digit analysis are provided below, see Figure \ref{fig:imdb_fullUQ_v1}, and the full data table in Table \ref{tab:imdb_fullUQ_v1}.

\begin{figure}[H]
    \centering
    \includegraphics[width=.8\linewidth]{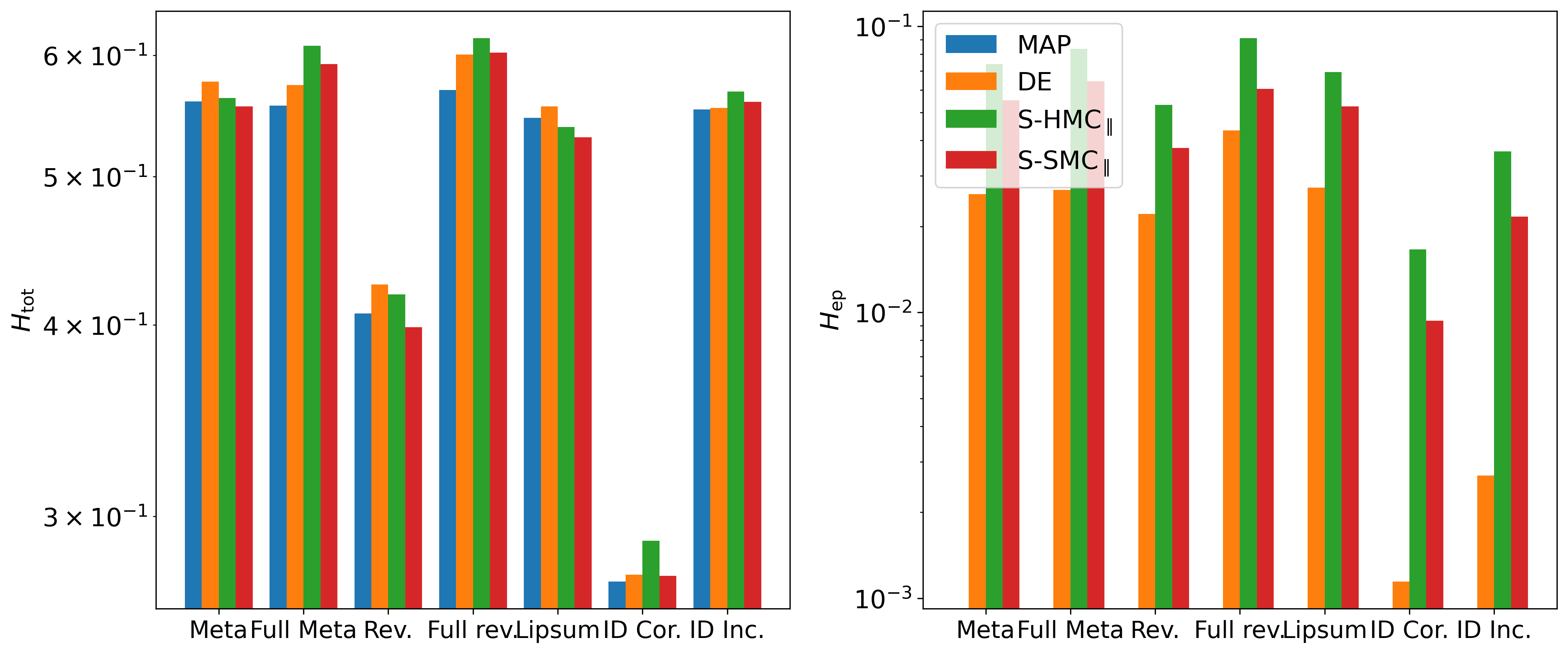}
    \caption{Comparison of average total and epistemic entropy over four out-of-domain classes and correct/incorrect predictions in-domain for IMDb. S-SMC\(_\parallel\) (\(P=8\) chain with \(N=10\)), S-HMC\(_{\parallel}\) (\(NP\) chains), DE (\(N\) models) and MAP, with fixed number of leapfrog \(L=1\), \(B=26\), \(M=2\), \(v=1\) and \(s=0.35\) (\(5\) realizations).}
\label{fig:imdb_entropyAll_v1}
\end{figure}

\begin{figure}[H]
    \centering
    \begin{subfigure}[b]{0.24\textwidth}
    \includegraphics[width=\linewidth]{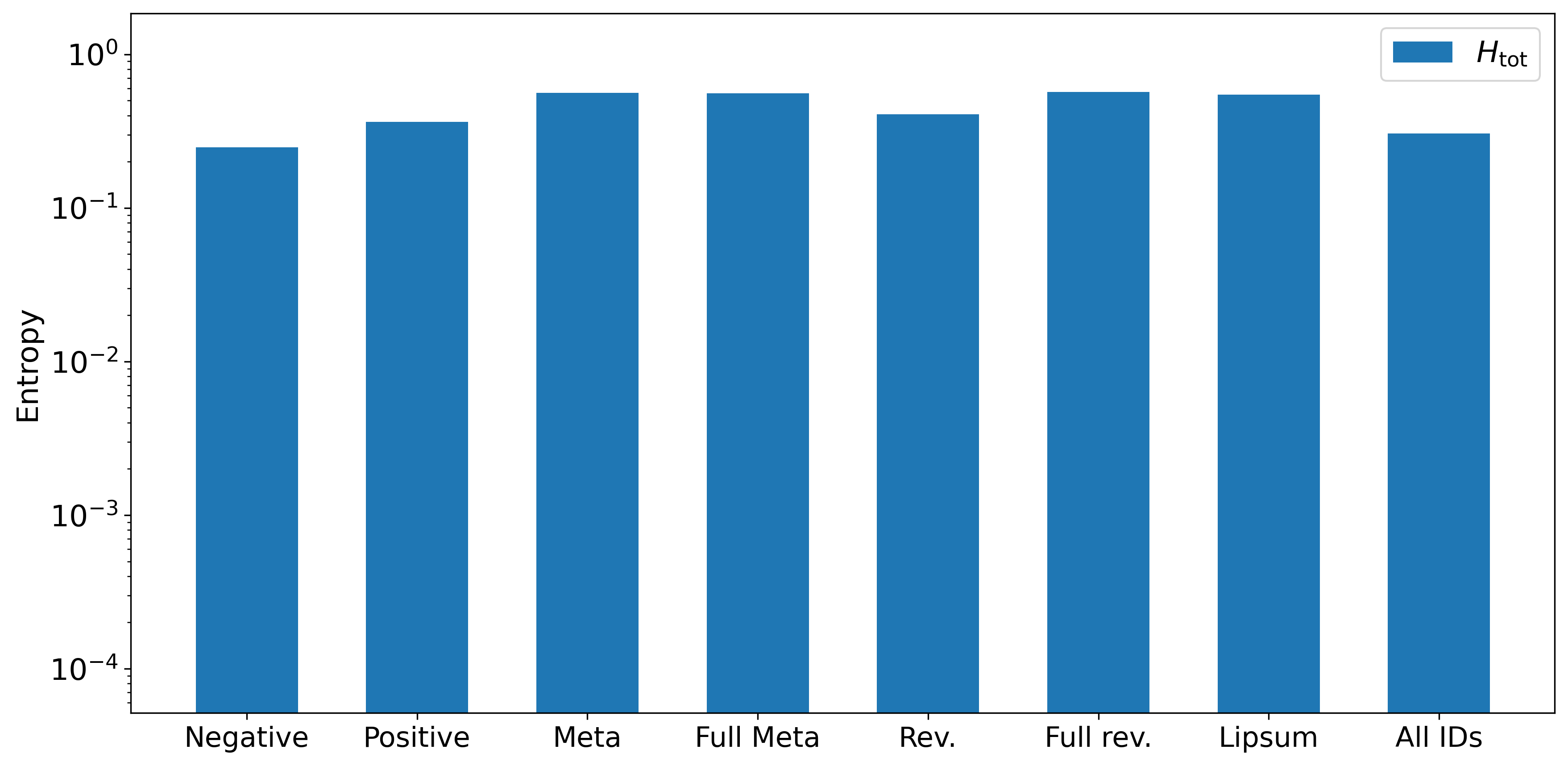}
    \caption{MAP}
  \end{subfigure}
  \begin{subfigure}[b]{0.24\textwidth}
    \includegraphics[width=\linewidth]{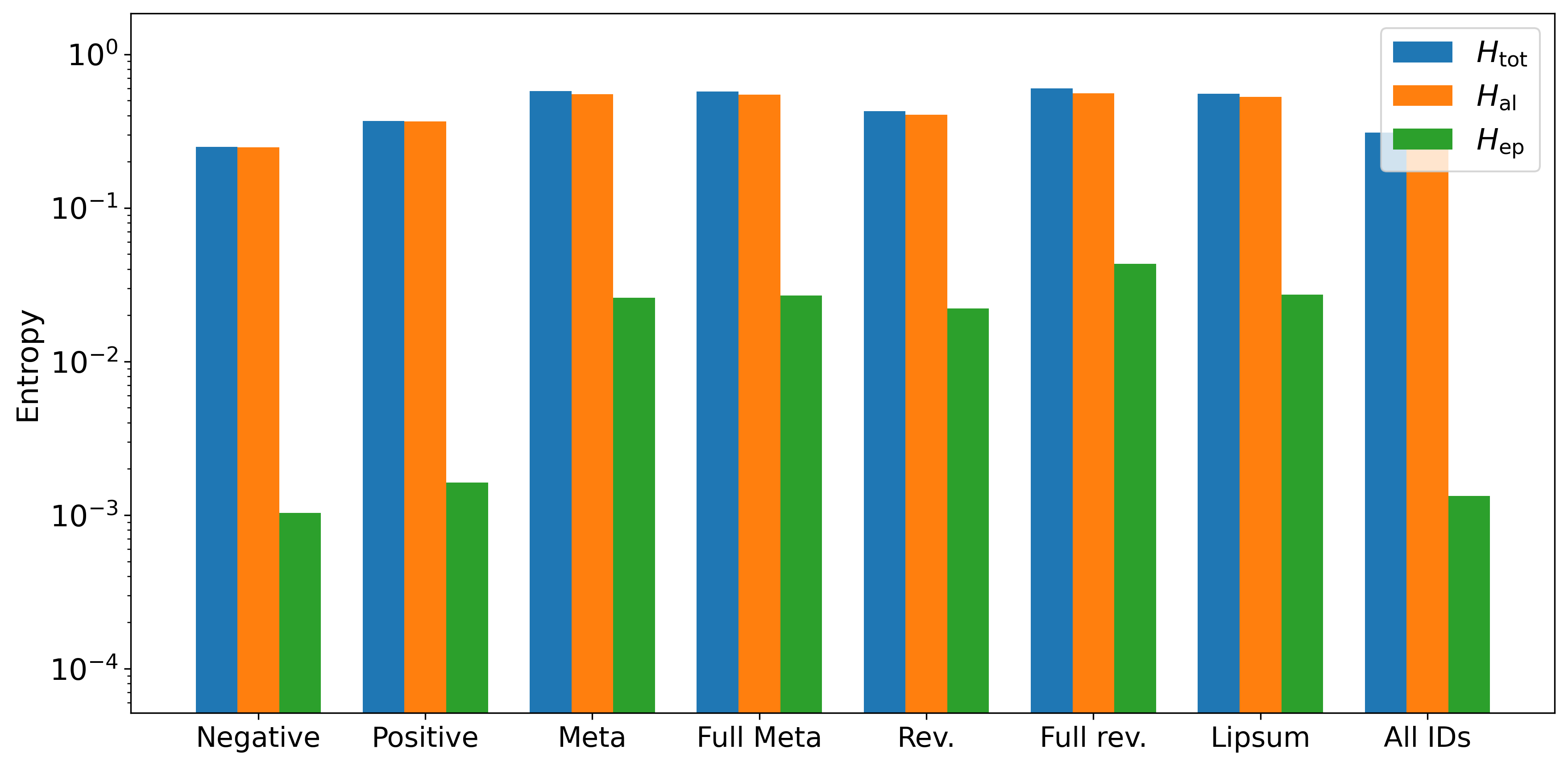}
    \caption{DE}
  \end{subfigure}
  \begin{subfigure}[b]{0.24\textwidth}
    \includegraphics[width=\linewidth]{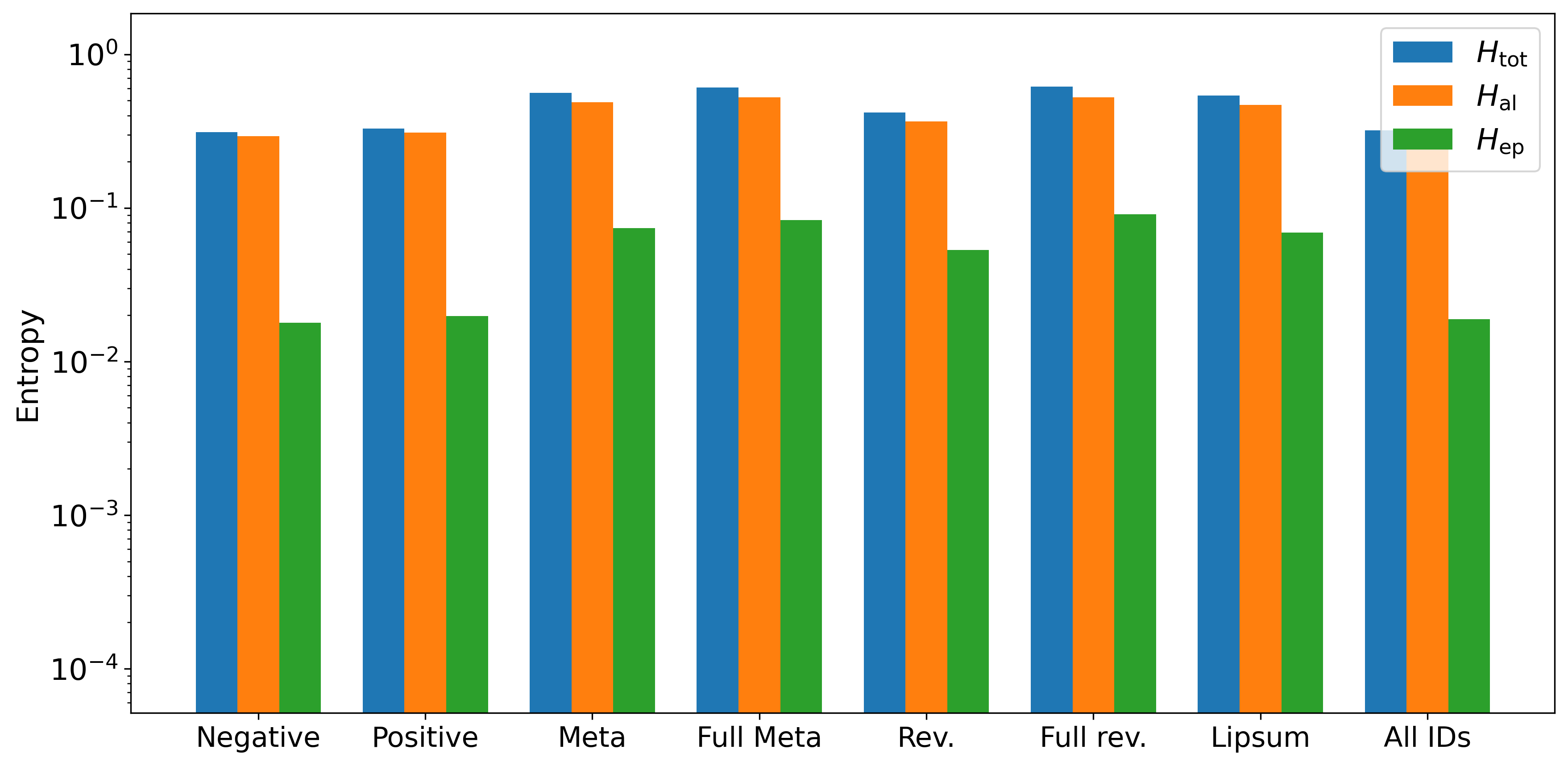}
    \caption{S-HMC\(_\parallel\)}
  \end{subfigure}
  \begin{subfigure}[b]{0.24\textwidth}
    \includegraphics[width=\linewidth]{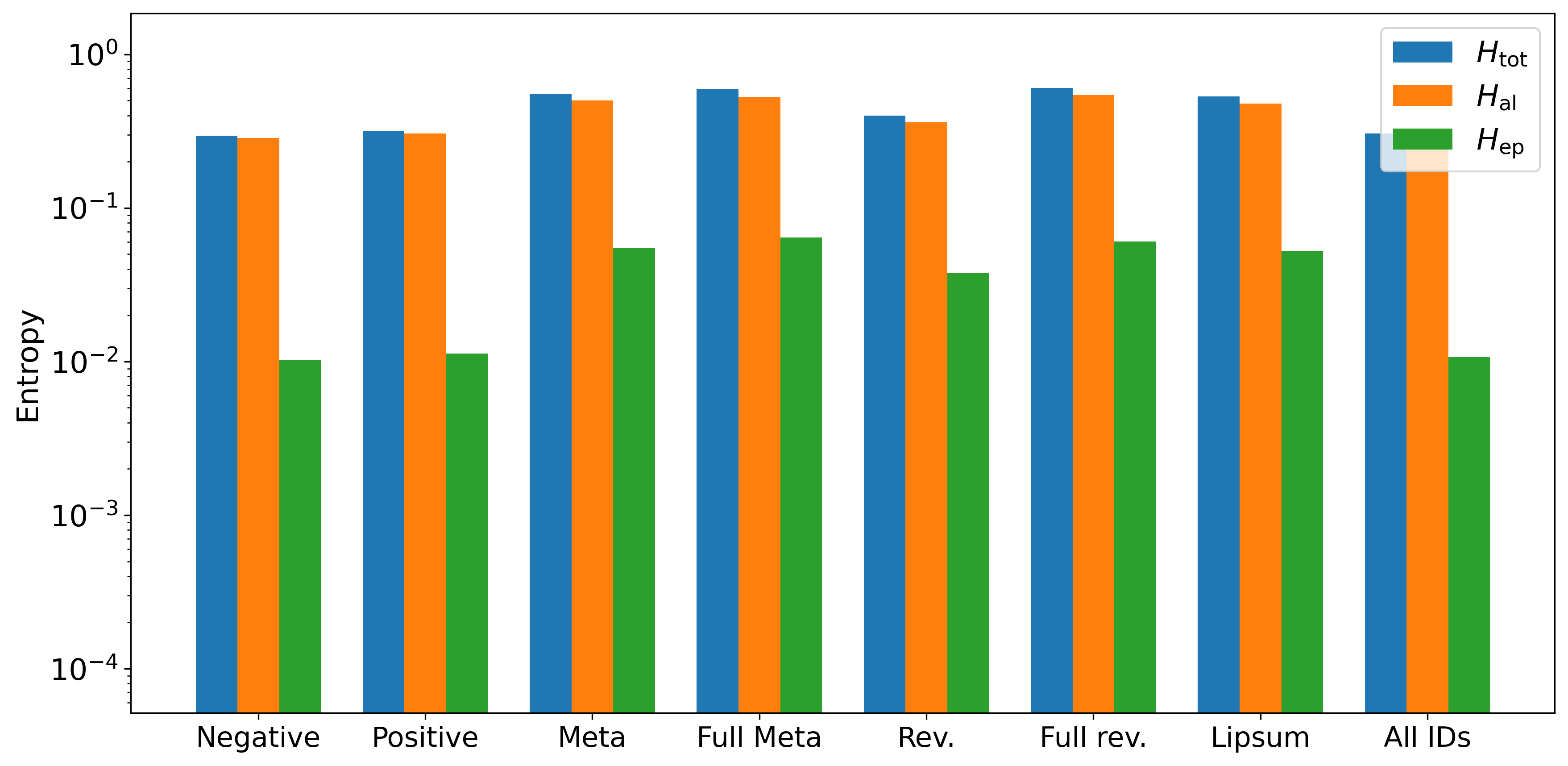}
    \caption{S-SMC\(_\parallel\)}
  \end{subfigure}
    \caption{Comparison of entropy across groups for IMDb. S-SMC\(_\parallel\) (\(P=8\) chain with \(N=10\)), S-HMC\(_{\parallel}\) (\(NP\) chains), DE (\(N\) models) and MAP, with fixed number of leapfrog \(L=1\), \(B=26\), \(M=2\), \(v=1\) and \(s=0.35\) (\(5\) realizations).}
\label{fig:imdb_fullUQ_v1}
\end{figure}

\subsection{CIFAR10}
\label{app:cifar_uq}

In the CIFAR10 case, the full setting is described in Appendix \ref{app:cifar}, where we let $N_{\sf id}=10000$ and $N_{\sf ood}=300$, and each dataset has $100$ data points. Results of entropy comparison among MAP, DE and SBMCs are given in Figure \ref{fig:cifar_entropyAll}, showing comparison in the OOD datasets and the correct/incorrect prediction in the ID domain.

\begin{figure}[H]
    \centering
    \includegraphics[width=.8\columnwidth]{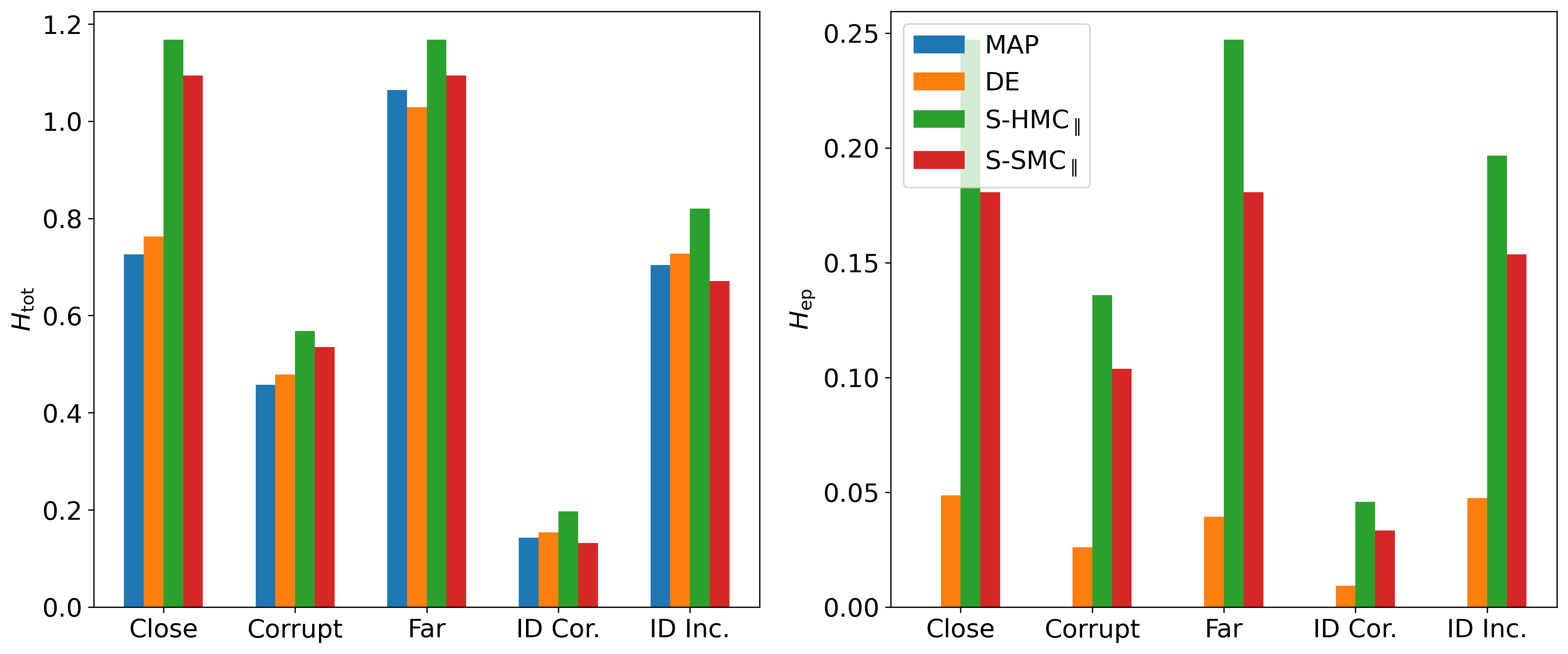}
    \caption{Comparison of average total and epistemic entropy over three out-of-domain classes and correctly/incorrectly predicted ID for CIFAR10. S-SMC\(_\parallel\) (\(P=8\) chains with \(N=10\)), S-HMC\(_{\parallel}\) (\(NP\) chains), DE (\(N\)) and MAP, with fixed number of leapfrog \(L=1\), \(B=200\), \(M=4\), \(v=0.2\) and \(s=0.05\) (\(5\) realizations).}
\label{fig:cifar_entropyAll}
\end{figure}

\section{Further results of OOD inference}
\label{app:ood}

Establishment of the meta-classifier of incorrect/OOD
data is given in the main text under Out-of-domain inference. Here, the OOD detection is performed in the default and optimal $F_1$ decision rule, respectively.

\paragraph{The default decision rule} treats the output probability of "abstain" (out-of-domain or likely misclassified) in the meta-classifier as a binary decision with a fixed cut‐off at \(0.5\). That is, if the model predicts that there is at least a \(50\%\) probability of the data being OOD or incorrectly predicted, it abstains; otherwise, it classifies the data as correctly predicted ID. This rule requires no adjustment beyond the choice of \(0.5\). Its behaviour is totally dependent on whether the model's confidence in abstention exceeds the halfway level.

\paragraph{The optimal \(F_1\) decision rule} adapts the abstention threshold to maximize the \(F_1\) score on a held‐out set. In practices, the meta-classifier's probabilities are assessed over a grid of potential thresholds ranging from \(0\) to \(1\), the F1 score are calculated for each threshold, and the threshold with the highest \(F_1\) score is chosen as the optimal \(F_1\) threshold. This customised threshold balances false positives and false negatives in the most effective way for the given data distribution, at the cost of requiring a representative validation set. It often outperforms the default decision rule when class proportions or costs of errors are skewed.

\subsection{MNIST7}
\label{app:mnist_ood}

In the MNIST7 case, the full setting is described in Appendix \ref{app:mnist}, where we let \(N_{\sf id}=2000\) and $N_{\sf ood}=2000$, where each dataset has $500$ data. Metrics of Precision, Recall, F1 and AUC-ROC metrics are given in Table \ref{tab:mnist_perf_optimal_avg}, the normalized confusion rate matrices to show how the OOD domain has been detected from the ID domain are given in Figure \ref{fig:mnist_confusion}. Plots for ROC curve and 2-level estimator accuracy are given in Figure \ref{fig:mnist_roc}.

\begin{table}[H]
  \centering
  \caption{Evaluation Metrics using thresholds. S-SMC\(_\parallel\) (\(P=1,8\) chains with \(N=10\)) and S-HMC\(_{\parallel}\) (\(NP\) chains), with fixed number of leapfrog \(L=1\), \(B=160\), \(M=10\), \(v = 0.1\) and \(s=0.1\), on MNIST (\(5\) realizations, \(\pm\) s.e. in metrics and bold the first $30\%$ data in mean).}
  \label{tab:mnist_perf_optimal_avg}
  \begin{subtable}[t]{0.49\textwidth}
    \scriptsize
    \centering
  \subcaption{Default decision threshold (0.5).} 
  \label{tab:mnist_ood_default}
  \begin{adjustbox}{width=\textwidth}
  \begin{tabular}{ll|c|c|c|c}
    \toprule
    \(P\) & Method & Precision & Recall & F1 & AUC-ROC \\
    \midrule
    – & MAP & 0.846\(\pm\)0.014 & 0.162\(\pm\)0.016 & 0.271\(\pm\)0.024 & 0.828\(\pm\)0.013 \\
    – & DE  & 0.876\(\pm\)0.007 & 0.213\(\pm\)0.011 & 0.342\(\pm\)0.015 & 0.855\(\pm\)0.003 \\
    1 & S-SMC\(_\parallel\) & 0.845\(\pm\)0.020 & 0.216\(\pm\)0.037 & 0.338\(\pm\)0.049 & 0.824\(\pm\)0.017 \\
    8 & S-SMC\(_\parallel\) & 0.894\(\pm\)0.015 & 0.389\(\pm\)0.046 & 0.537\(\pm\)0.052 & 0.884\(\pm\)0.006 \\
    1 & S-HMC\(_\parallel\) & {\bf 0.906\(\pm\)0.004} & {\bf 0.432\(\pm\)0.020} & {\bf 0.584\(\pm\)0.020} & {\bf 0.885\(\pm\)0.002} \\
    8 & S-HMC\(_\parallel\) & {\bf 0.907\(\pm\)0.001} & {\bf 0.470\(\pm\)0.007} & {\bf 0.619\(\pm\)0.006} & {\bf 0.892\(\pm\)0.001} \\
    \bottomrule
  \end{tabular}
  \end{adjustbox}
  \end{subtable}
\hfill
  \begin{subtable}[t]{0.49\textwidth}
    \scriptsize
    \centering
  \subcaption{Optimal \(F_1\) decision threshold.}
  \label{tab:mnist_ood_optimal}
  \begin{adjustbox}{width=\textwidth}
  \begin{tabular}{ll|c|c|c|c}
    \toprule
    \(P\) & Method & Precision & Recall & F1 & AUC-ROC \\
    \midrule
    – & MAP & 0.707\(\pm\)0.013 & 0.898\(\pm\)0.006 & 0.791\(\pm\)0.009 & 0.828\(\pm\)0.013 \\
    – & DE  & 0.734\(\pm\)0.004 & 0.897\(\pm\)0.003 & 0.807\(\pm\)0.002 & 0.855\(\pm\)0.003 \\
    1 & S-SMC\(_\parallel\) & 0.701\(\pm\)0.021 & 0.890\(\pm\)0.015 & 0.783\(\pm\)0.010 & 0.824\(\pm\)0.017 \\
    8 & S-SMC\(_\parallel\) & {\bf 0.753\(\pm\)0.004} & {\bf 0.915\(\pm\)0.004} & {\bf 0.826\(\pm\)0.001} & 0.884\(\pm\)0.006 \\
    1 & S-HMC\(_\parallel\) & 0.750\(\pm\)0.003 & 0.906\(\pm\)0.004 & 0.820\(\pm\)0.002 & {\bf 0.885\(\pm\)0.002} \\
    8 & S-HMC\(_\parallel\) & {\bf 0.752\(\pm\)0.003} & {\bf 0.913\(\pm\)0.004} & {\bf 0.825\(\pm\)0.001} & {\bf 0.892\(\pm\)0.001} \\
    \bottomrule
  \end{tabular}
  \end{adjustbox}
  \end{subtable}
\end{table}

\begin{figure}[H]
  \centering
  \begin{subfigure}[b]{0.16\textwidth}
    \includegraphics[width=\linewidth]{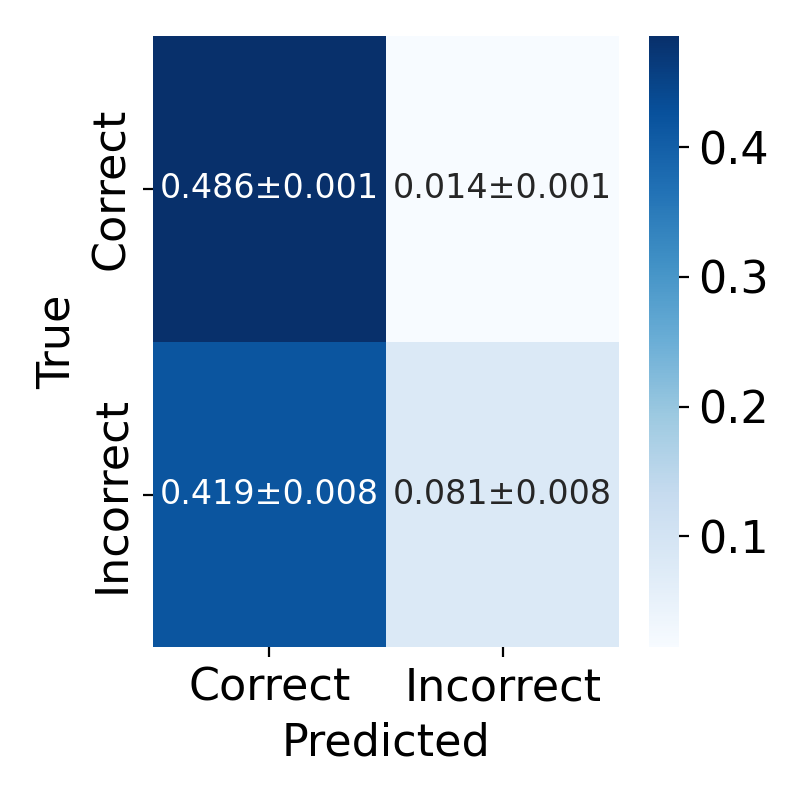}
    \caption{\scriptsize MAP}
  \end{subfigure}
  \hfill
  \begin{subfigure}[b]{0.16\textwidth}
    \includegraphics[width=\linewidth]{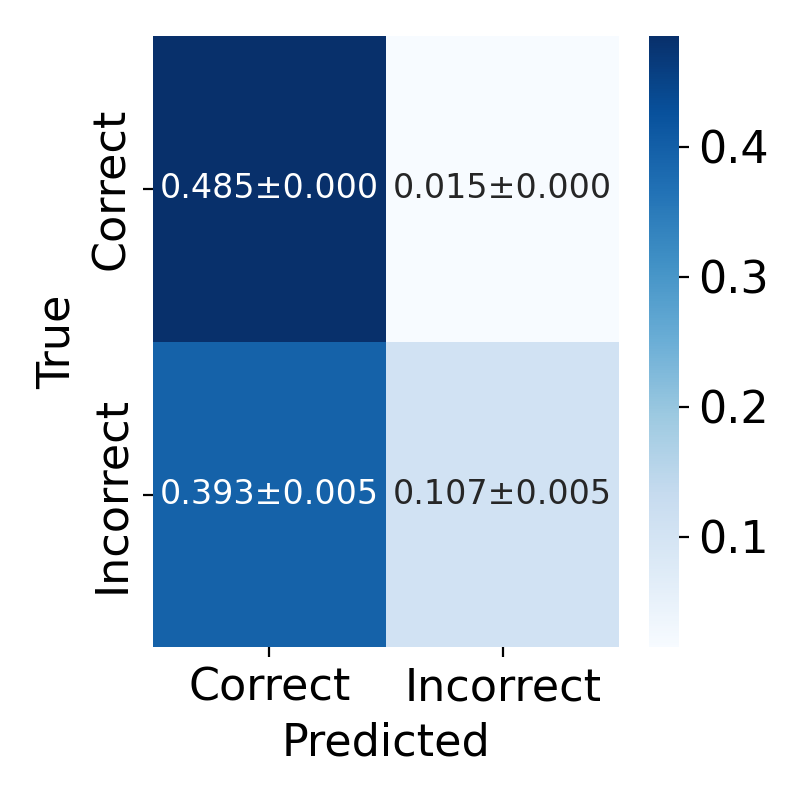}
    \caption{\scriptsize DE}
  \end{subfigure}
  \hfill
  \begin{subfigure}[b]{0.16\textwidth}
    \includegraphics[width=\linewidth]{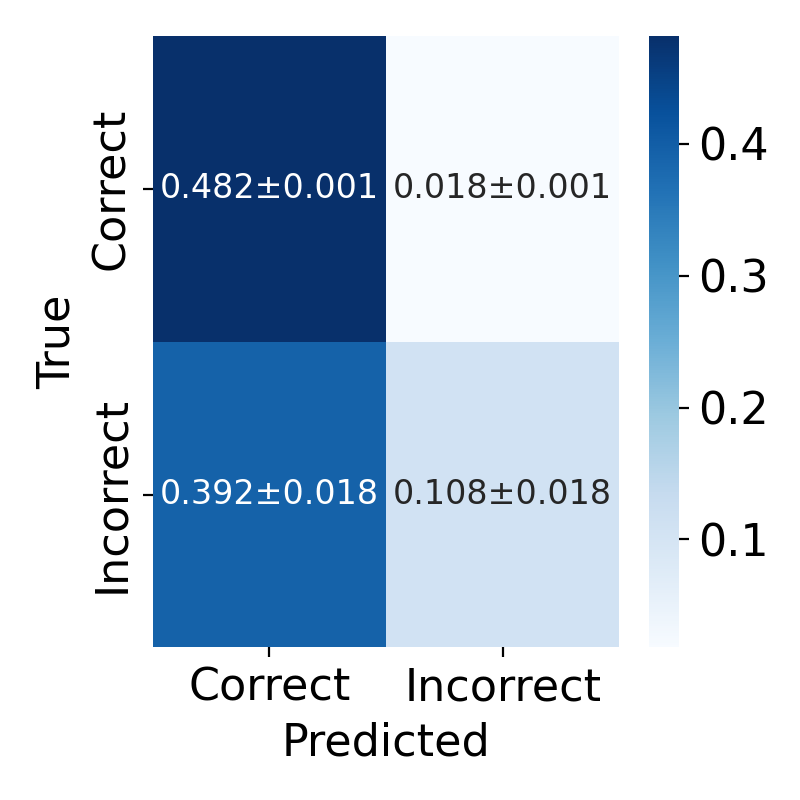}
    \caption{\scriptsize S-SMC\(_\parallel\)(\(P=1\))}
   \end{subfigure}
   \hfill
  \begin{subfigure}[b]{0.16\textwidth}
    \includegraphics[width=\linewidth]{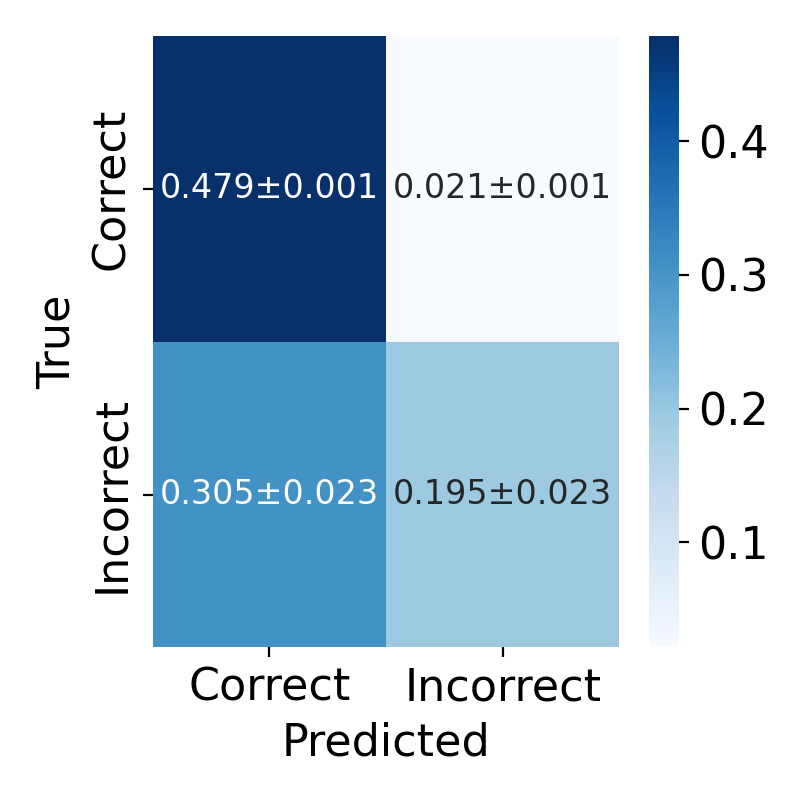}
    \caption{\scriptsize S-SMC\(_\parallel\)(\(P=8\))}
   \end{subfigure}
  \hfill
  \begin{subfigure}[b]{0.16\textwidth}
    \includegraphics[width=\linewidth]{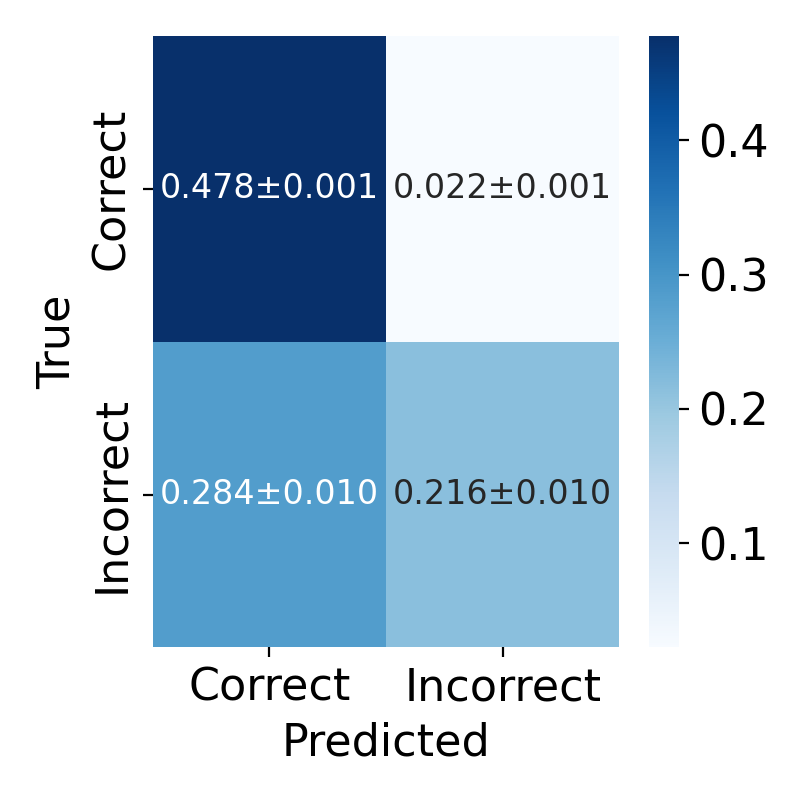}
    \caption{\scriptsize S-HMC\(_\parallel\)(\(P=1\))}
  \end{subfigure}
  \hfill
  \begin{subfigure}[b]{0.16\textwidth}
    \includegraphics[width=\linewidth]{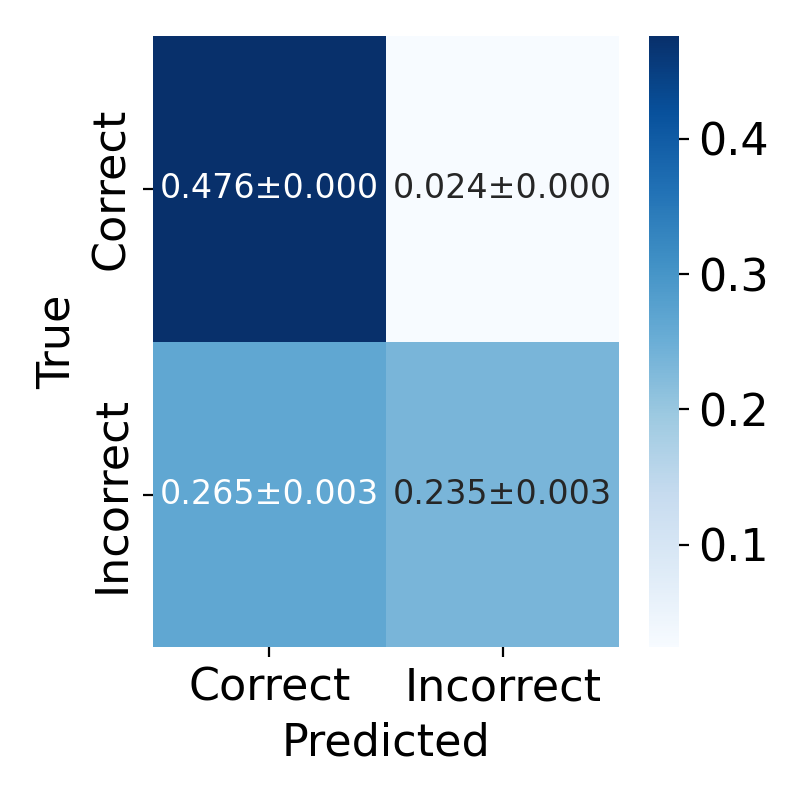}
    \caption{\scriptsize S-HMC\(_\parallel\)(\(P=8\))}
  \end{subfigure}

  \vspace{1em}

  \begin{subfigure}[b]{0.16\textwidth}
    \includegraphics[width=\linewidth]{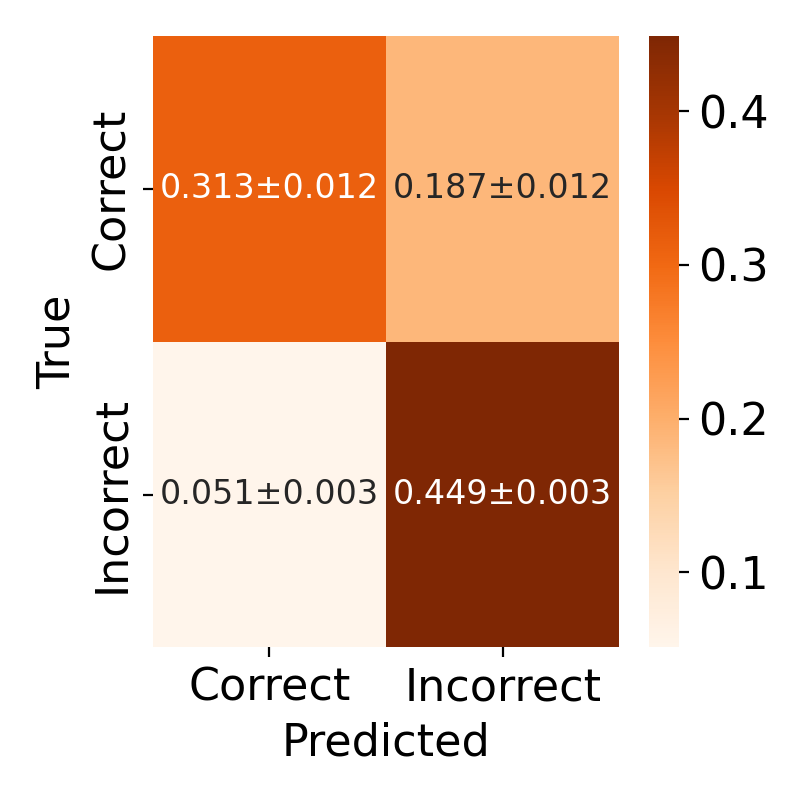}
    \caption{\scriptsize MAP}
  \end{subfigure}
  \hfill
  \begin{subfigure}[b]{0.16\textwidth}
    \includegraphics[width=\linewidth]{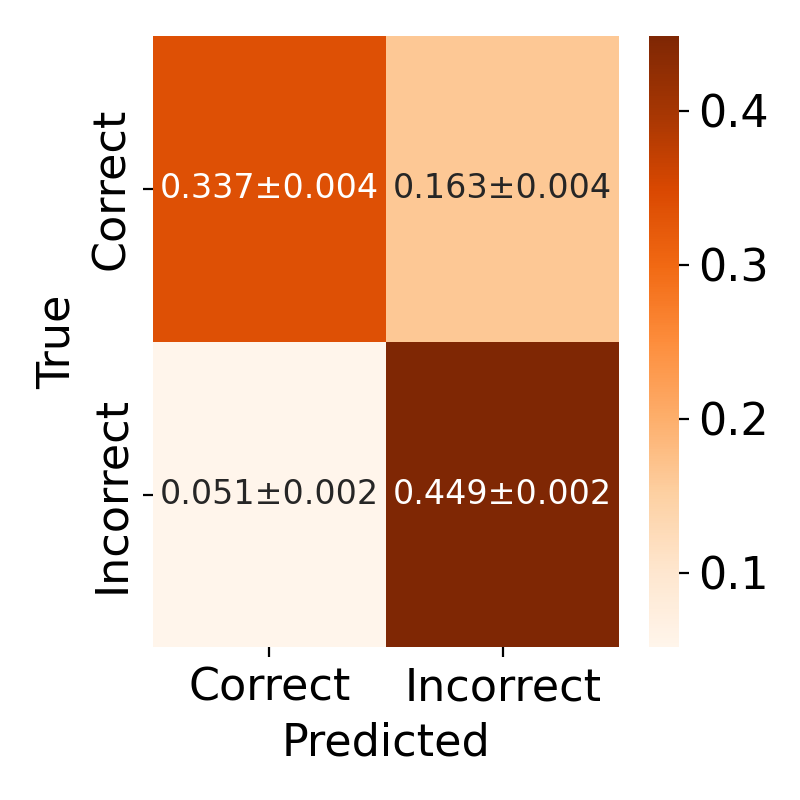}
    \caption{\scriptsize DE}
  \end{subfigure}
  \hfill
  \begin{subfigure}[b]{0.16\textwidth}
    \includegraphics[width=\linewidth]{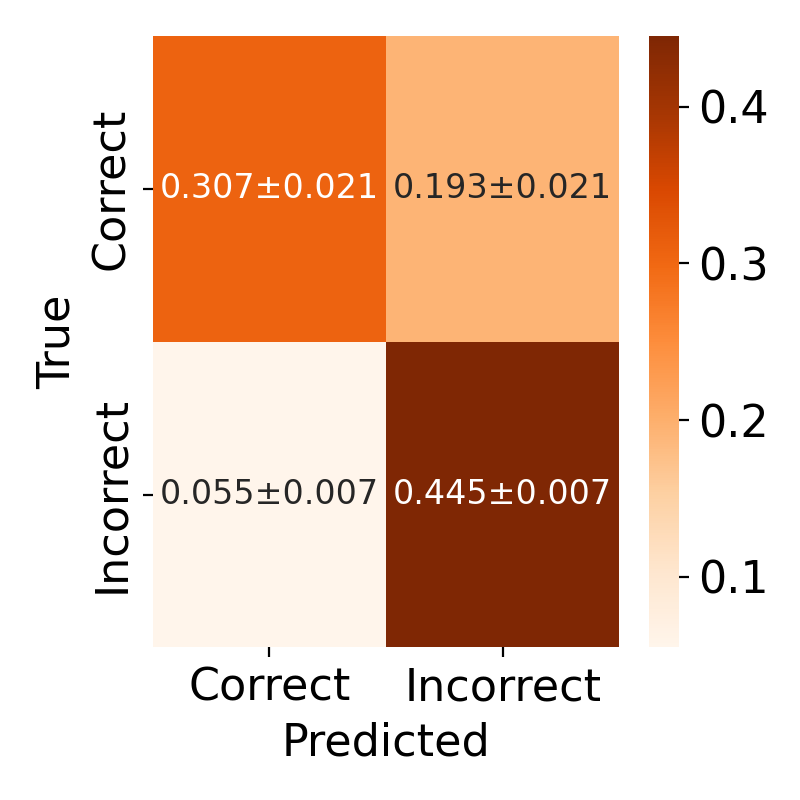}
    \caption{\scriptsize S-SMC\(_\parallel\)(\(P=1\))}
  \end{subfigure}
  \hfill
  \begin{subfigure}[b]{0.16\textwidth}
    \includegraphics[width=\linewidth]{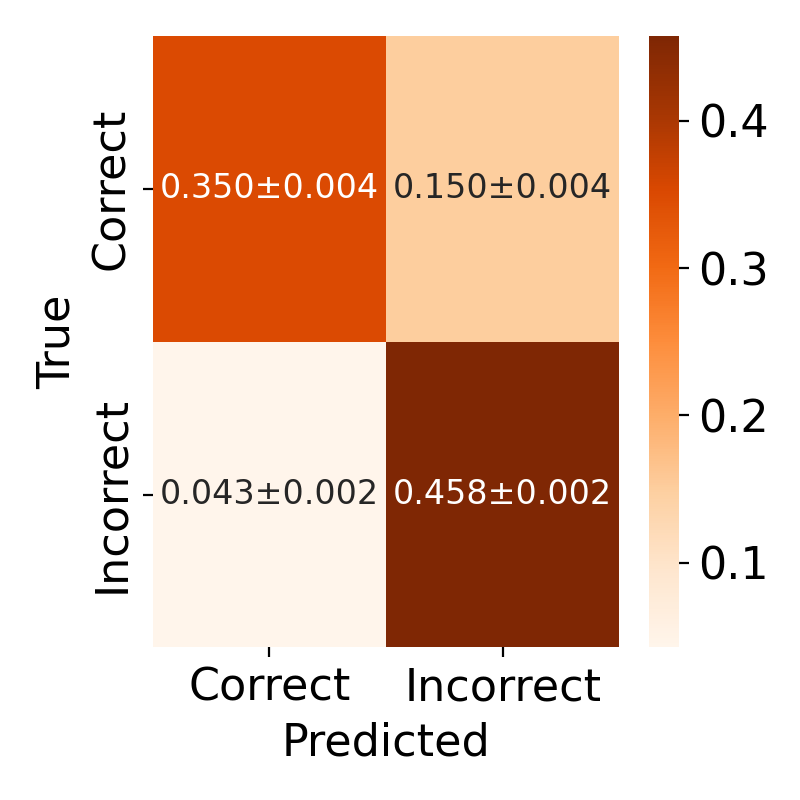}
    \caption{\scriptsize S-SMC\(_\parallel\)(\(P=8\))}
  \end{subfigure}
  \hfill
  \begin{subfigure}[b]{0.16\textwidth}
    \includegraphics[width=\linewidth]{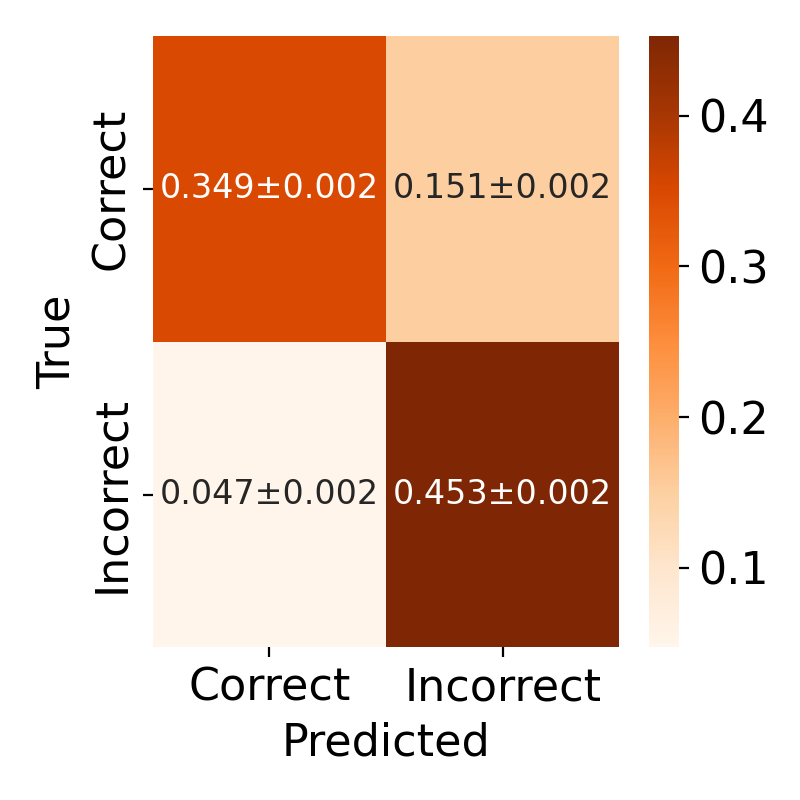}
    \caption{\scriptsize S-HMC\(_\parallel\)(\(P=1\))}
  \end{subfigure}
  \hfill
  \begin{subfigure}[b]{0.16\textwidth}
    \includegraphics[width=\linewidth]{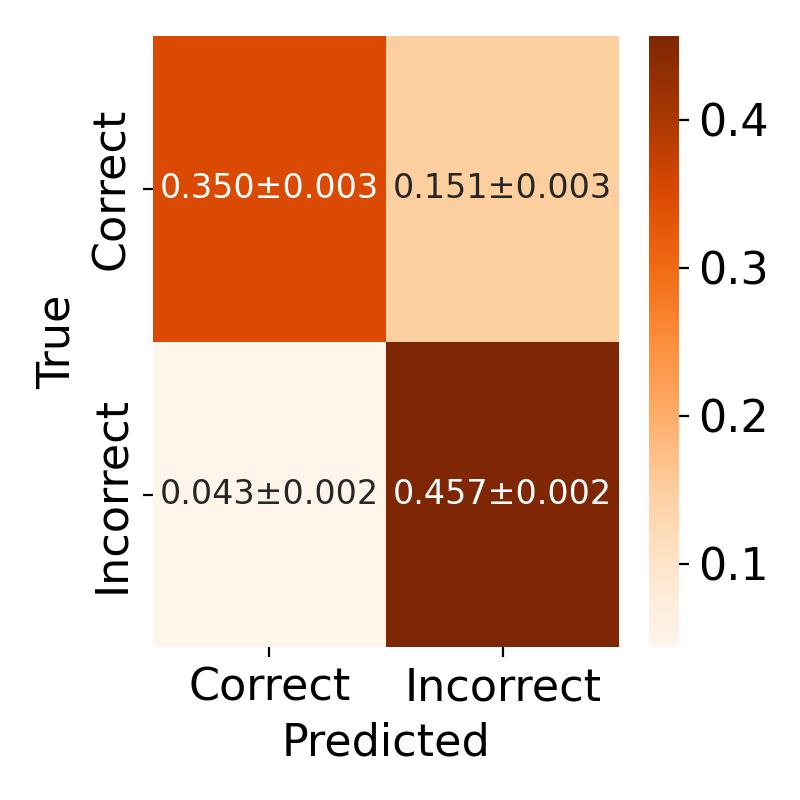}
    \caption{\scriptsize S-HMC\(_\parallel\)(\(P=8\))}
  \end{subfigure}

  \caption{Averaged confusion rate matrices for OOD prediction on MNIST7, with default decision threshold (top) and optimal \(F_1\) decision threshold (bottom). S-SMC\(_\parallel\) (\(P=1,8\) chains with \(N=10\)) and S-HMC\(_{\parallel}\) (\(NP\) chains), with fixed number of leapfrog \(L=1\), \(B=160\), \(M=10\), \(v = 0.1\) and \(s=0.1\), on MNIST (5 realizations and \(\pm\) s.e. in metrics).}
  \label{fig:mnist_confusion}
\end{figure}

\begin{figure}[H]
  \centering
  \begin{subfigure}[b]{0.49\textwidth}
  \centering
    \includegraphics[width=\linewidth]{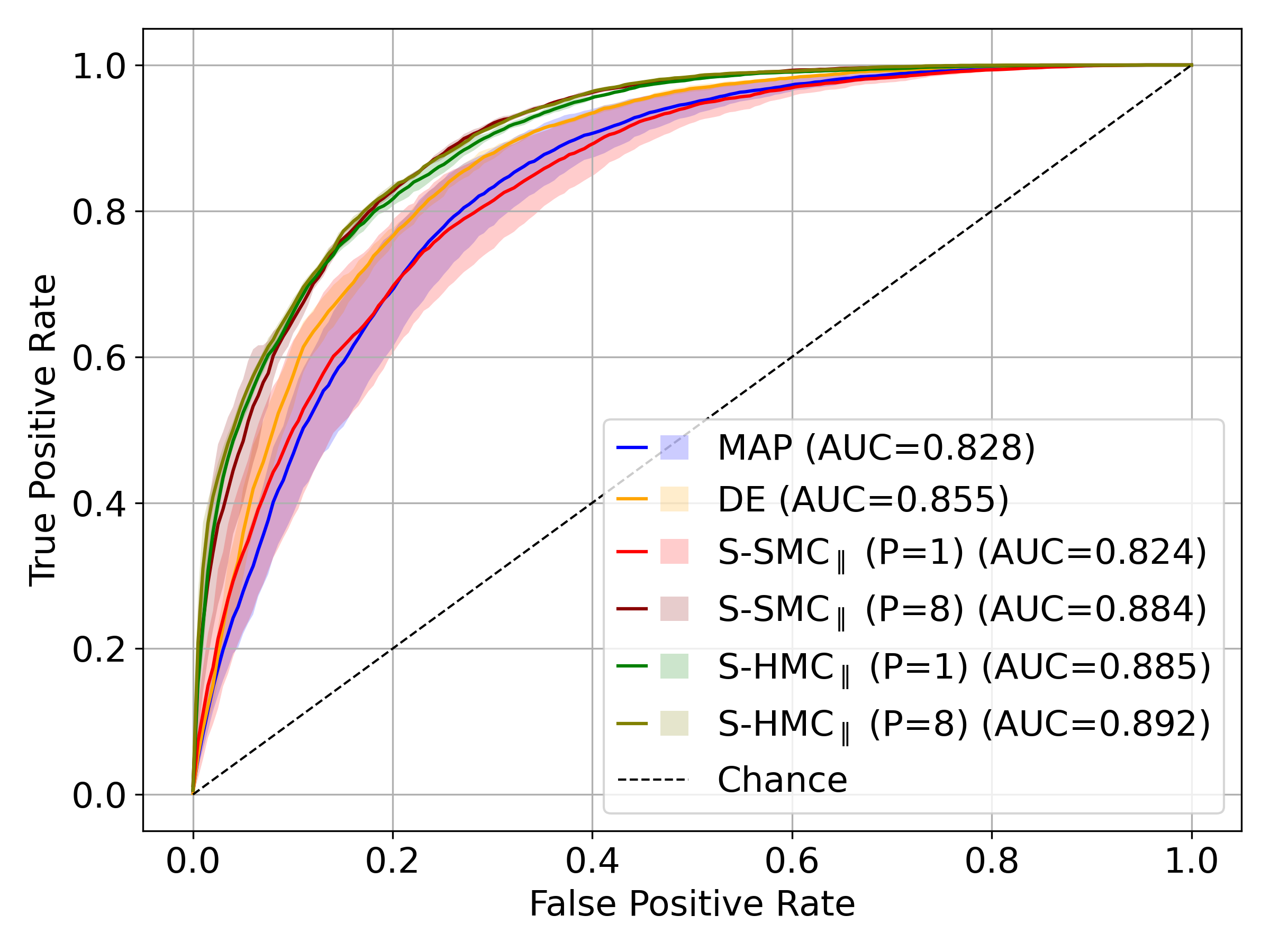}
    \caption{ROC curve}
  \end{subfigure}
  \hfill
  \begin{subfigure}[b]{0.49\textwidth}
  \centering
    \includegraphics[width=\linewidth]{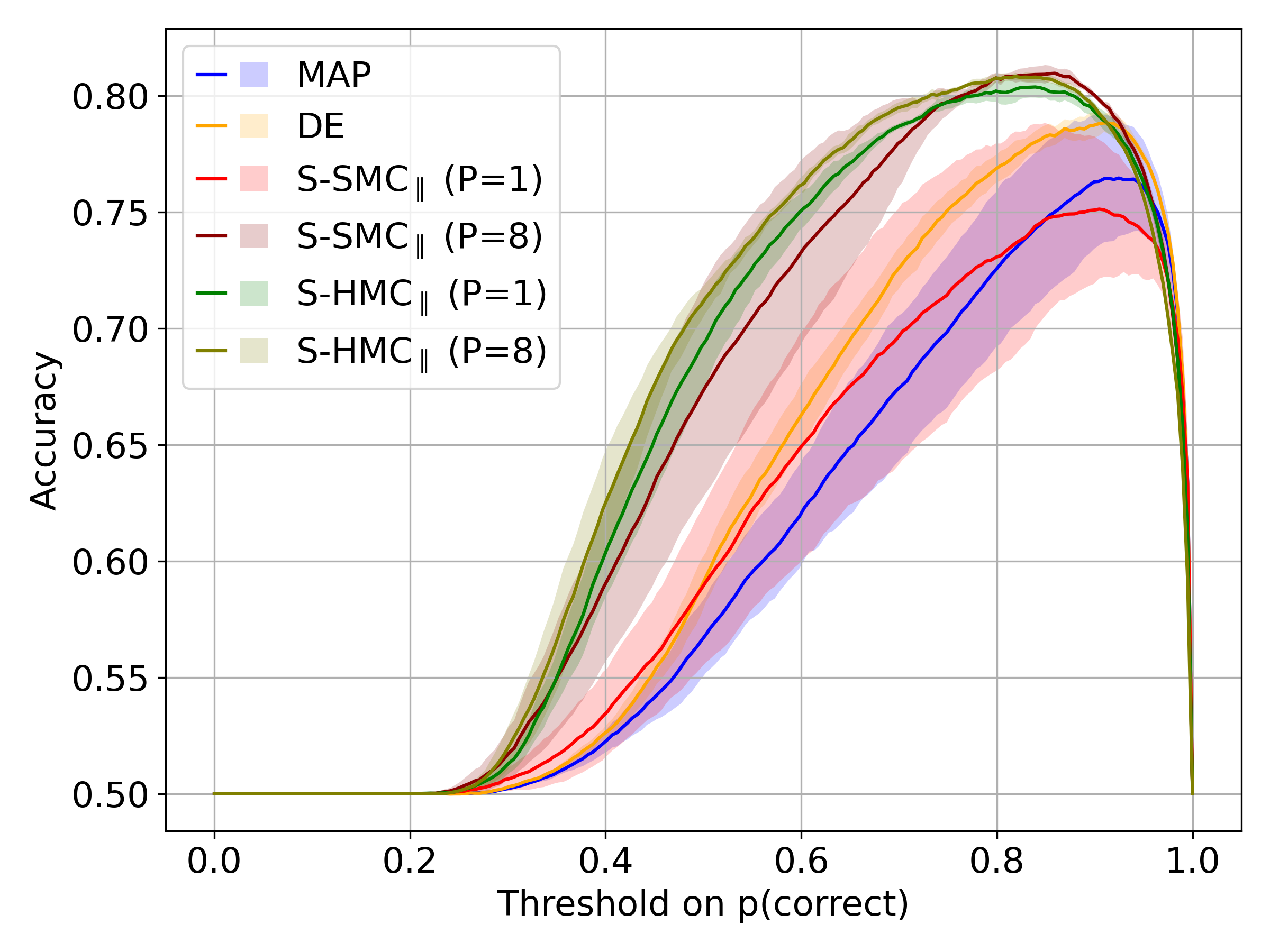}
    \caption{2-level estimator accuracy}
  \end{subfigure}
  \caption{Averaged curve plots for OOD detection on MNIST7. S-SMC\(_\parallel\) (\(P=1,8\) chains with \(N=10\)) and S-HMC\(_{\parallel}\) (\(NP\) chains), with fixed number of leapfrog \(L=1\), \(B=160\), \(M=10\), \(v = 0.1\) and \(s=0.1\), on MNIST (5 realizations and \(\pm\) s.e. in metrics).}
  \label{fig:mnist_roc}
\end{figure}

\subsection{IMDB}

In the IMDb case, the full setting is described in Appendix \ref{app:imdb}, where we let $N_{\sf ood}=25000$, and each dataset has $5000$ data points.

\paragraph{Experiment with $v=\frac{1}{40}$} Metrics of Precision, Recall, F1 and AUC-ROC metrics are given in Table \ref{tab:imdb_perf_optimal_avg_v0025}, the normalized confusion rate matrices to show how the OOD domain has been detected from ID domain are given in Figure \ref{fig:imdb_confusion}. Plots for ROC curve and 2-level estimator accuracy are given in Figure \ref{fig:imdb_roc}.

\begin{figure}[H]
  \centering
  \begin{subfigure}[b]{0.16\textwidth}
    \includegraphics[width=\linewidth]{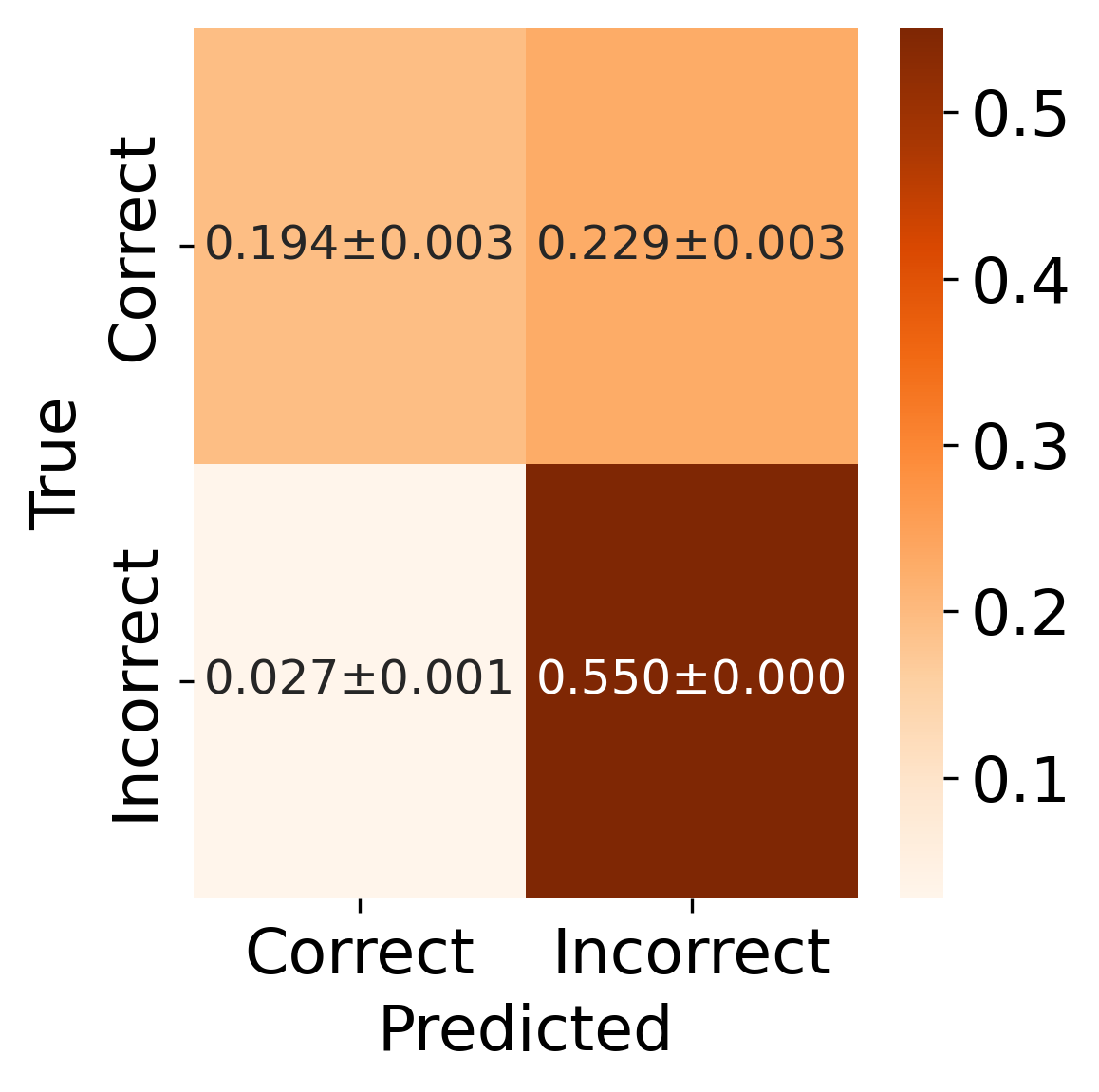}
    \caption{\scriptsize MAP}
  \end{subfigure}
  \hfill
  \begin{subfigure}[b]{0.16\textwidth}
    \includegraphics[width=\linewidth]{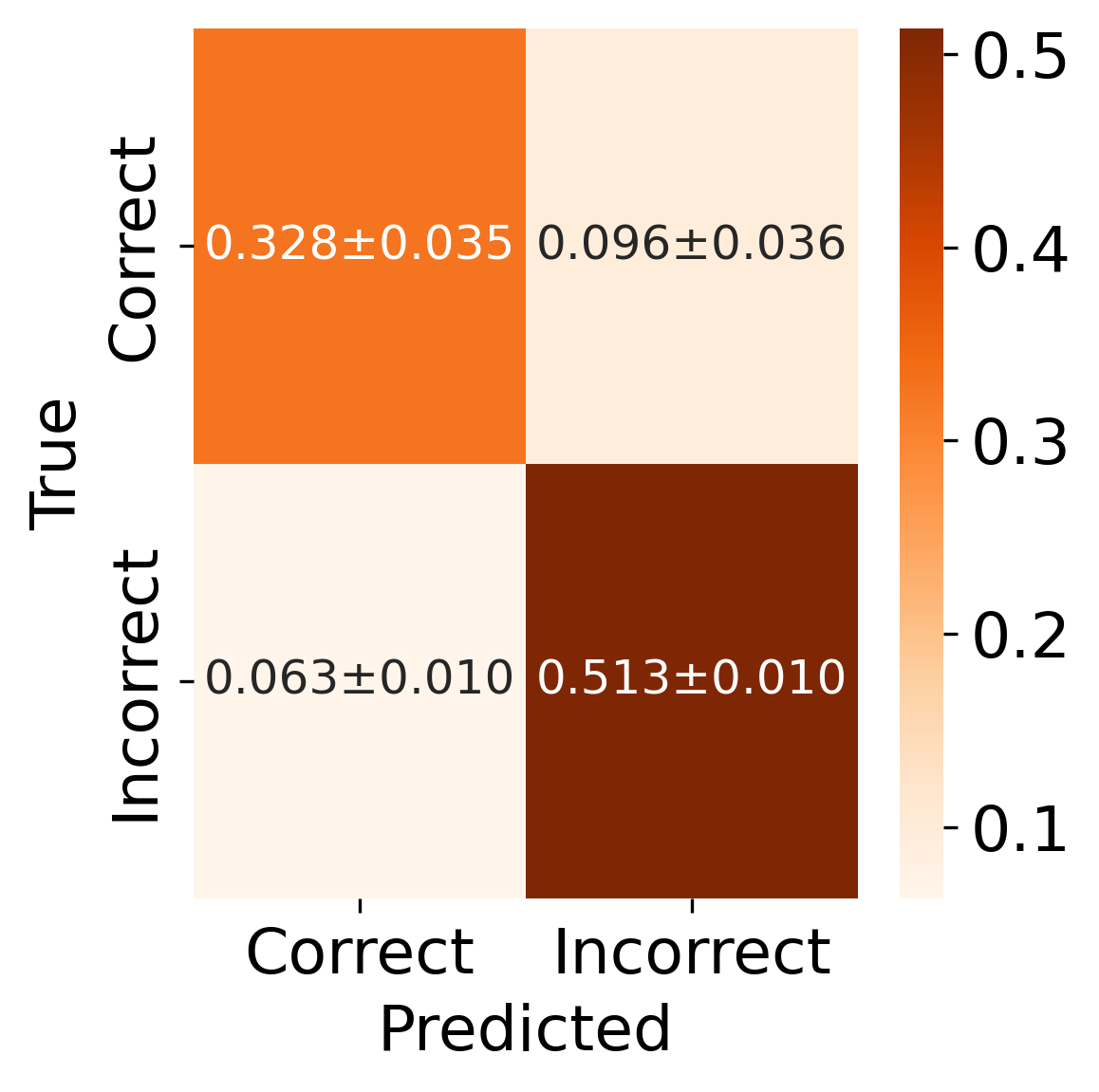}
    \caption{\scriptsize DE}
  \end{subfigure}
  \hfill
  \begin{subfigure}[b]{0.16\textwidth}
    \includegraphics[width=\linewidth]{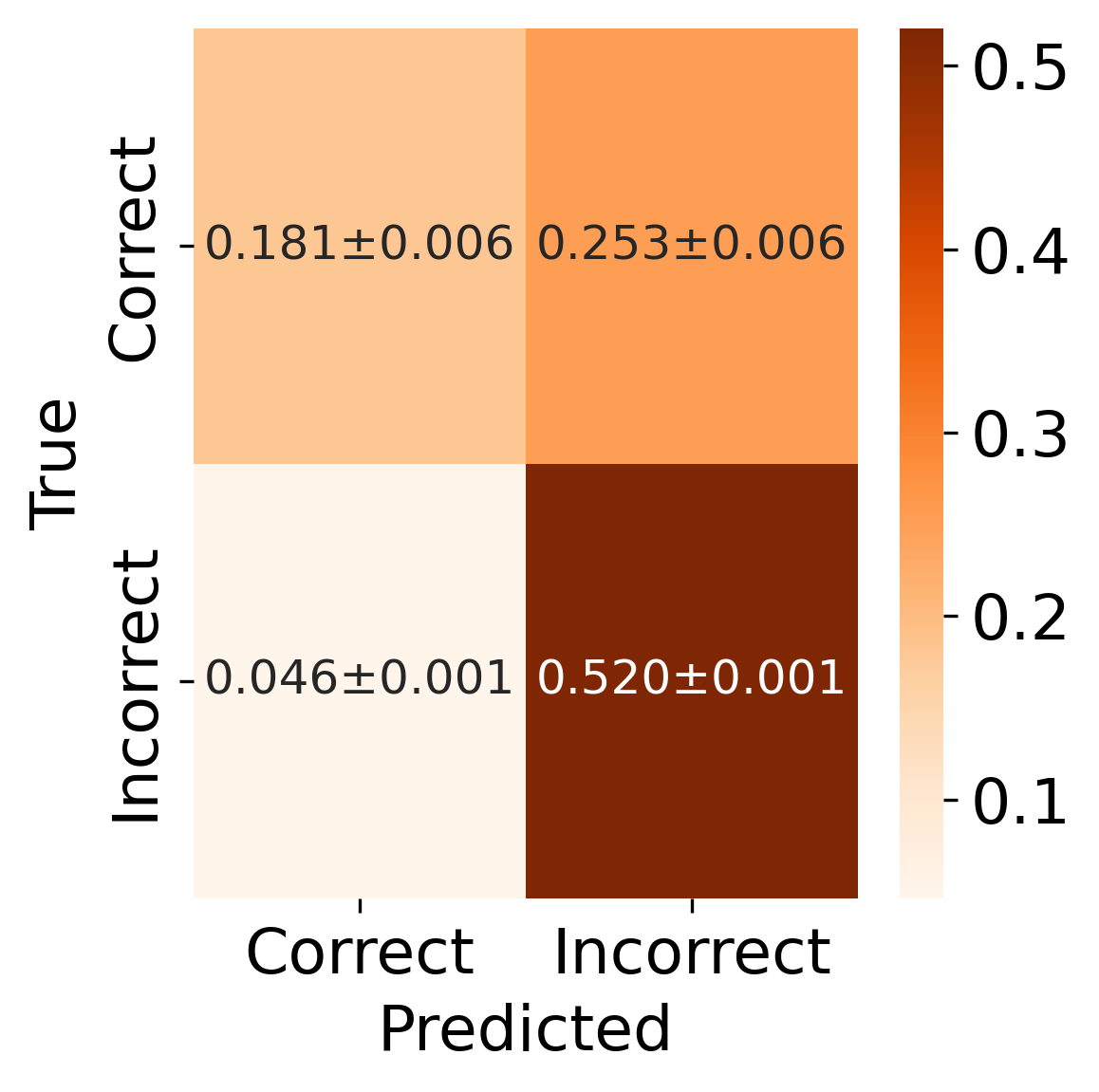}
    \caption{\scriptsize S-SMC\(_\parallel\)(\(P=1\))}
  \end{subfigure}
  \hfill
  \begin{subfigure}[b]{0.16\textwidth}
    \includegraphics[width=\linewidth]{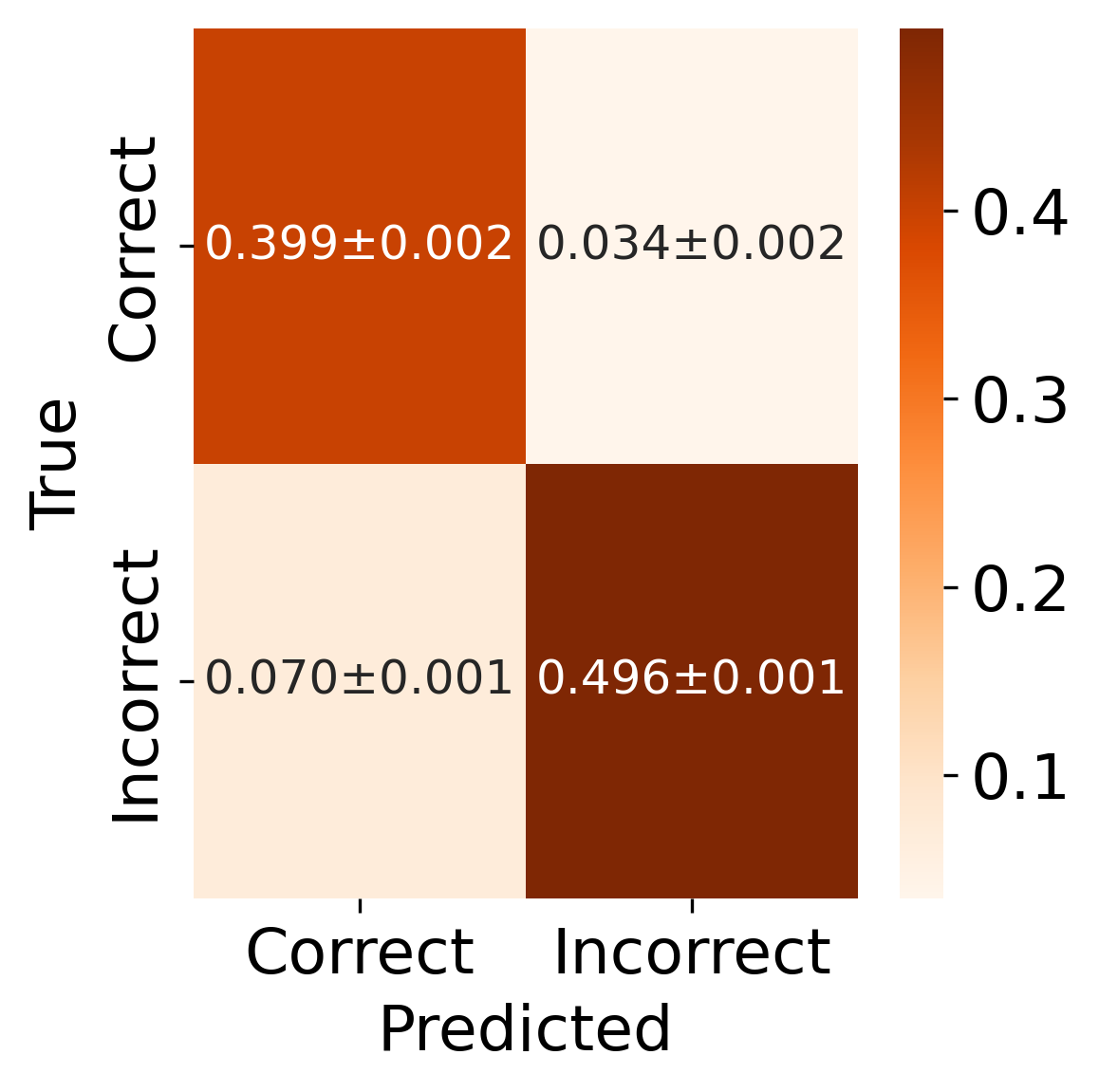}
    \caption{\scriptsize S-SMC\(_\parallel\)(\(P=8\))}
  \end{subfigure}
  \hfill
  \begin{subfigure}[b]{0.16\textwidth}
    \includegraphics[width=\linewidth]{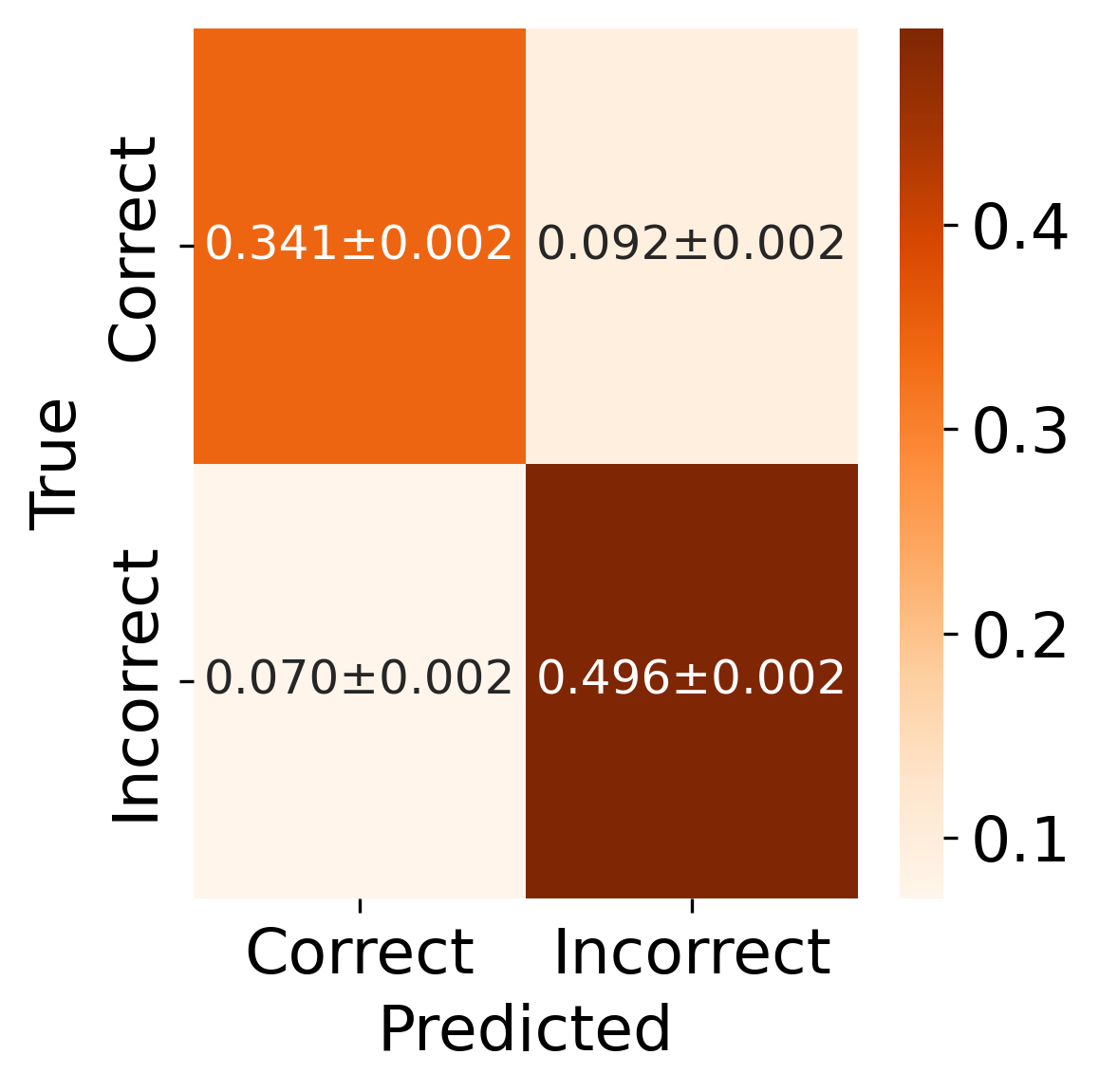}
    \caption{\scriptsize S-HMC\(_\parallel\)(\(P=1\))}
  \end{subfigure}
\hfill
  \begin{subfigure}[b]{0.16\textwidth}
    \includegraphics[width=\linewidth]{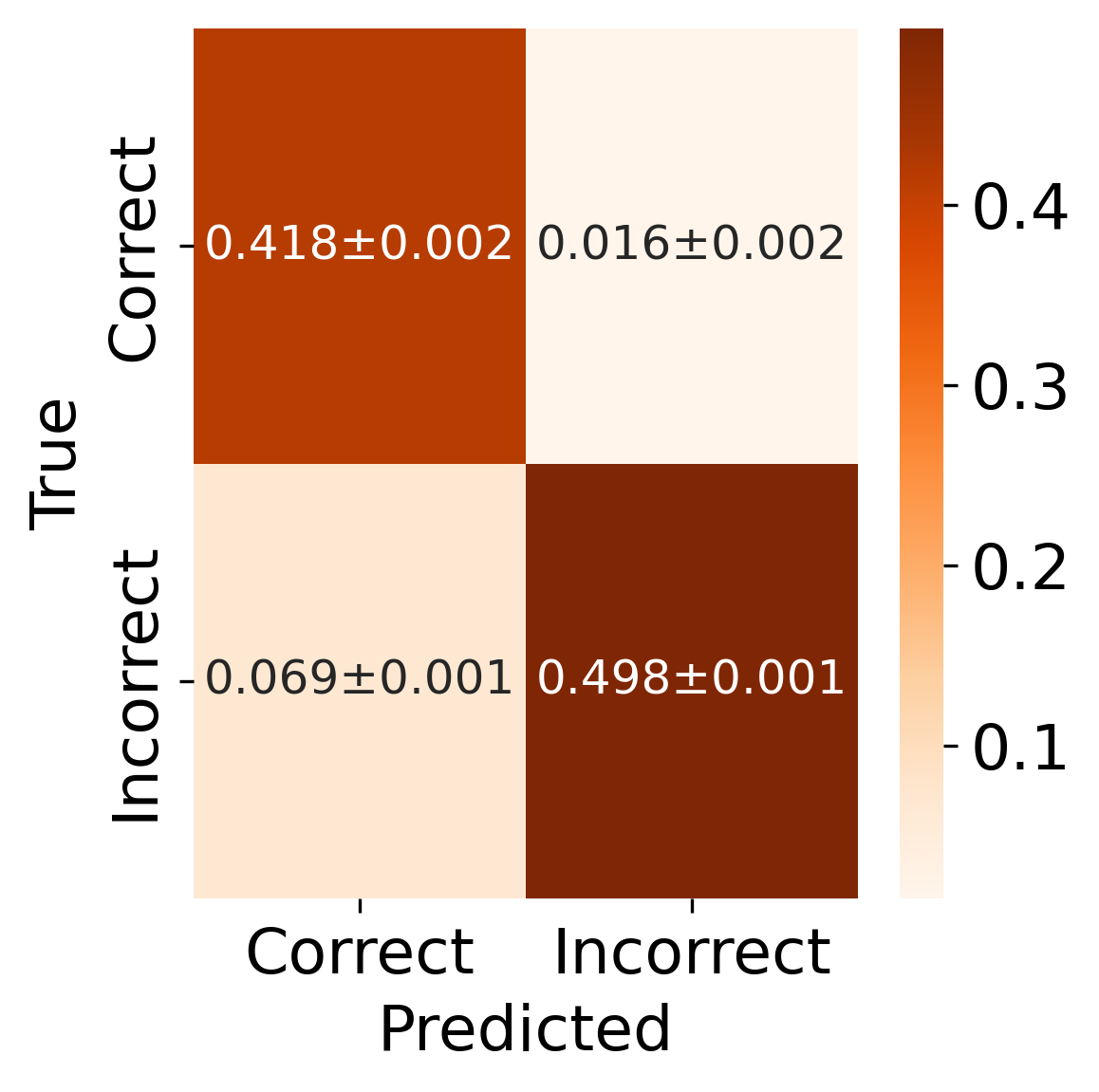}
    \caption{\scriptsize S-HMC\(_\parallel\)(\(P=8\))}
  \end{subfigure}
  \caption{Averaged confusion rate matrices for OOD prediction on IMDb, with optimal \(F_1\) decision threshold. S-SMC\(_\parallel\) (\(P=1,8\) chain with \(N=10\)), S-HMC\(_{\parallel}\) (\(NP\) chains), DE (\(N\) models) and MAP, with fixed number of leapfrog \(L=1\), \(B=25\), \(M=1\), \(v=0.025\) and \(s=0.1\) (\(5\) realizations and \(\pm\) s.e. in metrics).}
  \label{fig:imdb_confusion}
\end{figure}

\begin{table}[H]
\footnotesize
  \centering
  \caption{Performance at the optimal \(F_1\) decision threshold. S-SMC\(_\parallel\) (\(P=1\) chain with \(N=10\)), S-HMC\(_{\parallel}\) (\(NP\) chains), DE (\(N\) models) and MAP, with fixed number of leapfrog \(L=1\), \(B\), \(M\), \(v=0.025\) and \(s=0.1\) (\(5\) realizations, \(\pm\) s.e. in metrics and bold the first $30\%$ data in mean).}
  \label{tab:imdb_perf_optimal_avg_v0025}
  \begin{tabular}{ll|c|c|c|c}
    \toprule
    \(P\) & Method & Precision & Recall & F1 & AUC-ROC \\
    \midrule
    – & MAP & 0.707 ± 0.003 & \textbf{0.953 ± 0.001} & 0.811 ± 0.001 & 0.768 ± 0.005\\
    – & DE & 0.856 ± 0.041 & 0.890 ± 0.017 & 0.869 ± 0.016 & 0.896 ± 0.025\\
    1 & S-SMC\(_\parallel\) & 0.673 ± 0.005 & {\bf 0.919 ± 0.002} & 0.777 ± 0.004 & 0.777 ± 0.002\\
    8 & S-SMC\(_\parallel\) & \textbf{0.935 ± 0.003} & 0.876 ± 0.002 & \textbf{0.905 ± 0.000} & \textbf{0.935 ± 0.002}\\
    1 & S-HMC\(_\parallel\) & 0.844 ± 0.002 & 0.876 ± 0.004 & 0.859 ± 0.001 & 0.905 ± 0.001\\
    8 & S-HMC\(_\parallel\) & \textbf{0.970 ± 0.003} & 0.879 ± 0.003 & \textbf{0.922 ± 0.000} & \textbf{0.944 ± 0.002}\\
    \bottomrule
  \end{tabular}
\end{table}

\begin{figure}[H]
  \centering
  \begin{subfigure}[b]{0.49\textwidth}
  \centering
    \includegraphics[width=\linewidth]{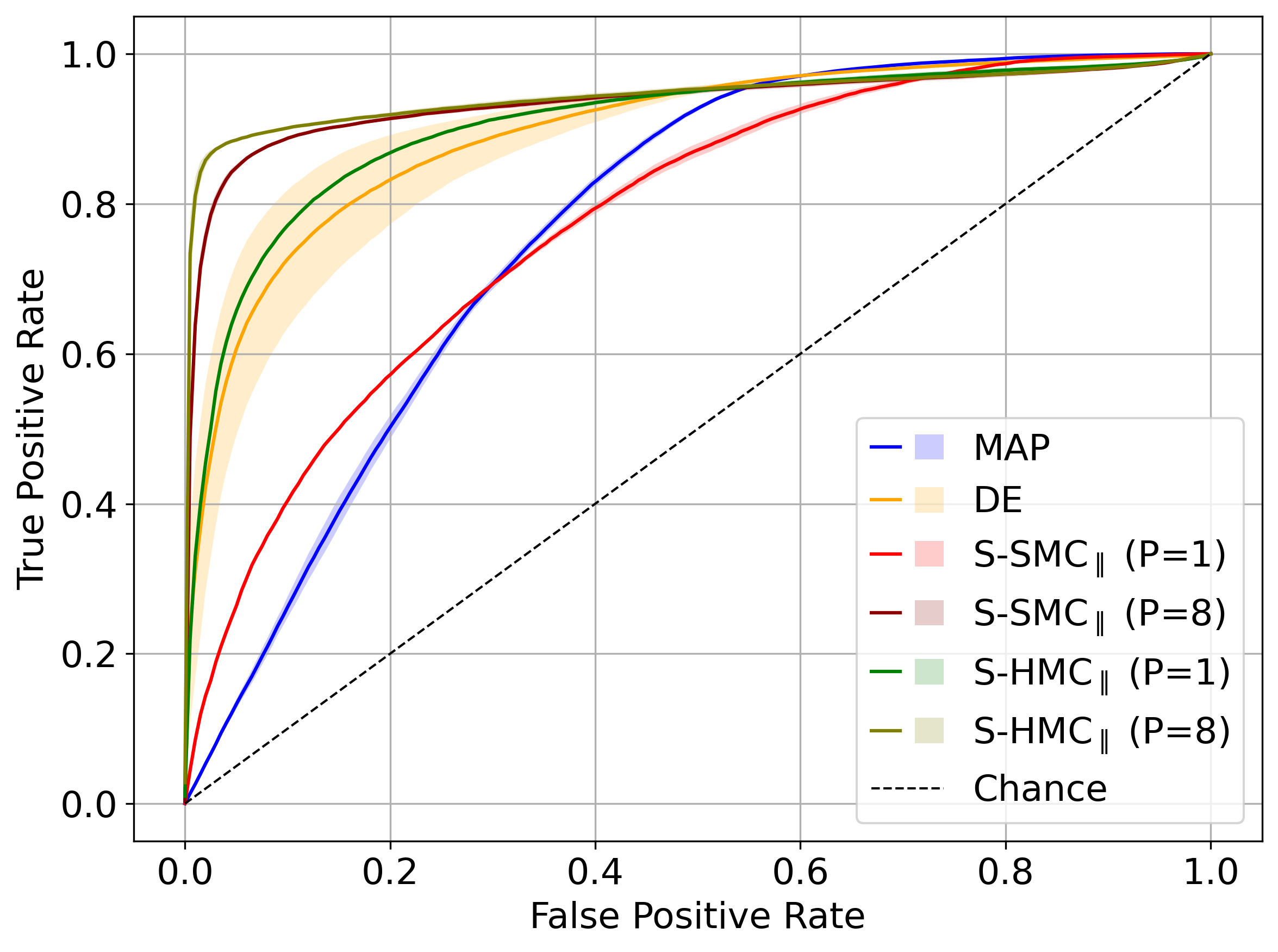}
    \caption{ROC curve}
  \end{subfigure}
  \hfill
  \begin{subfigure}[b]{0.49\textwidth}
  \centering
    \includegraphics[width=\linewidth]{figs/imdb_mean_accuracy_vs_pcorrect.png}
    \caption{2-level estimator accuracy}
  \end{subfigure}
  \caption{Averaged curve plots for OOD detection in IMDb. S-SMC\(_\parallel\) (\(P=1,8\) chain with \(N=10\)), S-HMC\(_{\parallel}\) (\(NP\) chains), DE (\(N\) models) and MAP, with fixed number of leapfrog \(L=1\), \(B=25\), \(M=1\), \(v=0.025\) and \(s=0.1\) (\(5\) realizations and \(\pm\) s.e. in metrics).}
  \label{fig:imdb_roc}
\end{figure}

\paragraph{Experiment with $v=1$.} Metrics of Precision, Recall, F1 and AUC-ROC metrics are given in Table \ref{tab:imdb_perf_optimal_avg_v1}, the normalized confusion rate matrices to show how the OOD domain has been detected from the ID domain are given in Figure \ref{fig:imdb_confusion_v1}. The plots for the ROC curve and 2-level estimator accuracy are given in Figure \ref{fig:imdb_roc_v1}.

\begin{table}[H]
\footnotesize
  \centering
  \caption{Performance at the optimal \(F_1\) decision threshold. S-SMC\(_\parallel\) (\(P=1\) chain with \(N=10\)), S-HMC\(_{\parallel}\) (\(NP\) chains), DE (\(N\) models) and MAP, with fixed number of leapfrog \(L=1\), \(B\), \(M\), \(v=1\) and \(s=0.35\) (\(5\) realizations, \(\pm\) s.e. in metrics and bold the first $30\%$ data in mean).}
  \label{tab:imdb_perf_optimal_avg_v1}
  \begin{tabular}{ll|c|c|c|c}
    \toprule
    \(P\) & Method & Precision & Recall & F1 & AUC-ROC \\
    \midrule
    – & MAP & 0.733\(\pm\)0.004 & {\bf 0.897\(\pm\)0.003} & 0.807\(\pm\)0.002 & 0.809\(\pm\)0.006 \\
    – & DE & {\bf 0.968\(\pm\)0.004} & 0.880\(\pm\)0.004 & {\bf 0.922\(\pm\)0.001} & {\bf 0.959\(\pm\)0.003} \\
    1 & S-SMC\(_\parallel\) & 0.791\(\pm\)0.000 & 0.871\(\pm\)0.003 & 0.829\(\pm\)0.001 & 0.889\(\pm\)0.001 \\
    8 & S-SMC\(_\parallel\) & 0.947\(\pm\)0.003 & 0.878\(\pm\)0.003 & 0.911\(\pm\)0.000 & 0.944\(\pm\)0.002 \\
    1 & S-HMC\(_\parallel\) & 0.923\(\pm\)0.002 & {\bf 0.885\(\pm\)0.001} & 0.904\(\pm\)0.001 & 0.943\(\pm\)0.002 \\
    8 & S-HMC\(_\parallel\) & {\bf 0.965\(\pm\)0.004} & 0.880\(\pm\)0.003 & {\bf 0.920\(\pm\)0.000} & {\bf 0.948\(\pm\)0.002} \\
    \bottomrule
  \end{tabular}
\end{table}

\begin{figure}[H]
  \centering
  \begin{subfigure}[b]{0.16\textwidth}
    \includegraphics[width=\linewidth]{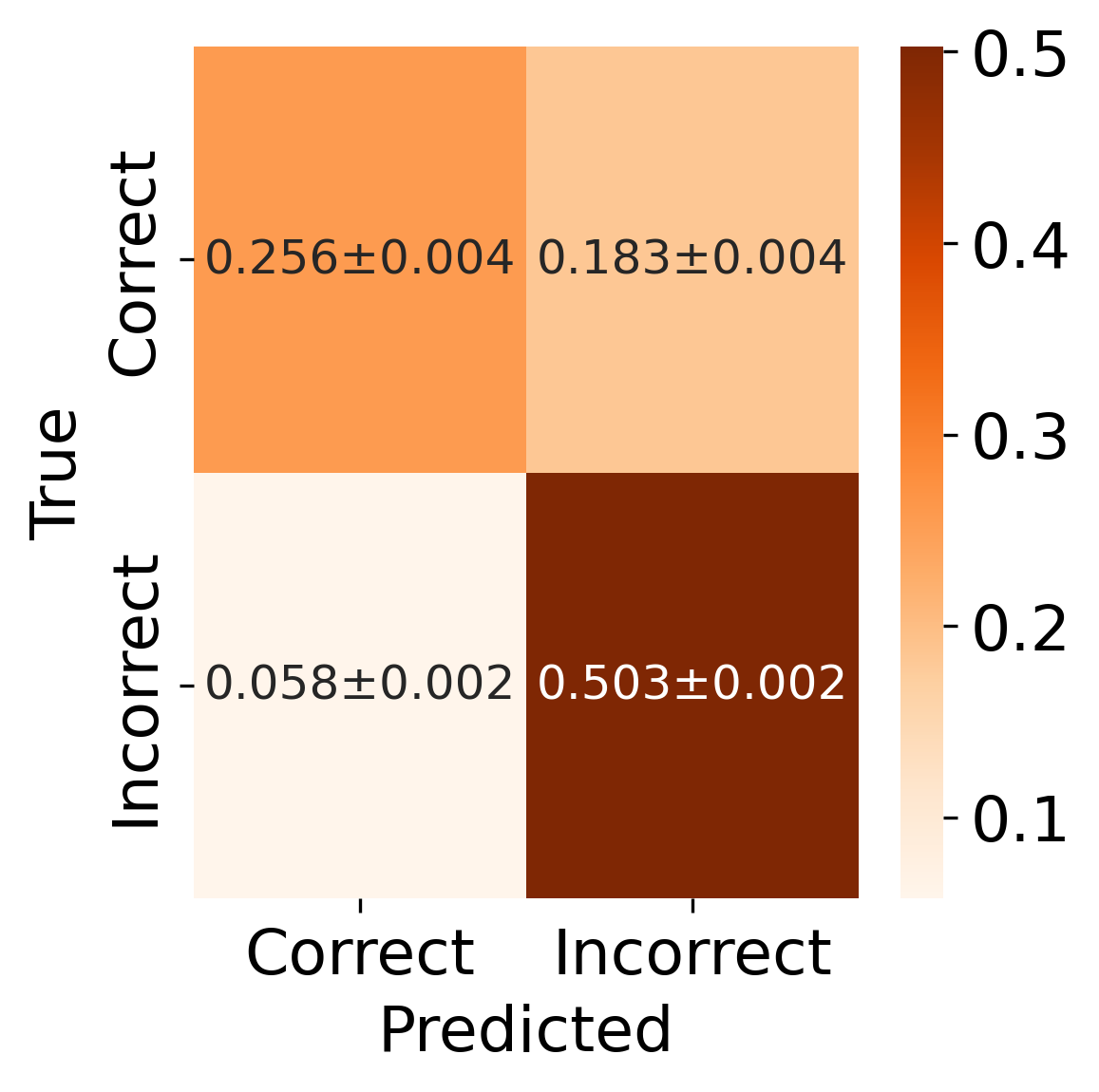}
    \caption{\scriptsize MAP}
  \end{subfigure}
  \hfill
  \begin{subfigure}[b]{0.16\textwidth}
    \includegraphics[width=\linewidth]{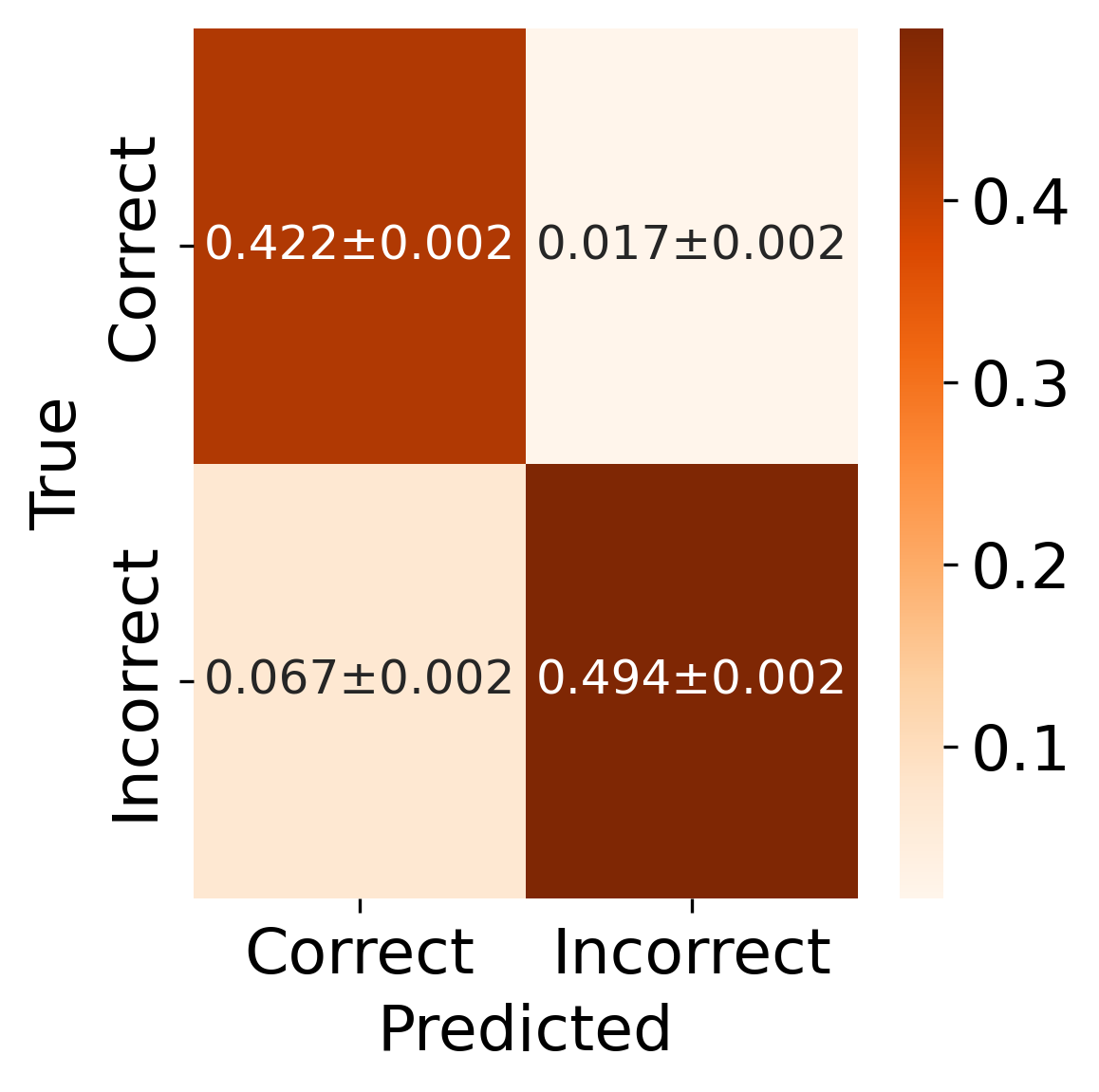}
    \caption{\scriptsize DE}
  \end{subfigure}
  \hfill
  \begin{subfigure}[b]{0.16\textwidth}
    \includegraphics[width=\linewidth]{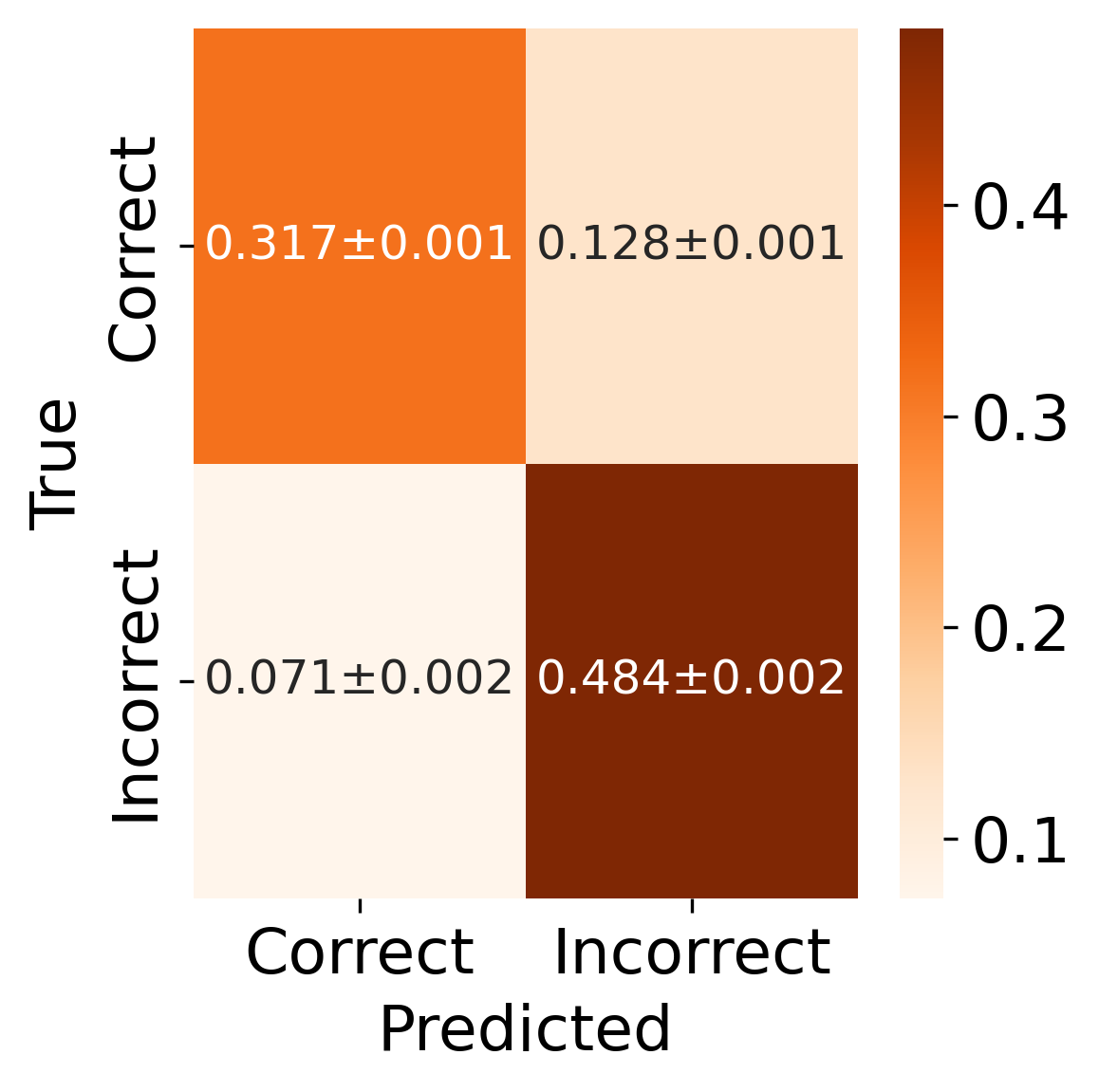}
    \caption{\scriptsize S-SMC\(_\parallel\)(\(P=1\))}
  \end{subfigure}
  \hfill
  \begin{subfigure}[b]{0.16\textwidth}
    \includegraphics[width=\linewidth]{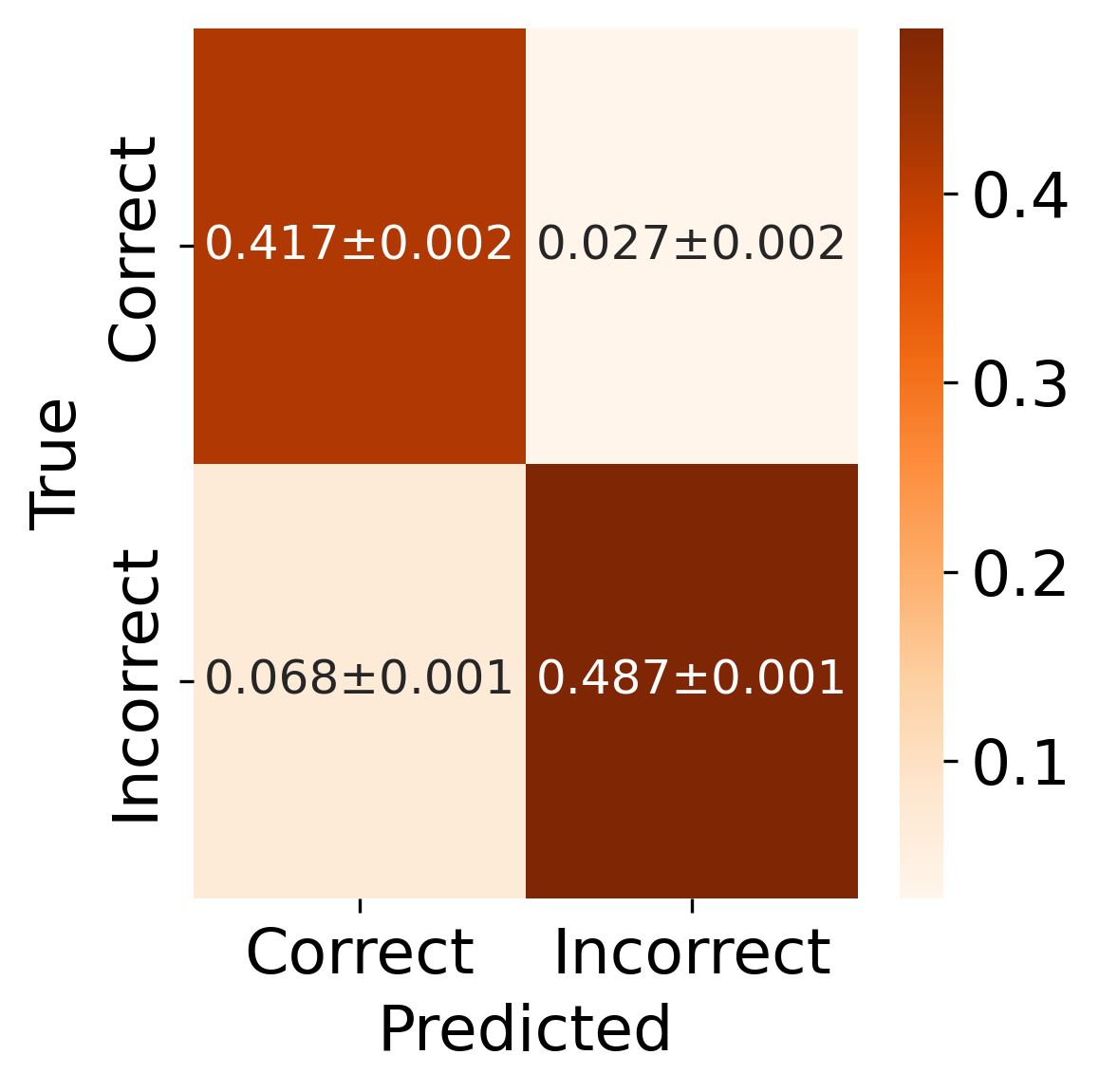}
    \caption{\scriptsize S-SMC\(_\parallel\)(\(P=8\))}
  \end{subfigure}
  \hfill
  \begin{subfigure}[b]{0.16\textwidth}
    \includegraphics[width=\linewidth]{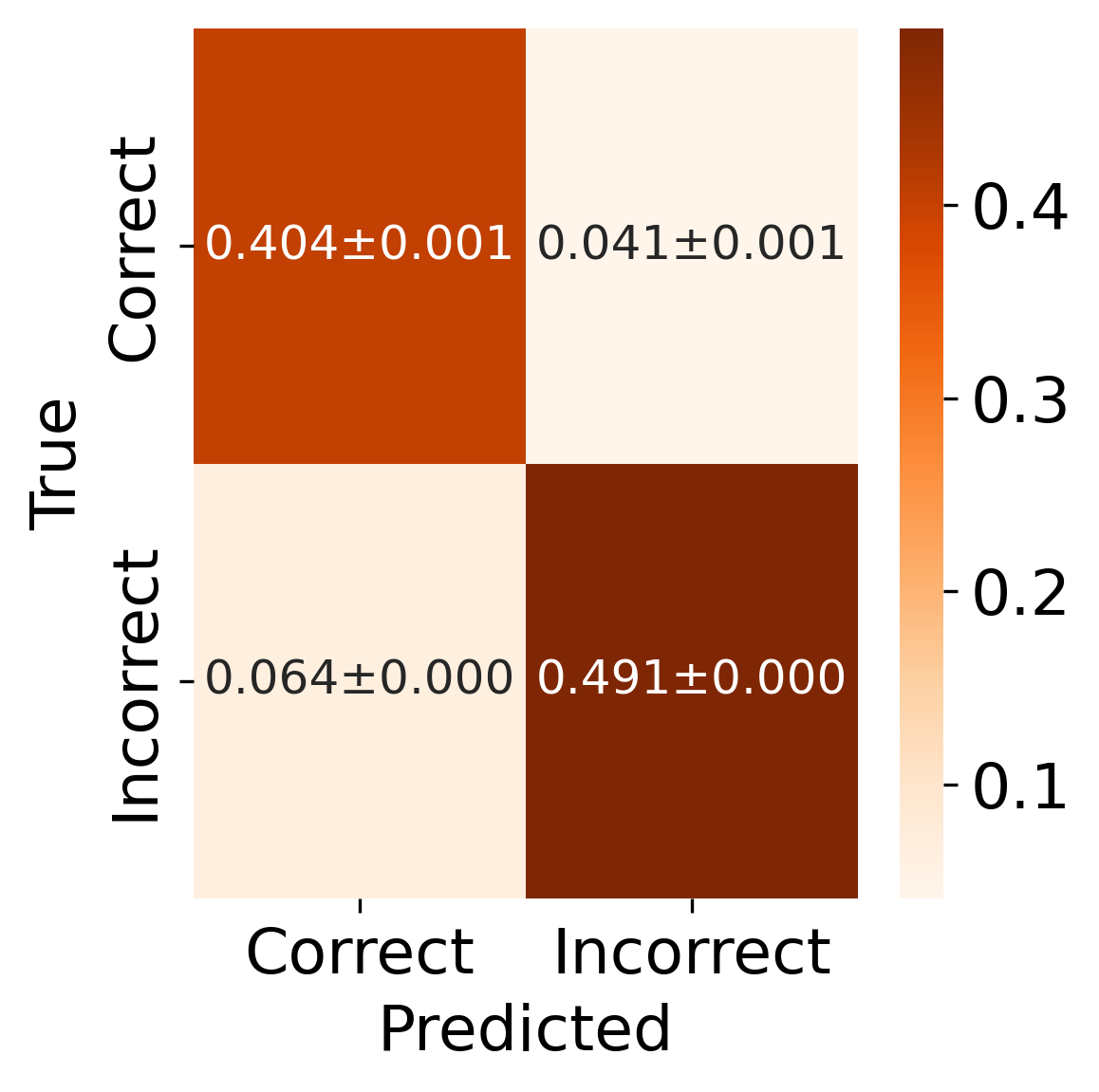}
    \caption{\scriptsize S-HMC\(_\parallel\)(\(P=1\))}
  \end{subfigure}
    \hfill
  \begin{subfigure}[b]{0.16\textwidth}
    \includegraphics[width=\linewidth]{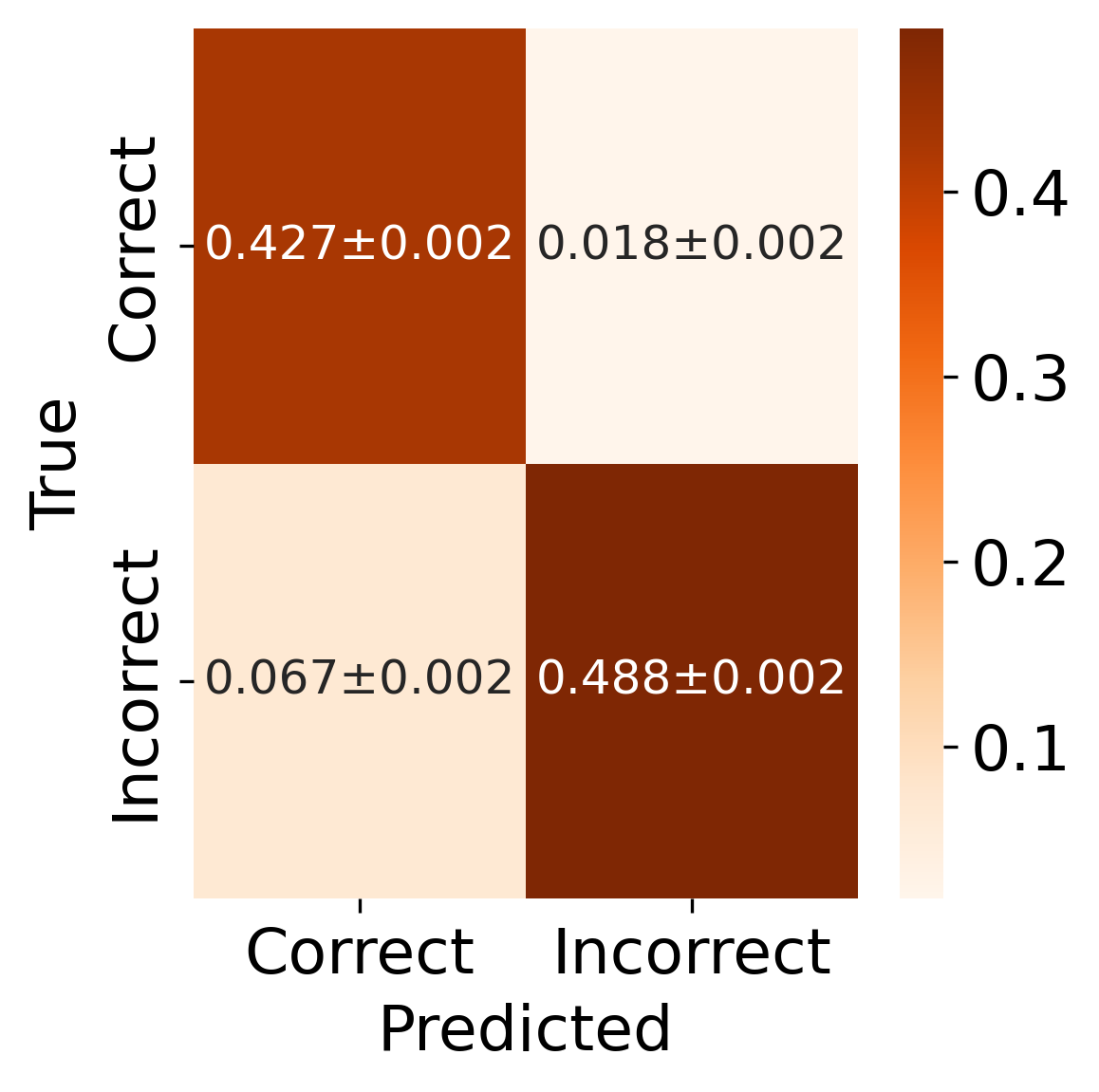}
    \caption{\scriptsize S-HMC\(_\parallel\)(\(P=8\))}
  \end{subfigure}
  \caption{Averaged confusion rate matrices for OOD prediction on IMDb, with optimal \(F_1\) decision threshold. S-SMC\(_\parallel\) (\(P=1,8\) chain with \(N=10\)), S-HMC\(_{\parallel}\) (\(NP\) chains), DE (\(N\) models) and MAP, with fixed number of leapfrog \(L=1\), \(B=26\), \(M=2\), \(v=1\) and \(s=0.35\) (\(5\) realizations and \(\pm\) s.e. in metrics).}
  \label{fig:imdb_confusion_v1}
\end{figure}

\begin{figure}[H]
  \centering
  \begin{subfigure}[b]{0.49\textwidth}
  \centering
    \includegraphics[width=\textwidth]{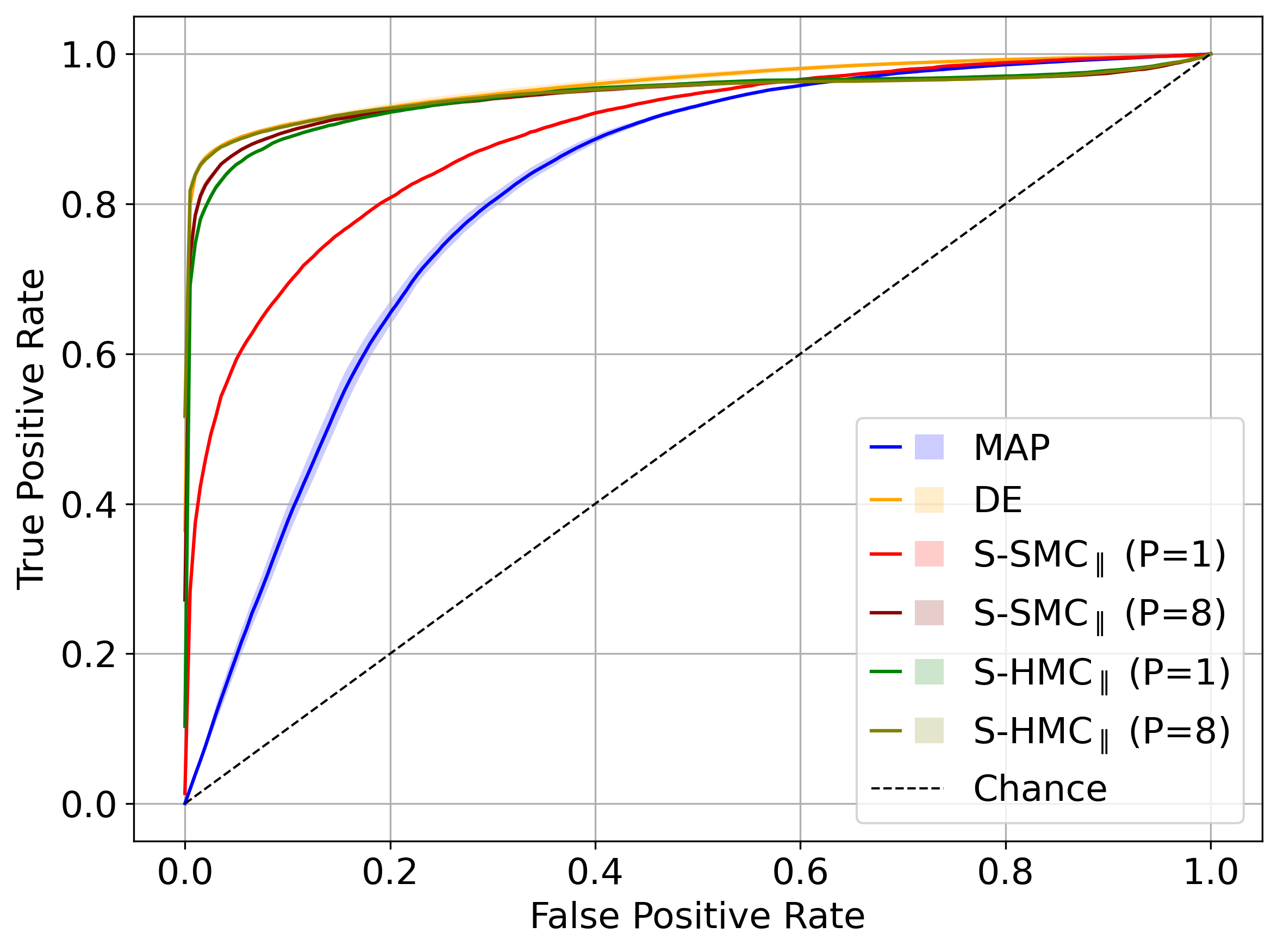}
    \caption{ROC curve}
  \end{subfigure}
  \hfill
  \begin{subfigure}[b]{0.49\textwidth}
  \centering
    \includegraphics[width=\textwidth]{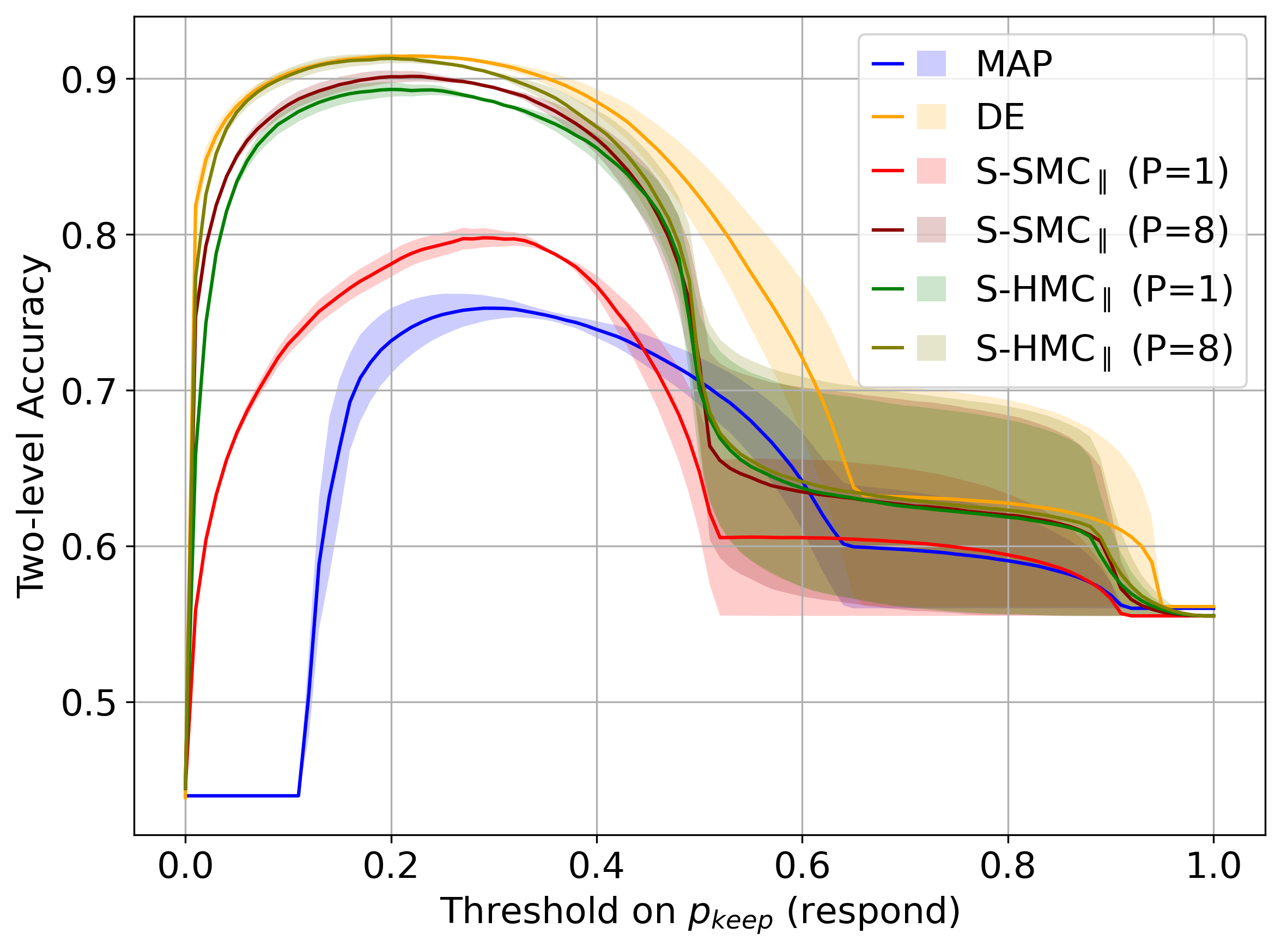}
    \caption{2-level estimator accuracy}
  \end{subfigure}
  \caption{Averaged curve plots for OOD detection in IMDb. S-SMC\(_\parallel\) (\(P=1,8\) chain with \(N=10\)), S-HMC\(_{\parallel}\) (\(NP\) chains), DE (\(N\) models) and MAP, with fixed number of leapfrog \(L=1\), \(B=26\), \(M=2\), \(v=1\) and \(s=0.35\) (\(5\) realizations and \(\pm\) s.e. in metrics).}
  \label{fig:imdb_roc_v1}
\end{figure}

\subsection{CIFAR10}
\label{cifar_ood}

In the CIFAR10 case, the full setting is described in Appendix \ref{app:cifar}, where we let $N_{\sf id}=9000$ and $N_{\sf ood}=9000$, and each dataset has $3000$ data points. Metrics of Precision, Recall, F1 and AUC-ROC metrics are given in Table \ref{tab:cifar_ood}, the normalized confusion rate matrices to show how the OOD domain has been detected from the ID domain are given in Figure \ref{fig:cifar_confusion}. Plots for ROC curve and 2-level estimator accuracy are given in Figure \ref{fig:cifar_roc}.

\begin{figure}[H]
  \centering
    \begin{subfigure}[b]{0.16\textwidth}
    \includegraphics[width=\linewidth]{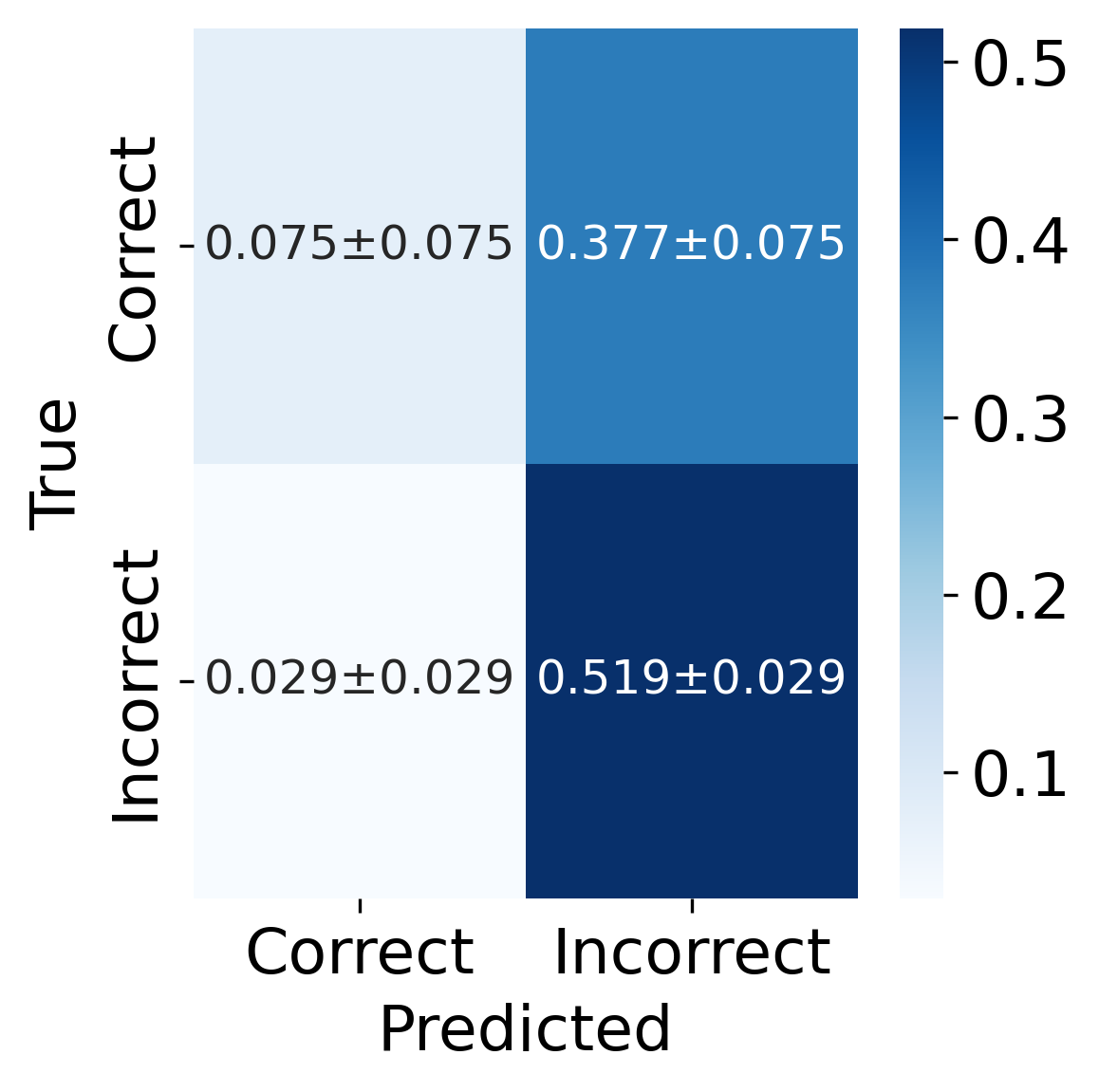}
    \caption{\scriptsize MAP}
  \end{subfigure}
  \hfill
  \begin{subfigure}[b]{0.16\textwidth}
    \includegraphics[width=\linewidth]{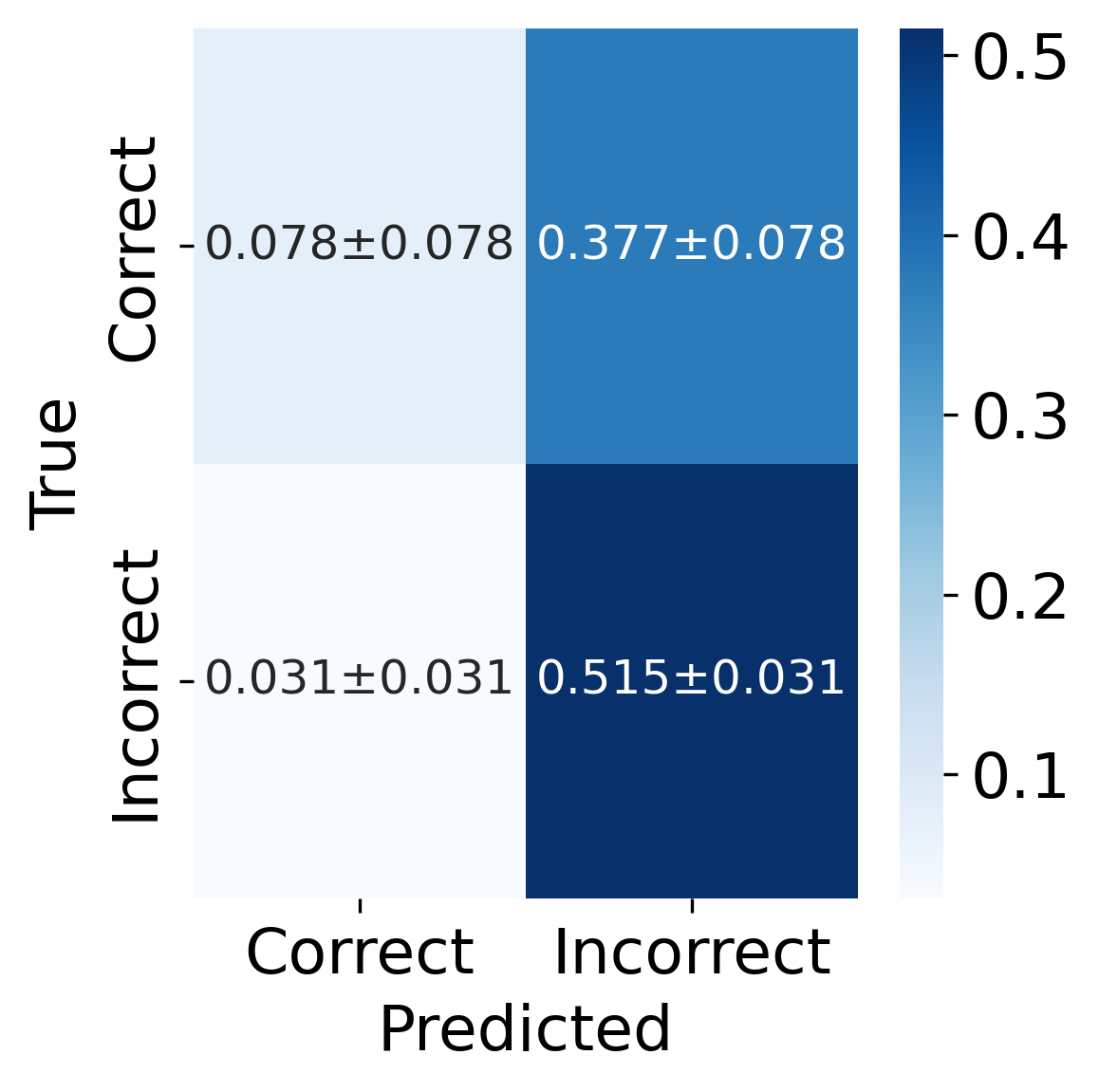}
    \caption{\scriptsize DE}
  \end{subfigure}
  \hfill
  \begin{subfigure}[b]{0.16\textwidth}
    \includegraphics[width=\linewidth]{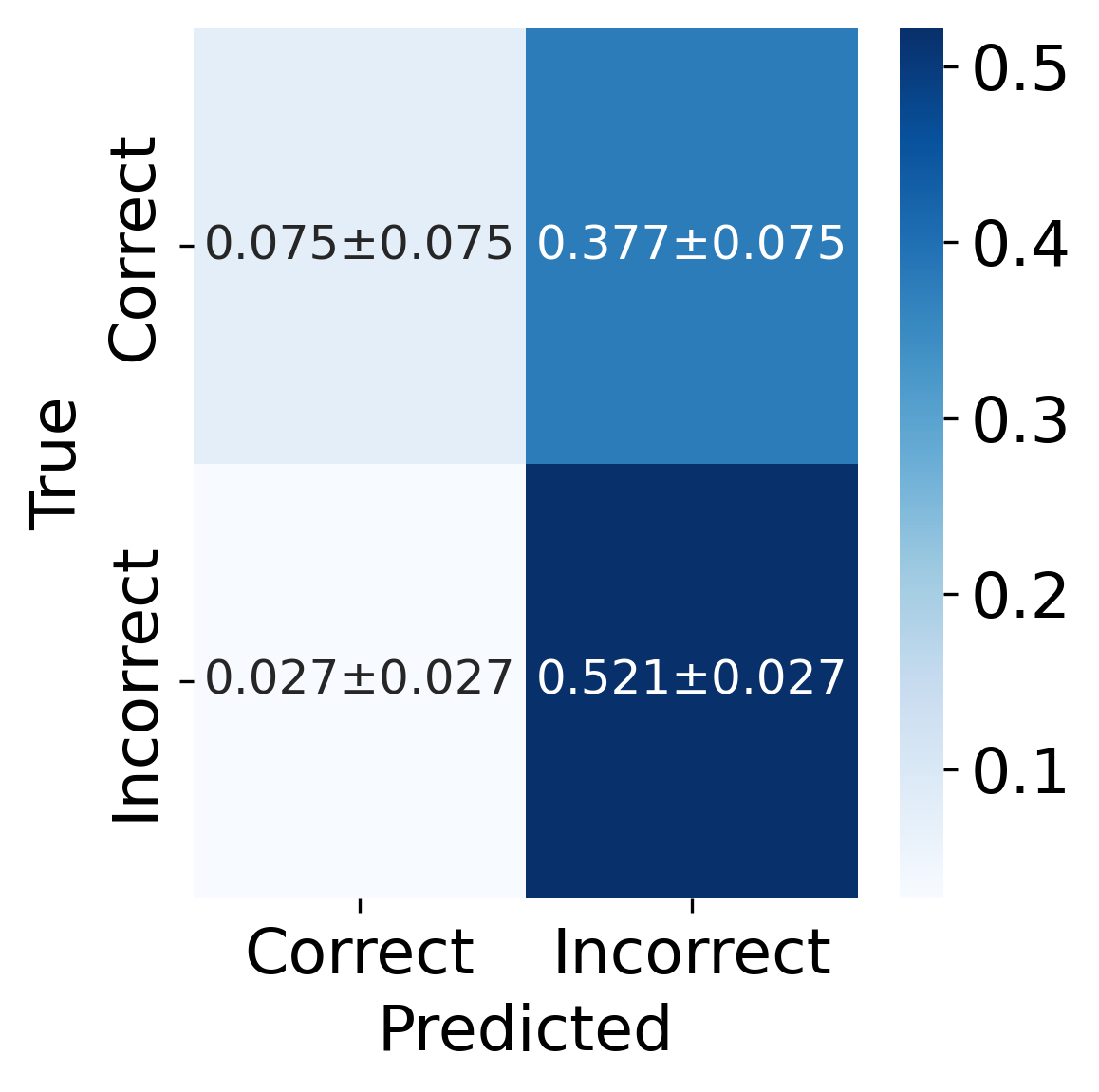}
    \caption{\scriptsize S-SMC\(_\parallel\)(\(P=1\))}
  \end{subfigure}
  \hfill
  \begin{subfigure}[b]{0.16\textwidth}
    \includegraphics[width=\linewidth]{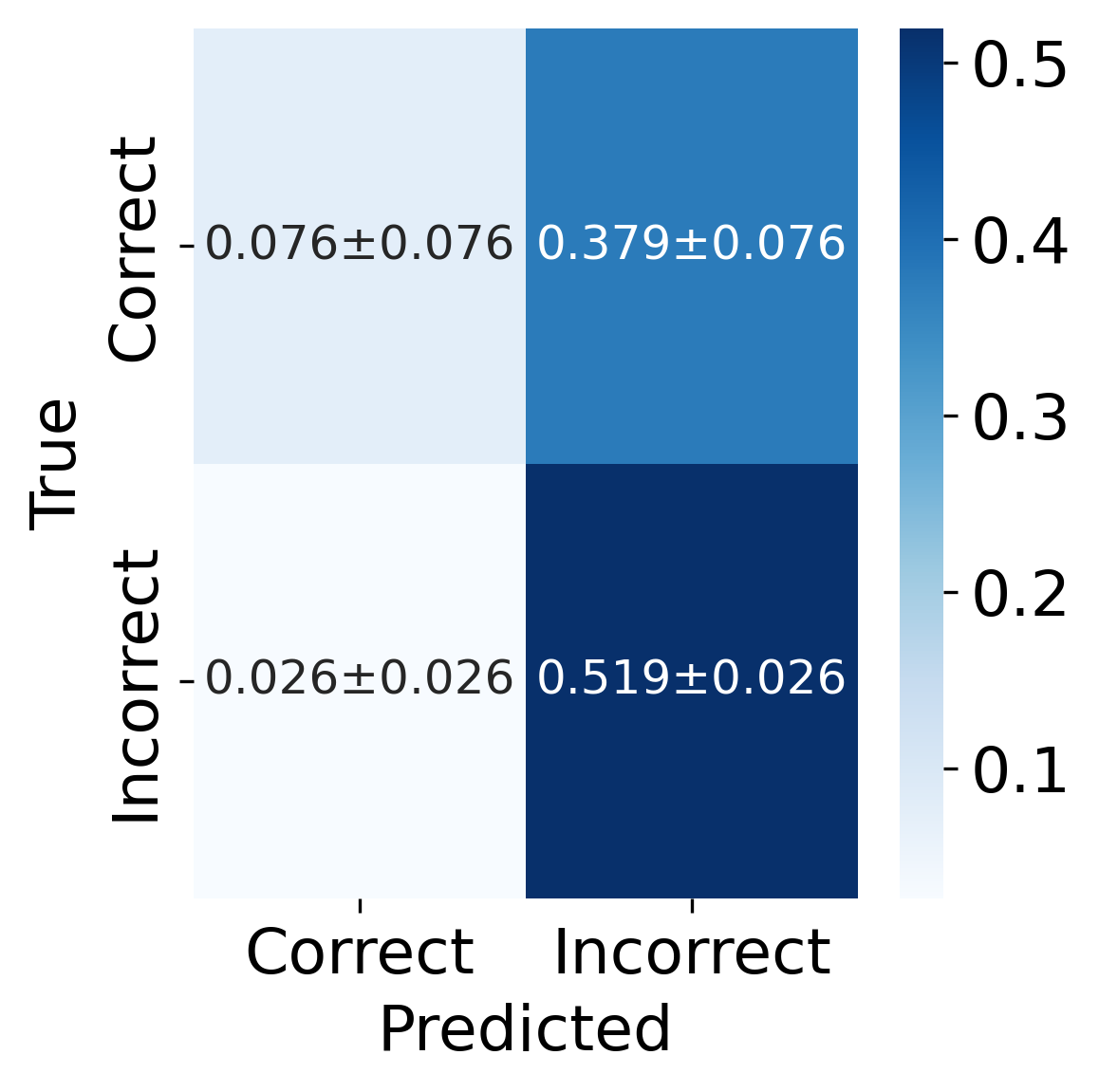}
    \caption{\scriptsize S-SMC\(_\parallel\)(\(P=8\))}
  \end{subfigure}
  \hfill
  \begin{subfigure}[b]{0.16\textwidth}
    \includegraphics[width=\linewidth]{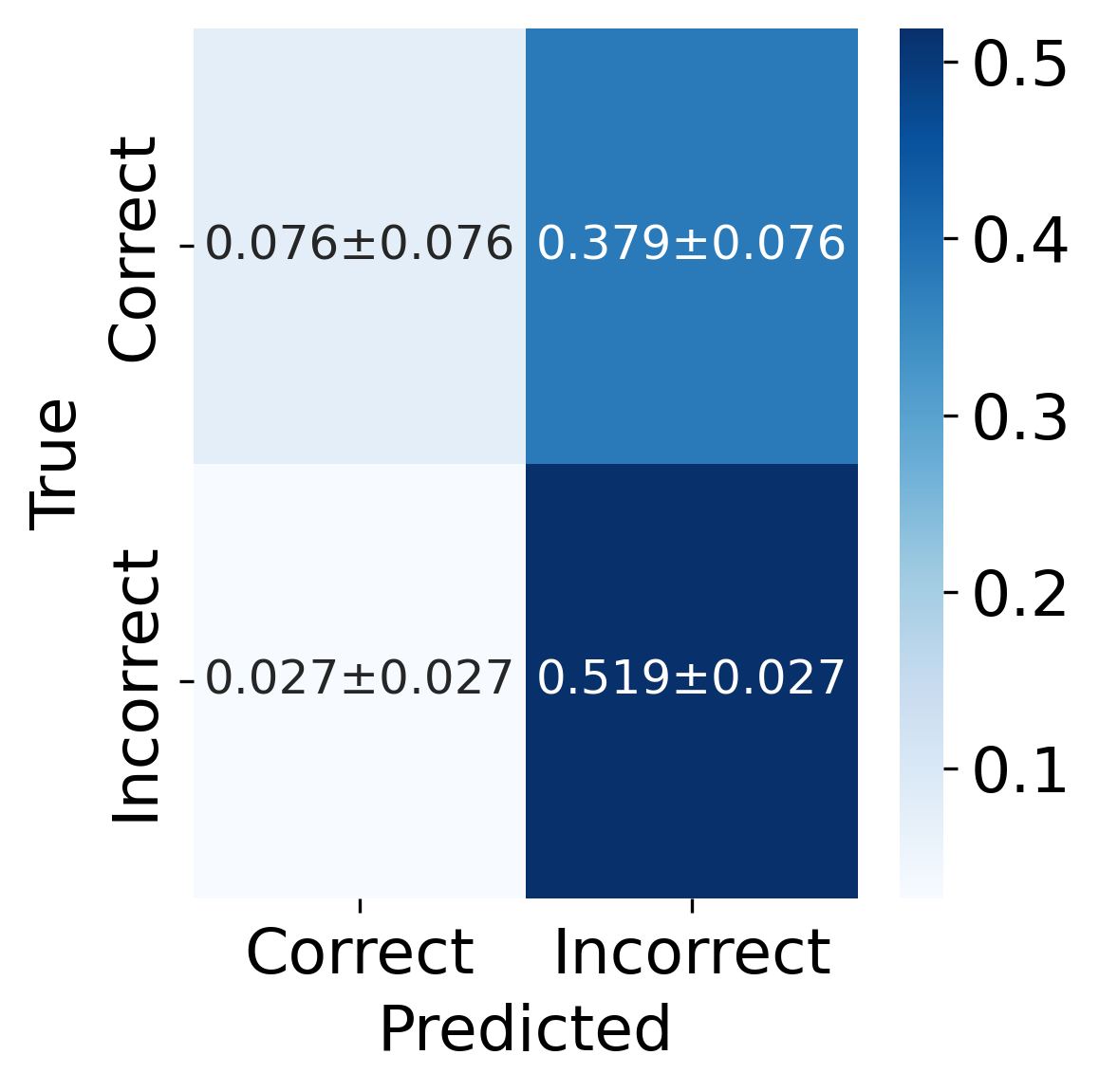}
    \caption{\scriptsize S-HMC\(_\parallel\)(\(P=1\))}
  \end{subfigure}
  \hfill
  \begin{subfigure}[b]{0.16\textwidth}
    \includegraphics[width=\linewidth]{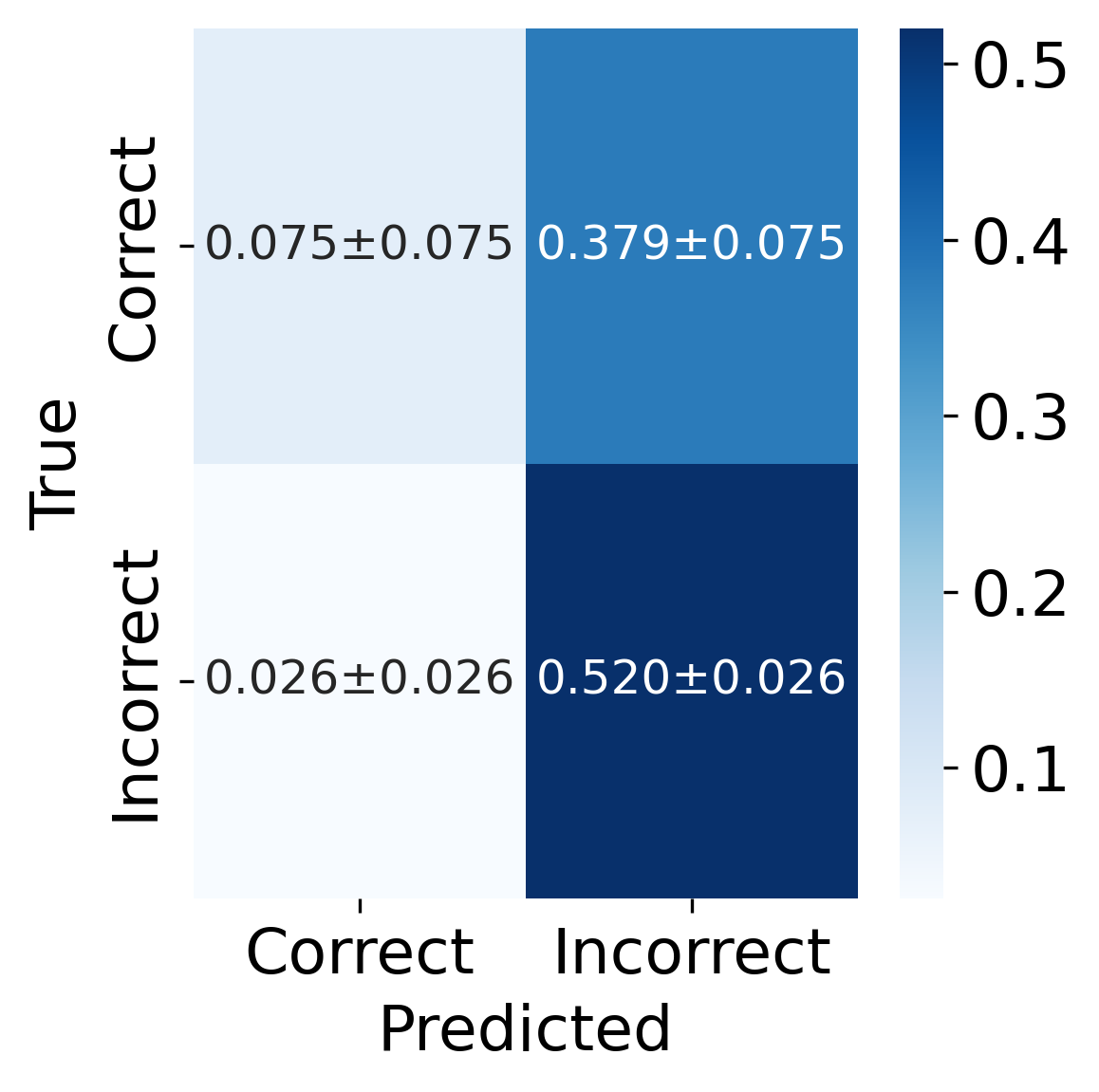}
    \caption{\scriptsize S-HMC\(_\parallel\)(\(P=8\))}
  \end{subfigure}

  \vspace{1em}
  
  \begin{subfigure}[b]{0.16\textwidth}
    \includegraphics[width=\linewidth]{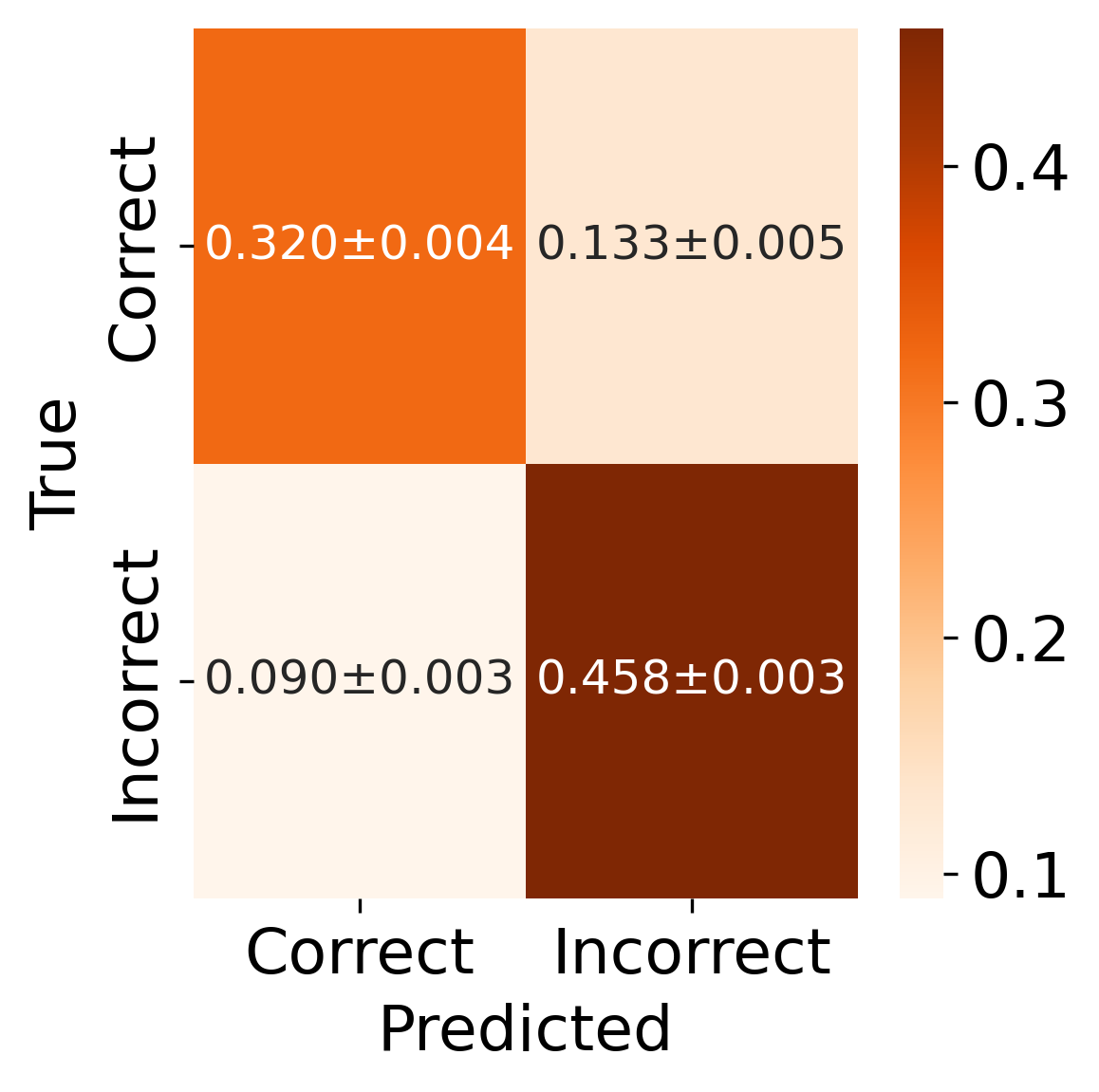}
    \caption{\scriptsize MAP}
  \end{subfigure}
  \hfill
  \begin{subfigure}[b]{0.16\textwidth}
    \includegraphics[width=\linewidth]{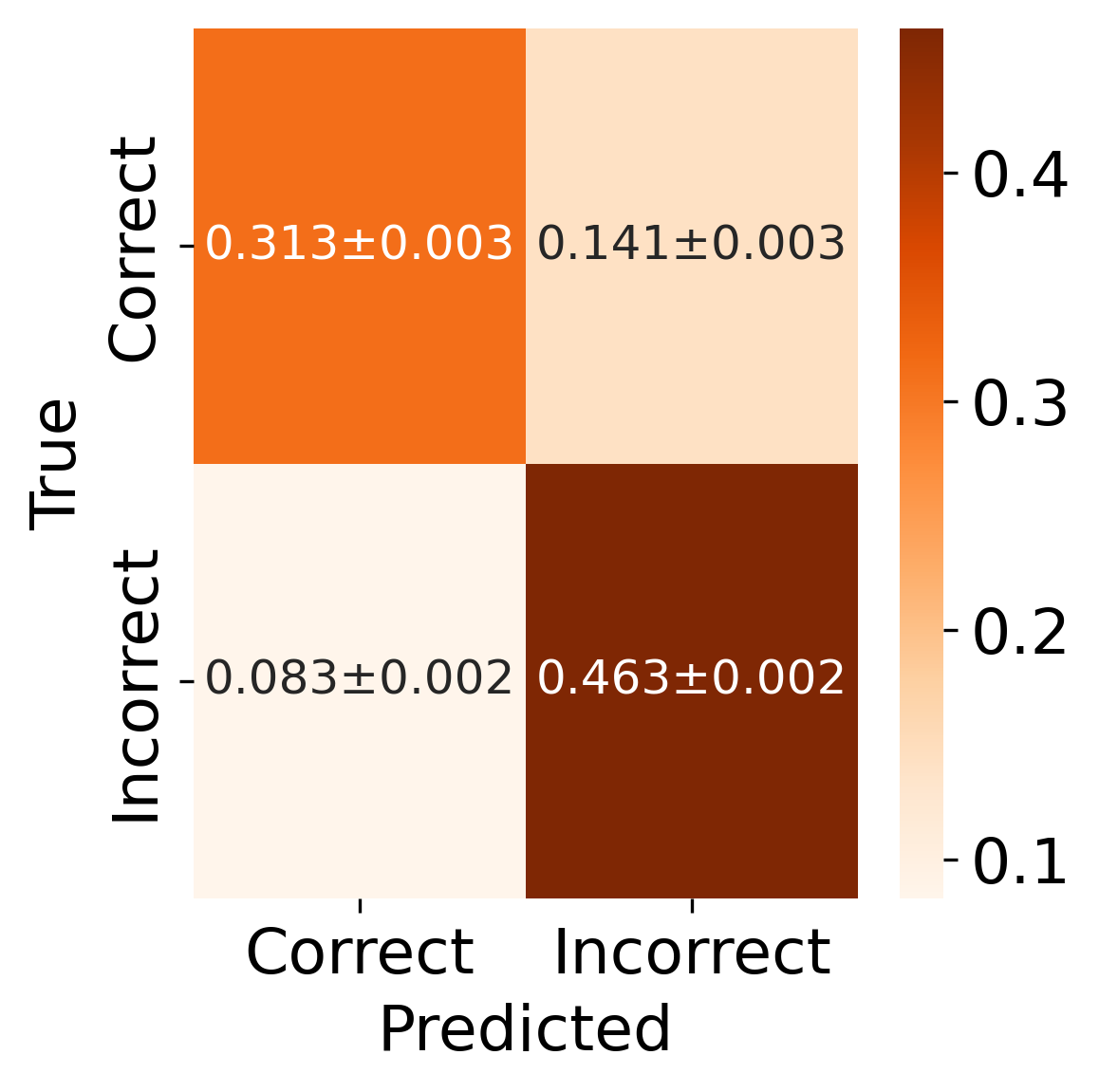}
    \caption{\scriptsize DE}
  \end{subfigure}
  \hfill
  \begin{subfigure}[b]{0.16\textwidth}
    \includegraphics[width=\linewidth]{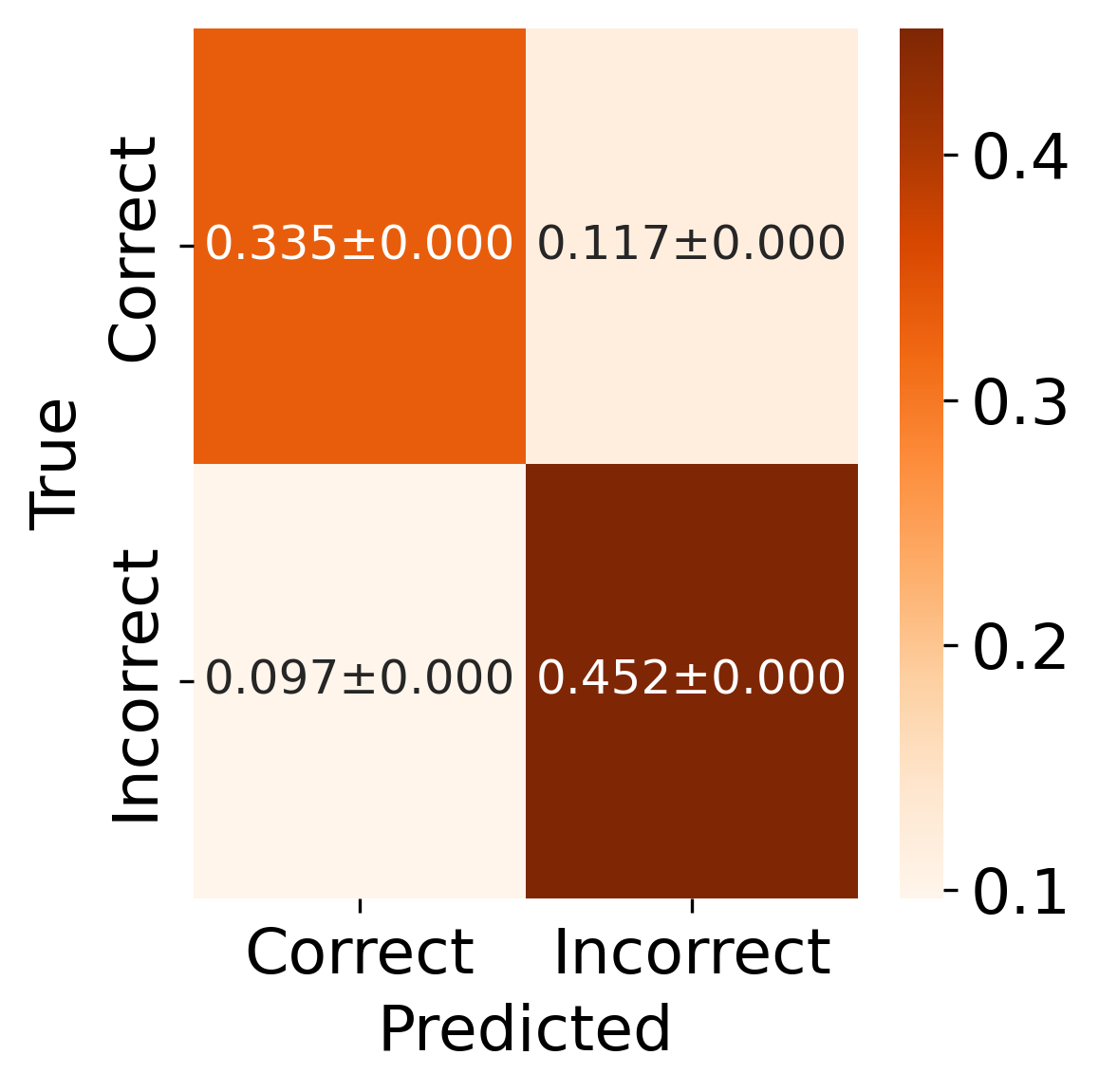}
    \caption{\scriptsize S-SMC\(_\parallel\)(\(P=1\))}
  \end{subfigure}
  \hfill
  \begin{subfigure}[b]{0.16\textwidth}
    \includegraphics[width=\linewidth]{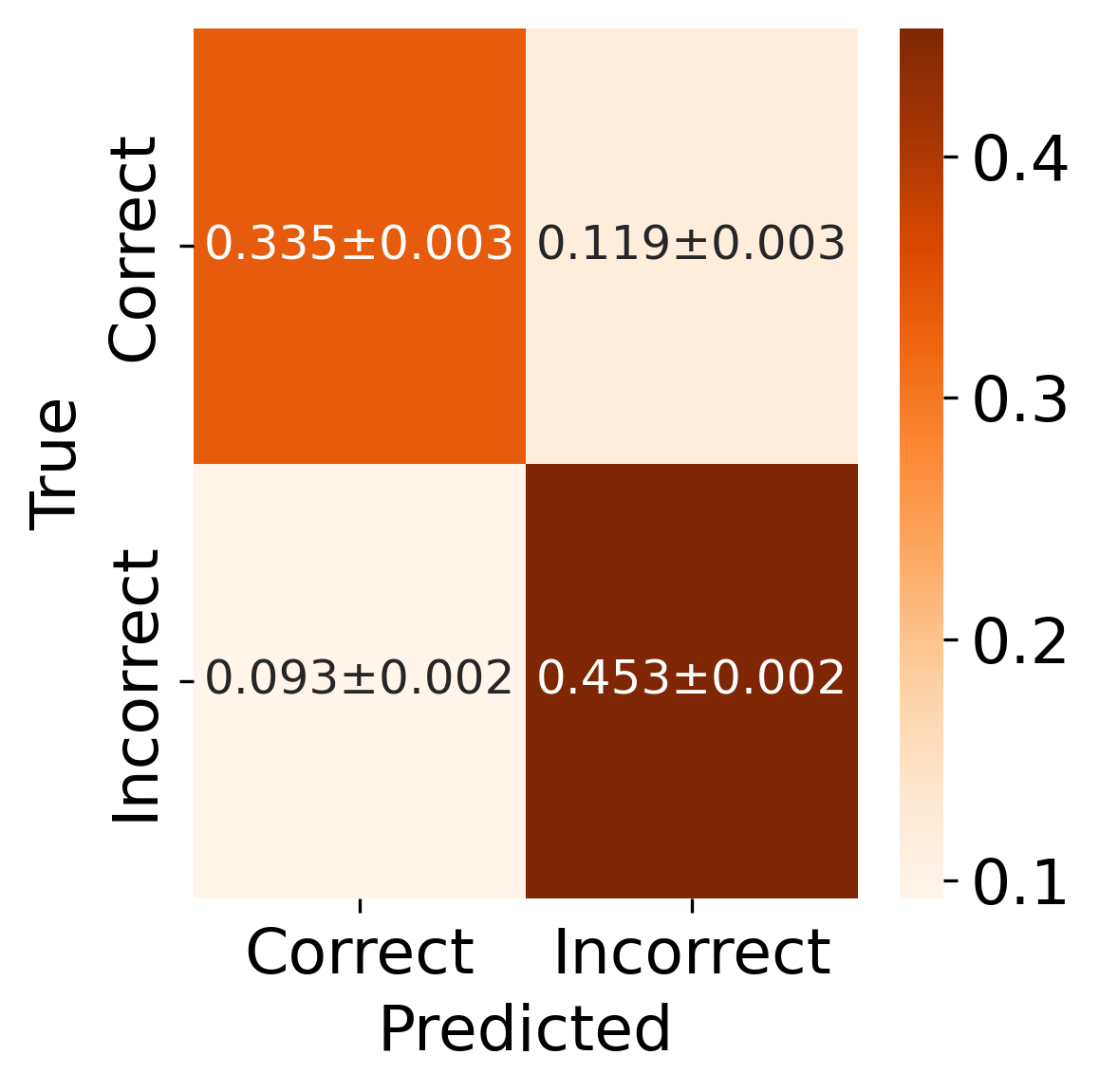}
    \caption{\scriptsize S-SMC\(_\parallel\)(\(P=8\))}
  \end{subfigure}
  \hfill
  \begin{subfigure}[b]{0.16\textwidth}
    \includegraphics[width=\linewidth]{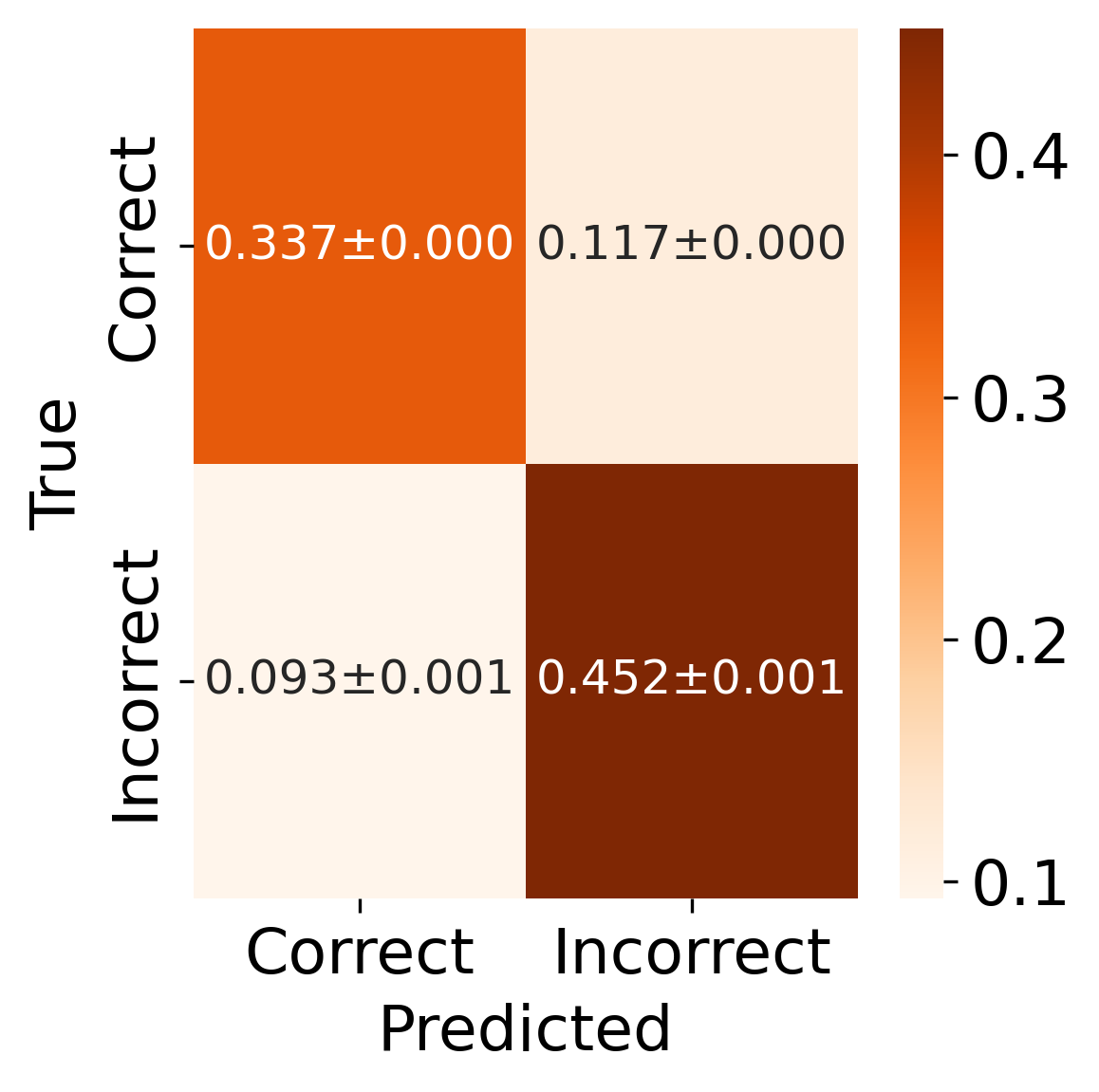}
    \caption{\scriptsize S-HMC\(_\parallel\)(\(P=1\))}
  \end{subfigure}
  \hfill
  \begin{subfigure}[b]{0.16\textwidth}
    \includegraphics[width=\linewidth]{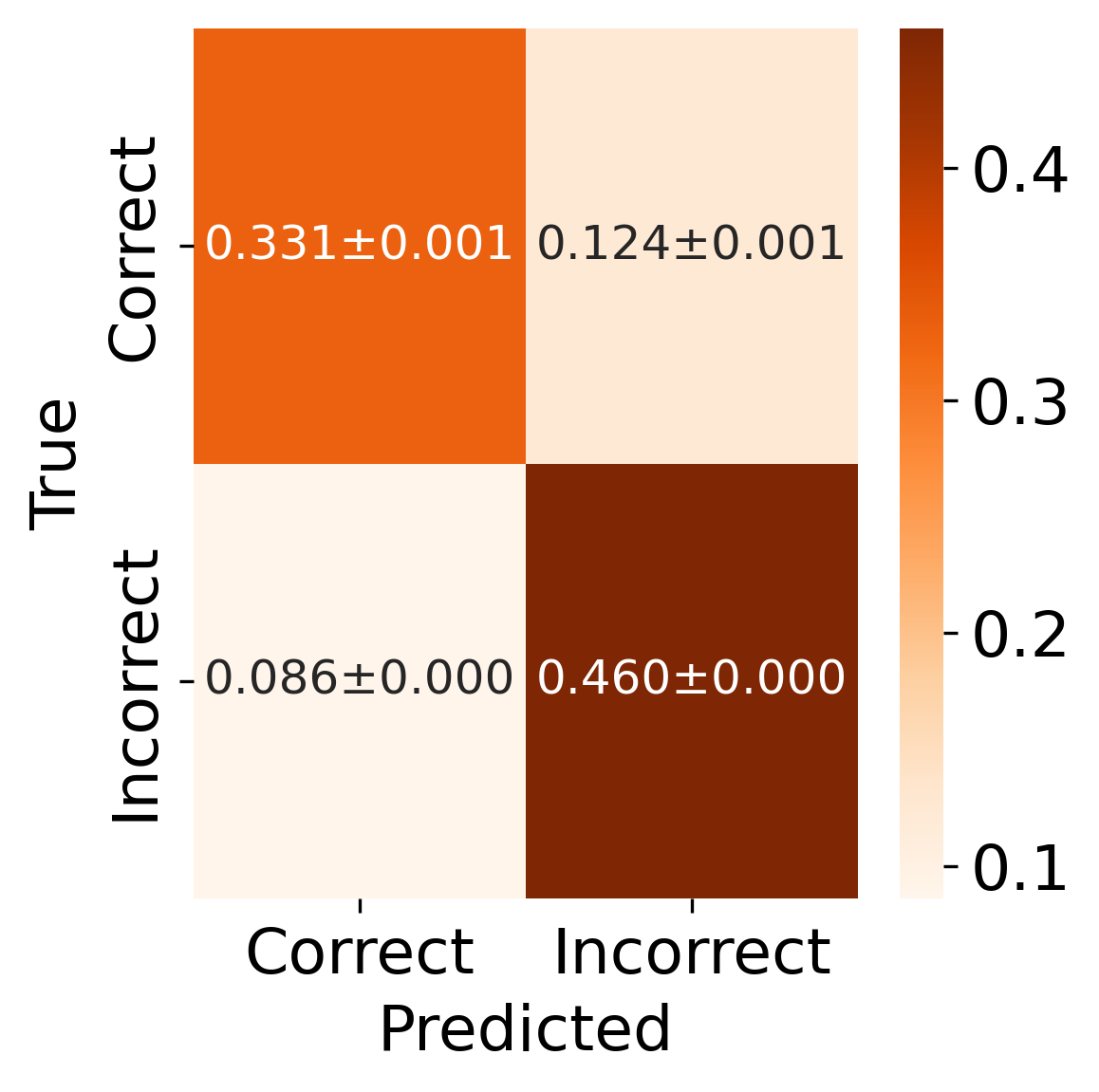}
    \caption{\scriptsize S-HMC\(_\parallel\)(\(P=8\))}
  \end{subfigure}

  \caption{Averaged confusion rate matrices for OOD prediction on CIFAR10, with default decision threshold (top) and optimal \(F_1\) decision threshold (bottom). S-SMC\(_\parallel\) (\(P=1,8\) chains with \(N=10\)), S-HMC\(_{\parallel}\) (\(NP\) chains), DE (\(N\)) and MAP, with fixed number of leapfrog \(L=1\), \(B=200\), \(M=4\), \(v=0.2\) and \(s=0.05\) (\(5\) realizations and \(\pm\) s.e. in metrics).}
  \label{fig:cifar_confusion}
\end{figure}

\begin{table}[H]
  \centering
  \caption{Evaluation Metrics using thresholds. S-SMC\(_\parallel\) (\(P=1,8\) chains with \(N=10\)), S-HMC\(_{\parallel}\) (\(NP\) chains), DE (\(N\)) and MAP, with fixed number of leapfrog \(L=1\), \(B=200\), \(M=4\), \(v=0.2\) and \(s=0.05\) (\(5\) realizations, \(\pm\) s.e. in metrics and bold the first $30\%$ data in mean).}
  \label{tab:cifar_ood}
\begin{subtable}[t]{0.49\textwidth}
    \scriptsize
    \centering
  \subcaption{Default decision threshold (0.5).} 
  \label{tab:cifar_ood_default}
  \begin{adjustbox}{width=\textwidth}
  \begin{tabular}{ll|c|c|c|c}
    \toprule
    \(P\) & Method & Precision & Recall & F1 & AUC-ROC \\
    \midrule
    – & MAP     & 0.606\(\pm\)0.058 & 0.947\(\pm\)0.053 & 0.723\(\pm\)0.015 & 0.856\(\pm\)0.001 \\
    – & DE      & {\bf 0.608\(\pm\)0.062} & 0.943\(\pm\)0.057 & 0.721\(\pm\)0.015 & 0.858\(\pm\)0.002 \\
    1 & S-SMC\(_\parallel\)     & {\bf 0.607\(\pm\)0.059} & 0.951\(\pm\)0.049 & {\bf 0.726\(\pm\)0.018} & 0.861\(\pm\)0.001 \\
    8 & S-SMC\(_\parallel\)     & 0.606\(\pm\)0.060 & {\bf 0.952\(\pm\)0.048} & {\bf 0.725\(\pm\)0.019} & {\bf 0.864\(\pm\)0.000} \\
    1 & S-HMC\(_\parallel\)     & 0.606\(\pm\)0.060 & 0.951\(\pm\)0.049 & 0.724\(\pm\)0.019 & {\bf 0.864\(\pm\)0.000} \\
    8 & S-HMC\(_\parallel\)     & 0.605\(\pm\)0.060 & {\bf 0.953\(\pm\)0.047} & {\bf 0.725\(\pm\)0.019} & {\bf 0.867\(\pm\)0.000} \\
    \bottomrule
  \end{tabular}
  \end{adjustbox}
  \end{subtable}
\hfill
  \begin{subtable}[t]{0.49\textwidth}
    \scriptsize
    \centering
  \subcaption{Optimal \(F_1\) decision threshold.}
  \label{tab:cifar_ood_optimal}
  \begin{adjustbox}{width=\textwidth}
  \begin{tabular}{ll|c|c|c|c}
    \toprule
    \(P\) & Method & Precision & Recall & F1 & AUC-ROC \\
    \midrule
    – & MAP     & 0.776\(\pm\)0.005 & 0.836\(\pm\)0.006 & 0.805\(\pm\)0.001 & 0.856\(\pm\)0.001 \\
    – & DE      & 0.767\(\pm\)0.003 & {\bf 0.848\(\pm\)0.003} & 0.805\(\pm\)0.001 & 0.858\(\pm\)0.002 \\
    1 & S-SMC\(_\parallel\)     & {\bf 0.794\(\pm\)0.000} & 0.824\(\pm\)0.001 & 0.809\(\pm\)0.000 & 0.861\(\pm\)0.001 \\
    8 & S-SMC\(_\parallel\)     & 0.792\(\pm\)0.003 & 0.830\(\pm\)0.003 & {\bf 0.811\(\pm\)0.000} & {\bf 0.864\(\pm\)0.000} \\
    1 & S-HMC\(_\parallel\)     & {\bf 0.794\(\pm\)0.000} & 0.829\(\pm\)0.001 & {\bf 0.811\(\pm\)0.000} & {\bf 0.864\(\pm\)0.000} \\
    8 & S-HMC\(_\parallel\)     & 0.788\(\pm\)0.001 & {\bf 0.842\(\pm\)0.001} & {\bf 0.814\(\pm\)0.000} & {\bf 0.867\(\pm\)0.000} \\
    \bottomrule
  \end{tabular}
  \end{adjustbox}
  \end{subtable}
\end{table}

\begin{figure}[H]
  \centering
  \begin{subfigure}[b]{0.49\textwidth}
  \centering
    \includegraphics[width=\linewidth]{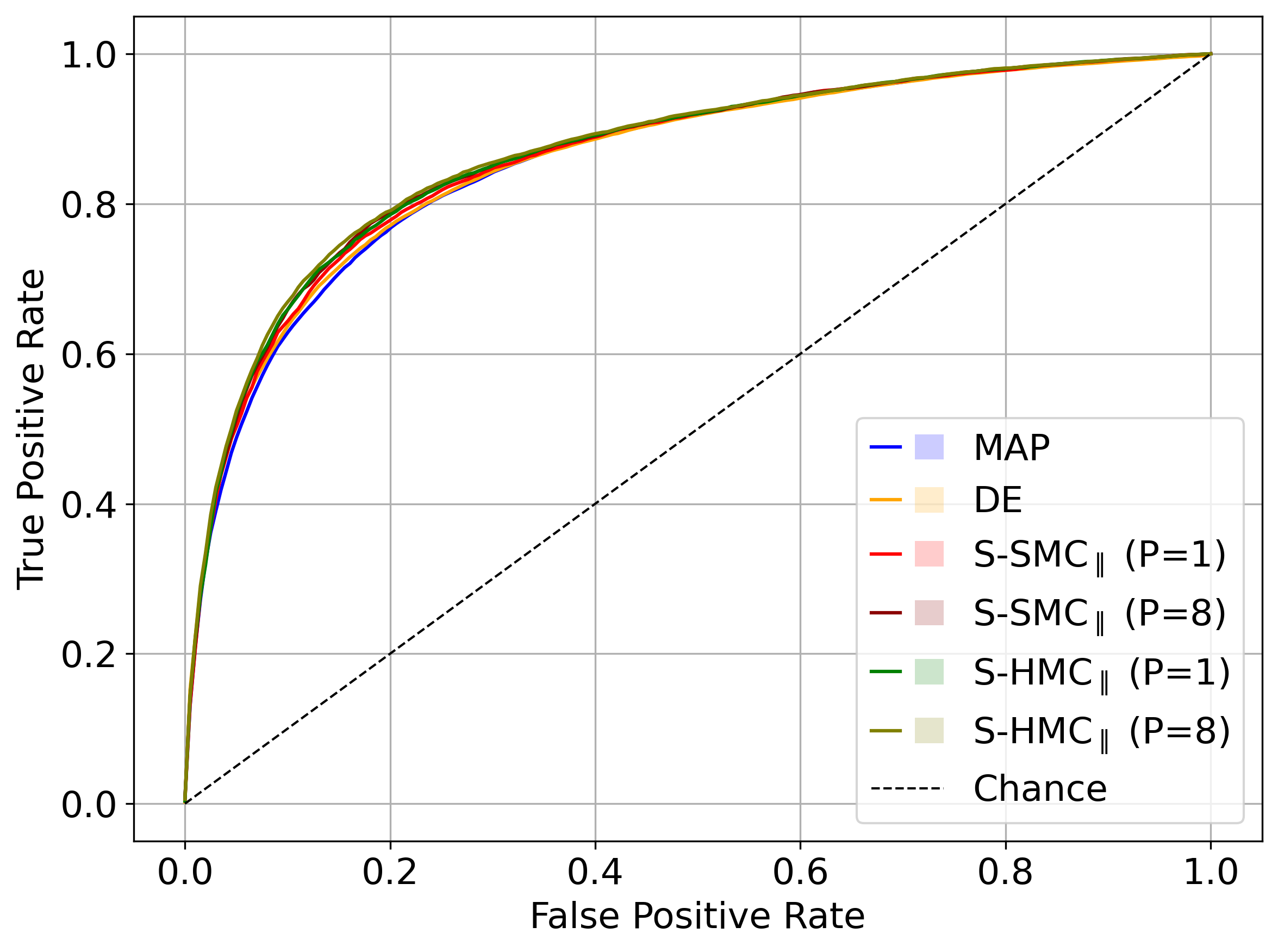}
    \caption{ROC curve}
  \end{subfigure}
  \hfill
  \begin{subfigure}[b]{0.49\textwidth}
  \centering
    \includegraphics[width=\linewidth]{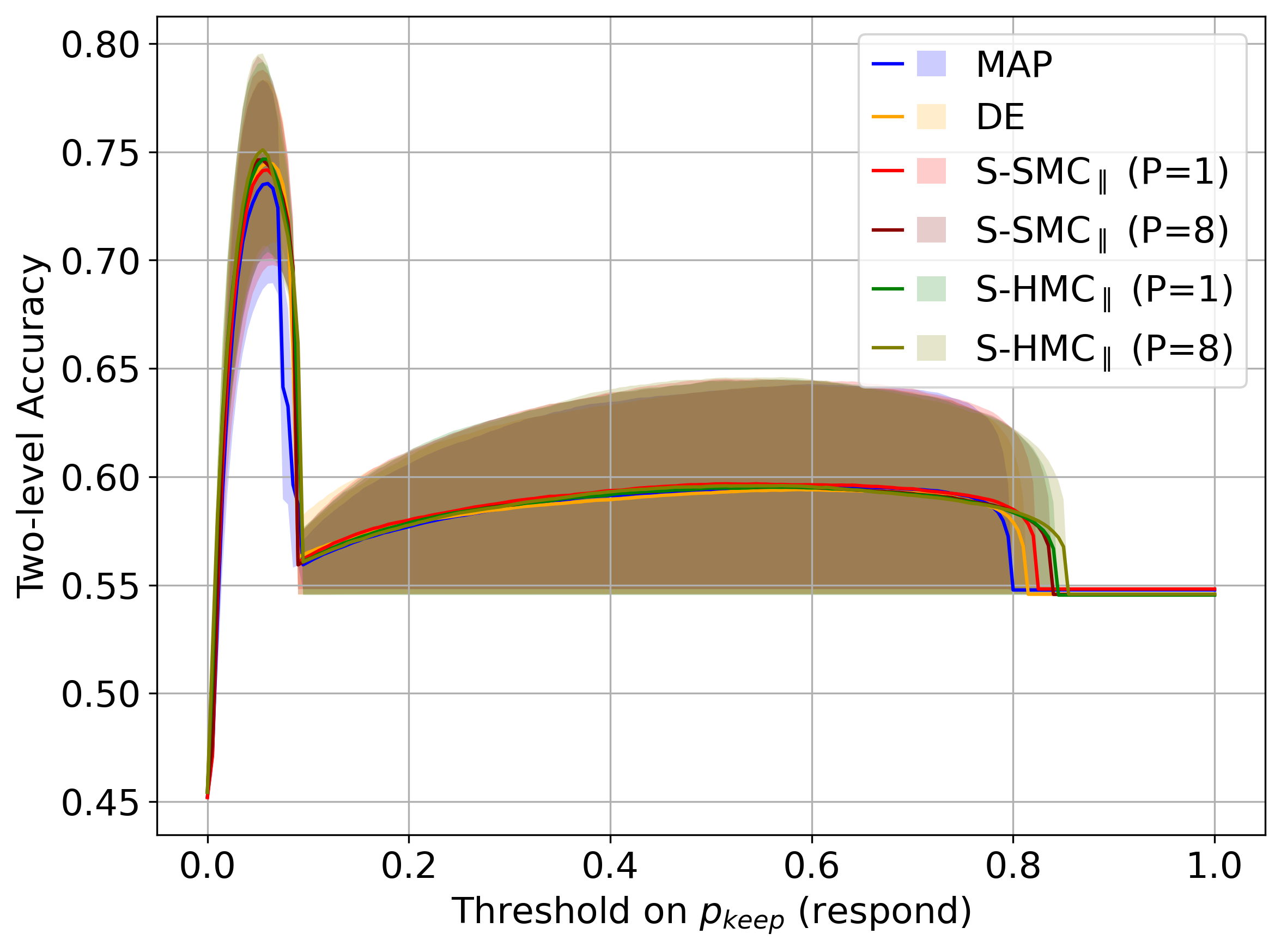}
    \caption{Total accuracy over thresholds}
  \end{subfigure}
  \caption{Averaged curve plots for OOD detection in CIFAR10. S-SMC\(_\parallel\) (\(P=1,8\) chains with \(N=10\)), S-HMC\(_{\parallel}\) (\(NP\) chains), DE (\(N\)) and MAP, with fixed number of leapfrog \(L=1\), \(B=200\), \(M=4\), \(v=0.2\) and \(s=0.05\) (\(5\) realizations and \(\pm\) s.e. in metrics).}
  \label{fig:cifar_roc}
\end{figure}

\section{Further results of Ablations in Practical SBMC($s<\frac{1}{2}$)}
\label{app:ablation}

\subsection{MNIST7}
\label{mnist_ablation}

Experiments in this section are tested on the (filtered) MNIST7 dataset with the model setting stated in Appendix \ref{app:mnist}. 

Table \ref{tab:metric_diffs_mnist} shows the performance as the tuning parameter \(s\) varies. 
Figure \ref{fig:metric_diffP_mnist_s025} and \ref{fig:metric_diffP_mnist_s01} show the trend of the SBMC$_\parallel$ in different values of the tuning parameter $s$ as $P$ increases. Table 
\ref{tab:metric_diffP_mnist_s025} 
 and \ref{tab:metric_diffP_mnist_s01} 
give the corresponding full data results of the below figures. 

\begin{table}[H]
\scriptsize
\caption{Comparison of different \(s\) of (S-)SMC\(_\parallel\) ($P=1,8$ chain with \(N=10\)), (S-)HMC\(_{\parallel}\) (\(NP\) chains), MAP and DE (\(NP\) models), with fixed number of leapfrog \(L=1\) and \(v = 0.1\), on MNIST7 (\(5\) realizations and \(\pm\) s.e. in accuracy).}
\label{tab:metric_diffs_mnist}
\centering
\begin{tabular}{lc|c|c|c|c|cc|cccc}
\toprule
\(s\) & Method & Ep. & Acc. & NL & Brier & \multicolumn{6}{c}{\(H_{\sf ep}\)}\\
\cmidrule{7-12}
& & & & & & \multicolumn{2}{c|}{ID} & \multicolumn{4}{c}{OD} \\
\cmidrule{7-8} \cmidrule{9-12}
& cor. & inc. & 8 & 9 & wn & per. \\
\midrule
\multirow{2}{*}{\(1\)} 
 & HMC (GS) & 2e4 & 93.61\(\pm\)0.41 &  2.224e-1 & 1.015e-1 &  9.621e-2 &  4.097e-1 &  4.614e-1 &  3.119e-1 &  1.126e+0 &  7.919e-1\\ 
 & HMC (GS) & 2e5 &  94.77\(\pm\)0.21 &  1.942e-1 & 8.700e-2 & 1.204e-1 &  4.928e-1 &  5.635e-1 &  4.067e-1 &  1.602e+0 &  1.031e+0 \\
 & HMC (GS) & 1.8e6 &  95.13\(\pm\)0.02 &  1.882e-1 & 8.345e-2 &  1.281e-1 &  5.185e-1 &  5.856e-1 &  4.244e-1 &  1.682e+0 &  1.112e+0 \\
\midrule
\multirow{2}{*}{\(1\)} 
 & SMC\(_\parallel\) & 173.0 & 79.74\(\pm\)2.71 & 6.230e-1 & 2.920e-1 & 1.337e-2 & 3.339e-2 & 3.321e-2 & 2.775e-2 & 6.482e-2 & 5.512e-2\\
 & HMC\(_\parallel\) & 160 & 78.41\(\pm\)2.39 & 1.273e+0 & 5.799e-1 & 3.026e-1 &  3.247e-1 &  3.173e-1 &  2.988e-1 &  6.626e-1 &  1.099e+0 \\
\midrule
\multirow{2}{*}{\(0.5\)} 
 & S-SMC\(_\parallel\) & 161.0 & 84.18\(\pm\)0.64 & 4.827e-1 & 2.304e-1 & 1.556e-2 & 4.082e-2 & 4.000e-2 & 3.234e-2 & 1.238e-1 & 6.418e-2\\
 & S-HMC\(_\parallel\) & 160 & 85.26\(\pm\)1.06 & 8.234e-1 & 3.672e-1 & 2.993e-1 &  3.704e-1 &  3.627e-1 &  3.271e-1 &  8.832e-1 &  8.030e-1\\
\midrule
\multirow{4}{*}{\(0.25\)} 
 & S-SMC\(_\parallel\) & 166.6 & 90.35\(\pm\)0.26 & 3.300e-1 & 1.441e-1 &  2.257e-2 & 1.094e-1 & 1.146e-1 & 7.791e-2 & 3.888e-1 & 1.996e-1\\
  & $P=8$ & 161.5 & 93.00\(\pm\)0.11 & 2.366e-1 & 1.096e-1 & 8.828e-2 & 3.717e-1 & 2.984e-1 & 2.089e-1 & 7.488e-1 & 4.585e-1\\
 & S-HMC\(_\parallel\) & 160 & 92.79\(\pm\)0.19 & 2.571e-1 & 1.156e-1 &  1.133e-1 &  4.232e-1 &  4.985e-1 &  3.225e-1 &  1.289e+0 &  6.280e-1\\
   & $P=8$ & 160 & 93.15\(\pm\)0.05 & 2.490e-1 & 1.127e-1 & 1.384e-1 & 4.788e-1 & 5.572e-1 & 3.678e-1 & 1.349e+0 & 7.311e-1 \\
\midrule
\multirow{4}{*}{\(0.1\)} 
 & S-SMC\(_\parallel\) & 170.0 & 92.17\(\pm\)0.37 & 2.671e-1 & 1.186e-1 &  2.642e-2 &  1.288e-1 & 1.384e-1 & 9.406e-2 & 3.943e-1 & 1.832e-1\\
   & $P=8$ & 178.0 & 93.26\(\pm\)0.16 & 2.259e-1 & 1.025e-1  & 5.871e-2 & 2.725e-1 & 2.440e-1 & 1.637e-1 & 7.238e-1 & 3.823e-1 \\
 & S-HMC\(_\parallel\) & 160 & 92.96\(\pm\)0.17 & 2.326e-1 & 1.071e-1 &  5.624e-2 &  2.645e-1 &  3.072e-1 &  1.941e-1 &  9.304e-1 &  4.216e-1\\
   & $P=8$ & 160 & 93.12\(\pm\)0.08 & 2.310e-1 & 1.072e-1 & 6.982e-2 & 2.993e-1 & 3.524e-1 & 2.258e-1 & 1.067e+0 & 4.780e-1 \\
\midrule
\multirow{2}{*}{\(0.01\)} 
 & S-SMC\(_\parallel\) & 183.6 &  92.57\(\pm\)0.37 &  2.439e-1 & 1.121e-1 & 1.149e-2 & 5.904e-2 & 6.445e-2 & 4.602e-2 & 2.187e-1 & 1.008e-1\\
 & S-HMC\(_\parallel\) & 162 &  92.95\(\pm\)0.10 &  2.289e-1 & 1.069e-1 & 1.912e-2 & 1.015e-1 & 1.238e-1 & 7.814e-2 & 4.678e-1 & 1.945e-1\\
\midrule
\multirow{3}{*}{\(0\)} 
 & MAP & 160.2 & 92.32\(\pm\)0.37 & 2.527e-1 & 1.163e-1 & 0 & 0 & 0 & 0 & 0 & 0\\
 & DE (\(N\)) & 176.5 & 92.40\(\pm\)0.15 & 2.455e-1 & 1.148e-1 & 1.059e-2 & 5.646e-2 & 7.433e-2 & 3.468e-2 &  2.690e-1 & 1.1056e-1 \\
 & DE (\(8N\)) & 178.38 & 92.54\(\pm\)0.06 & 2.393e-1 & 1.124e-1  & 1.111e-2 & 5.980e-2 & 7.846e-2 & 4.016e-2 & 2.935e-1 & 1.188e-1\\
\bottomrule
\end{tabular}

\vspace{0.2cm}

\begin{tabular}{lc|cc|cccc}
\toprule
\(s\) & Method & \multicolumn{6}{c}{\(H_{\sf tot}\)} \\
\cmidrule{3-8}
& & \multicolumn{2}{c|}{ID} & \multicolumn{4}{c}{OD} \\
\cmidrule{3-4} \cmidrule{5-8}
& & cor. & inc. & 8 & 9 & wn & per.\\
\midrule
\multirow{2}{*}{\(1\)} 
 & HMC (GS) &  2.621e-1 &  9.652e-1 &  1.110e+0 &  8.198e-1 &  1.492e+0 &  1.081e+0 \\
 & HMC (GS) &  2.852e-1 &  1.033e+0 &  1.204e+0 &  9.322e-1 &  1.915e+0 &  1.296e+0 \\ 
 & HMC (GS) &  2.948e-1 &  1.057e+0 &  1.223e+0 &  9.532e-1 &  2.012e+0 &  1.384e+0  \\
\midrule
\multirow{2}{*}{\(1\)} 
 & SMC\(_\parallel\) & 5.506e-1 &  1.078e+0 &  1.138e+0 &  9.851e-1 & 6.426e-1 &  9.171e-1 \\
 & HMC\(_\parallel\) & 1.854e+0 & 1.962e+0 & 1.988e+0 & 1.927e+0 &  1.965e+0 &  1.844e+0 \\
\midrule
\multirow{2}{*}{\(0.5\)} 
 & S-SMC\(_\parallel\)  &  4.363e-1 &  1.019e+0 &  1.127e+0 &  9.294e-1 & 8.128e-1 & 8.712e-1 \\
 & S-HMC\(_\parallel\) & 1.427e+0 & 1.752e+0 & 1.837e+0 & 1.667e+0 &  1.857e+0 &  1.694e+0 \\
\midrule
\multirow{4}{*}{\(0.25\)} 
 & S-SMC\(_\parallel\) & 1.508e-1 & 6.679e-1 & 8.384e-1 & 5.945e-1 & 8.495e-1 & 7.445e-1 \\
   & $P=8$  & 1.247e-1 & 6.641e-1 & 1.000e+0 & 7.354e-1 & 1.177e+0 & 9.931e-1 \\
 & S-HMC\(_\parallel\) &  3.149e-1 &  1.026e+0 &  1.220e+0 &  8.606e-1 &  1.624e+0 &  1.019e+0 \\
   & $P=8$ & 3.456e-1 & 1.070e+0 & 1.267e+0 & 9.025e-1 & 1.714e+0 & 1.111e+0 \\
\midrule
\multirow{4}{*}{\(0.1\)} 
 & S-SMC\(_\parallel\) & 1.536e-1 & 7.042e-1 & 8.975e-1 & 6.591e-1 & 9.743e-1 & 8.001e-1 \\
   & $P=8$ & 1.374e-1 & 7.075e-1 & 1.001e+0 & 7.307e-1 & 1.216e+0 & 9.805e-1 \\
 & S-HMC\(_\parallel\) &  2.343e-1 &  9.132e-1 &  1.091e+0 & 7.567e-1 &  1.452e+0 & 8.443e-1 \\
   & $P=8$  & 2.553e-1 & 9.380e-1 & 1.127e+0 & 7.893e-1 & 1.543e+0 & 8.937e-1 \\
\midrule
\multirow{2}{*}{\(0.01\)} 
 & S-SMC\(_\parallel\) & 1.737e-1 & 7.571e-1 & 9.632e-1 & 6.607e-1 & 9.254e-1 & 8.511e-1 \\
 & S-HMC\(_\parallel\) &  1.995e-1 & 8.232e-1 & 1.003e+0 & 6.991e-1 & 1.234e+0 & 6.726e-1 \\
\midrule
\multirow{3}{*}{\(0\)} 
 & MAP & 1.833e-1 & 7.645e-1 & 9.507e-1 & 6.157e-1 & 7.682e-1 & 7.839e-1 \\
 & DE (\(N\))  & 1.919e-1 & 7.899e-1 & 9.821e-1 &  6.393e-1 & 9.806e-1 & 8.112e-1  \\
 & DE (\(8N\)) &  1.938e-1 & 7.988e-1 & 1.001e+0 & 6.532e-1  & 1.067e+0 & 8.133e-1  \\
\bottomrule
\end{tabular}
\end{table}

\begin{figure}[H]
  \centering
  \begin{subfigure}[b]{0.32\textwidth}
    \includegraphics[width=\linewidth]{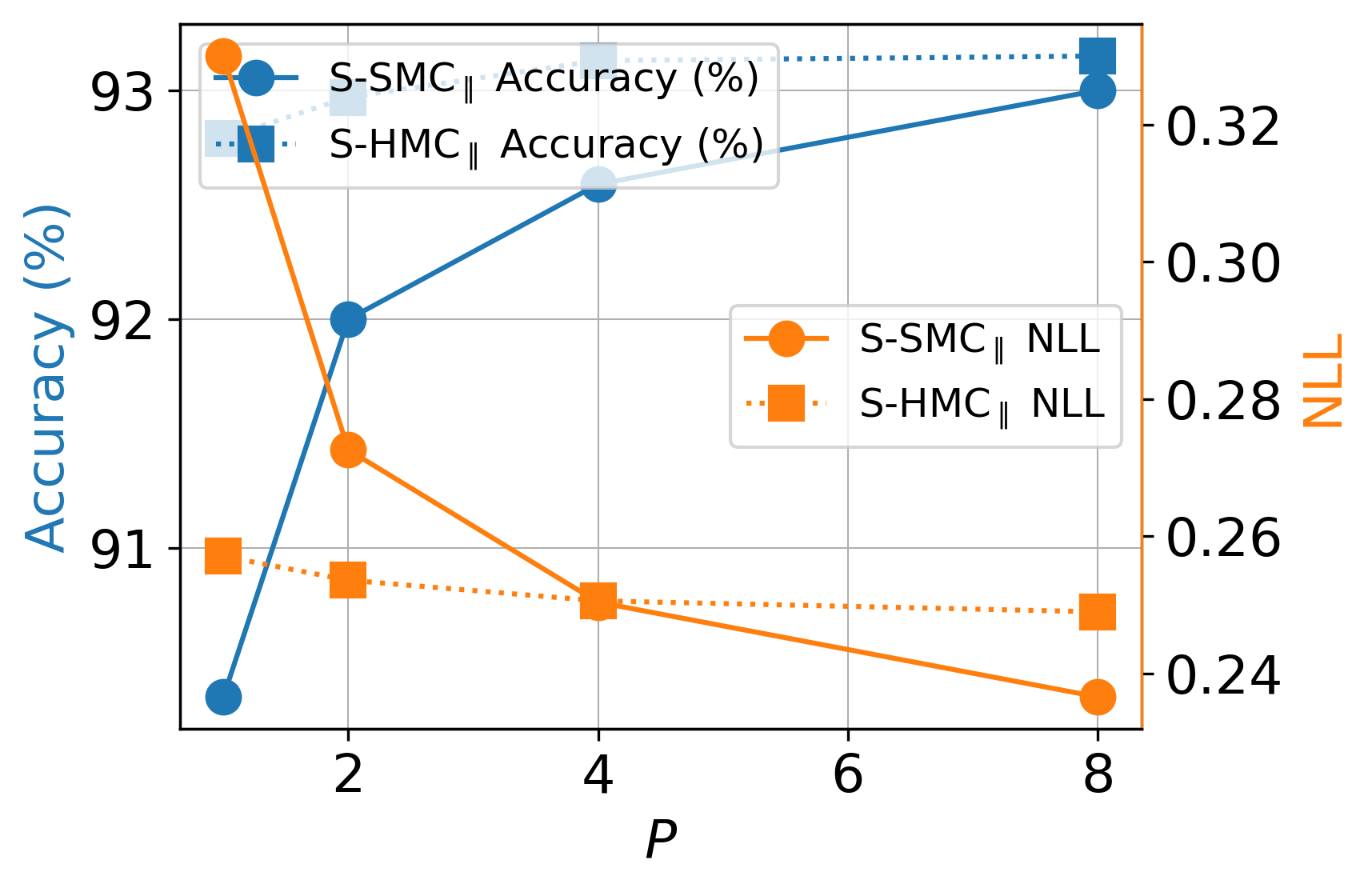}
  \end{subfigure}
  \hfill
  \begin{subfigure}[b]{0.32\textwidth}
    \includegraphics[width=\linewidth]{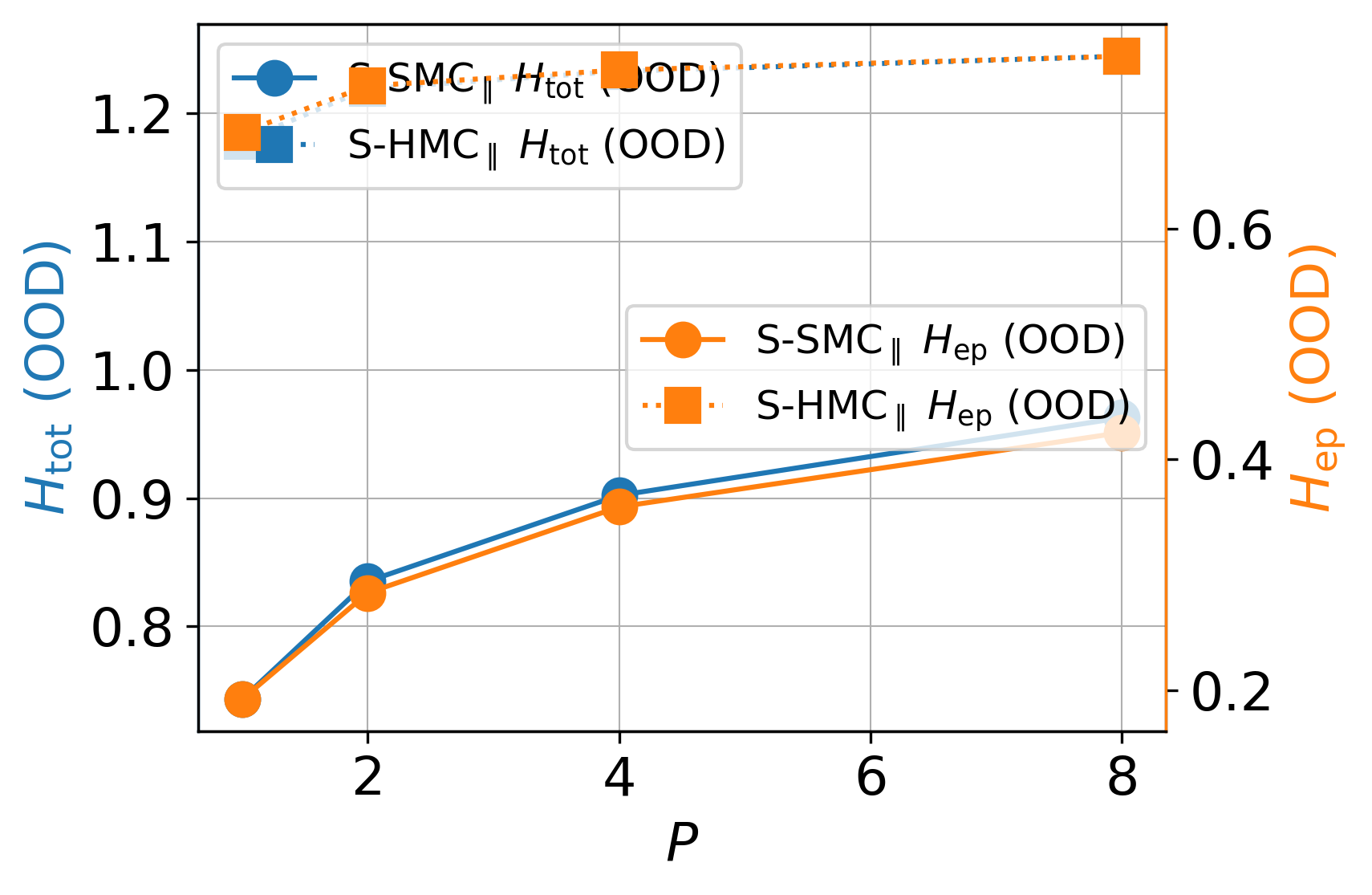}
  \end{subfigure}
  \hfill
  \begin{subfigure}[b]{0.32\textwidth}
    \includegraphics[width=\linewidth]{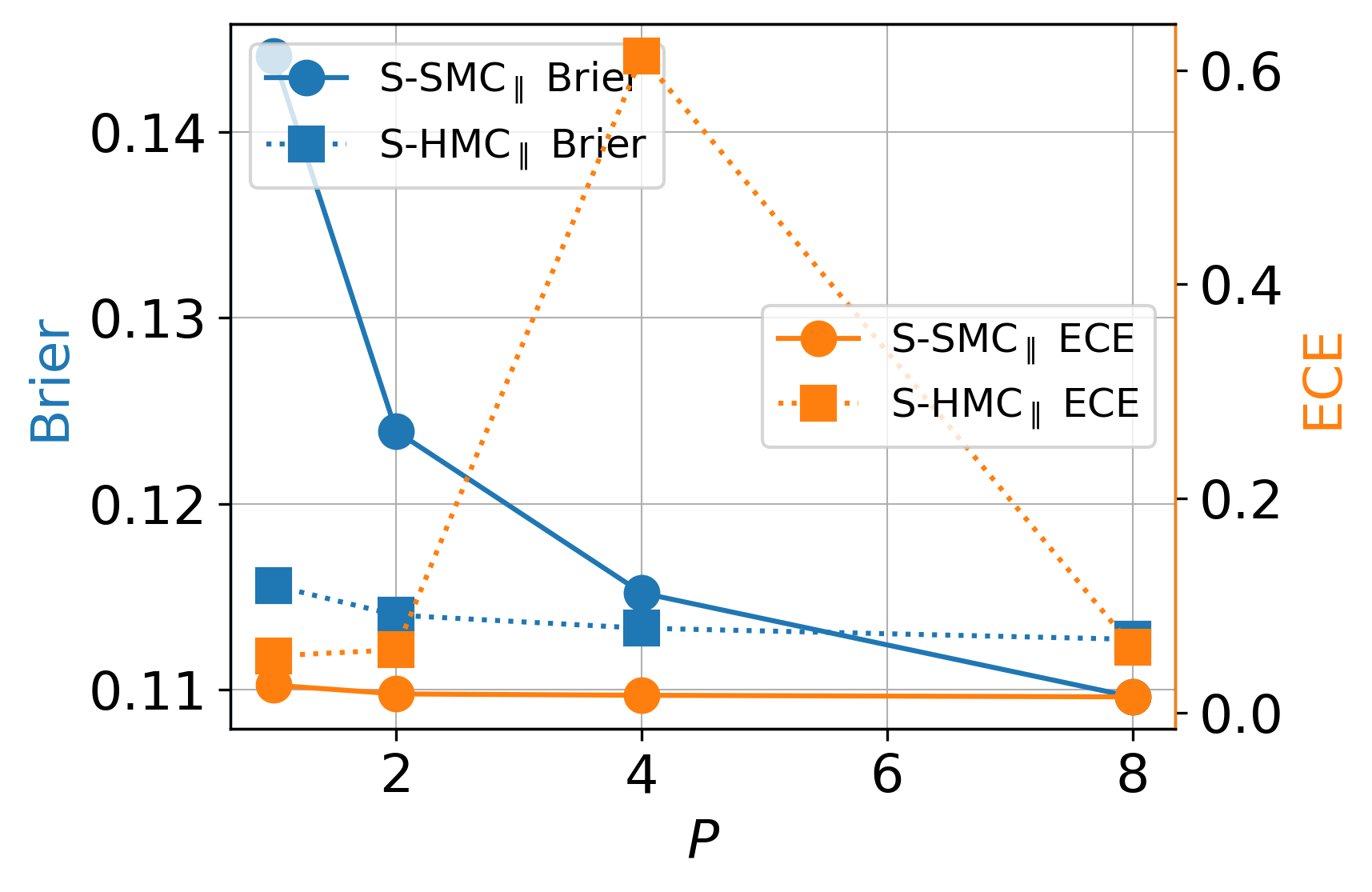}
  \end{subfigure}
  \caption{Comparison of S-SMC\(_\parallel\) (\(P\) chains with \(N=10\)) and S-HMC\(_{\parallel}\) (\(NP\) chains), with fixed number of leapfrog \(L=1\), \(B=160\), \(M=7\), \(v = 0.1\) and \(s=0.25\), on MNIST7 (5 realizations).}
  \label{fig:metric_diffP_mnist_s025}
\end{figure}

\begin{figure}[H]
  \centering
  \begin{subfigure}[b]{0.32\textwidth}
    \includegraphics[width=\linewidth]{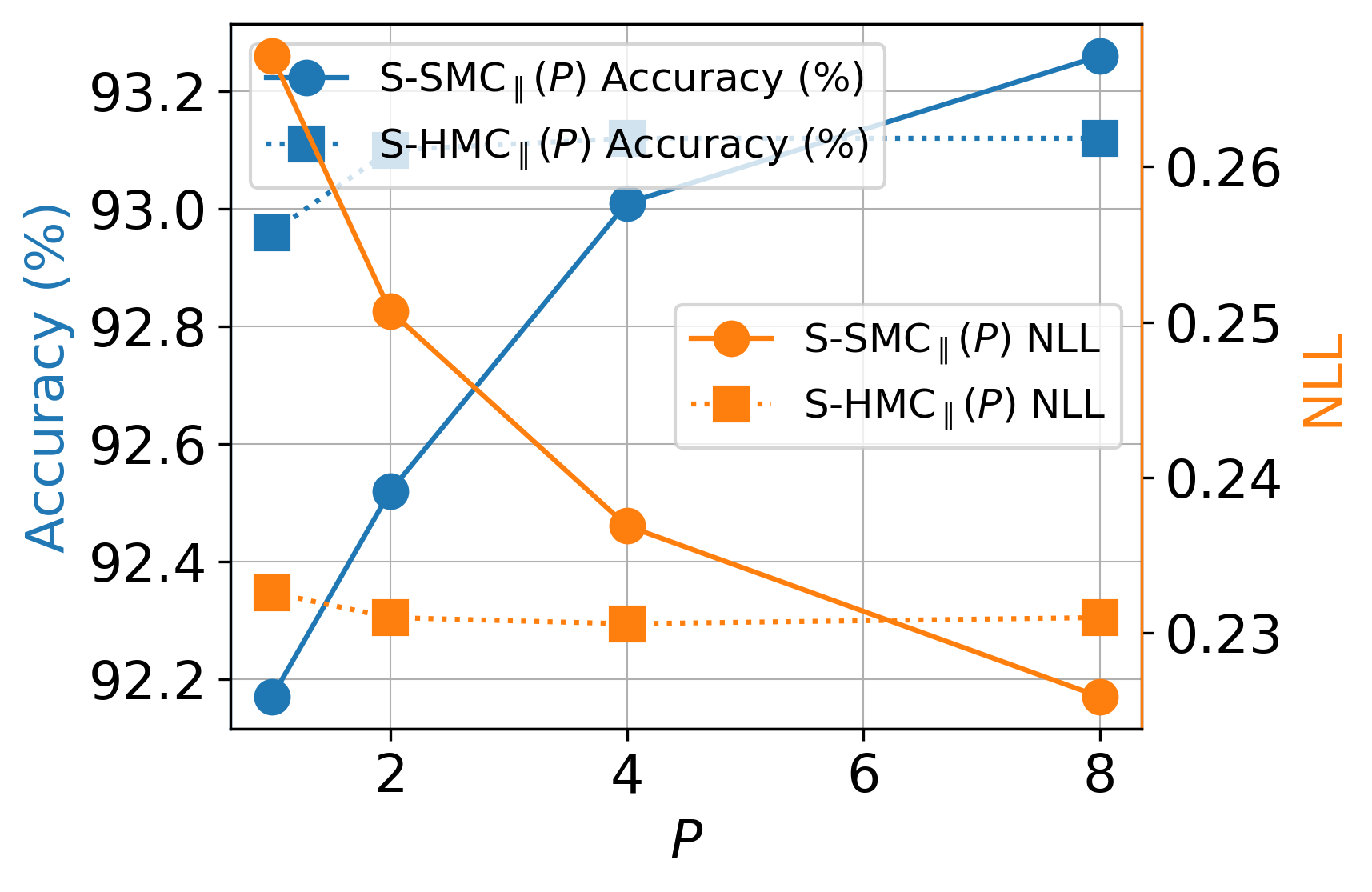}
  \end{subfigure}
  \hfill
  \begin{subfigure}[b]{0.32\textwidth}
    \includegraphics[width=\linewidth]{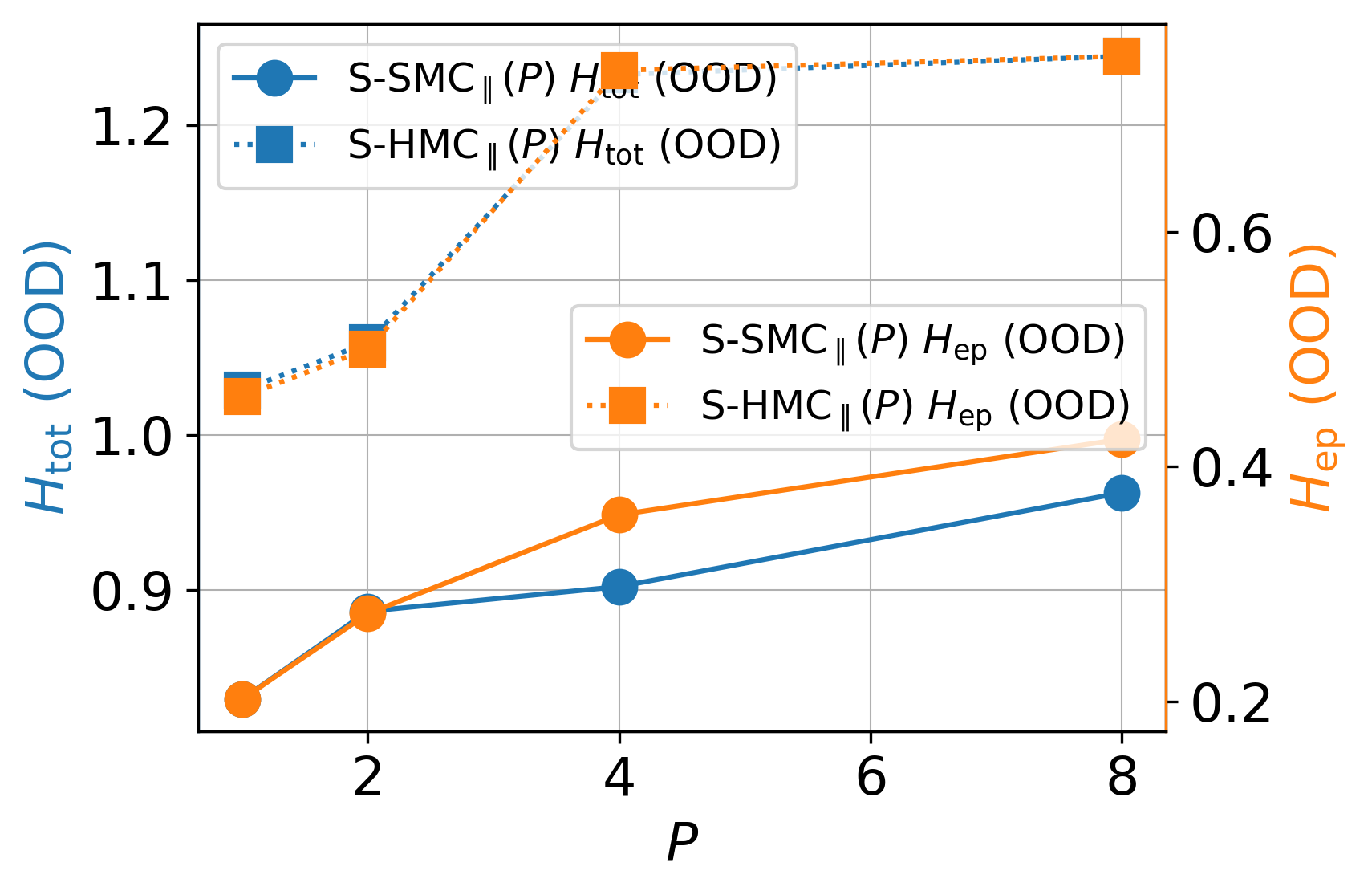}
  \end{subfigure}
  \hfill
  \begin{subfigure}[b]{0.32\textwidth}
    \includegraphics[width=\linewidth]{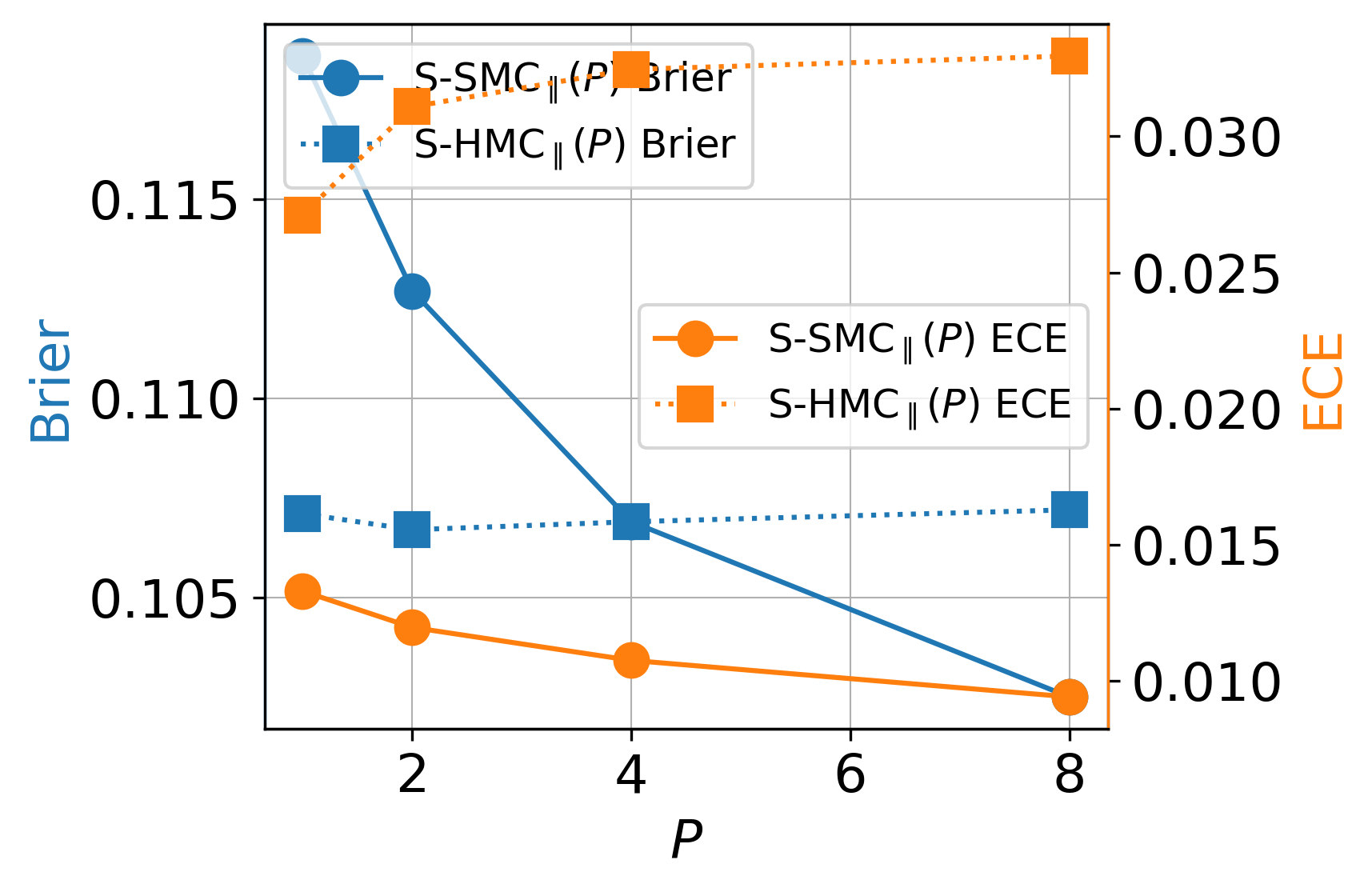}
  \end{subfigure}
  \caption{Comparison of S-SMC\(_\parallel\) (\(P\) chains with \(N=10\)) and S-HMC\(_{\parallel}\) (\(NP\) chains), with fixed number of leapfrog \(L=1\), \(B=160\), \(M=10\), \(v = 0.1\) and \(s=0.1\), on MNIST7 (5 realizations).}
  \label{fig:metric_diffP_mnist_s01}
\end{figure}

\subsection{IMDB}
\label{app:imdb_ablation}

The experiments in this section are tested on the IMDb dataset with the model setting stated in Appendix \ref{app:imdb}. 

\paragraph{Experiment with ($v=\frac1{40}$).} Summary metrics of IMDb dataset with $v=0.025$ and $s=0.1$ is shown in the left spider-plot in Figure \ref{fig:spiders}. Table \ref{tab:metric_diffmethod_imdb_v0025_s01} shows the performance as the tuning parameter $s$ varies. Figure \ref{fig:metric_diffP_imdb_v0025_s01} shows the trend for the SBMC methods as $P$ increases, where the full data can be found in Table \ref{tab:metric_diffP_imdb_v0025_s01}.

\begin{table}[H]
\scriptsize
\caption{Comparison of S-SMC\(_\parallel\) (\(N=10\)), S-HMC\(_{\parallel}\) (\(N\) chains), DE (\(N\) models) and MAP, with fixed number of leapfrog \(L=1\), \(B=25\), \(M=1\) and \(v=0.025\), on IMDb (\(5\) realizations and \(\pm\) s.e.\ in accuracy).}
\label{tab:metric_diffmethod_imdb_v0025_s01}
\centering
\begin{tabular}{ll|c|c|c|cc|ccccc}
\toprule
\(s\) & Method & Ep. & Acc. & NL & \multicolumn{7}{c}{$H_{\sf ep}$} \\
 \cmidrule(lr){6-12}
& &  &  &  & \multicolumn{2}{c|}{ID} & \multicolumn{5}{c}{OD}\\
\cmidrule(lr){6-7} \cmidrule(lr){8-12}
& &  &  &  & cor. & inc. & reviews & meta & lipsum & full reviews & full meta \\
\midrule
\multirow{4}{*}{\(0.1\)} 
& S-SMC\(_\parallel\) & 18.60 & 86.70$\pm$0.03 & 3.655e-1 & 1.122e-4 & 1.664e-4 & 1.792e-4 & 2.200e-4 & 1.749e-4 & 3.285e-4 & 3.212e-4 \\
& \(P=8\) & 19.15 & 86.69$\pm$0.01 & 3.653e-1 & 2.531e-4 & 3.697e-4 & 4.744e-4 & 4.971e-4 & 4.862e-4 & 5.876e-4 & 6.105e-4 \\
& S-HMC\(_\parallel\) & 25 & 86.70$\pm$0.01 & 3.634e-1 & 2.418e-4 & 3.565e-4 & 4.598e-4 & 4.633e-4 & 4.260e-4 & 5.057e-4 & 5.410e-4 \\
& \(P=8\) & 25 & 86.72$\pm$0.00 & 3.633e-1 & 2.766e-4 & 4.022e-4 & 5.694e-4 & 6.062e-4 & 5.637e-4 & 7.438e-4 & 7.051e-4 \\
\midrule
\multirow{2}{*}{\(0\)}
 & MAP & 25.00 & 84.47\(\pm\)0.09 &  3.911e-1 & 0 & 0 & 0 & 0 & 0 & 0 & 0 \\
& DE (\(N\)) & 25.86 & 84.76\(\pm\)0.08 & 3.888e-1 & 1.005e-04 & 1.366e-04 & 5.064e-5 & 5.026e-5 & 4.909e-5 & 6.302e-5 & 5.548e-5  \\
\bottomrule
\end{tabular}

\vspace{0.2cm}

\begin{tabular}{ll|c|c|cc|ccccc}
\toprule
\(s\) & Method & Brier & ECE & \multicolumn{7}{c}{$H_{\sf tot}$} \\
\cmidrule(lr){5-11}
& & & & \multicolumn{2}{c|}{ID} & \multicolumn{5}{c}{OD} \\
\cmidrule(lr){5-6} \cmidrule(lr){7-11}
& & & & cor. & inc. & reviews & meta & lipsum & full reviews & full meta \\
\midrule
\multirow{4}{*}{\(0.1\)} 
& S-SMC\(_\parallel\) & 1.093e-1 & 3.699e-1 & 4.792e-1 & 6.357e-1 & 5.251e-1 & 5.463e-1 & 5.142e-1 & 6.457e-1 & 6.261e-1 \\
& \(P=8\) & 1.092e-1 & 3.698e-1 & 4.788e-1 & 6.355e-1 & 5.213e-1 & 5.406e-1 & 5.115e-1 & 6.430e-1 & 6.236e-1 \\
 & S-HMC\(_\parallel\) & 1.086e-1 & 3.694e-1 &4.750e-1 & 6.340e-1 & 5.165e-1 & 5.331e-1 & 5.079e-1 & 6.400e-1 & 6.186e-1 \\
& \(P=8\) & 1.086e-1 & 3.701e-1 & 4.752e-1 & 6.341e-1 & 5.184e-1 & 5.359e-1 & 5.085e-1 & 6.426e-1 & 6.209e-1 \\
\midrule
\multirow{2}{*}{\(0\)} 
 & MAP & 1.204e-1 & 4.389e-1 & 4.800e-1 & 6.306e-1 & 5.814e-1 & 6.117e-1 & 5.894e-1 & 6.705e-1 & 6.658e-1 \\
 & DE (\(N\)) & 1.193e-1 & 4.340e-1 & 4.819e-1 & 6.319e-1 & 5.793e-1 & 6.098e-1 & 5.882e-1 & 6.702e-1 & 6.649e-1 \\
\bottomrule
\end{tabular}
\end{table}

\begin{figure}[H]
  \centering
  \begin{subfigure}[b]{0.32\textwidth}
    \includegraphics[width=\linewidth]{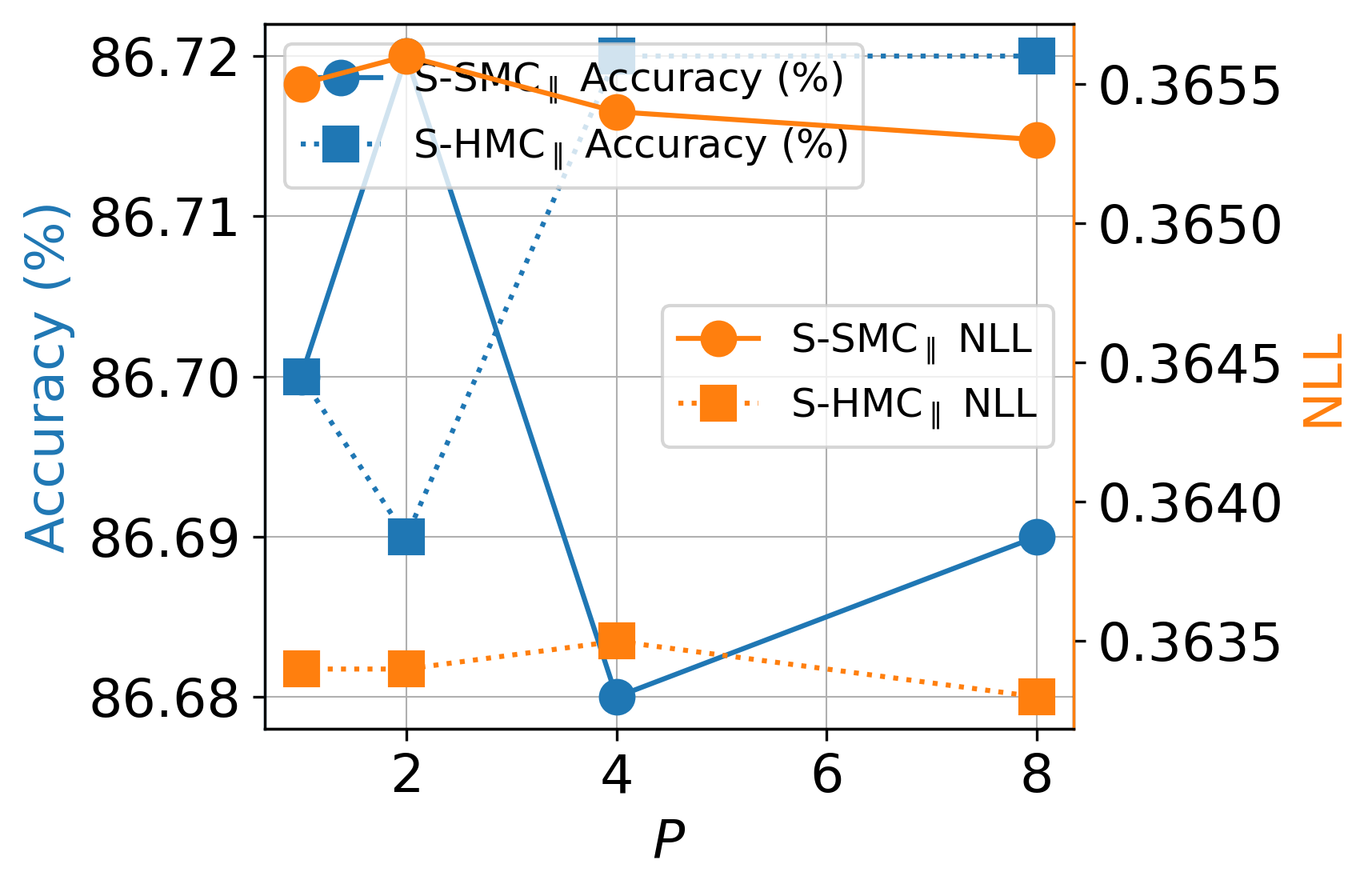}
  \end{subfigure}
  \hfill
  \begin{subfigure}[b]{0.32\textwidth}
    \includegraphics[width=\linewidth]{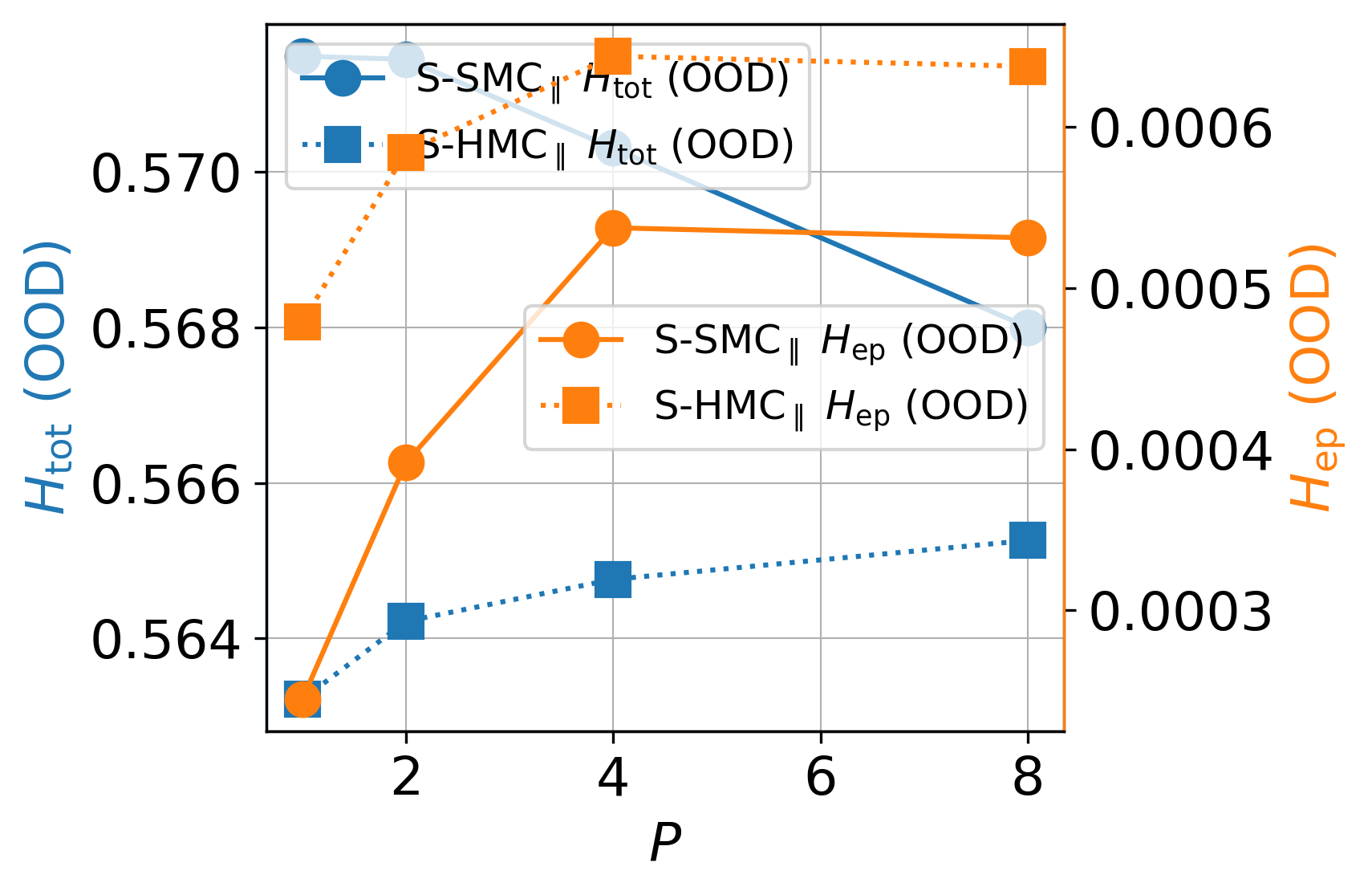}
  \end{subfigure}
  \hfill
  \begin{subfigure}[b]{0.32\textwidth}
    \includegraphics[width=\linewidth]{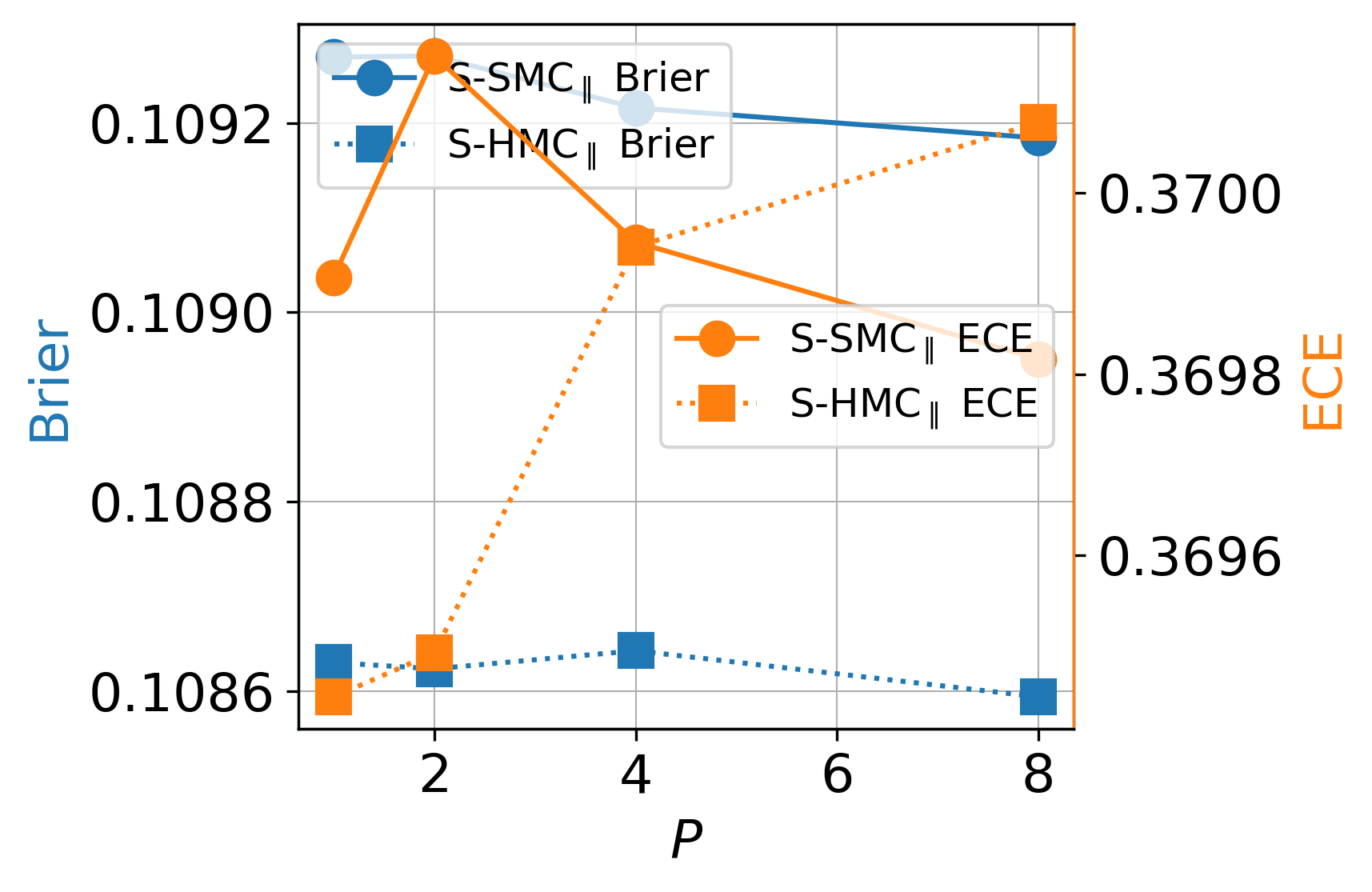}
  \end{subfigure}
  \caption{Comparison of S-SMC\(_\parallel\) ($P$ chains with $N=10$) and S-HMC\(_\parallel\) ($NP$ chains), with fixed $L=1$, $B=25$, $M=1$, $v=0.025$, $s=0.1$, on IMDb (5 realizations)}
  \label{fig:metric_diffP_imdb_v0025_s01}
\end{figure}

\paragraph{Experiments with $v=1$. }Summary metrics of IMDb dataset with $v=15$ and $s=0.35$ are shown in the spider-plot in Figure \ref{fig:imdb_spider_v1}. Table \ref{tab:metric_diffs_imdb} shows the performance as the tuning parameter \(s\) vary. Figure \ref{fig:metric_diffP_imdb_v1_s035}, \ref{fig:metric_diffP_imdb_v1_s025} and \ref{fig:metric_diffP_imdb_v1_s01} give the full convergence of SBMC$_\parallel$ with increasing $P$. The corresponding full data results are given in the Table 
\ref{tab:metric_diffP_imdb_s035}
, \ref{tab:metric_diffP_imdb_s025} 
and \ref{tab:metric_diffP_imdb_s01}.

\begin{table}[H]
\scriptsize
\caption{Comparison of different \(s\) of S-SMC\(_\parallel\) ($N=10$) and S-HMC\(_\parallel\) ($N$ chains), with fixed number of leapfrog $L=1$ and $v=1$, on IMDb (5 realizations and \(\pm\) s.e.\ in accuracy).}
\label{tab:metric_diffs_imdb}
\centering
\setlength{\tabcolsep}{3pt}
\begin{tabular}{lc|c|c|c|cc|ccccc}
\toprule
$s$ & Method & Ep. & Acc. & NL & \multicolumn{7}{c}{$H_{\sf ep}$} \\
 \cmidrule(lr){6-12}
& &  &  &  & \multicolumn{2}{c|}{ID} & \multicolumn{5}{c}{OD}\\
\cmidrule(lr){6-7} \cmidrule(lr){8-12}
& &  &  &  & cor. & inc. & reviews & meta & lipsum & full reviews & full meta \\
\midrule
\multirow{2}{*}{\(0.35\)} 
 & S-SMC\(_\parallel\) & 27.40 & 88.27$\pm$0.07 & 2.803e-1 & 6.177e-4 & 1.460e-3 & 1.581e-3 & 2.495e-3 & 2.187e-3 & 2.279e-3 & 2.504e-3 \\
 & $P=8$ & 27.63 & 88.88$\pm$0.03 & 2.714e-1 & 9.342e-3 & 2.164e-2 & 3.756e-2 & 5.515e-2 & 5.260e-2 & 6.049e-2 & 6.435e-2 \\
 & S-HMC\(_\parallel\) & 26    & 88.81$\pm$0.01 & 2.750e-1 & 1.565e-2 & 3.463e-2 & 5.414e-2 & 7.021e-2 & 6.872e-2 & 8.407e-2 & 7.930e-2 \\
  & $P=8$ & 26 & 88.93$\pm$0.02 & 2.737e-1 & 1.662e-2 & 3.651e-2 & 5.315e-2 & 7.391e-2 & 6.917e-2 & 9.098e-2 & 8.360e-2 \\
\midrule
\multirow{2}{*}{\(0.25\)} 
 & S-SMC\(_\parallel\) & 29.6  & 88.27\(\pm\)0.10 & 2.807e-1 & 2.512e-4 & 6.069e-4 & 4.733e-4 & 7.166e-4 & 8.590e-4 & 9.313e-4 & 9.520e-4 \\
  & $P=8$ & 28.5 & 88.87\(\pm\)0.03 & 2.720e-1 & 8.124e-3 & 1.872e-2 & 3.066e-2 & 5.057e-2 & 4.691e-2 & 6.403e-2 & 5.777e-2 \\
 & S-HMC\(_\parallel\) & 26    & 88.83\(\pm\)0.02 & 2.745e-1 & 1.269e-2 & 2.830e-2 & 4.522e-2 & 5.927e-2 & 5.886e-2 & 7.342e-2 & 6.803e-2 \\
  & $P=8$ & 26 & 88.92\(\pm\)0.02 & 2.734e-1 & 1.337e-2 & 2.964e-2 & 4.442e-2 & 6.215e-2 & 5.863e-2 & 7.869e-2 & 7.108e-2 \\
\midrule
\multirow{2}{*}{\(0.1\)} 
 & S-SMC\(_\parallel\) & 24    & 88.54\(\pm\)0.11 & 2.762e-1 & 1.820e-4 & 4.315e-4 & 4.819e-4 & 5.915e-4 & 6.766e-4 & 7.711e-4 & 6.500e-4  \\
  & $P=8$ & 23.7  & 88.92\(\pm\)0.01 & 2.711e-1 & 4.207e-3 & 9.768e-3 & 2.182e-2 & 3.140e-2 & 2.700e-2 & 3.352e-2 & 3.540e-2  \\
 & S-HMC\(_\parallel\) & 26    & 88.86\(\pm\)0.02 & 2.726e-1 & 5.753e-3 & 1.319e-2 & 2.712e-2 & 3.551e-2 & 3.561e-2 & 5.110e-2 & 4.289e-2  \\
  & $P=8$ & 26    & 88.93\(\pm\)0.01 & 2.721e-1 & 6.065e-3 & 1.386e-2 & 2.792e-2 & 3.888e-2 & 3.653e-2 & 5.408e-2 & 4.638e-2  \\
\midrule
\multirow{3}{*}{\(0\)} 
 & MAP              & 52    & 87.97\(\pm\)0.04 & 2.854e-1 & –        & 0        & 0        & 0        & 0        & 0        & 0         \\
 & DE (\(N\))       & 26.52 & 87.75\(\pm\)0.02 & 2.921e-1 & 3.144e-3 & 7.055e-3 & 4.514e-2 & 5.054e-2 & 5.386e-2 & 7.963e-2 & 5.318e-2  \\
 & DE (\(8N\))      & 25.86 & 87.70 \(\pm\)0.01 & 2.925e-1 & 3.469e-3 & 7.608e-3 & 4.394e-2 & 5.622e-2 & 5.089e-2 & 7.435e-2 & 6.198e-2  \\
\bottomrule
\end{tabular}

\vspace{0.2cm}

\begin{tabular}{lc|c|c|cc|ccccc}
\toprule
\(s\) & Method & Brier & ECE & \multicolumn{7}{c}{$H_{\sf tot}$} \\
\cmidrule(lr){5-11}
& & && \multicolumn{2}{c|}{ID} & \multicolumn{5}{c}{OD} \\
\cmidrule(lr){5-6} \cmidrule(lr){7-11}
& & && cor. & inc. & reviews & meta & lipsum & full reviews & full meta \\
\midrule
\multirow{2}{*}{\(0.35\)} 
 & S-SMC\(_\parallel\) & 8.547e-2 & 3.832e-1 & 2.643e-1 & 5.482e-1 & 3.802e-1 & 5.116e-1 & 5.172e-1 & 5.581e-1 & 5.286e-1 \\
  & $P=8$ & 8.206e-2 & 3.899e-1 & 2.744e-1 & 5.596e-1 & 3.987e-1 & 5.555e-1 & 5.304e-1 & 6.025e-1 & 5.920e-1 \\
 & S-HMC\(_\parallel\) & 8.298e-2 & 3.889e-1 & 2.890e-1 & 5.681e-1 & 4.289e-1 & 5.583e-1 & 5.556e-1 & 6.133e-1 & 6.120e-1 \\
  & $P=8$ & 8.254e-2 & 3.904e-1 & 2.893e-1 & 5.683e-1 & 4.188e-1 & 5.626e-1 & 5.386e-1 & 6.156e-1 & 6.088e-1 \\
\midrule
\multirow{2}{*}{\(0.25\)} 
 & S-SMC\(_\parallel\) & 8.548e-2 & 3.825e-1 & 2.673e-1 & 5.500e-1 & 3.562e-1 & 4.872e-1 & 4.727e-1 & 5.548e-1 & 5.609e-1 \\
  & $P=8$ & 8.220e-2 & 3.901e-1 & 2.772e-1 & 5.609e-1 & 3.891e-1 & 5.447e-1 & 5.340e-1 & 6.036e-1 & 5.874e-1 \\
 & S-HMC\(_\parallel\) & 8.286e-2 & 3.896e-1 & 2.873e-1 & 5.667e-1 & 4.239e-1 & 5.585e-1 & 5.533e-1 & 6.117e-1 & 6.076e-1 \\
  & $P=8$ & 8.249e-2 & 3.904e-1 & 2.871e-1 & 5.665e-1 & 4.138e-1 & 5.595e-1 & 5.338e-1 & 6.135e-1 & 6.047e-1 \\
\midrule
\multirow{2}{*}{\(0.1\)} 
 & S-SMC\(_\parallel\) & 8.387e-2 & 3.869e-1 & 2.721e-1 & 5.544e-1 & 3.970e-1 & 5.202e-1 & 5.063e-1 & 5.686e-1 & 5.684e-1  \\
  & $P=8$ & 8.190e-2 & 3.906e-1 & 2.752e-1 & 5.588e-1 & 3.920e-1 & 5.407e-1 & 5.143e-1 & 5.959e-1 & 5.865e-1  \\\
 & S-HMC\(_\parallel\) & 8.232e-2 & 3.898e-1 & 2.802e-1 & 5.618e-1 & 4.088e-1 & 5.505e-1 & 5.384e-1 & 6.066e-1 & 5.975e-1  \\
  & $P=8$ & 8.217e-2 & 3.906e-1 & 2.801e-1 & 5.613e-1 & 4.028e-1 & 5.510e-1 & 5.231e-1 & 6.081e-1 & 5.961e-1  \\
\midrule
\multirow{3}{*}{\(0\)} 
& MAP            & 8.714e-2 & 4.350e-1 & 2.721e-1 & 5.531e-1 & 4.071e-1 & 5.598e-1 & 5.462e-1 & 5.695e-1 & 5.561e-1  \\
& DE (\(N\))         & 8.928e-2 &  4.343e-1 & 2.850e-1 & 5.580e-1 & 4.497e-1 & 5.808e-1 & 5.609e-1 & 6.185e-1 & 6.032e-1  \\
& DE (\(8N\))         & 8.941e-2 &  4.354e-1  & 2.859e-1 & 5.588e-1 & 4.533e-1 & 5.888e-1 & 5.708e-1 & 6.187e-1 & 6.068e-1  \\
\bottomrule
\end{tabular}
\end{table}


\begin{figure}[H]
\centering
    \includegraphics[width=.45\linewidth]{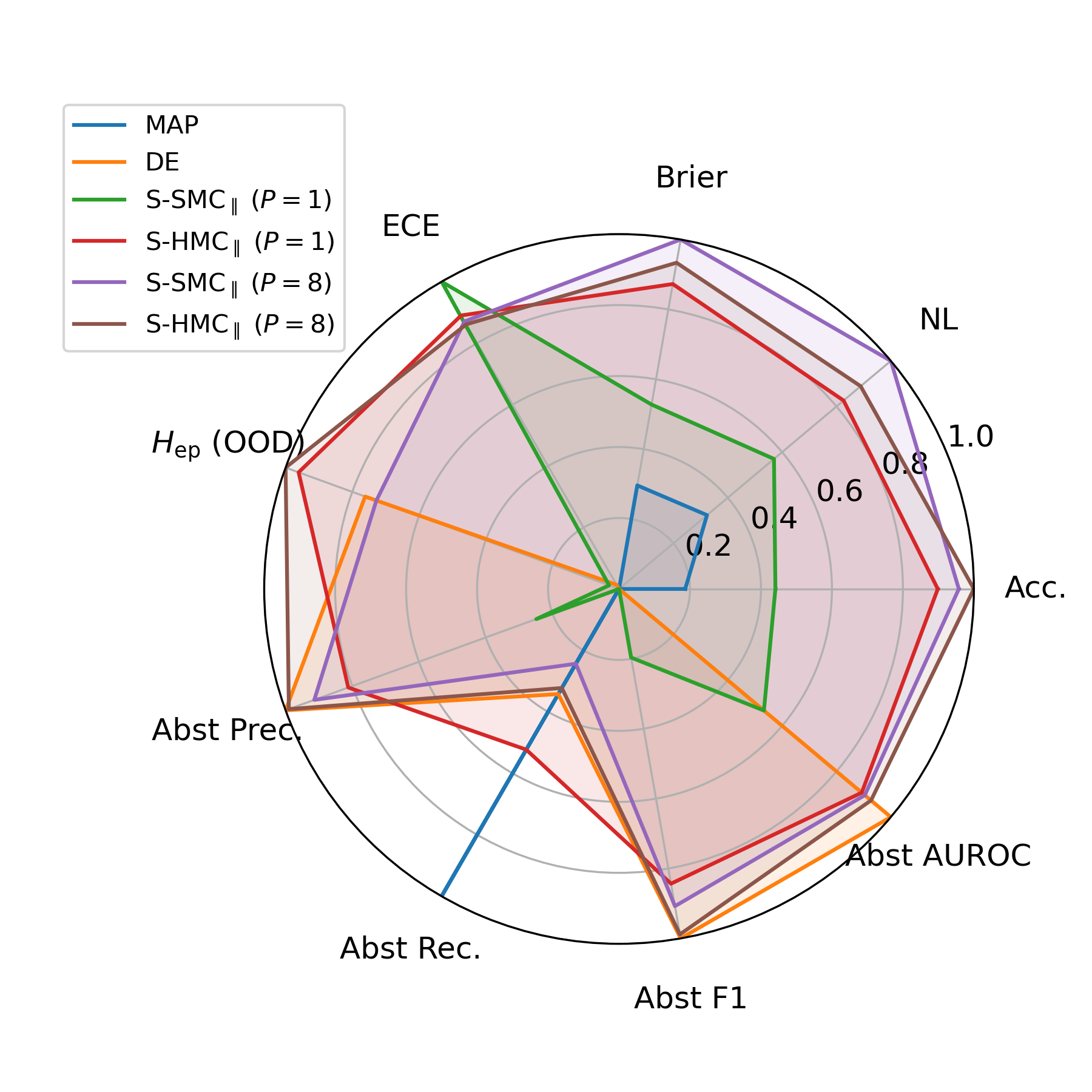}
    \caption{Summary metrics for IMDb in all methods. S-SMC\(_\parallel\) (\(P=1\) chain with \(N=10\)), S-HMC\(_{\parallel}\) (\(NP\) chains), DE (\(N\) models) and MAP, with fixed number of leapfrog \(L=1\), \(B=26\), \(M=2\), \(v=1\) and \(s=0.35\) (\(5\) realizations).}
\label{fig:imdb_spider_v1}
\end{figure}

\begin{figure}[H]
  \centering
  \begin{subfigure}[b]{0.32\textwidth}
    \includegraphics[width=\linewidth]{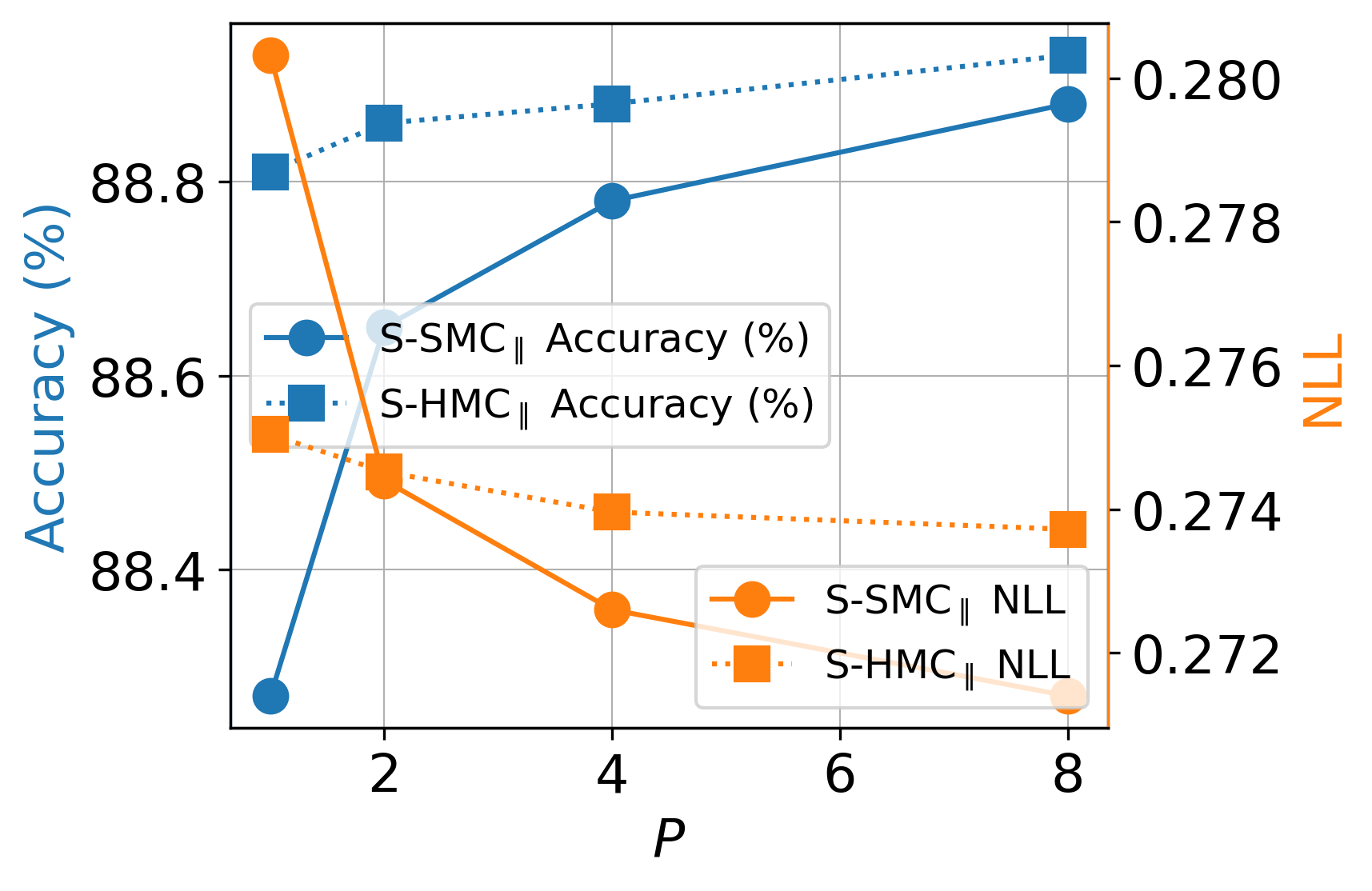}
  \end{subfigure}
  \hfill
  \begin{subfigure}[b]{0.32\textwidth}
    \includegraphics[width=\linewidth]{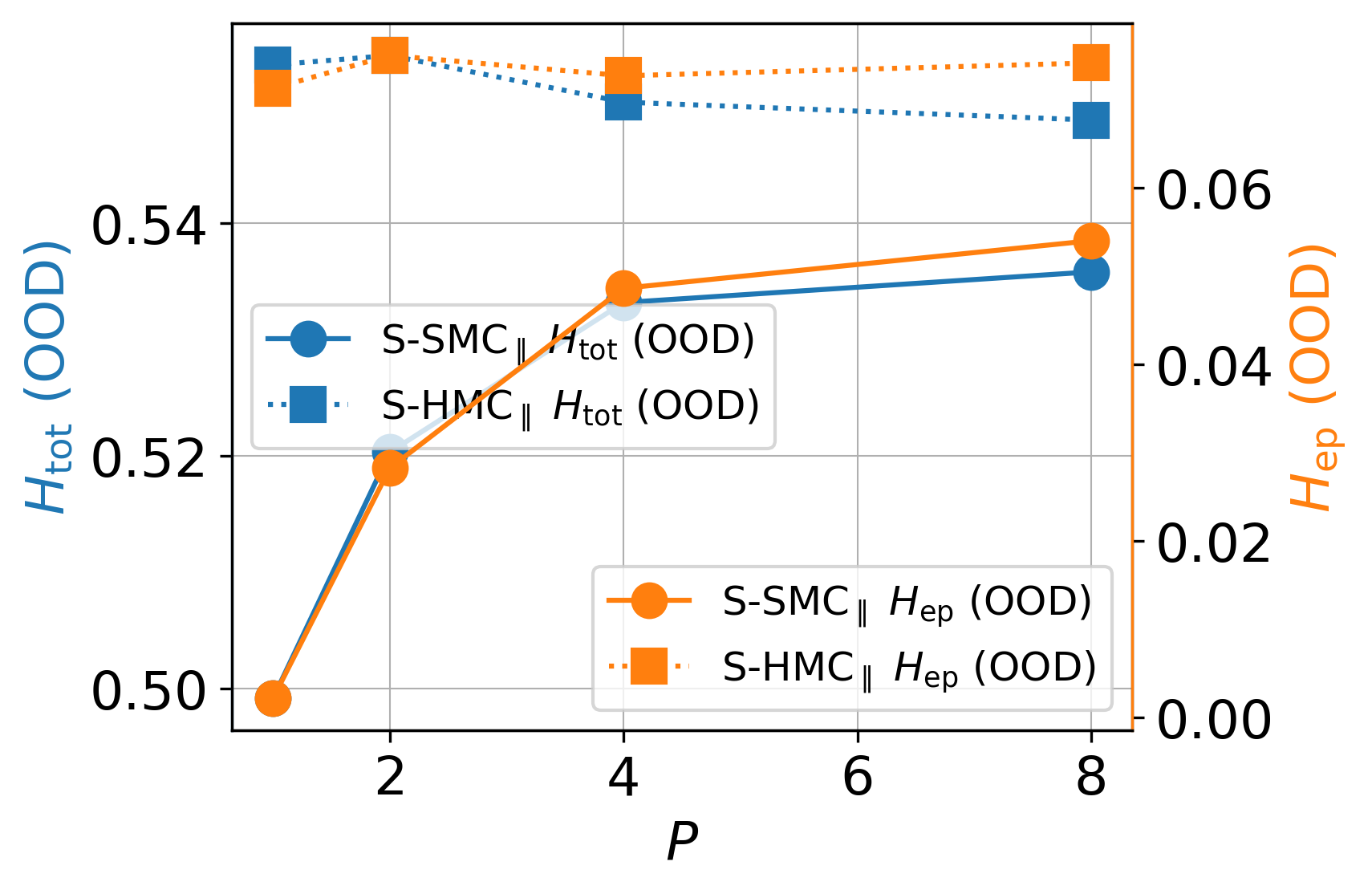}
  \end{subfigure}
  \hfill
  \begin{subfigure}[b]{0.32\textwidth}
    \includegraphics[width=\linewidth]{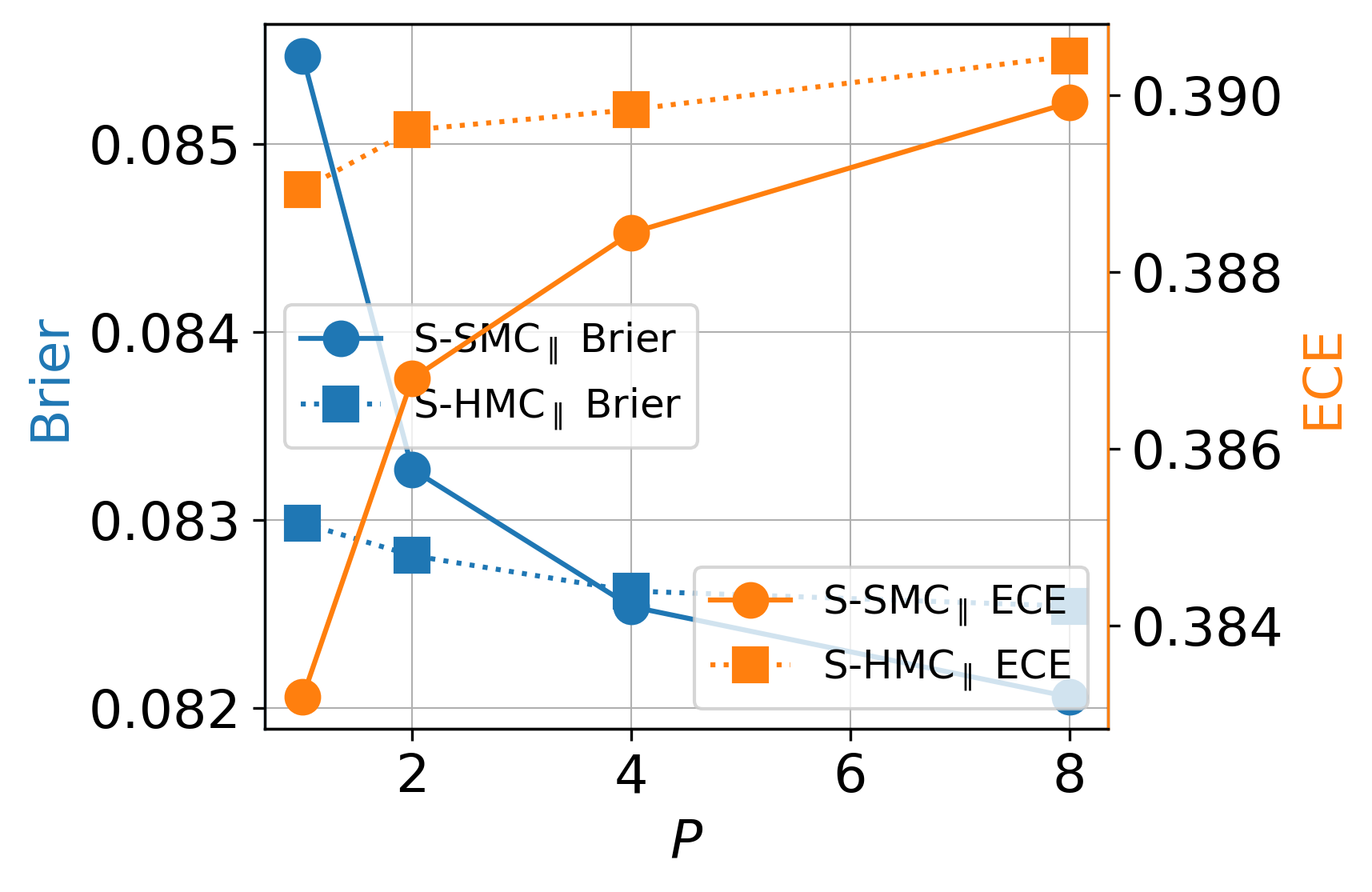}
  \end{subfigure}
  \caption{Comparison of S-SMC\(_\parallel\) ($P$ chains with $N=10$) and S-HMC\(_\parallel\) ($NP$ chains), with fixed number of leapfrog $L=1$, $B=26$, $M=1$, $v=1$ and \(s=0.35\), on IMDb (5 realizations and \(\pm\)s.e. in accuracy ).}
  \label{fig:metric_diffP_imdb_v1_s035}
\end{figure}

\begin{figure}[H]
  \centering
  \begin{subfigure}[b]{0.32\textwidth}
    \includegraphics[width=\linewidth]{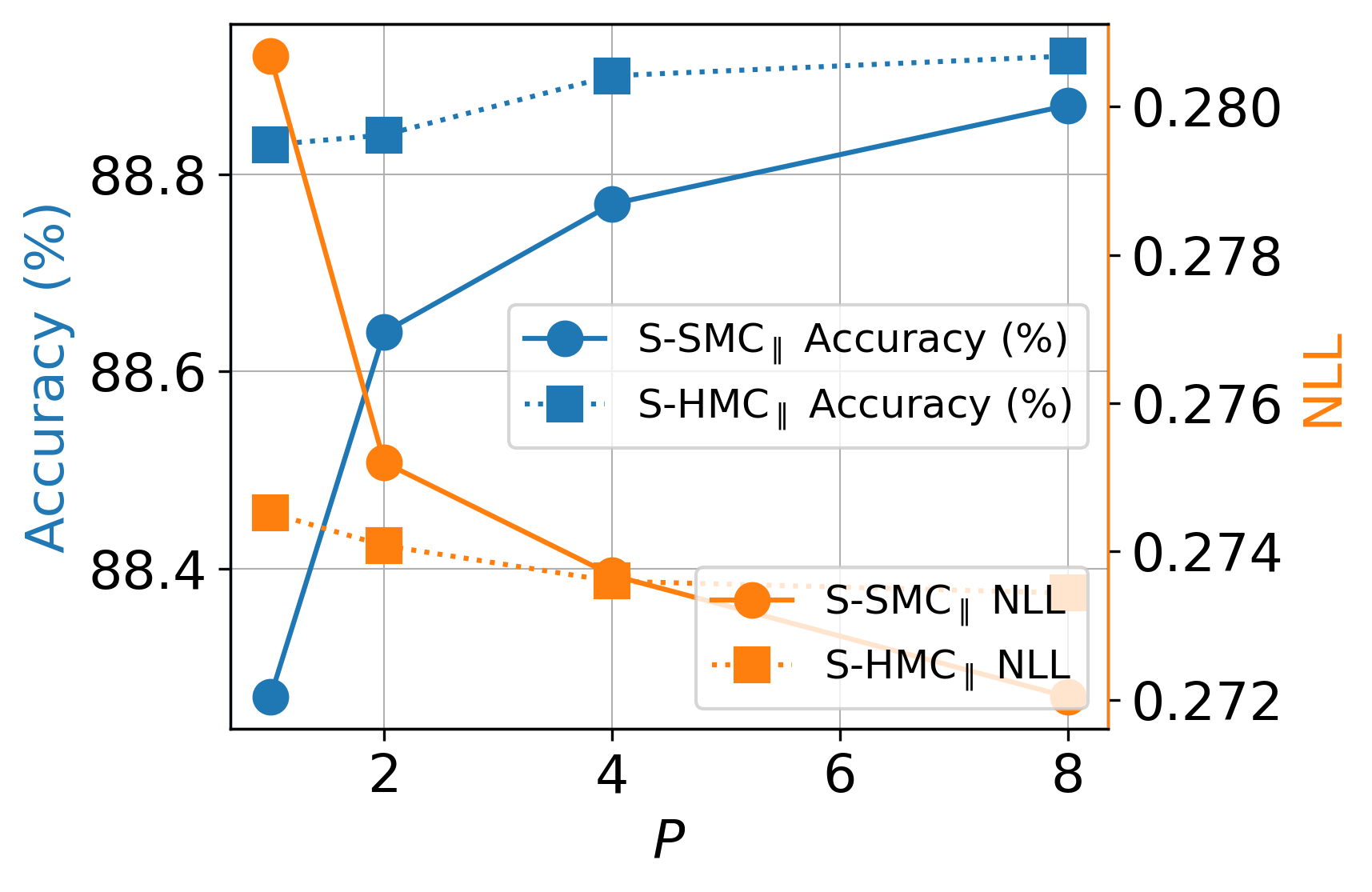}
  \end{subfigure}
  \hfill
  \begin{subfigure}[b]{0.32\textwidth}
    \includegraphics[width=\linewidth]{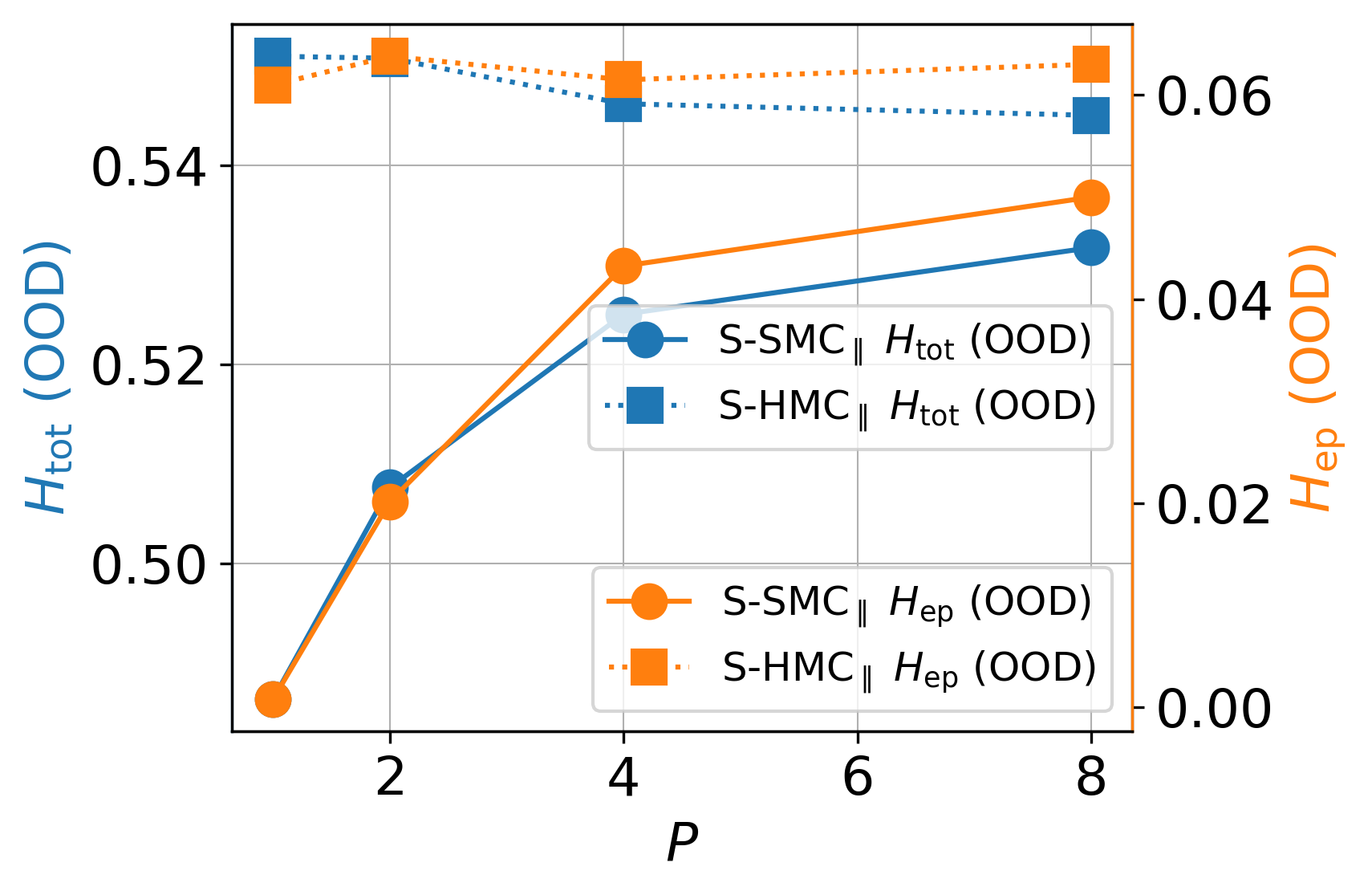}
  \end{subfigure}
  \hfill
  \begin{subfigure}[b]{0.32\textwidth}
    \includegraphics[width=\linewidth]{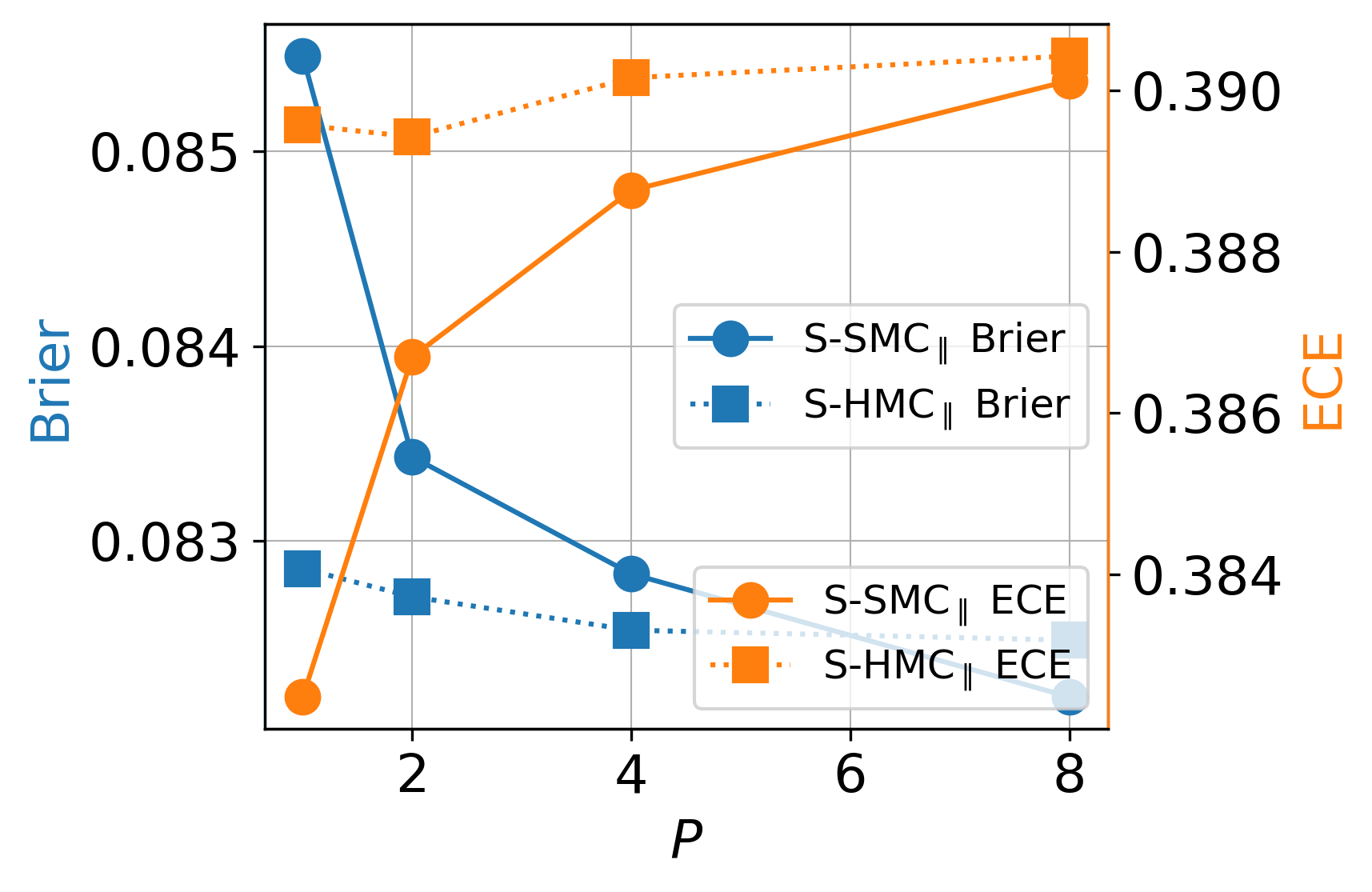}
  \end{subfigure}
  \caption{Comparison of S-SMC\(_\parallel\) ($P$ chains with $N=10$) and S-HMC\(_\parallel\) ($NP$ chains), with fixed number of leapfrog $L=1$, $B=26$, $M=2$, $v=1$ and \(s=0.25\), on IMDb (5 realizations and \(\pm\) s.e.\ in accuracy).}
  \label{fig:metric_diffP_imdb_v1_s025}
\end{figure}

\begin{figure}[H]
  \centering
  \begin{subfigure}[b]{0.32\textwidth}
    \includegraphics[width=\linewidth]{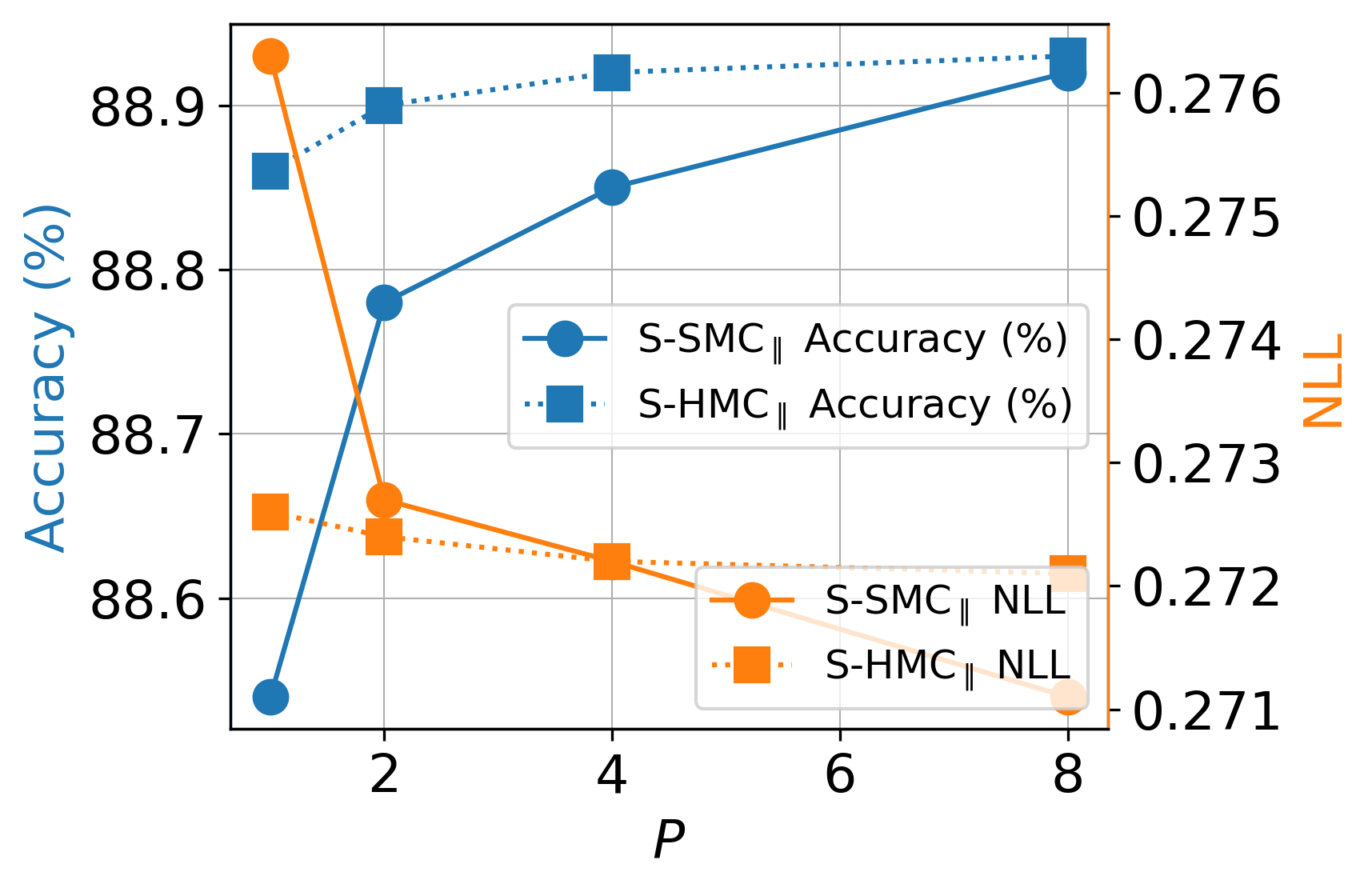}
  \end{subfigure}
  \hfill
  \begin{subfigure}[b]{0.32\textwidth}
    \includegraphics[width=\linewidth]{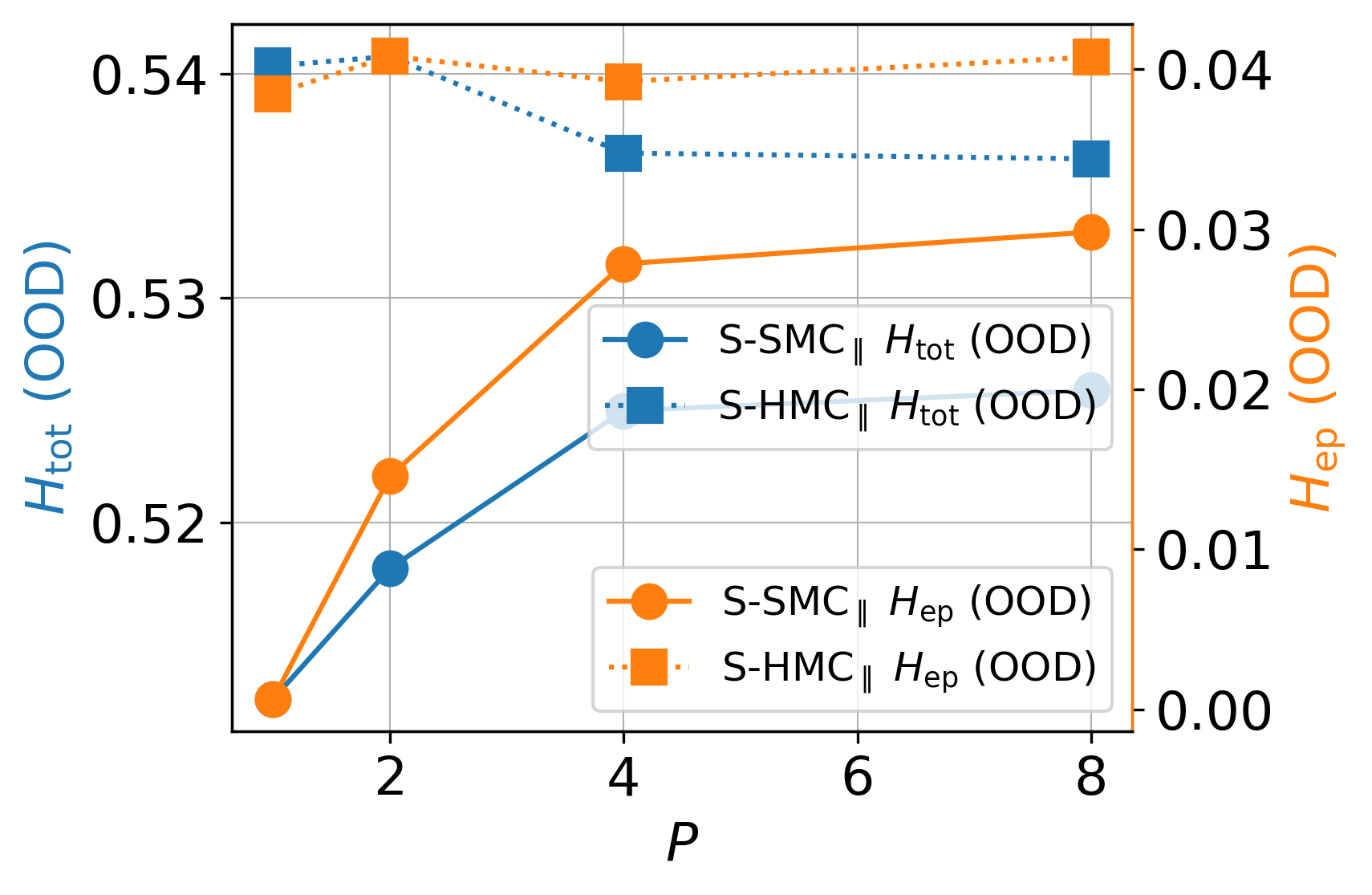}
  \end{subfigure}
  \hfill
  \begin{subfigure}[b]{0.32\textwidth}
    \includegraphics[width=\linewidth]{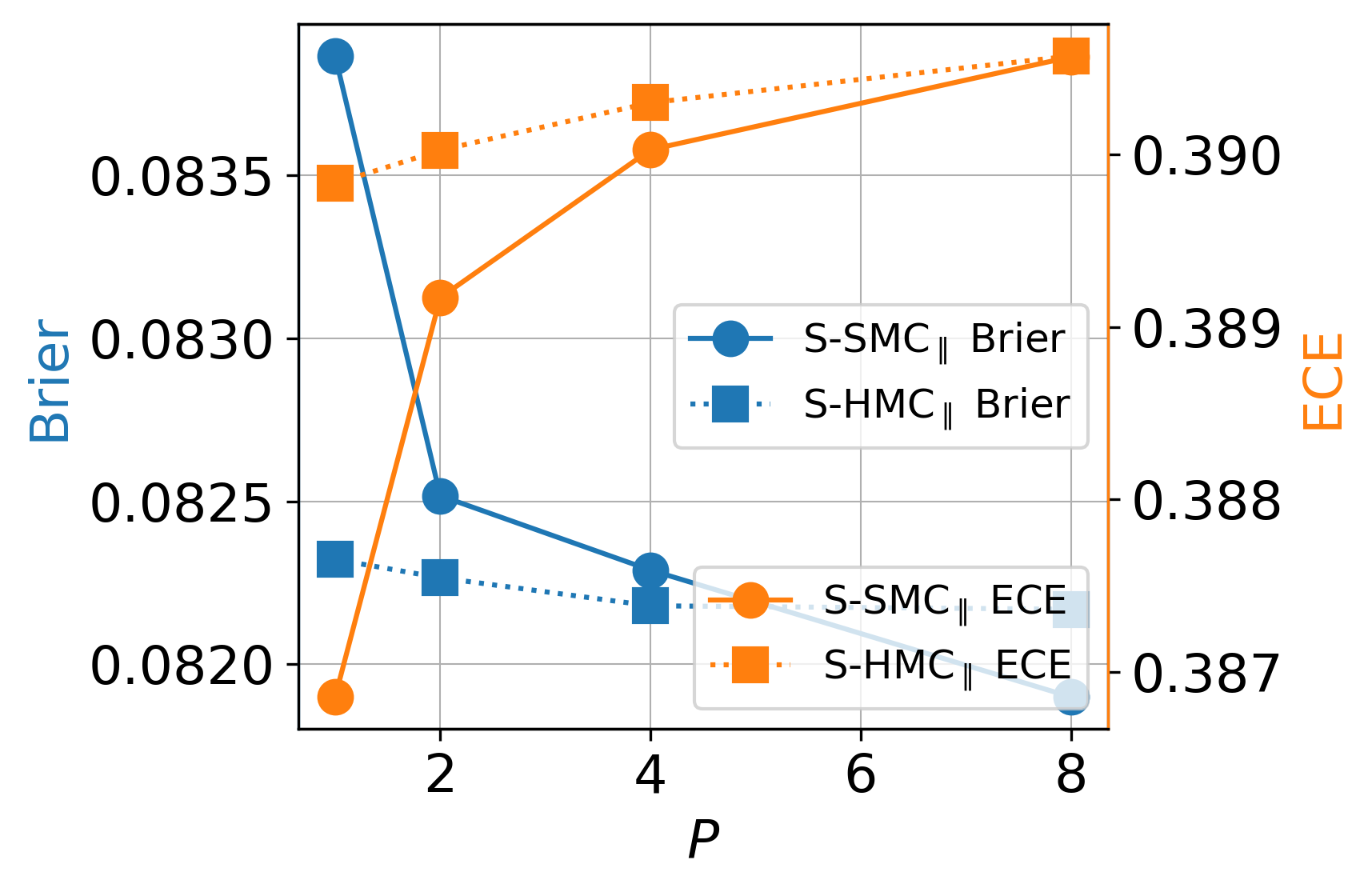}
  \end{subfigure}
  \caption{Comparison of S-SMC\(_\parallel\) ($P$ chains with $N=10$) and S-HMC\(_\parallel\) ($NP$ chains), with fixed number of leapfrog $L=1$, $B=26$, $M=2$, $v=1$ and \(s=0.1\), on IMDb (5 realizations).}
  \label{fig:metric_diffP_imdb_v1_s01}
\end{figure}

\subsection{CIFAR10}

Experiments in this section are tested on the CIFAR10 dataset with the model setting stated in Appendix \ref{app:cifar}. 

The summary metrics on CIFAR10 are shown in a spider-plot in Figure \ref{fig:spiders}. Table \ref{tab:metric_diffs_cifar} shows the performance as the tuning parameter \(s\) vary. 
Figure \ref{fig:metric_diffP_cifar_v02_s005}, \ref{fig:metric_diffP_cifar_v02_s01} and \ref{fig:metric_diffP_cifar_v02_s02} give the full convergence of SBMC$_\parallel$ with increasing $P$.
 The corresponding full data results are given in Table \ref{tab:metric_diffP_cifar_s005}
 , \ref{tab:metric_diffP_cifar_s01}
  and \ref{tab:metric_diffP_cifar_s02}.

\begin{figure}[H]
  \centering
  \begin{subfigure}[b]{0.32\textwidth}
    \includegraphics[width=\linewidth]{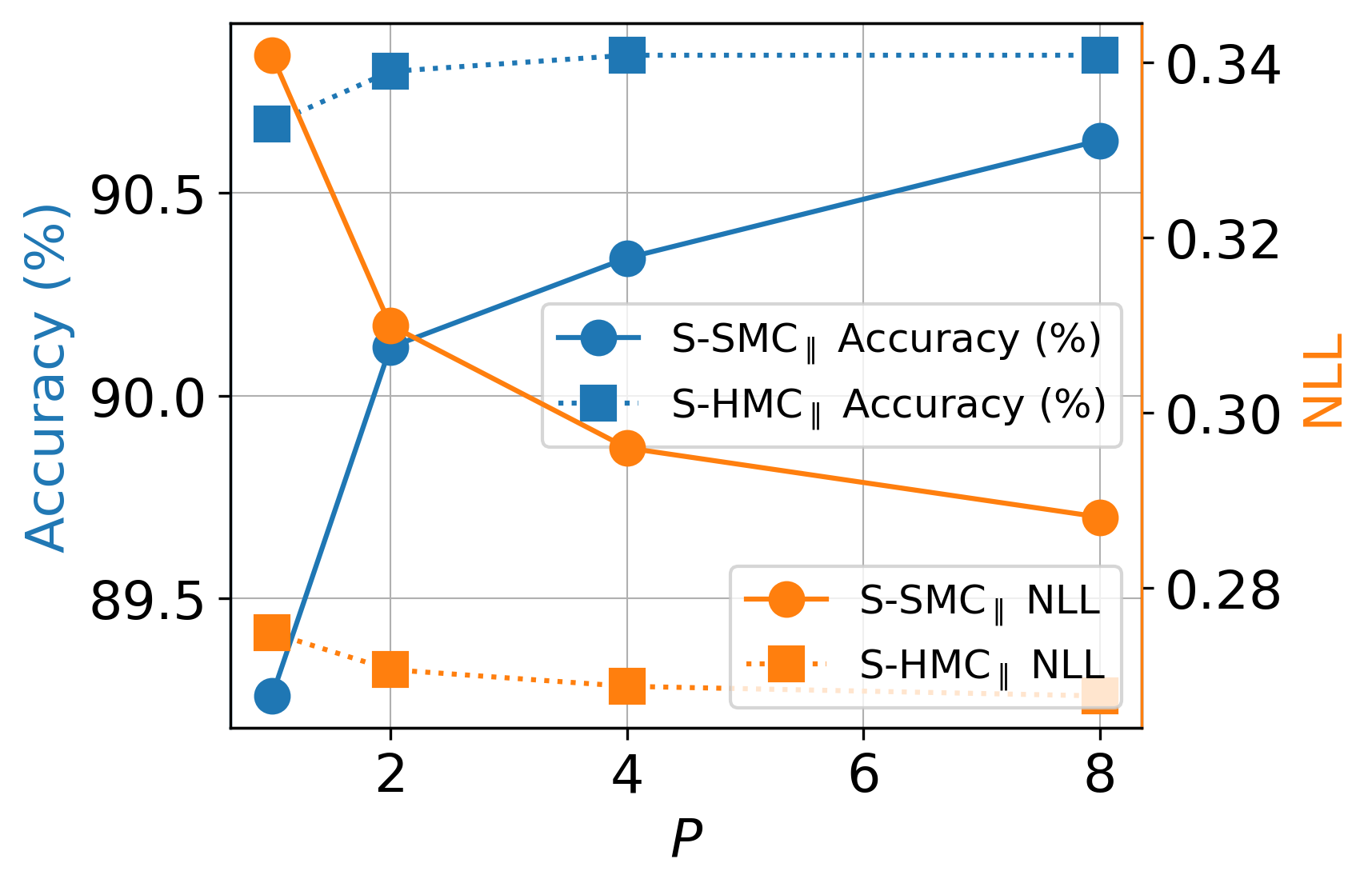}
  \end{subfigure}
  \hfill
  \begin{subfigure}[b]{0.32\textwidth}
    \includegraphics[width=\linewidth]{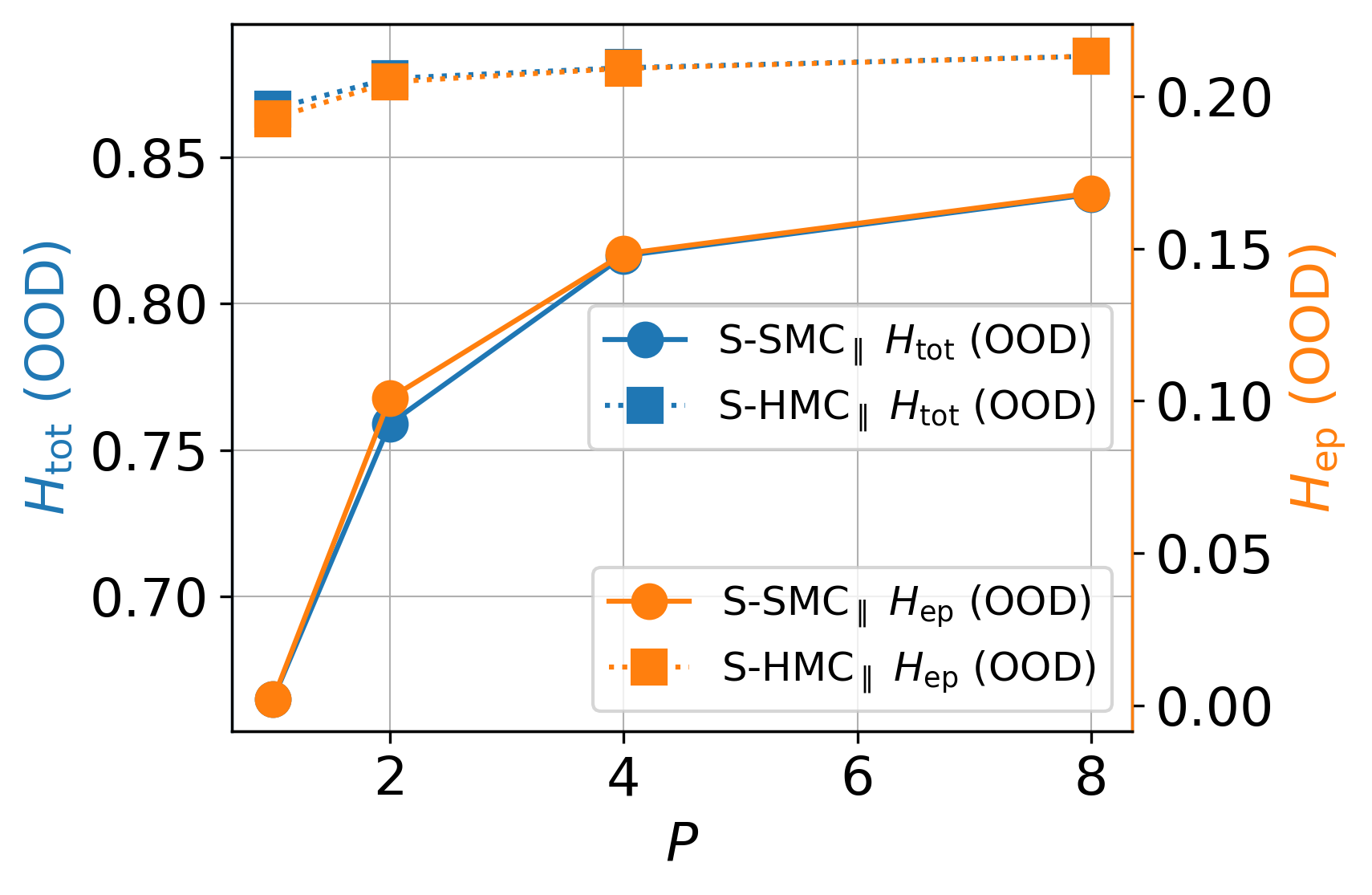}
  \end{subfigure}
  \hfill
  \begin{subfigure}[b]{0.32\textwidth}
    \includegraphics[width=\linewidth]{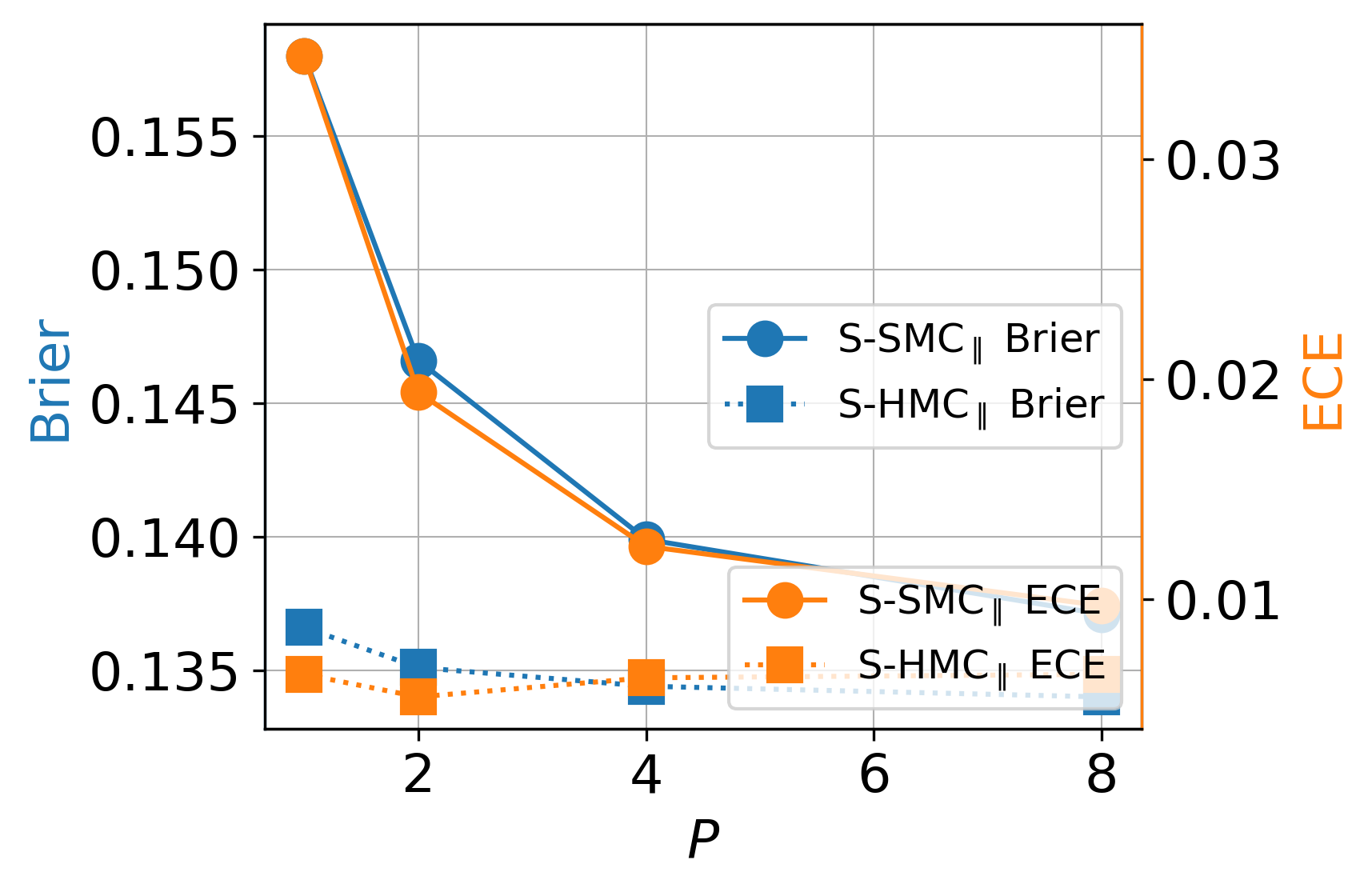}
  \end{subfigure}
  \caption{Comparison of S-SMC\(_\parallel\) ($P$ chains with $N=10$) and S-HMC\(_\parallel\) ($NP$ chains), with fixed number of leapfrog $L=1$, $B=200$, $M=4$, $v=0.2$ and \(s=0.05\), on CIFAR10 (5 realizations).}
  \label{fig:metric_diffP_cifar_v02_s005}
\end{figure}

\begin{table}[H]
\scriptsize
\caption{Comparison of S-SMC\(_\parallel\) (\(N=10\)), S-HMC\(_{\parallel}\) (\(N\) chains), DE (\(N\)) and MAP, with fixed number of leapfrog \(L=1\), \(B=200\), \(M=4\) and \(v=0.2\), on CIFAR10 (\(5\) realizations and \(\pm\) s.e.\ in accuracy).}
\label{tab:metric_diffs_cifar}
\centering
\setlength{\tabcolsep}{3pt}
\begin{tabular}{lc|c|c|c|c|c}
\toprule
$s$ & Method & Ep. & Acc. & NL & Brier & ECE \\
\midrule
\multirow{2}{*}{\(0.2\)} 
 & S-SMC\(_\parallel\) & 289.6 & 86.99 \(\pm\) 0.08 & 4.710e-1 & 2.007e-1 & 6.462e-2\\
 & $P=8$  & 289.3 & 90.30 \(\pm\) 0.03 & 3.217e-1 & 1.445e-1 & 1.180e-2\\
 & S-HMC\(_\parallel\) & 200 & 90.23 \(\pm\) 0.08 & 2.990e-1 & 1.466e-1 & 2.518e-2\\
  & $P=8$ & 200 & 90.82 \(\pm\) 0.03 & 2.810e-1 & 1.395e-1 & 3.481e-2\\
 \midrule
 \multirow{2}{*}{\(0.1\)} 
 & S-SMC\(_\parallel\) & 229.6 & 88.26 \(\pm\) 0.07 & 3.855e-1 & 1.770e-1 & 4.593e-2\\
  & $P=8$ & 225.3 & 90.45 \(\pm\) 0.06 & 2.980e-1 & 1.400e-1 & 7.737e-3\\
 & S-HMC\(_\parallel\) & 200 & 90.57 \(\pm\) 0.04 & 2.823e-1 & 1.398e-1 & 1.073e-2\\
  & $P=8$ & 200 & 90.83 \(\pm\) 0.03 & 2.701e-1 & 1.353e-1 & 1.517e-2\\
\midrule
\multirow{2}{*}{\(0.05\)} 
 & S-SMC\(_\parallel\)      & 168.8 & 89.26 \(\pm\) 0.07 & 3.408e-1 & 1.580e-1 & 3.470e-2 \\
  & $P=8$ & 174.3 & 90.63 \(\pm\) 0.05 & 2.881e-1 & 1.371e-1 & 9.720e-3 \\
 & S-HMC\(_\parallel\)      & 200   & 90.67 \(\pm\) 0.03 & 2.749e-1 & 1.366e-1 & 6.598e-3 \\
  & $P=8$ & 200   & 90.84 \(\pm\) 0.03 & 2.677e-1 & 1.340e-1 & 6.601e-3 \\
\midrule
 \multirow{2}{*}{\(0\)} 
 & MAP & 200 & 90.39\(\pm\)0.07 & 2.913e-1 & 1.420e-1 & 2.502e-2 \\
 & DE (\(N\)) & 200 & 90.81\(\pm\)0.03 & 2.741e-1 & 1.355e-1 & 1.770e-2\\
\bottomrule
\end{tabular}

\vspace{0.2cm}

\begin{tabular}{lc|cc|ccc|cc|ccc}
\toprule
\(s\) & Method & \multicolumn{5}{c|}{$H_{\sf ep}$} & \multicolumn{5}{c}{$H_{\sf tot}$} \\
\cmidrule(lr){3-7} \cmidrule(lr){8-12}
& & \multicolumn{2}{c|}{ID} & \multicolumn{3}{c|}{OOD} & \multicolumn{2}{c|}{ID} & \multicolumn{3}{c}{OOD} \\
\cmidrule(lr){3-4} \cmidrule(lr){5-7} \cmidrule(lr){8-9} \cmidrule(lr){10-12}
& & cor. & inc. & close & corrupt & far & cor. & inc. & close & corrupt & far \\
\midrule
\multirow{2}{*}{\(0.2\)} 
 & S-SMC\(_\parallel\) & 3.682e-4 & 1.947e-3 & 2.063e-3 & 1.092e-3 & 1.629e-3 & 1.136e-1 & 5.613e-1 & 5.440e-1 & 3.630e-1 & 7.756e-1 \\
  & $P=8$ & 8.362e-2 & 3.326e-1 & 4.244e-1 & 2.361e-1 & 4.065e-1 & 1.071e-1 & 5.954e-1 & 9.675e-1 & 6.080e-1 & 1.160e+0 \\
 & S-HMC\(_\parallel\) & 1.159e-1 & 4.091e-1 & 5.195e-1 & 2.993e-1 & 5.333e-1 & 2.676e-1 & 9.231e-1 & 1.059e+0 & 6.768e-1 & 1.297e+0 \\
  & $P=8$ & 1.405e-1 & 4.604e-1 & 6.042e-1 & 3.476e-1 & 6.129e-1 & 2.945e-1 & 9.687e-1 & 1.146e+0 & 7.256e-1 & 1.364e+0 \\
\midrule
\multirow{2}{*}{\(0.1\)} 
 & S-SMC\(_\parallel\) & 2.948e-4 & 1.507e-3 & 1.603e-3 & 8.256e-4 & 1.450e-3 & 1.309e-1 & 6.364e-1 & 6.121e-1 & 4.027e-1 & 9.110e-1 \\
  & $P=8$ & 5.539e-2 & 2.369e-1 & 3.054e-1 & 1.636e-1 & 2.937e-1 & 1.217e-1 & 6.453e-1 & 9.184e-1 & 5.690e-1 & 1.156e+0 \\
 & S-HMC\(_\parallel\) & 7.055e-2 & 2.795e-1 & 3.591e-1 & 1.972e-1 & 3.581e-1 & 2.241e-1 & 8.596e-1 & 9.703e-1 & 6.102e-1 & 1.219e+0 \\
  & $P=8$ & 1.035e-1 & 3.121e-1 & 4.095e-1 & 2.244e-1 & 4.031e-1 & 2.367e-1 & 8.884e-1 & 1.023e+0 & 6.371e-1 & 1.257e+0 \\
\midrule
\multirow{2}{*}{\(0.05\)} 
 & S-SMC\(_\parallel\) & 4.258e-4 & 2.273e-3 & 2.515e-3 & 1.297e-3 & 2.185e-3 & 1.351e-1 & 6.620e-1 & 6.639e-1 & 4.300e-1 & 9.008e-1 \\
  & $P=8$ & 3.507e-2 & 1.584e-1 & 2.027e-1 & 1.058e-1 & 1.961e-1 & 1.311e-1 & 6.684e-1 & 8.643e-1 & 5.368e-1 & 1.111e+0 \\
 & S-HMC\(_\parallel\) & 4.060e-2 & 1.804e-1 & 2.308e-1 & 1.228e-1 & 2.243e-1 & 1.917e-1 & 8.073e-1 & 8.900e-1 & 5.558e-1 & 1.154e+0 \\
  & $P=8$ & 4.579e-2 & 1.966e-1 & 2.564e-1 & 1.358e-1 & 2.472e-1 & 1.971e-1 & 8.203e-1 & 9.173e-1 & 5.684e-1 & 1.168e+0 \\
\midrule
\multirow{2}{*}{\(0\)} 
 & MAP & 0 & 0 & 0 & 0 & 0 & 1.423e-1 & 7.037e-1 & 7.258e-1 & 4.577e-1 & 1.065e+0 \\
 & DE (\(N\)) & 9.291e-3 & 4.753e-2 & 4.861e-2 & 2.603e-2 & 3.930e-2 & 1.541e-1& 7.275e-1 & 7.629e-1 & 4.786e-1 & 1.029e+0\\
\bottomrule
\end{tabular}
\end{table}

\begin{figure}[H]
  \centering
  \begin{subfigure}[b]{0.32\textwidth}
    \includegraphics[width=\linewidth]{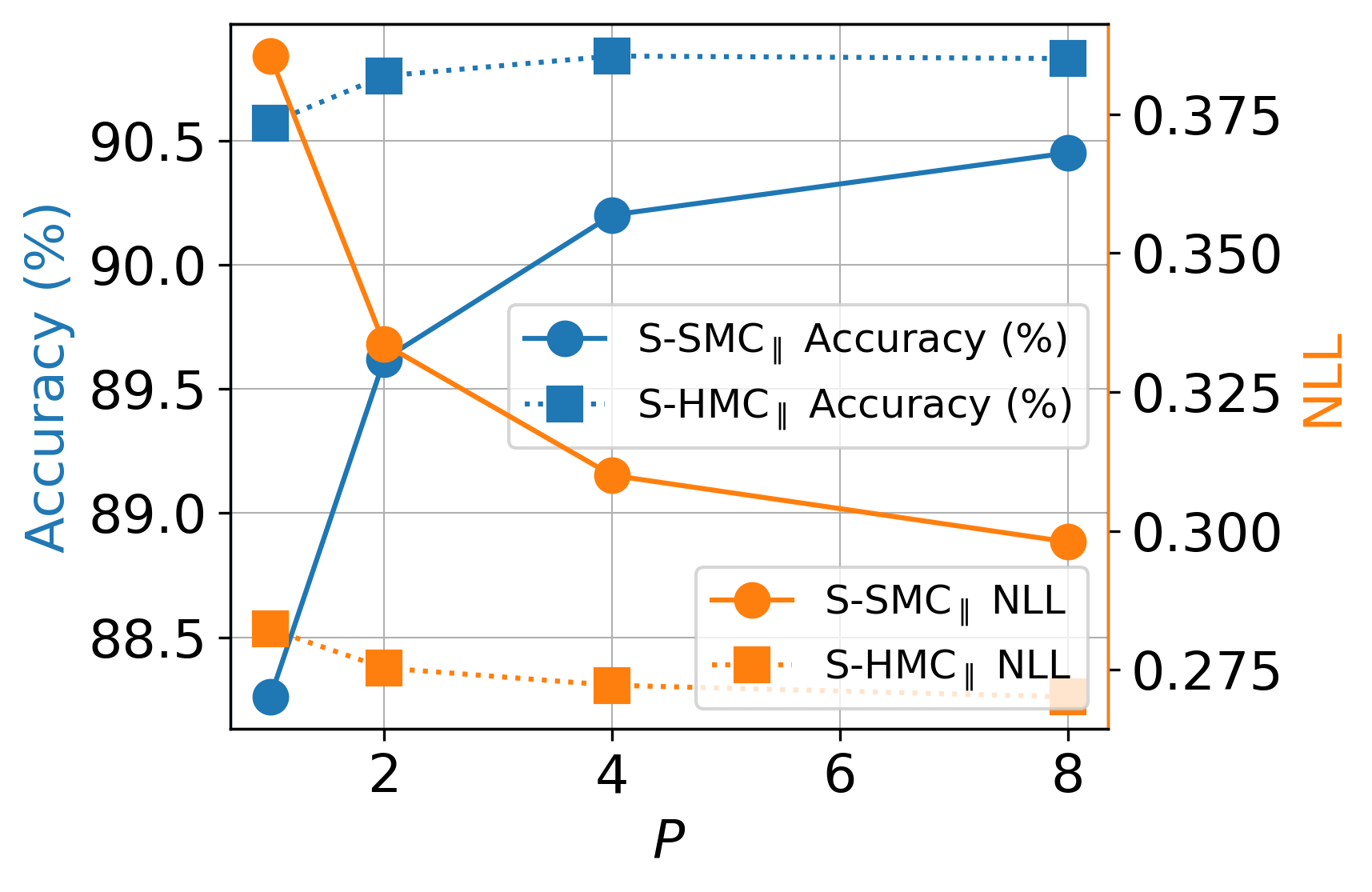}
  \end{subfigure}
  \hfill
  \begin{subfigure}[b]{0.32\textwidth}
    \includegraphics[width=\linewidth]{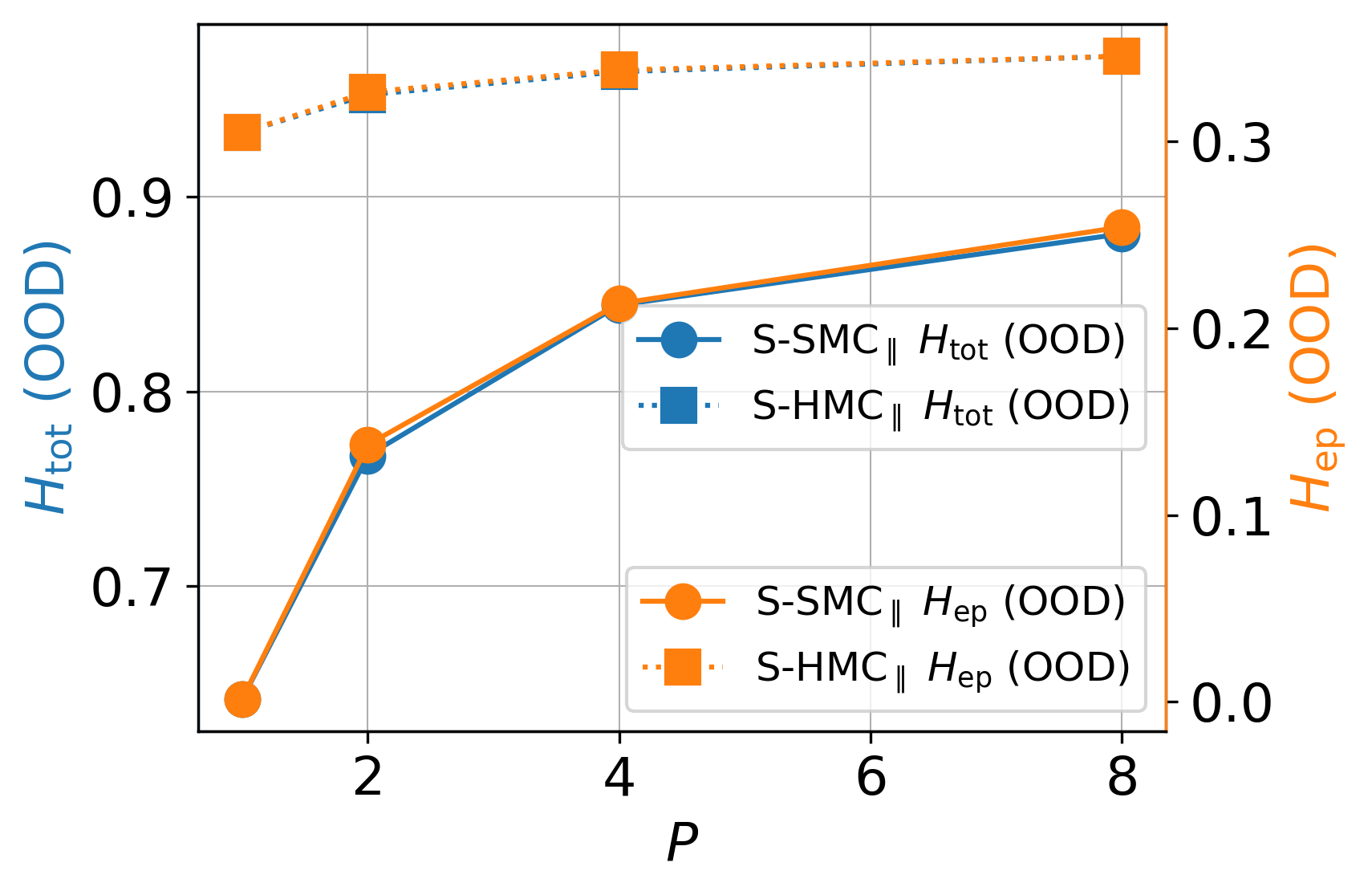}
  \end{subfigure}
  \hfill
  \begin{subfigure}[b]{0.32\textwidth}
    \includegraphics[width=\linewidth]{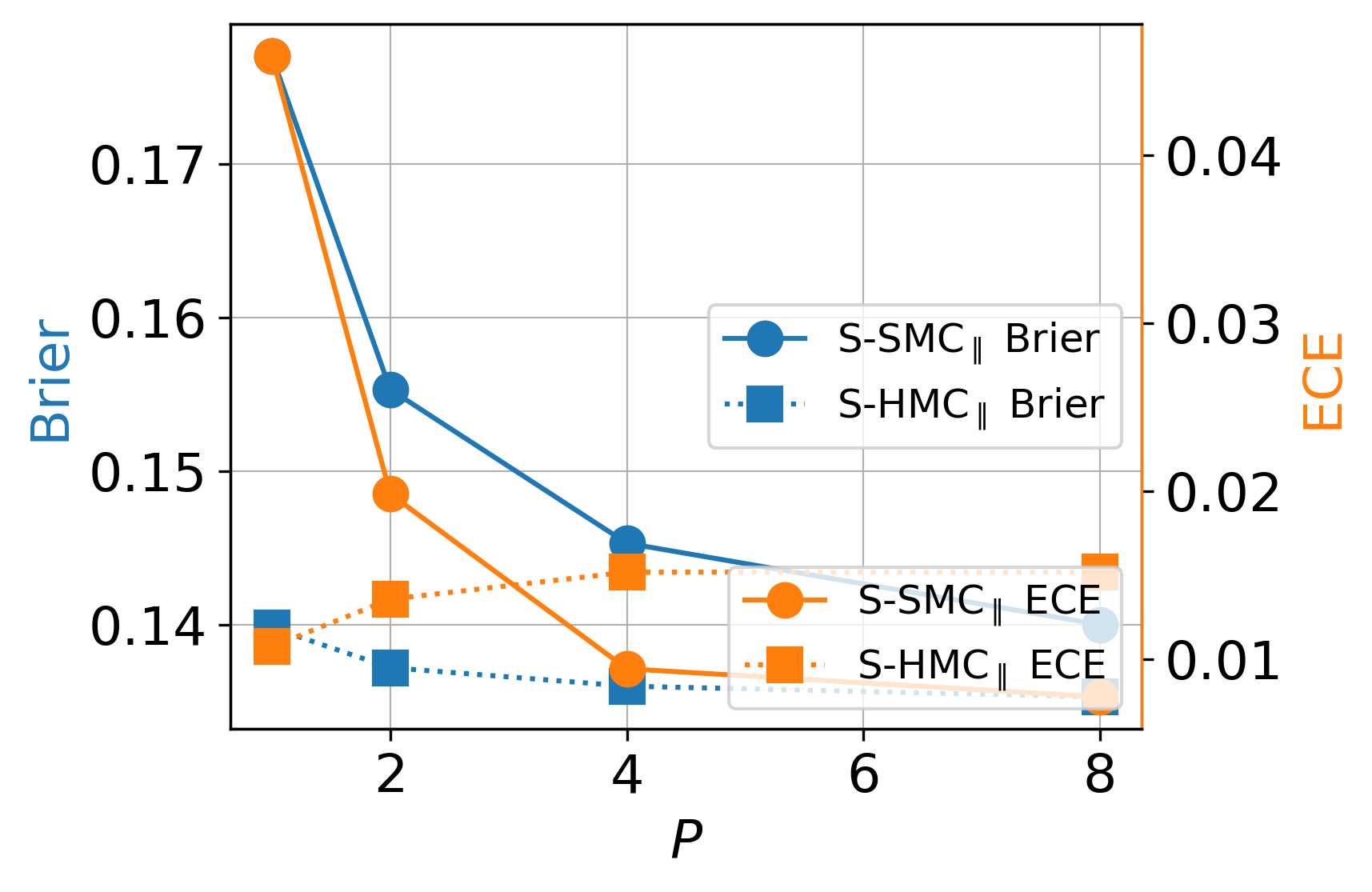}
  \end{subfigure}
  \caption{Comparison of S-SMC\(_\parallel\) ($P$ chains with $N=10$) and S-HMC\(_\parallel\) ($NP$ chains), with fixed number of leapfrog $L=1$, $B=200$, $M=4$, $v=0.2$ and \(s=0.1\), on CIFAR10 (5 realizations).}
  \label{fig:metric_diffP_cifar_v02_s01}
\end{figure}

\begin{figure}[H]
  \centering
  \begin{subfigure}[b]{0.32\textwidth}
    \includegraphics[width=\linewidth]{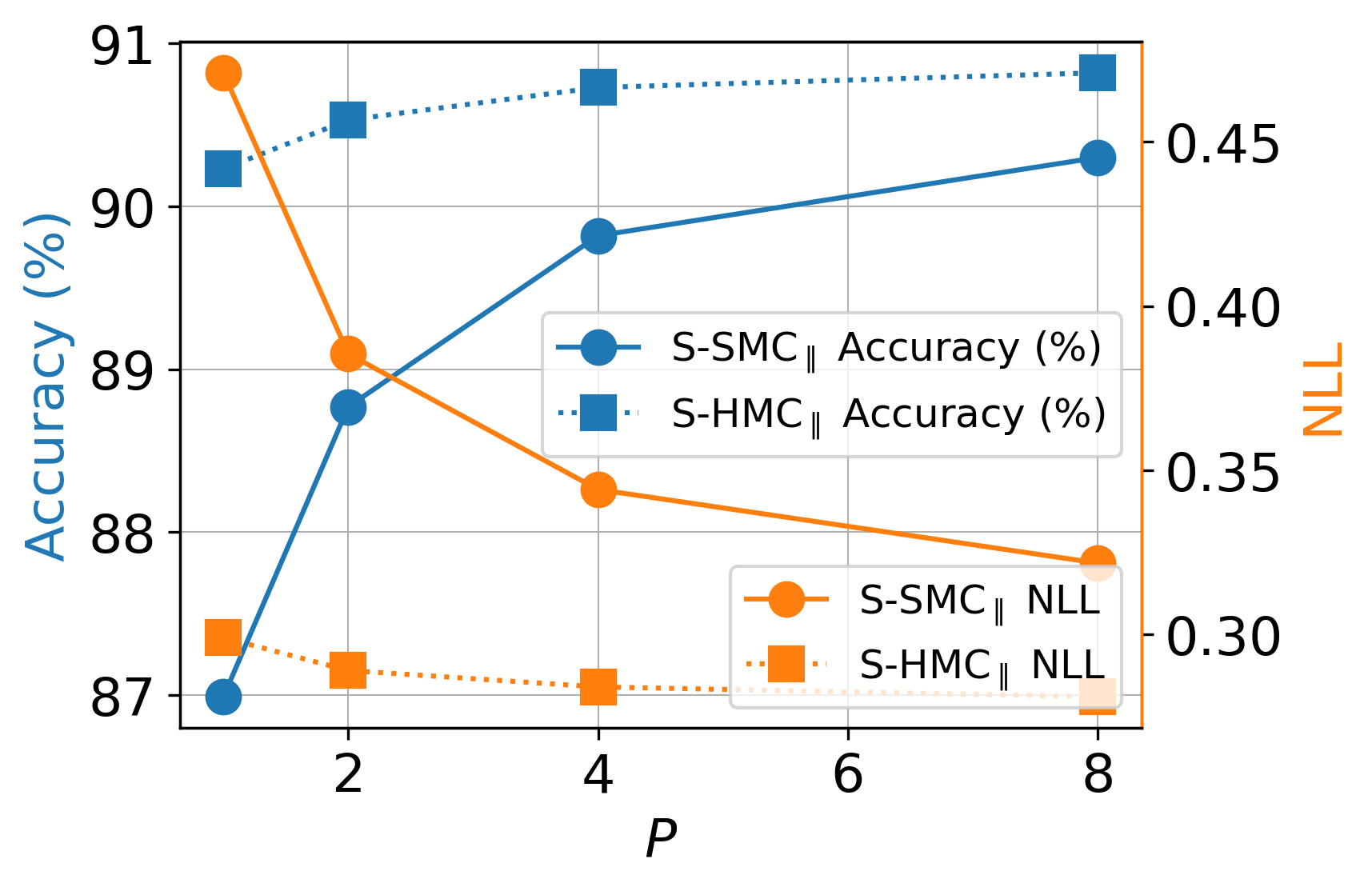}
  \end{subfigure}
  \hfill
  \begin{subfigure}[b]{0.32\textwidth}
    \includegraphics[width=\linewidth]{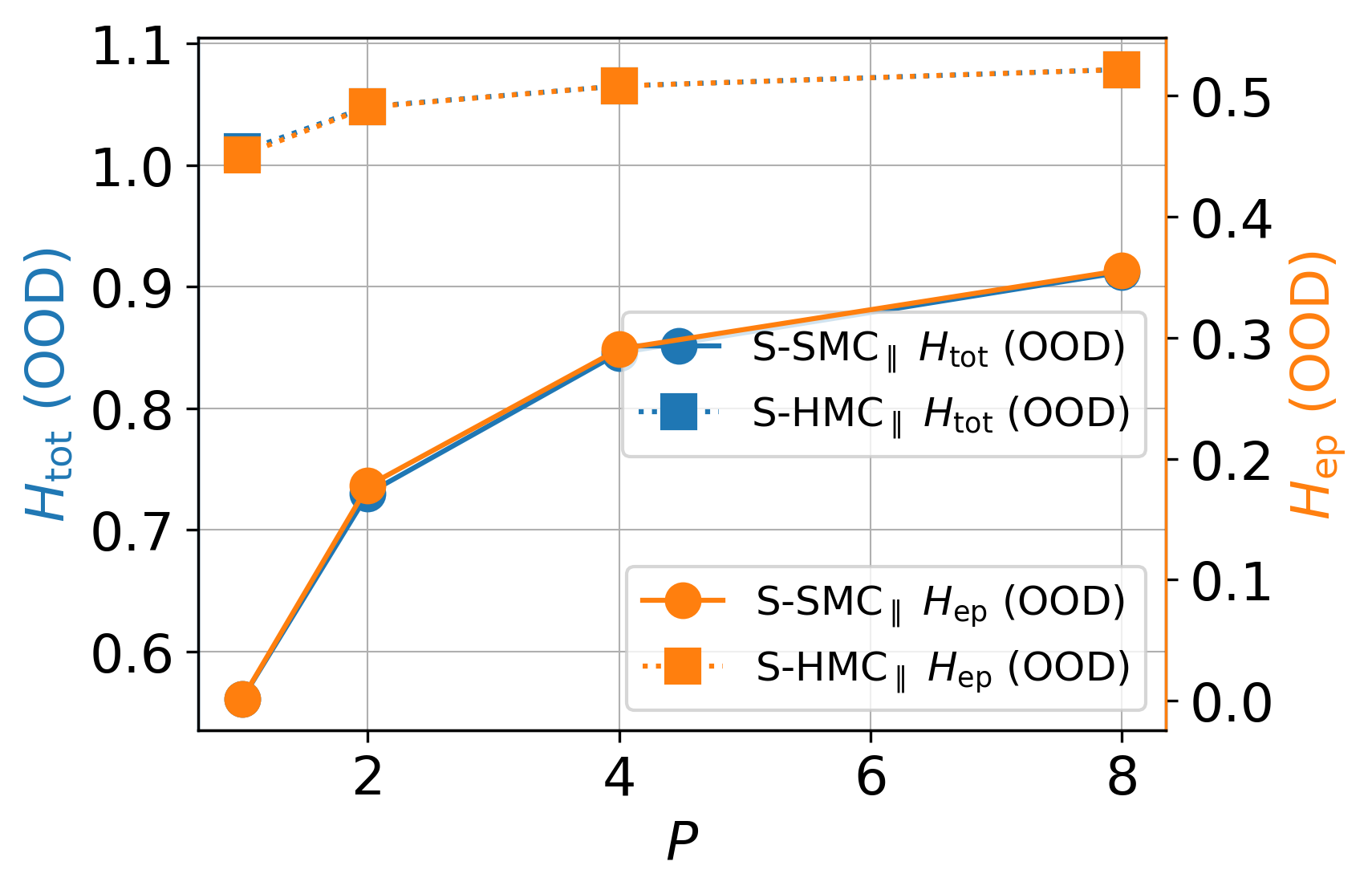}
  \end{subfigure}
  \hfill
  \begin{subfigure}[b]{0.32\textwidth}
    \includegraphics[width=\linewidth]{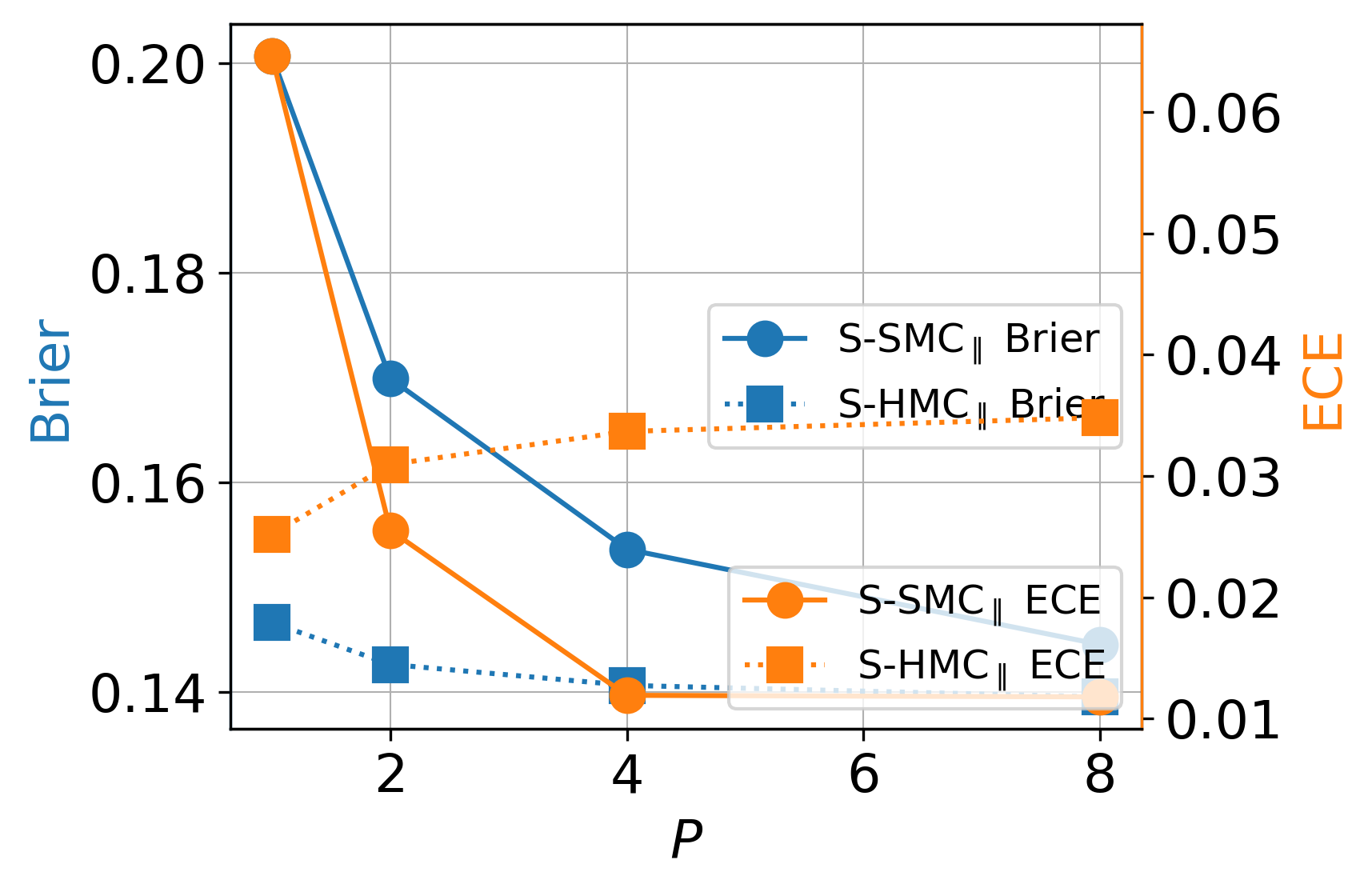}
  \end{subfigure}
  \caption{Comparison of S-SMC\(_\parallel\) ($P$ chains with $N=10$) and S-HMC\(_\parallel\) ($NP$ chains), with fixed number of leapfrog $L=1$, $B=200$, $M=4$, $v=0.2$ and \(s=0.2\), on CIFAR10 (5 realizations).}
  \label{fig:metric_diffP_cifar_v02_s02}
\end{figure}

\section{All-inclusive data tables}
\label{app:tables}

\begin{table}[h]
\centering
\caption{Comparison in all domains among S-SMC$_\parallel$ ($P=1$ chain with $N=10$), S-HMC$_{\parallel}$ ($NP$ chains), HMC (GS) ($2e4$ samples), DE ($N$ models) and MAP, with fixed number of leapfrog $L=1$, $v=0.1$ and $s=0.1$, on MNIST7 (5 realizations and $\pm$ s.e. in entropy).}
\label{tab:full_digits_mnist}
\begin{adjustbox}{width=\textwidth}
\begin{tabular}{l|c|c|c|c|c|c|c|c|c|c|c|c|c}
\toprule
Group & \multicolumn{1}{c|}{MAP} & \multicolumn{3}{c|}{DE} & \multicolumn{3}{c|}{S-HMC (\(s=0.1\))} & \multicolumn{3}{c|}{S-SMC (\(s=0.1\))} & \multicolumn{3}{c}{HMC (GS) (\(s=1\))}\\
\cmidrule(rl){2-2} \cmidrule(rl){3-5} \cmidrule(rl){6-8} \cmidrule(rl){9-11} \cmidrule(rl){12-14}
 & $H_{\mathsf{tot}}$ & $H_{\mathsf{tot}}$ & $H_{\mathsf{al}}$ & $H_{\mathsf{ep}}$ & $H_{\mathsf{tot}}$ & $H_{\mathsf{al}}$ & $H_{\mathsf{ep}}$ & $H_{\mathsf{tot}}$ & $H_{\mathsf{al}}$ & $H_{\mathsf{ep}}$ & $H_{\mathsf{tot}}$ & $H_{\mathsf{al}}$ & $H_{\mathsf{ep}}$ \\
\midrule
Digit 0 & 1.276e-1 & 1.307e-1 & 1.237e-1 & 7.076e-3 & 1.671e-1 & 1.228e-1 & 4.427e-2 & 1.110e-1 & 8.528e-2 & 2.574e-2 & 2.157e-1 & 1.268e-1 & 8.893e-2 \\
\midrule
Digit 1 & 1.768e-1 & 1.840e-1 & 1.781e-1 & 5.823e-3 & 1.889e-1 & 1.585e-1 & 3.038e-2 & 1.408e-1 & 1.255e-1 & 1.535e-2 & 2.124e-1 & 1.621e-1 & 5.025e-2 \\
\midrule
Digit 2 & 2.294e-1 & 2.266e-1 & 2.142e-1 & 1.236e-2 & 2.980e-1 & 2.168e-1 & 8.122e-2 & 2.090e-1 & 1.679e-1 & 4.118e-2 & 3.410e-1 & 1.988e-1 & 1.423e-1 \\
\midrule
Digit 3 & 3.168e-1 & 3.493e-1 & 3.222e-1 & 2.711e-2 & 3.883e-1 & 2.896e-1 & 9.873e-2 & 2.686e-1 & 2.245e-1 & 4.404e-2 & 4.229e-1 & 2.616e-1 & 1.613e-1 \\
\midrule
Digit 4 & 2.158e-1 & 2.221e-1 & 2.103e-1 & 1.182e-2 & 2.753e-1 & 2.095e-1 & 6.583e-2 & 1.981e-1 & 1.654e-1 & 3.272e-2 & 2.925e-1 & 1.847e-1 & 1.078e-1 \\
\midrule
Digit 5 & 3.993e-1 & 4.058e-1 & 3.787e-1 & 2.712e-2 & 4.395e-1 & 3.224e-1 & 1.171e-1 & 3.056e-1 & 2.520e-1 & 5.366e-2 & 4.428e-1 & 2.655e-1 & 1.773e-1 \\
\midrule
Digit 6 & 1.856e-1 & 2.045e-1 & 1.927e-1 & 1.180e-2 & 2.836e-1 & 2.047e-1 & 7.891e-2 & 1.856e-1 & 1.495e-1 & 3.605e-2 & 2.968e-1 & 1.785e-1 & 1.182e-1 \\
\midrule
Digit 7 & 1.897e-1 & 1.957e-1 & 1.859e-1 & 9.730e-3 & 2.403e-1 & 1.802e-1 & 6.008e-2 & 1.693e-1 & 1.396e-1 & 2.967e-2 & 2.528e-1 & 1.569e-1 & 9.589e-2 \\
\midrule
Digit 8 & 9.507e-1 & 9.821e-1 & 9.078e-1 & 7.433e-2 & 1.091e+0 & 7.836e-1 & 3.072e-1 & 8.975e-1 & 7.591e-1 & 1.384e-1 & 1.121e+0 & 6.333e-1 & 4.873e-1 \\
\midrule
Digit 9 & 6.157e-1 & 6.393e-1 & 6.046e-1 & 3.468e-2 & 7.567e-1 & 5.626e-1 & 1.941e-1 & 6.591e-1 & 5.651e-1 & 9.406e-2 & 9.210e-1 & 5.554e-1 & 3.657e-1 \\
\midrule
Perturbed & 7.528e-1 & 8.112e-1 & 7.006e-1 & 1.106e-1 & 8.443e-1 & 4.227e-1 & 4.216e-1 & 8.001e-1 & 6.169e-1 & 1.832e-1 & 1.228e+0 & 2.819e-1 & 9.466e-1 \\
\midrule
White Noise & 7.703e-1 & 9.806e-1 & 7.117e-1 & 2.690e-1 & 1.453e+0 & 5.221e-1 & 9.304e-1 & 9.744e-1 & 5.800e-1 & 3.944e-1 & 1.398e+0 & 3.444e-1 & 1.053e+0 \\
\midrule
All ID & 2.301e-1 & 2.398e-1 & 2.069e-1 & 3.291e-2 & 2.821e-1 & 2.111e-1 & 7.090e-2 & 1.966e-1 & 1.623e-1 & 3.429e-2 & 1.021e+0 & 5.943e-1 & 4.265e-1 \\
\bottomrule
\end{tabular}
\end{adjustbox}
\end{table}

\begin{table}[h]
\centering
\scriptsize
\caption{Comparison in all domains among S-SMC\(_\parallel\) (\(P=1\) chain with \(N=10\)), S-HMC\(_{\parallel}\) (\(NP\) chains), DE (\(N\) models) and MAP, with fixed number of leapfrog \(L=1\), \(B=25\), \(M=1\), \(v=0.025\) and \(s=0.1\), on IMDb (\(5\) realizations).}
\label{tab:imdb_fullUQ}
\begin{adjustbox}{width=\textwidth}
\begin{tabular}{l|c|c|c|c|c|c|c|c|c|c}
\toprule
Group & \multicolumn{1}{c|}{MAP} & \multicolumn{3}{c|}{DE} & \multicolumn{3}{c|}{S-HMC$_\parallel$} & \multicolumn{3}{c}{S-SMC$_\parallel$} \\
\cmidrule(rl){2-2} \cmidrule(rl){3-5} \cmidrule(rl){6-8} \cmidrule(rl){9-11}
 & $H_{\mathsf{tot}}$ & $H_{\mathsf{tot}}$ & $H_{\mathsf{al}}$ & $H_{\mathsf{ep}}$ & $H_{\mathsf{tot}}$ & $H_{\mathsf{al}}$ & $H_{\mathsf{ep}}$ & $H_{\mathsf{tot}}$ & $H_{\mathsf{al}}$ & $H_{\mathsf{ep}}$ \\
\midrule
Negative & 4.352e-1 & 4.407e-1 & 4.406e-1 & 9.314e-5 & 4.892e-1 & 4.889e-1 & 2.482e-4 & 4.929e-1 & 4.927e-1 & 1.148e-4 \\
\midrule
Positive & 5.716e-1 & 5.688e-1 & 5.687e-1 & 1.188e-4 & 5.031e-1 & 5.029e-1 & 2.659e-4 & 5.072e-1 & 5.071e-1 & 1.240e-4 \\
\midrule
Meta & 6.117e-1 & 6.098e-1 & 6.097e-1 & 5.026e-5 & 5.331e-1 & 5.326e-1 & 4.633e-4 & 5.463e-1 & 5.461e-1 & 2.200e-4 \\
\midrule
Full Meta & 6.658e-1 & 6.649e-1 & 6.649e-1 & 5.548e-5 & 6.185e-1 & 6.180e-1 & 5.410e-4 & 6.260e-1 & 6.257e-1 & 3.212e-4 \\
\midrule
Reviews & 5.814e-1 & 5.793e-1 & 5.793e-1 & 5.064e-5 & 5.165e-1 & 5.160e-1 & 4.598e-4 & 5.251e-1 & 5.249e-1 & 1.792e-4 \\
\midrule
Full reviews & 6.705e-1 & 6.702e-1 & 6.701e-1 & 6.302e-5 & 6.400e-1 & 6.395e-1 & 5.057e-4 & 6.457e-1 & 6.454e-1 & 3.285e-4 \\
\midrule
Lipsum & 5.894e-1 & 5.882e-1 & 5.881e-1 & 4.909e-5 & 5.079e-1 & 5.074e-1 & 4.260e-4 & 5.142e-1 & 5.140e-1 & 1.749e-4 \\
\midrule
All ID & 5.034e-1 & 5.048e-1 & 5.046e-1 & 1.060e-4 & 4.962e-1 & 4.959e-1 & 2.570e-4 & 5.000e-1 & 4.999e-1 & 1.194e-4 \\
\bottomrule
\end{tabular}
\end{adjustbox}
\end{table}

\begin{table}[h]
\centering
\caption{Comparison in all domains among S-SMC\(_\parallel\) (\(P=8\) chain with \(N=10\)), S-HMC\(_{\parallel}\) (\(NP\) chains), DE (\(N\) models) and MAP, with fixed number of leapfrog \(L=1\), \(B=26\), \(M=2\), \(v=1\) and \(s=0.35\), on IMDb (\(5\) realizations).}
\label{tab:imdb_fullUQ_v1}
\begin{adjustbox}{width=\textwidth}
\begin{tabular}{l|c|c|c|c|c|c|c|c|c|c}
\toprule
Group & \multicolumn{1}{c|}{MAP} & \multicolumn{3}{c|}{DE} & \multicolumn{3}{c|}{S-HMC\(_\parallel\)} & \multicolumn{3}{c}{S-SMC\(_\parallel\)} \\
\cmidrule(rl){2-2} \cmidrule(rl){3-5} \cmidrule(rl){6-8} \cmidrule(rl){9-11}
 & $H_{\mathsf{tot}}$ & $H_{\mathsf{tot}}$ & $H_{\mathsf{al}}$ & $H_{\mathsf{ep}}$ & $H_{\mathsf{tot}}$ & $H_{\mathsf{al}}$ & $H_{\mathsf{ep}}$ & $H_{\mathsf{tot}}$ & $H_{\mathsf{al}}$ & $H_{\mathsf{ep}}$ \\
\midrule
Negative & 2.489e-1 & 2.494e-1 & 2.483e-1 & 1.033e-3 & 3.107e-1 & 2.928e-1 & 1.794e-2 & 2.962e-1 & 2.860e-1 & 1.016e-2 \\
\midrule
Positive & 3.629e-1 & 3.675e-1 & 3.659e-1 & 1.631e-3 & 3.296e-1 & 3.099e-1 & 1.971e-2 & 3.160e-1 & 3.048e-1 & 1.126e-2 \\
\midrule
Meta & 5.598e-1 & 5.767e-1 & 5.507e-1 & 2.594e-2 & 5.626e-1 & 4.887e-1 & 7.391e-2 & 5.555e-1 & 5.004e-1 & 5.514e-2 \\
\midrule
Full Meta & 5.561e-1 & 5.737e-1 & 5.469e-1 & 2.683e-2 & 6.088e-1 & 5.251e-1 & 8.360e-2 & 5.920e-1 & 5.277e-1 & 6.435e-2 \\
\midrule
Reviews & 4.071e-1 & 4.251e-1 & 4.030e-1 & 2.212e-2 & 4.188e-1 & 3.657e-1 & 5.315e-2 & 3.987e-1 & 3.611e-1 & 3.756e-2 \\
\midrule
Full reviews & 5.695e-1 & 6.007e-1 & 5.574e-1 & 4.331e-2 & 6.156e-1 & 5.246e-1 & 9.098e-2 & 6.025e-1 & 5.420e-1 & 6.049e-2 \\
\midrule
Lipsum & 5.462e-1 & 5.556e-1 & 5.283e-1 & 2.733e-2 & 5.386e-1 & 4.695e-1 & 6.917e-2 & 5.304e-1 & 4.778e-1 & 5.260e-2 \\
\midrule
All ID & 3.059e-1 & 3.084e-1 & 3.071e-1 & 1.332e-3 & 3.202e-1 & 3.013e-1 & 1.883e-2 & 3.061e-1 & 2.954e-1 & 1.071e-2 \\
\bottomrule
\end{tabular}
\end{adjustbox}
\end{table}

\begin{table}[h]
\tiny
\caption{Comparison of S-SMC\(_\parallel\) (\(P\) chains with \(N=10\)) and S-HMC\(_{\parallel}\) (\(NP\) chains), with fixed number of leapfrog \(L=1\), \(B=160\), \(M=7\), \(v = 0.1\) and \(s=0.25\), on MNIST7 (5 realizations and \(\pm\) s.e. in accuracy).}
\label{tab:metric_diffP_mnist_s025}
\centering
\begin{tabular}{lc|c|c|c|c}
\toprule
\(P\) & Method & Ep. & Acc. & NL & Brier \\
\midrule
1 & S-SMC\(_\parallel\) & 166.6 & 90.35\(\pm\)0.26 & 3.300e-1 & 1.441e-1 \\ 
1 & S-HMC\(_\parallel\) & 160 & 92.79\(\pm\)0.19 & 2.571e-1 & 1.156e-1 \\
\midrule
2 & S-SMC\(_\parallel\) & 160.3 & 92.00\(\pm\)0.24 & 2.726e-1 & 1.239e-1\\ 
2 & S-HMC\(_\parallel\) & 160 & 92.97\(\pm\)0.14 & 2.536e-1 & 1.140e-1 \\
\midrule
4 & S-SMC\(_\parallel\) & 164.5 & 92.59\(\pm\)0.15 & 2.504e-1 & 1.152e-1 \\
4 & S-HMC\(_\parallel\) & 160 & 93.13\(\pm\)0.07 & 2.506e-1 & 1.133e-1 \\
\midrule
8 & S-SMC\(_\parallel\) & 161.5 & 93.00\(\pm\)0.11 & 2.366e-1 & 1.096e-1 \\
8 & S-HMC\(_\parallel\) & 160 & 93.15\(\pm\)0.05 & 2.490e-1 & 1.127e-1 \\
\midrule
& HMC (GS) & 2e4 & 92.87\(\pm\)0.48 & 2.376e-1 & 1.079e-1 \\
\bottomrule
\end{tabular}

\vspace{0.2cm}

\begin{adjustbox}{width=\textwidth}
\begin{tabular}{lc|cc|cccc|cc|cccc}
\toprule
\(P\) & Method & \multicolumn{6}{c|}{\(H_{\sf ep}\)} & \multicolumn{6}{c}{\(H_{\sf tot}\)}\\
  \cmidrule{3-8} \cmidrule{9-14}
 & & \multicolumn{2}{c|}{ID} & \multicolumn{4}{c|}{OOD} 
  & \multicolumn{2}{c|}{ID} & \multicolumn{4}{c}{OOD} \\
  \cmidrule{3-4} \cmidrule{5-8} \cmidrule{9-10} \cmidrule{11-14}
 & & cor. & inc. & 8 & 9 & wn & per. 
  & cor. & inc. & 8 & 9 & wn & per. \\
\midrule
1 & S-SMC\(_\parallel\) & 2.257e-2 & 1.094e-1 & 1.146e-1 & 7.791e-2 & 3.847e-1 & 1.914e-1 & 1.508e-1 & 6.679e-1 & 8.384e-1 & 5.945e-1 & 8.290e-1 & 7.110e-1 \\
1 & S-HMC\(_\parallel\) & 1.133e-1 & 4.232e-1 & 4.985e-1 & 3.225e-1 & 1.281e+0 & 6.314e-1 & 3.149e-1 & 1.026e+0 & 1.220e+0 & 8.606e-1 & 1.614e+0 & 1.019e+0 \\
\midrule
2 & S-SMC\(_\parallel\) & 5.445e-2 & 2.442e-1 & 1.975e-1 & 1.402e-1 & 5.073e-1 & 2.923e-1 & 1.371e-1 & 6.768e-1 & 9.108e-1 & 6.629e-1 & 9.559e-1 & 8.111e-1 \\
2 & HMC\(_\parallel\) & 1.298e-1 & 4.568e-1 & 5.380e-1 & 3.638e-1 & 1.321e+0 & 6.720e-1 & 3.358e-1 & 1.050e+0 & 1.250e+0 & 8.975e-1 & 1.680e+0 & 1.050e+0 \\
\midrule
4 & S-SMC\(_\parallel\) & 7.492e-2 & 3.240e-1 & 2.544e-1 & 1.797e-1 & 6.264e-1 & 3.749e-1 & 1.287e-1 & 6.669e-1 & 9.591e-1 & 7.064e-1 & 1.051e+0 & 8.923e-1 \\
4 & HMC\(_\parallel\) & 1.356e-1 & 4.718e-1 & 5.508e-1 & 3.634e-1 & 1.327e+0 & 7.084e-1 & 3.443e-1 & 1.065e+0 & 1.261e+0 & 9.031e-1 & 1.689e+0 & 1.078e+0 \\
\midrule
8 & S-SMC\(_\parallel\) & 8.828e-2 & 3.717e-1 & 2.984e-1 & 2.089e-1 & 7.384e-1 & 4.475e-1 & 1.247e-1 & 6.641e-1 & 1.001e+0 & 7.354e-1 & 1.152e+0 & 9.627e-1 \\
8 & S-HMC\(_\parallel\) & 1.384e-1 & 4.788e-1 & 5.572e-1 & 3.678e-1 & 1.349e+0 & 7.235e-1 & 3.456e-1 & 1.070e+0 & 1.267e+0 & 9.025e-1 & 1.715e+0 & 1.094e+0 \\
\midrule
& HMC (GS) & 7.199e-2 & 3.432e-1 & 3.887e-1 & 2.748e-1 & 1.169e+0 & 5.579e-1 & 2.045e-1 & 8.566e-1 & 9.984e-1 & 7.425e-1 & 1.574e+0 & 8.725e-1 \\
\bottomrule
\end{tabular}
\end{adjustbox}
\end{table}

\begin{table}[h]
\tiny
\caption{Comparison of S-SMC\(_\parallel\) (\(P\) chains with \(N=10\)) and S-HMC\(_{\parallel}\) (\(NP\) chains), with fixed number of leapfrog \(L=1\), \(B=160\), \(M=10\), \(v = 0.1\) and \(s=0.1\), on MNIST7 (5 realizations and \(\pm\) s.e. in accuracy).} 
\label{tab:metric_diffP_mnist_s01}
\centering
\begin{tabular}{lc|c|c|c|c}
\toprule
\(P\) & Method & Ep. & Acc. & NL & Brier \\
\midrule
1 & S-SMC\(_\parallel\) & 170.0 & 92.17\(\pm\)0.37 & 2.671e-1 & 1.186e-1 \\
1 & S-HMC\(_\parallel\) & 160 & 92.96\(\pm\)0.17 & 2.326e-1 & 1.071e-1 \\
\midrule
2 & S-SMC\(_\parallel\) & 179.0 & 92.52\(\pm\)0.30 & 2.507e-1 & 1.127e-1 \\
2 & S-HMC\(_\parallel\) & 160 & 93.10\(\pm\)0.12 & 2.310e-1 & 1.067e-1 \\
\midrule
4 & S-SMC\(_\parallel\) & 180.5 & 93.01\(\pm\)0.29 & 2.369e-1 & 1.069e-1 \\
4 & HMC\(_\parallel\) & 160 & 93.12\(\pm\)0.09 & 2.306e-1 & 1.069e-1 \\
\midrule
8 & S-SMC\(_\parallel\) & 178.0 & 93.26\(\pm\)0.16 & 2.259e-1 & 1.025e-1 \\
8 & S-HMC\(_\parallel\) & 160 & 93.12\(\pm\)0.09 & 2.310e-1 & 1.072e-1 \\
\midrule
& HMC (GS) & 2e4 & 92.92\(\pm\)0.41 & 2.366e-1 & 1.084e-1 \\
\bottomrule
\end{tabular}

\vspace{0.2cm}

\begin{adjustbox}{width=\textwidth}
\begin{tabular}{lc|cc|cccc|cc|cccc}
\toprule
\(P\) & Method & \multicolumn{6}{c|}{$H_{\mathsf{ep}}$} & \multicolumn{6}{c}{$H_{\mathsf{tot}}$} \\
\cmidrule{3-8} \cmidrule{9-14}
 & & \multicolumn{2}{c|}{ID} & \multicolumn{4}{c|}{OOD} & \multicolumn{2}{c|}{ID} & \multicolumn{4}{c}{OOD} \\
\cmidrule{3-4} \cmidrule{5-8} \cmidrule{9-10} \cmidrule{11-14}
 & & cor. & inc. & 8 & 9 & wn & per. & cor. & inc. & 8 & 9 & wn & per. \\
\midrule
1 & S-SMC\(_\parallel\) & 2.642e-2 & 1.288e-1 & 1.384e-1 & 9.406e-2 & 3.972e-1 & 1.776e-1 & 1.536e-1 & 7.042e-1 & 8.975e-1 & 6.591e-1 & 9.753e-1 & 7.859e-1 \\
1 & S-HMC\(_\parallel\) & 5.624e-2 & 2.645e-1 & 3.072e-1 & 1.941e-1 & 9.259e-1 & 4.100e-1 & 2.343e-1 & 9.132e-1 & 1.091e+0 & 7.567e-1 & 1.443e+0 & 8.248e-1 \\
\midrule
2 & S-SMC\(_\parallel\) & 4.233e-2 & 2.042e-1 & 1.836e-1 & 1.227e-1 & 5.306e-1 & 2.620e-1 & 1.449e-1 & 7.053e-1 & 9.422e-1 & 6.870e-1 & 1.058e+0 & 8.577e-1 \\
2 & S-HMC\(_\parallel\) & 6.494e-2 & 2.844e-1 & 3.395e-1 & 2.198e-1 & 9.977e-1 & 4.432e-1 & 2.472e-1 & 9.223e-1 & 1.112e+0 & 7.769e-1 & 1.506e+0 & 8.436e-1 \\
\midrule
4 & S-SMC\(_\parallel\) & 5.315e-2 & 2.471e-1 & 2.200e-1 & 1.465e-1 & 6.363e-1 & 3.288e-1 & 1.403e-1 & 7.019e-1 & 9.791e-1 & 7.130e-1 & 1.126e+0 & 9.159e-1 \\
4 & S-HMC\(_\parallel\) & 6.733e-2 & 2.924e-1 & 3.479e-1 & 2.179e-1 & 1.010e+0 & 4.740e-1 & 2.520e-1 & 9.300e-1 & 1.119e+0 & 7.823e-1 & 1.493e+0 & 8.866e-1 \\
\midrule
8 & S-SMC\(_\parallel\) & 5.872e-2 & 2.725e-1 & 2.440e-1 & 1.637e-1 & 7.309e-1 & 3.750e-1 & 1.374e-1 & 7.075e-1 & 1.001e+0 & 7.307e-1 & 1.220e+0 & 9.649e-1 \\
8 & S-HMC\(_\parallel\) & 6.982e-2 & 2.993e-1 & 3.524e-1 & 2.258e-1 & 1.066e+0 & 4.817e-1 & 2.553e-1 & 9.380e-1 & 1.127e+0 & 7.893e-1 & 1.539e+0 & 8.946e-1 \\
\midrule
& HMC (GS) & 5.034e-2 & 2.417e-1 & 2.713e-1 & 1.900e-1 & 9.303e-1 & 3.683e-1  & 2.076e-1 & 8.477e-1 & 1.001e+0 & 7.264e-1 & 1.430e+0 & 7.518e-1 \\
\bottomrule
\end{tabular}
\end{adjustbox}
\end{table}

\begin{table}[H]
\scriptsize
\caption{Comparison of S-SMC\(_\parallel\) ($P$ chains with $N=10$) and S-HMC\(_\parallel\) ($NP$ chains), with fixed $L=1$, $B=25$, $M=1$, $v=0.025$, $s=0.1$, on IMDb (5 realizations and $\pm$ s.e.\ in accuracy).}
\label{tab:metric_diffP_imdb_v0025_s01}
\centering
\setlength{\tabcolsep}{3pt}
\begin{tabular}{lc|c|c|c|cc|ccccc}
\toprule
$P$ & Method & Ep. & Acc. & NL & \multicolumn{7}{c}{$H_{\sf ep}$} \\
\cmidrule(lr){6-12}
& &  &  &  & \multicolumn{2}{c|}{ID} & \multicolumn{5}{c}{OOD}\\
\cmidrule(lr){6-7}\cmidrule(lr){8-12}
& &  &  &  & cor. & inc. & reviews & meta & lipsum & full reviews & full meta \\
\midrule
1 & S-SMC\(_\parallel\) & 18.60 & 86.70$\pm$0.03 & 3.655e-1 & 1.122e-4 & 1.664e-4 & 1.792e-4 & 2.200e-4 & 1.749e-4 & 3.285e-4 & 3.212e-4 \\
1 & S-HMC\(_\parallel\) & 25 & 86.70$\pm$0.01 & 3.634e-1 & 2.418e-4 & 3.565e-4 & 4.598e-4 & 4.633e-4 & 4.260e-4 & 5.057e-4 & 5.410e-4 \\
\midrule
2 & S-SMC\(_\parallel\) & 18.70 & 86.72$\pm$0.03 & 3.656e-1 & 1.936e-4 & 2.798e-4 & 3.061e-4 & 3.358e-4 & 4.211e-4 & 4.491e-4 & 4.465e-4 \\
2 & S-HMC\(_\parallel\) & 25 & 86.69$\pm$0.01 & 3.634e-1 & 2.697e-4 & 3.955e-4 & 5.604e-4 & 5.852e-4 & 4.916e-4 & 6.022e-4 & 6.819e-4 \\
\midrule
4 & S-SMC\(_\parallel\) & 19.10 & 86.68$\pm$0.02 & 3.654e-1 & 2.370e-4 & 3.452e-4 & 4.433e-4 & 4.874e-4 & 5.515e-4 & 5.848e-4 & 6.201e-4 \\
4 & S-HMC\(_\parallel\) & 25 & 86.72$\pm$0.01 & 3.635e-1 & 2.776e-4 & 4.042e-4 & 5.940e-4 & 6.264e-4 & 5.629e-4 & 7.074e-4 & 7.288e-4 \\
\midrule
8 & S-SMC\(_\parallel\) & 19.15 & 86.69$\pm$0.01 & 3.653e-1 & 2.531e-4 & 3.697e-4 & 4.744e-4 & 4.971e-4 & 4.862e-4 & 5.876e-4 & 6.105e-4 \\
8 & S-HMC\(_\parallel\) & 25 & 86.72$\pm$0.00 & 3.633e-1 & 2.766e-4 & 4.022e-4 & 5.694e-4 & 6.062e-4 & 5.637e-4 & 7.438e-4 & 7.051e-4 \\
\bottomrule
\end{tabular}

\vspace{0.2cm}

\begin{tabular}{lc|c|c|cc|ccccc}
\toprule
$P$ & Method & Brier & ECE & \multicolumn{7}{c}{$H_{\sf tot}$} \\
\cmidrule(lr){5-11}
& &&& \multicolumn{2}{c|}{ID} & \multicolumn{5}{c}{OOD} \\
\cmidrule(lr){5-6} \cmidrule(lr){7-11}
& & &&cor. & inc. & reviews & meta & lipsum & full reviews & full meta \\
\midrule
1 & S-SMC\(_\parallel\) & 1.093e-1 & 3.699e-1 & 4.792e-1 & 6.357e-1 & 5.251e-1 & 5.463e-1 & 5.142e-1 & 6.457e-1 & 6.261e-1 \\
1 & S-HMC\(_\parallel\) & 1.086e-1 & 3.694e-1 & 4.750e-1 & 6.340e-1 & 5.165e-1 & 5.331e-1 & 5.079e-1 & 6.400e-1 & 6.186e-1 \\
\midrule
2 & S-SMC\(_\parallel\) & 1.093e-1 & 3.702e-1 & 4.793e-1 & 6.356e-1 & 5.243e-1 & 5.466e-1 & 5.131e-1 & 6.462e-1 & 6.270e-1 \\
2 & S-HMC\(_\parallel\) & 1.086e-1 & 3.695e-1 & 4.752e-1 & 6.342e-1 & 5.172e-1 & 5.342e-1 & 5.092e-1 & 6.409e-1 & 6.196e-1 \\
\midrule
4 & S-SMC\(_\parallel\) & 1.092e-1 & 3.699e-1 & 4.787e-1 & 6.355e-1 & 5.230e-1 & 5.440e-1 & 5.127e-1 & 6.459e-1 & 6.259e-1 \\
4 & S-HMC\(_\parallel\) & 1.086e-1 & 3.699e-1 & 4.754e-1 & 6.342e-1 & 5.180e-1 & 5.351e-1 & 5.085e-1 & 6.419e-1 & 6.203e-1 \\
\midrule
8 & S-SMC\(_\parallel\) & 1.092e-1 & 3.698e-1 & 4.788e-1 & 6.355e-1 & 5.213e-1 & 5.406e-1 & 5.115e-1 & 6.430e-1 & 6.236e-1 \\
8 & S-HMC\(_\parallel\) & 1.086e-1 & 3.701e-1 & 4.752e-1 & 6.341e-1 & 5.184e-1 & 5.359e-1 & 5.085e-1 & 6.426e-1 & 6.209e-1 \\
\bottomrule
\end{tabular}
\end{table}

\begin{table}[h]
\scriptsize
\caption{Comparison of S-SMC\(_\parallel\) ($P$ chains with $N=10$) and S-HMC\(_\parallel\) ($NP$ chains), with fixed number of leapfrog $L=1$, $B=26$, $M=1$, $v=1$ and \(s=0.35\), on IMDb (5 realizations and \(\pm\)s.e. in accuracy ).}
\label{tab:metric_diffP_imdb_s035}
\centering
\setlength{\tabcolsep}{3pt}
\begin{tabular}{lc|c|c|c|cc|ccccc}
\toprule
$P$ & Method & Ep. & Acc. & NL & \multicolumn{7}{c}{$H_{\sf ep}$} \\
 \cmidrule(lr){6-12}
& &  &  &  & \multicolumn{2}{c|}{ID} & \multicolumn{5}{c}{OOD}\\
\cmidrule(lr){6-7} \cmidrule(lr){8-12}
& &  &  &  & cor. & inc. & reviews & meta & lipsum & full reviews & full meta \\
\midrule
1 & S-SMC\(_\parallel\) & 27.40 & 88.27$\pm$0.07 & 2.803e-1 & 6.177e-4 & 1.460e-3 & 1.581e-3 & 2.495e-3 & 2.187e-3 & 2.279e-3 & 2.504e-3 \\
1 & S-HMC\(_\parallel\) & 26.00 & 88.81$\pm$0.01 & 2.750e-1 & 1.565e-2 & 3.463e-2 & 5.414e-2 & 7.021e-2 & 6.872e-2 & 8.407e-2 & 7.930e-2 \\
\midrule
2 & S-SMC\(_\parallel\) & 27.70 & 88.65$\pm$0.04 & 2.744e-1 & 5.513e-3 & 1.323e-2 & 2.132e-2 & 3.443e-2 & 2.751e-2 & 2.232e-2 & 3.585e-2 \\
2 & S-HMC\(_\parallel\) & 26.00 & 88.86$\pm$0.03 & 2.745e-1 & 1.620e-2 & 3.584e-2 & 5.510e-2 & 7.622e-2 & 7.118e-2 & 8.766e-2 & 8.490e-2 \\
\midrule
4 & S-SMC\(_\parallel\) & 28.55 & 88.78$\pm$0.03 & 2.726e-1 & 8.040e-3 & 1.881e-2 & 3.057e-2 & 4.854e-2 & 5.097e-2 & 5.547e-2 & 5.766e-2 \\
4 & S-HMC\(_\parallel\) & 26.00 & 88.88$\pm$0.01 & 2.740e-1 & 1.640e-2 & 3.610e-2 & 5.388e-2 & 7.488e-2 & 6.705e-2 & 8.588e-2 & 8.184e-2 \\
\midrule
8 & S-SMC\(_\parallel\) & 27.63 & 88.88$\pm$0.03 & 2.714e-1 & 9.342e-3 & 2.164e-2 & 3.756e-2 & 5.515e-2 & 5.260e-2 & 6.049e-2 & 6.435e-2 \\
8 & S-HMC\(_\parallel\) & 26.00 & 88.93$\pm$0.02 & 2.737e-1 & 1.662e-2 & 3.651e-2 & 5.315e-2 & 7.391e-2 & 6.917e-2 & 9.098e-2 & 8.360e-2 \\
\bottomrule
\end{tabular}

\vspace{0.2cm}

\begin{tabular}{lc|c|c|cc|ccccc}
\toprule
\(P\) & Method & Brier & ECE & \multicolumn{7}{c}{$H_{\sf tot}$} \\
\cmidrule(lr){5-11}
& & &&\multicolumn{2}{c|}{ID} & \multicolumn{5}{c}{OOD} \\
\cmidrule(lr){5-6} \cmidrule(lr){7-11}
& & &&cor. & inc. & reviews & meta & lipsum & full reviews & full meta \\
\midrule
1 & S-SMC\(_\parallel\) & 8.547e-2 & 3.832e-1 & 2.643e-1 & 5.482e-1 & 3.802e-1 & 5.116e-1 & 5.172e-1 & 5.581e-1 & 5.286e-1 \\
1 & S-HMC\(_\parallel\) & 8.298e-2 & 3.889e-1 & 2.890e-1 & 5.681e-1 & 4.289e-1 & 5.583e-1 & 5.556e-1 & 6.133e-1 & 6.120e-1 \\
\midrule
2 & S-SMC\(_\parallel\) & 8.327e-2 & 3.868e-1 & 2.694e-1 & 5.548e-1 & 3.937e-1 & 5.456e-1 & 5.091e-1 & 5.790e-1 & 5.742e-1 \\
2 & S-HMC\(_\parallel\) & 8.281e-2 & 3.896e-1 & 2.891e-1 & 5.681e-1 & 4.285e-1 & 5.684e-1 & 5.521e-1 & 6.125e-1 & 6.107e-1 \\
\midrule
4 & S-SMC\(_\parallel\) & 8.254e-2 & 3.884e-1 & 2.720e-1 & 5.576e-1 & 3.985e-1 & 5.601e-1 & 5.295e-1 & 5.963e-1 & 5.815e-1 \\
4 & S-HMC\(_\parallel\) & 8.262e-2 & 3.898e-1 & 2.891e-1 & 5.684e-1 & 4.232e-1 & 5.673e-1 & 5.377e-1 & 6.139e-1 & 6.100e-1 \\
\midrule
8 & S-SMC\(_\parallel\) & 8.206e-2 & 3.899e-1 & 2.744e-1 & 5.596e-1 & 3.987e-1 & 5.555e-1 & 5.304e-1 & 6.025e-1 & 5.920e-1 \\
8 & S-HMC\(_\parallel\) & 8.254e-2 & 3.904e-1 & 2.893e-1 & 5.683e-1 & 4.188e-1 & 5.626e-1 & 5.386e-1 & 6.156e-1 & 6.088e-1 \\
\bottomrule
\end{tabular}
\end{table}

\begin{table}[h]
\scriptsize
\caption{Comparison of S-SMC\(_\parallel\) ($P$ chains with $N=10$) and S-HMC\(_\parallel\) ($NP$ chains), with fixed number of leapfrog $L=1$, $B=26$, $M=2$, $v=1$ and \(s=0.25\), on IMDb (5 realizations and \(\pm\) s.e.\ in accuracy).}
\label{tab:metric_diffP_imdb_s025}
\centering
\setlength{\tabcolsep}{3pt}
\begin{tabular}{lc|c|c|c|cc|ccccc}
\toprule
$P$ & Method & Ep. & Acc. & NL & \multicolumn{7}{c}{$H_{\sf ep}$} \\
 \cmidrule(lr){6-12}
& &  &  &  & \multicolumn{2}{c|}{ID} & \multicolumn{5}{c}{OOD}\\
\cmidrule(lr){6-7} \cmidrule(lr){8-12}
& &  &  &  & cor. & inc. & reviews & meta & lipsum & full reviews & full meta \\
\midrule
1 & S-SMC\(_\parallel\) & 29.6 & 88.27\(\pm\)0.10 & 2.807e-1 & 2.512e-4 & 6.069e-4 & 4.733e-4 & 7.166e-4 & 8.590e-4 & 9.313e-4 & 9.520e-4 \\
1 & S-HMC\(_\parallel\) & 26 & 88.83\(\pm\)0.02 & 2.745e-1 & 1.269e-2 & 2.830e-2 & 4.522e-2 & 5.927e-2 & 5.886e-2 & 7.342e-2 & 6.803e-2 \\
\midrule
2 & S-SMC\(_\parallel\) & 27.2 & 88.64\(\pm\)0.05 & 2.752e-1 & 4.948e-3 & 1.174e-2 & 1.344e-2 & 2.120e-2 & 1.410e-2 & 2.411e-2 & 2.792e-2 \\
2 & S-HMC\(_\parallel\) & 26 & 88.84\(\pm\)0.03 & 2.741e-1 & 1.304e-2 & 2.918e-2 & 4.586e-2 & 6.400e-2 & 6.096e-2 & 7.596e-2 & 7.216e-2 \\
\midrule
4 & S-SMC\(_\parallel\) & 28.3 & 88.77\(\pm\)0.04 & 2.737e-1 & 7.211e-3 & 1.662e-2 & 2.329e-2 & 4.029e-2 & 4.913e-2 & 5.544e-2 & 4.798e-2 \\
4 & S-HMC\(_\parallel\) & 26 & 88.90\(\pm\)0.1 & 2.736e-1 & 1.321e-2 & 2.932e-2 & 4.487e-2 & 6.264e-2 & 5.689e-2 & 7.369e-2 & 6.930e-2 \\
\midrule
8 & S-SMC\(_\parallel\) & 28.5 & 88.87\(\pm\)0.03 & 2.720e-1 & 8.124e-3 & 1.872e-2 & 3.066e-2 & 5.057e-2 & 4.691e-2 & 6.403e-2 & 5.777e-2 \\
8 & S-HMC\(_\parallel\) & 26 & 88.92\(\pm\)0.02 & 2.734e-1 & 1.337e-2 & 2.964e-2 & 4.442e-2 & 6.215e-2 & 5.863e-2 & 7.869e-2 & 7.108e-2 \\
\bottomrule
\end{tabular}

\vspace{0.2cm}

\begin{tabular}{lc|c|c|cc|ccccc}
\toprule
\(P\) & Method & Brier & ECE & \multicolumn{7}{c}{$H_{\sf tot}$} \\
\cmidrule(lr){5-11}
& & &&\multicolumn{2}{c|}{ID} & \multicolumn{5}{c}{OOD} \\
\cmidrule(lr){5-6} \cmidrule(lr){7-11}
& &&& cor. & inc. & reviews & meta & lipsum & full reviews & full meta \\
\midrule
1 & S-SMC\(_\parallel\) & 8.548e-2 & 3.825e-1 & 2.673e-1 & 5.500e-1 & 3.562e-1 & 4.872e-1 & 4.727e-1 & 5.548e-1 & 5.609e-1 \\
1 & S-HMC\(_\parallel\) & 8.286e-2 & 3.896e-1 & 2.873e-1 & 5.667e-1 & 4.239e-1 & 5.585e-1 & 5.533e-1 & 6.117e-1 & 6.076e-1 \\
\midrule
2 & S-SMC\(_\parallel\) & 8.343e-2 & 3.867e-1 & 2.723e-1 & 5.560e-1 & 3.696e-1 & 5.182e-1 & 4.861e-1 & 5.794e-1 & 5.849e-1 \\
2 & S-HMC\(_\parallel\) & 8.271e-2 & 3.894e-1 & 2.870e-1 & 5.668e-1 & 4.228e-1 & 5.661e-1 & 5.478e-1 & 6.112e-1 & 6.061e-1 \\
\midrule
4 & S-SMC\(_\parallel\) & 8.283e-2 & 3.888e-1 & 2.758e-1 & 5.589e-1 & 3.823e-1 & 5.388e-1 & 5.321e-1 & 5.920e-1 & 5.801e-1 \\
4 & S-HMC\(_\parallel\) & 8.254e-2 & 3.902e-1 & 2.870e-1 & 5.665e-1 & 4.174e-1 & 5.633e-1 & 5.328e-1 & 6.118e-1 & 6.058e-1 \\
\midrule
8 & S-SMC\(_\parallel\) & 8.220e-2 & 3.901e-1 & 2.772e-1 & 5.609e-1 & 3.891e-1 & 5.447e-1 & 5.340e-1 & 6.036e-1 & 5.874e-1 \\
8 & S-HMC\(_\parallel\) & 8.249e-2 & 3.904e-1 & 2.871e-1 & 5.665e-1 & 4.138e-1 & 5.595e-1 & 5.338e-1 & 6.135e-1 & 6.047e-1 \\
\bottomrule
\end{tabular}
\end{table}

\begin{table}[H]
\scriptsize
\caption{Comparison of S-SMC\(_\parallel\) ($P$ chains with $N=10$) and S-HMC\(_\parallel\) ($NP$ chains), with fixed number of leapfrog $L=1$, $B=26$, $M=2$, $v=1$ and \(s=0.1\), on IMDb (5 realizations and \(\pm\) s.e.\ in accuracy).}
\label{tab:metric_diffP_imdb_s01}
\centering
\setlength{\tabcolsep}{3pt}
\begin{tabular}{lc|c|c|c|cc|ccccc}
\toprule
$P$ & Method & Ep. & Acc. & NL & \multicolumn{7}{c}{$H_{\sf ep}$} \\
 \cmidrule(lr){6-12}
& &  &  &  & \multicolumn{2}{c|}{ID} & \multicolumn{5}{c}{OOD}\\
\cmidrule(lr){6-7} \cmidrule(lr){8-12}
& &  &  &  & cor. & inc. & reviews & meta & lipsum & full reviews & full meta \\
\midrule
1 & S-SMC\(_\parallel\) & 24    & 88.54\(\pm\)0.11 & 2.763e-1 & 1.820e-4 & 4.315e-4 & 4.819e-4 & 5.915e-4 & 6.766e-4 & 7.711e-4 & 6.500e-4  \\
1 & S-HMC\(_\parallel\) & 26    & 88.86\(\pm\)0.02 & 2.726e-1 & 5.753e-3 & 1.319e-2 & 2.712e-2 & 3.551e-2 & 3.561e-2 & 5.110e-2 & 4.289e-2  \\
\midrule
2 & S-SMC\(_\parallel\) & 23.2  & 88.78\(\pm\)0.08 & 2.727e-1 & 2.476e-3 & 5.724e-3 & 1.160e-2 & 1.973e-2 & 9.401e-3 & 1.314e-2 & 1.894e-2  \\
2 & S-HMC\(_\parallel\) & 26    & 88.90\(\pm\)0.02 & 2.724e-1 & 5.934e-3 & 1.365e-2 & 2.817e-2 & 3.918e-2 & 3.793e-2 & 5.296e-2 & 4.581e-2  \\
\midrule
4 & S-SMC\(_\parallel\) & 23.5  & 88.85\(\pm\)0.05 & 2.722e-1 & 3.730e-3 & 8.583e-3 & 1.937e-2 & 2.851e-2 & 2.377e-2 & 3.216e-2 & 3.538e-2  \\
4 & S-HMC\(_\parallel\) & 26    & 88.92\(\pm\)0.01 & 2.722e-1 & 5.992e-3 & 1.368e-2 & 2.753e-2 & 3.839e-2 & 3.531e-2 & 5.049e-2 & 4.444e-2  \\
\midrule
8 & S-SMC\(_\parallel\) & 23.7  & 88.92\(\pm\)0.01 & 2.711e-1 & 4.207e-3 & 9.768e-3 & 2.182e-2 & 3.140e-2 & 2.700e-2 & 3.352e-2 & 3.540e-2  \\
8 & S-HMC\(_\parallel\) & 26    & 88.93\(\pm\)0.01 & 2.721e-1 & 6.065e-3 & 1.386e-2 & 2.792e-2 & 3.888e-2 & 3.653e-2 & 5.408e-2 & 4.638e-2  \\
\bottomrule
\end{tabular}

\vspace{0.2cm}

\begin{tabular}{lc|c|c|cc|ccccc}
\toprule
\(P\) & Method & Brier & ECE & \multicolumn{7}{c}{$H_{\sf tot}$} \\
\cmidrule(lr){5-11}
& & &&\multicolumn{2}{c|}{ID} & \multicolumn{5}{c}{OOD} \\
\cmidrule(lr){5-6} \cmidrule(lr){7-11}
& & &&cor. & inc. & reviews & meta & lipsum & full reviews & full meta \\
\midrule
1 & S-SMC\(_\parallel\) & 8.387e-2 & 3.869e-1 & 2.721e-1 & 5.544e-1 & 3.970e-1 & 5.202e-1 & 5.063e-1 & 5.686e-1 & 5.684e-1  \\
1 & S-HMC\(_\parallel\) & 8.232e-2 & 3.898e-1 & 2.802e-1 & 5.618e-1 & 4.088e-1 & 5.505e-1 & 5.384e-1 & 6.066e-1 & 5.975e-1  \\
\midrule
2 & S-SMC\(_\parallel\) & 8.252e-2 & 3.892e-1 & 2.734e-1 & 5.571e-1 & 3.939e-1 & 5.263e-1 & 5.028e-1 & 5.812e-1 & 5.855e-1  \\
2 & S-HMC\(_\parallel\) & 8.227e-2 & 3.900e-1 & 2.801e-1 & 5.614e-1 & 4.092e-1 & 5.567e-1 & 5.347e-1 & 6.068e-1 & 5.966e-1  \\
\midrule
4 & S-SMC\(_\parallel\) & 8.229e-2 & 3.900e-1 & 2.750e-1 & 5.583e-1 & 3.944e-1 & 5.389e-1 & 5.114e-1 & 5.948e-1 & 5.854e-1  \\
4 & S-HMC\(_\parallel\)  & 8.218e-2 & 3.903e-1 & 2.800e-1 & 5.612e-1 & 4.041e-1 & 5.526e-1 & 5.222e-1 & 6.068e-1 & 5.967e-1  \\
\midrule
8 & S-SMC\(_\parallel\) & 8.190e-2 & 3.906e-1 & 2.752e-1 & 5.588e-1 & 3.920e-1 & 5.407e-1 & 5.143e-1 & 5.959e-1 & 5.865e-1  \\
8 & S-HMC\(_\parallel\) & 8.217e-2 & 3.906e-1 & 2.801e-1 & 5.613e-1 & 4.028e-1 & 5.510e-1 & 5.231e-1 & 6.081e-1 & 5.961e-1  \\
\bottomrule
\end{tabular}
\end{table}

\begin{table}[h]
\scriptsize
\caption{Comparison of S-SMC\(_\parallel\) ($P$ chains with $N=10$) and S-HMC\(_\parallel\) ($NP$ chains), with fixed number of leapfrog $L=1$, $B=200$, $M=4$, $v=0.2$ and \(s=0.05\), on CIFAR10 (5 realizations and \(\pm\)s.e. in accuracy).}
\label{tab:metric_diffP_cifar_s005}
\centering
\setlength{\tabcolsep}{3pt}
\begin{tabular}{lc|c|c|c|c|c}
\toprule
$P$ & Method & Ep. & Acc. & NL & Brier & ECE \\
\midrule
1 & S-SMC\(_\parallel\)      & 168.8 & 89.26 \(\pm\) 0.07 & 3.408e-1 & 1.580e-1 & 3.470e-2 \\
1 & S-HMC\(_\parallel\)      & 200   & 90.67 \(\pm\) 0.03 & 2.749e-1 & 1.366e-1 & 6.598e-3 \\
\midrule
2 & S-SMC\(_\parallel\)      & 172.0 & 90.12 \(\pm\) 0.06 & 3.100e-1 & 1.466e-1 & 1.942e-2 \\
2 & S-HMC\(_\parallel\)      & 200   & 90.80 \(\pm\) 0.02 & 2.707e-1 & 1.351e-1 & 5.574e-3 \\
\midrule
4 & S-SMC\(_\parallel\)      & 173.4 & 90.34 \(\pm\) 0.04 & 2.960e-1 & 1.399e-1 & 1.242e-2 \\
4 & S-HMC\(_\parallel\)      & 200   & 90.84 \(\pm\) 0.04 & 2.688e-1 & 1.344e-1 & 6.442e-3 \\
\midrule
8 & S-SMC\(_\parallel\)      & 174.3 & 90.63 \(\pm\) 0.05 & 2.881e-1 & 1.371e-1 & 9.720e-3 \\
8 & S-HMC\(_\parallel\)      & 200   & 90.84 \(\pm\) 0.03 & 2.677e-1 & 1.340e-1 & 6.601e-3 \\
\bottomrule
\end{tabular}

\vspace{0.2cm}

\begin{tabular}{lc|cc|ccc|cc|ccc}
\toprule
\(P\) & Method & \multicolumn{5}{c|}{$H_{\sf ep}$} & \multicolumn{5}{c}{$H_{\sf tot}$} \\
\cmidrule(lr){3-7} \cmidrule(lr){8-12}
& & \multicolumn{2}{c|}{ID} & \multicolumn{3}{c|}{OOD} & \multicolumn{2}{c|}{ID} & \multicolumn{3}{c}{OOD} \\
\cmidrule(lr){3-4} \cmidrule(lr){5-7} \cmidrule(lr){8-9} \cmidrule(lr){10-12}
& & cor. & inc. & close & corrupt & far & cor. & inc. & close & corrupt & far \\
\midrule
1 & S-SMC\(_\parallel\) & 4.258e-4 & 2.273e-3 & 2.515e-3 & 1.297e-3 & 2.185e-3 & 1.351e-1 & 6.620e-1 & 6.639e-1 & 4.300e-1 & 9.008e-1 \\
1 & S-HMC\(_\parallel\) & 4.060e-2 & 1.804e-1 & 2.308e-1 & 1.228e-1 & 2.243e-1 & 1.917e-1 & 8.073e-1 & 8.900e-1 & 5.558e-1 & 1.154e+0 \\
\midrule
2 & S-SMC\(_\parallel\) & 1.796e-2 & 9.094e-2 & 1.115e-1 & 5.702e-2 & 1.340e-1 & 1.321e-1 & 6.646e-1 & 7.737e-1 & 4.864e-1 & 1.017e+0 \\
2 & S-HMC\(_\parallel\) & 4.376e-2 & 1.896e-1 & 2.454e-1 & 1.300e-1 & 2.387e-1 & 1.956e-1 & 8.144e-1 & 9.064e-1 & 5.626e-1 & 1.162e+0 \\
\midrule
4 & S-SMC\(_\parallel\) & 2.855e-2 & 1.356e-1 & 1.700e-1 & 8.859e-2 & 1.862e-1 & 1.304e-1 & 6.700e-1 & 8.293e-1 & 5.188e-1 & 1.101e+0 \\
4 & S-HMC\(_\parallel\) & 4.523e-2 & 1.937e-1 & 2.520e-1 & 1.337e-1 & 2.417e-1 & 1.970e-1 & 8.161e-1 & 9.129e-1 & 5.660e-1 & 1.163e+0 \\
\midrule
8 & S-SMC\(_\parallel\) & 3.507e-2 & 1.584e-1 & 2.027e-1 & 1.058e-1 & 1.961e-1 & 1.311e-1 & 6.684e-1 & 8.643e-1 & 5.368e-1 & 1.111e+0 \\
8 & S-HMC\(_\parallel\) & 4.579e-2 & 1.966e-1 & 2.564e-1 & 1.358e-1 & 2.472e-1 & 1.971e-1 & 8.203e-1 & 9.173e-1 & 5.684e-1 & 1.168e+0 \\
\bottomrule
\end{tabular}
\end{table}

\begin{table}[h]
\scriptsize
\caption{Comparison of S-SMC\(_\parallel\) ($P$ chains with $N=10$) and S-HMC\(_\parallel\) ($NP$ chains), with fixed number of leapfrog $L=1$, $B=200$, $M=4$, $v=0.2$ and \(s=0.1\), on CIFAR10 (5 realizations and \(\pm\)s.e. in accuracy).}
\label{tab:metric_diffP_cifar_s01}
\centering
\setlength{\tabcolsep}{3pt}
\begin{tabular}{lc|c|c|c|c|c}
\toprule
$P$ & Method & Ep. & Acc. & NL & Brier & ECE \\
\midrule
1 & S-SMC\(_\parallel\) & 229.6 & 88.26 \(\pm\) 0.07 & 3.855e-1 & 1.770e-1 & 4.593e-2\\
1 & S-HMC\(_\parallel\) & 200 & 90.57 \(\pm\) 0.04 & 2.823e-1 & 1.398e-1 & 1.073e-2\\
\midrule
2 & S-SMC\(_\parallel\) & 226.0 & 89.62 \(\pm\) 0.09 & 3.336e-1 & 1.553e-1 & 1.983e-2\\
2 & S-HMC\(_\parallel\) & 200 & 90.76 \(\pm\) 0.04 & 2.753e-1 & 1.372e-1 & 1.356e-2\\
\midrule
4 & S-SMC\(_\parallel\) & 224.8e & 90.20 \(\pm\) 0.08 & 3.100e-1 & 1.453e-1 & 9.413e-3\\
4 & S-HMC\(_\parallel\) & 200 & 90.84 \(\pm\) 0.04 & 2.722e-1 & 1.360e-1 & 1.517e-2 \\
\midrule
8 & S-SMC\(_\parallel\) & 225.3 & 90.45 \(\pm\) 0.06 & 2.980e-1 & 1.400e-1 & 7.737e-3\\
8 & S-HMC\(_\parallel\) & 200 & 90.83 \(\pm\) 0.03 & 2.701e-1 & 1.353e-1 & 1.517e-2\\
\bottomrule
\end{tabular}

\vspace{0.2cm}

\begin{tabular}{lc|cc|ccc|cc|ccc}
\toprule
\(P\) & Method & \multicolumn{5}{c|}{$H_{\sf ep}$} & \multicolumn{5}{c}{$H_{\sf tot}$} \\
\cmidrule(lr){3-7} \cmidrule(lr){8-12}
& & \multicolumn{2}{c|}{ID} & \multicolumn{3}{c|}{OOD} & \multicolumn{2}{c|}{ID} & \multicolumn{3}{c}{OOD} \\
\cmidrule(lr){3-4} \cmidrule(lr){5-7} \cmidrule(lr){8-9} \cmidrule(lr){10-12}
& & cor. & inc. & close & corrupt & far & cor. & inc. & close & corrupt & far \\
\midrule
1 & S-SMC\(_\parallel\) & 2.948e-4 & 1.507e-3 & 1.603e-3 & 8.256e-4 & 1.450e-3 & 1.309e-1 & 6.364e-1 & 6.121e-1 & 4.027e-1 & 9.110e-1 \\
1 & S-HMC\(_\parallel\) & 7.055e-2 & 2.795e-1 & 3.591e-1 & 1.972e-1 & 3.581e-1 & 2.241e-1 & 8.596e-1 & 9.703e-1 & 6.102e-1 & 1.219e+0 \\
\midrule
2 & S-SMC\(_\parallel\) & 2.733e-2 & 1.343e-1 & 1.594e-1 & 8.581e-2 & 1.677e-1 & 1.256e-1 & 6.382e-1 & 7.735e-1 & 4.910e-1 & 1.037e+0 \\
2 & S-HMC\(_\parallel\) & 9.734e-2 & 2.968e-1 & 3.862e-1 & 2.112e-1 & 3.821e-1 & 2.316e-1 & 8.732e-1 & 9.990e-1 & 6.238e-1 & 1.235e+0 \\
\midrule
4 & S-SMC\(_\parallel\) & 4.484e-2 & 2.002e-1 & 2.495e-1 & 1.365e-1 & 2.536e-1 & 1.235e-1 & 6.432e-1 & 8.577e-1 & 5.408e-1 & 1.135e+0 \\
4 & S-HMC\(_\parallel\) & 1.018e-1 & 3.072e-1 & 4.011e-1 & 2.200e-1 & 3.936e-1 & 2.360e-1 & 8.822e-1 & 1.014e+0 & 6.323e-1 & 1.247e+0 \\
\midrule
8 & S-SMC\(_\parallel\) & 5.539e-2 & 2.369e-1 & 3.054e-1 & 1.636e-1 & 2.937e-1 & 1.217e-1 & 6.453e-1 & 9.184e-1 & 5.690e-1 & 1.156e+0 \\
8 & S-HMC\(_\parallel\) & 1.035e-1 & 3.121e-1 & 4.095e-1 & 2.244e-1 & 4.031e-1 & 2.367e-1 & 8.884e-1 & 1.023e+0 & 6.371e-1 & 1.257e+0 \\
\bottomrule
\end{tabular}
\end{table}

\begin{table}[h]
\scriptsize
\caption{Comparison of S-SMC\(_\parallel\) ($P$ chains with $N=10$) and S-HMC\(_\parallel\) ($NP$ chains), with fixed number of leapfrog $L=1$, $B=200$, $M=4$, $v=0.2$ and \(s=0.2\), on CIFAR10 (5 realizations and \(\pm\)s.e. in accuracy).}
\label{tab:metric_diffP_cifar_s02}
\centering
\setlength{\tabcolsep}{3pt}
\begin{tabular}{lc|c|c|c|c|c}
\toprule
$P$ & Method & Ep. & Acc. & NL & Brier & ECE  \\
\midrule
1 & S-SMC\(_\parallel\) & 289.6 & 86.99 \(\pm\) 0.08 & 4.710e-1 & 2.007e-1 & 6.462e-2\\
1 & S-HMC\(_\parallel\) & 200 & 90.23 \(\pm\) 0.08 & 2.990e-1 & 1.466e-1 & 2.518e-2\\
\midrule
2 & S-SMC\(_\parallel\) & 289.6 & 88.77 \(\pm\) 0.07 & 3.854e-1 & 1.699e-1 & 2.554e-2\\
2 & S-HMC\(_\parallel\) & 200 & 90.53 \(\pm\) 0.04 & 2.890e-1 & 1.426e-1 & 3.096e-2\\
\midrule
4 & S-SMC\(_\parallel\) & 289.4 & 89.82 \(\pm\) 0.04 & 3.441e-1 & 1.536e-1 & 1.193e-2\\
4 & S-HMC\(_\parallel\) & 200 & 90.73 \(\pm\) 0.02 & 2.840e-1 & 1.406e-1 & 3.368e-2\\
\midrule
8 & S-SMC\(_\parallel\) & 289.3 & 90.30 \(\pm\) 0.03 & 3.217e-1 & 1.445e-1 & 1.180e-2\\
8 & S-HMC\(_\parallel\) & 200 & 90.82 \(\pm\) 0.03 & 2.810e-1 & 1.395e-1 & 3.481e-2\\
\bottomrule
\end{tabular}

\vspace{0.2cm}

\begin{tabular}{lc|cc|ccc|cc|ccc}
\toprule
\(P\) & Method & \multicolumn{5}{c|}{$H_{\sf ep}$} & \multicolumn{5}{c}{$H_{\sf tot}$} \\
\cmidrule(lr){3-7} \cmidrule(lr){8-12}
& & \multicolumn{2}{c|}{ID} & \multicolumn{3}{c|}{OOD} & \multicolumn{2}{c|}{ID} & \multicolumn{3}{c}{OOD} \\
\cmidrule(lr){3-4} \cmidrule(lr){5-7} \cmidrule(lr){8-9} \cmidrule(lr){10-12}
& & cor. & inc. & close & corrupt & far & cor. & inc. & close & corrupt & far \\
\midrule
1 & S-SMC\(_\parallel\) & 3.682e-4 & 1.947e-3 & 2.063e-3 & 1.092e-3 & 1.629e-3 & 1.136e-1 & 5.613e-1 & 5.440e-1 & 3.630e-1 & 7.756e-1 \\
1 & S-HMC\(_\parallel\) & 1.159e-1 & 4.091e-1 & 5.195e-1 & 2.993e-1 & 5.333e-1 & 2.676e-1 & 9.231e-1 & 1.059e+0 & 6.768e-1 & 1.297e+0 \\
\midrule
2 & S-SMC\(_\parallel\) & 3.863e-2 & 1.856e-1 & 2.117e-1 & 1.180e-1 & 2.030e-1 & 1.108e-1 & 5.822e-1 & 7.557e-1 & 4.901e-1 & 9.434e-1 \\
2 & S-HMC\(_\parallel\) & 1.300e-1 & 4.377e-1 & 5.681e-1 & 3.260e-1 & 5.774e-1 & 2.836e-1 & 9.479e-1 & 1.109e+0 & 7.039e-1 & 1.331e+0 \\
\midrule
4 & S-SMC\(_\parallel\) & 6.722e-2 & 2.780e-1 & 3.470e-1 & 1.916e-1 & 3.332e-1 & 1.089e-1 & 5.885e-1 & 8.861e-1 & 5.634e-1 & 1.087e+0 \\
4 & S-HMC\(_\parallel\) & 1.368e-1 & 4.524e-1 & 5.899e-1 & 3.392e-1 & 5.947e-1 & 2.909e-1 & 9.611e-1 & 1.132e+0 & 7.169e-1 & 1.346e+0 \\
\midrule
8 & S-SMC\(_\parallel\) & 8.362e-2 & 3.326e-1 & 4.244e-1 & 2.361e-1 & 4.065e-1 & 1.071e-1 & 5.954e-1 & 9.675e-1 & 6.080e-1 & 1.160e+0 \\
8 & S-HMC\(_\parallel\) & 1.405e-1 & 4.604e-1 & 6.042e-1 & 3.476e-1 & 6.129e-1 & 2.945e-1 & 9.687e-1 & 1.146e+0 & 7.256e-1 & 1.364e+0 \\
\bottomrule
\end{tabular}
\end{table}

\end{document}